\documentclass[letterpaper,journal]{IEEEtran}
\usepackage{amsmath,amsfonts}
\usepackage{algorithmic}
\usepackage{algorithm}
\usepackage{array}
\usepackage[caption=false,font=footnotesize,labelfont=sf,textfont=sf]{subfig}
\usepackage{textcomp}
\usepackage{stfloats}
\usepackage{url}
\usepackage{verbatim}
\usepackage{graphicx}
\usepackage{cite}
\usepackage{booktabs}
\usepackage{amssymb}
\usepackage{longtable}
\usepackage[table]{xcolor}
\usepackage{placeins}
\usepackage{float}
\usepackage[hidelinks]{hyperref}

\definecolor{statusI}{RGB}{207,226,243}
\definecolor{statusP}{RGB}{217,234,211}
\definecolor{statusD}{RGB}{244,204,204}
\definecolor{statusX}{RGB}{217,217,217}
\newcommand{\statI}{\cellcolor{statusI}\textbf{I}}
\newcommand{\statP}{\cellcolor{statusP}\textbf{P}}
\newcommand{\statD}{\cellcolor{statusD}\textbf{D}}
\newcommand{\statX}{\cellcolor{statusX}\textbf{X}}
\begin{document}
\bstctlcite{IEEEtran:BSTcontrol}

\title{Semantics-Guided Automatic Tensorization for Multiobjective Evolutionary Algorithms: \\A Multi-Agent Framework}

\author{Zhenyu Liang,
        Beichen Huang,
        Bowen Zheng, 
        and Ran Cheng
        \thanks{Zhenyu Liang, Beichen Huang, and Bowen Zheng are with the Department of Data Science and Artificial Intelligence, The Hong Kong Polytechnic University, Hong Kong SAR, China. E-mails: \{zhenyuliang97, bill.huang2001\}@gmail.com and bowen.zheng@protonmail.com.} 
        \thanks{Ran Cheng is with the Department of Data Science and Artificial Intelligence, The Hong Kong Polytechnic University, Hung Hom, Kowloon, Hong Kong SAR, China; the Hong Kong Polytechnic University Shenzhen Research Institute, Shenzhen 518057, Guangdong, China; and the Hong Kong Polytechnic University-Daya Bay Technology and Innovation Research Institute, Huizhou 516083, Guangdong, China. (E-mail: ranchengcn@gmail.com)}
}

\markboth{}{}

\maketitle

\begin{abstract}
Multiobjective evolutionary algorithms (MOEAs) naturally expose population-level parallelism, but many mature implementations encode their computation in sequential program structures designed for central processing units. Exploiting modern tensor computing platforms therefore requires more than direct code translation: the implementation must be restructured without changing the defining optimization mechanism of the underlying MOEA. We formulate automatic tensorization for MOEAs as semantics-guided computational restructuring and develop Evolutionary Code Conversion (EvoCoCo)\footnote{Source code: \url{https://github.com/EMI-Group/evococo}.}, a multi-agent framework that realizes this formulation. EvoCoCo reconstructs algorithm-specific states, dependencies, operators, and update logic into a structured semantic representation and organizes them through a shared tensorization blueprint. Specialized transformation branches then explore alternative tensor realizations, while execution feedback guides validation, repair, and candidate selection. Experiments on a benchmark of 48 MOEAs evaluate migration reliability, optimization fidelity, and computational scalability. Under matched large language model backends, EvoCoCo attains higher migration reliability than direct one-shot translation. Across the benchmark suites, 88.2\% of valid comparisons satisfy the predefined optimization-fidelity criterion. The tensorized implementations also exhibit increasing acceleration on graphics processing units as population size or decision dimension grows, with median measured speedups ranging from $22.6\times$ under population scaling to $80.2\times$ under decision-dimension scaling. External-source and ablation studies further assess transfer beyond the main benchmark and the roles of the major framework components. 
\end{abstract}


\section{Introduction}

\IEEEPARstart{M}{ultiobjective} evolutionary algorithms (MOEAs) approximate Pareto-optimal trade-offs in problems with multiple conflicting objectives~\cite{coello2007evolutionary,deb2002nsga2,zhang2007moead,liu2023learnable}. Such problems arise in engineering design, resource scheduling, path planning, energy systems, industrial control, machine learning, and scientific computing. Their computational burden can be dominated by expensive function evaluations~\cite{chugh2019survey} and can also grow with high-dimensional problem scale~\cite{tian2022largescale}. These costs limit the practical scale at which MOEAs can be applied.

The population-based structure of MOEAs also exposes substantial computational parallelism. Objective evaluation, variation, ranking, environmental selection, and archive operations repeatedly process candidate solutions or population-level relations that can often be evaluated concurrently. Modern graphics processing units (GPUs) can exploit this regularity through batched tensor operations~\cite{klosko2022high}. Consequently, tensorization provides a natural route to scalable MOEA computation when the underlying algorithmic structure can be expressed in a tensor-compatible form.

However, mature MOEA implementations often fail to expose this algorithmic parallelism in a GPU-compatible form. Many implementations combine vectorized operations with individual-level loops, dynamic containers, conditional branches, and framework-specific state management designed for central processing units (CPUs). The parallelism belongs to the algorithmic structure, whereas the computational bottleneck often lies in its program realization. GPU execution therefore requires the data representation, control structure, and state updates to be reorganized into tensor computation while the defining mechanism of the underlying MOEA remains intact.

Recent work has established GPU-accelerated infrastructure for evolutionary computation~\cite{huang2023evox}, general tensorization principles for MOEA implementations~\cite{liang2025tensorization}, and tensorized realizations of specific algorithms such as RVEA~\cite{liang2024tensorrvea} and NSGA-III~\cite{li2025tensornsga}. Tensorized GPU computation has also been investigated in constrained multiobjective optimization~\cite{huang2026fully}. Taken together, these studies demonstrate that computational restructuring can expose substantial GPU parallelism, but existing approaches rely largely on manual or algorithm-specific redesign and therefore do not resolve the automation problem. An automatic method must instead operate across mature implementations with heterogeneous states, operators, auxiliary functions, update schedules, and coding conventions. Moreover, the same algorithmic mechanism can admit different tensor realizations, while similar source structures can encode different algorithmic roles. Consequently, automatic tensorization requires a representation that separates what must be preserved from how the target computation may be restructured.

Closing this automation gap requires several distinct reasoning functions: reconstructing source algorithmic semantics, planning target computation, exploring alternative tensor realizations, diagnosing runtime failures, and assessing executable candidates. Large language models (LLMs) have demonstrated code-generation capabilities~\cite{chen2021codex} and program-translation capabilities~\cite{lachaux2020transcoder}. Execution-aware systems further show the value of repository interaction and test execution~\cite{yang2024sweagent}, as well as execution and performance signals for iterative GPU-program refinement~\cite{ouyang2025kernelbench}. However, automatic tensorization requires these heterogeneous functions to remain coordinated under a common interpretation of the source MOEA. This requirement motivates a computational architecture in which specialized reasoning roles exchange structured intermediate representations under shared semantic constraints.

To address these requirements, we formulate automatic tensorization for MOEAs as \emph{semantics-guided computational restructuring} and develop \textbf{Evolutionary Code Conversion (EvoCoCo)}, a multi-agent framework that realizes this formulation. EvoCoCo reconstructs source algorithmic semantics, organizes them into a shared tensorization blueprint, explores alternative tensor realizations, and closes the transformation loop through execution-guided refinement and selection. Its agent organization mirrors this functional decomposition and coordinates specialized roles through shared intermediate representations. The framework also exhibits an evolution-like search structure in which alternative computational realizations are generated under shared semantic constraints, evaluated in the target environment, refined when necessary, and selected according to target-side evidence. We instantiate and evaluate EvoCoCo by transforming heterogeneous PlatEMO implementations into tensorized EvoX/PyTorch implementations. The main contributions are summarized as follows:

\begin{itemize}
\item We formulate automatic tensorization for MOEAs as semantics-guided computational restructuring. For a concrete MOEA, the formulation treats its software implementation as the transformation object and the underlying optimization mechanism as the semantic constraint.

\item We develop EvoCoCo, a multi-agent framework whose specialized roles mirror the functional decomposition of automatic tensorization. Shared intermediate representations coordinate semantic analysis, contextual rule retrieval, transformation planning, diversified restructuring, repair, and selection.

\item We introduce a shared tensorization blueprint, alternative computational restructuring strategies, and closed-loop execution feedback. The blueprint links reconstructed algorithmic semantics to target tensor computation, while the restructuring branches explore different realizations under the same semantic constraints.

\item We construct a benchmark comprising 48 MOEAs and evaluate the generated tensorized implementations in terms of migration reliability, optimization fidelity, computational scalability, transfer to external implementations, and component contributions.
\end{itemize}

\section{Related Work}

\subsection{MOEAs and Software Implementations}

MOEA diversity induces corresponding diversity in computational structure. NSGA-II combines nondominated sorting with crowding-based density estimation~\cite{deb2002nsga2}, whereas SPEA2 uses strength-based fitness assignment together with density estimation~\cite{zitzler2001spea2}. MOEA/D organizes search around weighted subproblems and scalarizing functions~\cite{zhang2007moead}; NSGA-III uses reference points to guide association and niching~\cite{deb2014nsga3}; and RVEA uses reference vectors to guide environmental selection~\cite{cheng2016rvea}. Indicator-based methods such as HypE introduce contribution estimation during environmental selection~\cite{bader2011hype}. Constrained extensions introduce additional constraint-handling mechanisms~\cite{CMOEAMS2022}, while multimodal methods may coevolve multiple populations~\cite{CoMMEA2023}. Particle-swarm methods add velocity-based population updates~\cite{CMOPSO2018}; large-scale methods use problem transformation~\cite{WOF2018} or reformulation~\cite{LSMOF2019} to manage high-dimensional decision spaces; and sparse optimization introduces sparsity-oriented decision-variable structures~\cite{SparseEA2020}. Consequently, automatic tensorization must accommodate heterogeneous state lifetimes, data dependencies, update schedules, and control structures rather than apply a fixed operator template.

Software platforms make this structural heterogeneity explicit through concrete implementations. jMetal~\cite{durillo2011jmetal}, DEAP~\cite{fortin2012deap}, pagmo/PyGMO~\cite{biscani2020pagmo}, pymoo~\cite{blank2020pymoo}, and PlatEMO~\cite{tian2017platemo} provide reusable implementations and common experimental infrastructure. PlatEMO is particularly suitable for studying automatic tensorization because it contains a broad range of established MOEAs under unified interfaces. Nevertheless, its MATLAB implementations differ in population abstractions, auxiliary functions, dynamic variables, set operations, loops, branches, and persistent states. These implementation-level differences define the structural heterogeneity that the transformation process must resolve.

\subsection{Tensorization for Multiobjective Evolutionary Algorithms}

Tensorization reformulates the data and operations of MOEA implementations for batched tensor computation~\cite{liang2025tensorization}. Pairwise relations can be expressed through broadcasting, conditional selection through Boolean masks, decomposition metrics through batched scalarization, and subset manipulation through tensor indexing. These formulations reduce individual-level host-side processing and expose population-level computation more directly to GPU execution.

Existing studies establish the feasibility of tensor-based GPU acceleration for evolutionary computation and selected MOEAs. EvoX provides GPU-accelerated infrastructure for evolutionary computation~\cite{huang2023evox}. TensorRVEA reformulates the data structures and environmental selection of RVEA for tensor execution~\cite{liang2024tensorrvea}, while EvoMO develops tensorization principles for data structures, basic operations, and control flow and applies them to several representative algorithms~\cite{liang2025tensorization}. TensorNSGA-III develops a tensorized realization of NSGA-III~\cite{li2025tensornsga}, and tensorized computation has also been extended to constrained multiobjective optimization~\cite{huang2026fully}. However, these lines of work rely largely on manually designed tensorization or algorithm-specific GPU implementations. The present work therefore focuses on deriving tensorized implementations automatically from heterogeneous source programs.

\subsection{LLM-Based Program Transformation}

LLM-based program transformation demonstrates that learned models can map source programs across software representations. Codex demonstrates general code synthesis capabilities~\cite{chen2021codex}, while TransCoder studies unsupervised translation among programming languages~\cite{lachaux2020transcoder}. CodeXGLUE provides benchmark tasks for code understanding and generation, including program translation~\cite{lu2021codexglue}, while CodeNet provides a large-scale code dataset for diverse coding tasks~\cite{puri2021codenet}. Later studies investigate automatic library migration~\cite{almeida2024librarymigration} and large-scale code migration in industrial settings~\cite{ziftci2025codemigration}. However, these approaches do not determine which computational structures must remain invariant and which may be reorganized during tensorization.

Execution feedback complements static program transformation by exposing target-side constraints that cannot be determined reliably before execution. CodeRL uses unit-test feedback to guide code generation~\cite{le2022coderl}, SWE-agent combines repository interaction with test execution~\cite{yang2024sweagent}, and KernelBench evaluates and refines GPU programs using execution and performance signals~\cite{ouyang2025kernelbench}. Automatic MOEA tensorization requires this feedback-oriented view, but it also introduces algorithm-specific states, population-level restructuring, and stochastic optimization outcomes. Consequently, the transformation must reconstruct source algorithmic semantics before generation and validate the resulting implementation in the target optimization environment.

\subsection{Evaluation of Generated MOEA Implementations}

Generated programs are commonly assessed through executability, functional correctness, and computational efficiency. HumanEval~\cite{chen2021codex} and EvalPlus~\cite{liu2023evalplus} use test-based evaluation to assess functional correctness, while KernelBench combines correctness with runtime performance for generated GPU kernels~\cite{ouyang2025kernelbench}. Such protocols are appropriate when deterministic outputs or reference computations are available.

MOEA implementations require an additional evaluation layer because stochastic optimization outcomes cannot generally be validated through exact output matching. MOEA studies therefore use repeated runs on established benchmark suites, including DTLZ~\cite{deb2005dtlz}, WFG~\cite{huband2006wfg}, LSMOP~\cite{cheng2017large}, and MaF~\cite{cheng2017maf}, together with indicators such as inverted generational distance (IGD)~\cite{igd} and hypervolume~\cite{zitzler2003performance}. Automatically tensorized implementations must therefore answer three separate questions: whether the transformed program executes, whether its optimization outcomes remain comparable under matched settings, and whether the restructuring yields computational benefit. Accordingly, our evaluation treats these questions separately rather than using any single criterion as a proxy for transformation quality.

\section{Semantics-Guided Automatic Tensorization}
\label{sec:framework}

\subsection{Problem Formulation}

Automatic tensorization for MOEAs operates on concrete algorithm implementations rather than abstract algorithms. Let $C_S$ denote a source implementation of an MOEA, and let $C_T$ denote a tensorized target implementation derived from $C_S$. The program $C_S$ is the transformation object, whereas the source MOEA denotes the underlying optimization mechanism encoded by that program. Consequently, $C_T$ may differ from $C_S$ in data representation, control flow, framework interfaces, and execution model, provided that it retains the principal operators, persistent states, dependencies, and update logic of the source MOEA.

In this work, the \emph{algorithmic semantics} of a source MOEA implementation refers to the principal operators, persistent states, dependencies, and update relations that characterize its optimization mechanism. This term does not imply formal program equivalence or identical stochastic trajectories.

Formally, automatic tensorization maps a source implementation to a target implementation as
\begin{equation}
C_T = \mathcal{T}(C_S),
\qquad
C_T \in \mathcal{P}_{T},
\end{equation}
where $\mathcal{P}_{T}$ denotes the space of programs admitted by the target tensor-computing environment. A feasible target implementation must satisfy the constraints $g_j(C_T;C_S)=0$, $j=1,\ldots,J$, where $g_j=0$ denotes satisfaction of the $j$th transformation constraint. These constraints cover program integrity, target-environment compatibility, runtime executability, numerical validity, and retention of the principal algorithmic mechanism. The feasible target space is therefore
\begin{equation}
\mathcal{F}(C_S)
=
\left\{
C_T \in \mathcal{P}_{T}
\,\big|\,
g_j(C_T;C_S)=0,\; j=1,\ldots,J
\right\}.
\end{equation}

Algorithmic retention and optimization fidelity constrain different aspects of the transformation. Algorithmic retention requires the target implementation to preserve the principal operators, persistent states, dependencies, and update sequence encoded by the source program. It does not require identical random trajectories or element-wise agreement between $C_S$ and $C_T$. Optimization outcomes are assessed separately through repeated benchmark runs under matched problem settings. This separation prevents stochastic performance similarity from serving as a substitute for implementation completeness and avoids imposing trajectory-level equivalence.

Beyond feasibility, tensorization introduces a target-side restructuring objective. Among feasible implementations, the transformation should replace avoidable individual-level or host-side population computation with batched tensor operations while remaining compatible with the target execution model. Consequently, automatic tensorization is not direct code translation. It is a constrained restructuring problem in which reconstructed algorithmic semantics constrain how the computational realization may change. The notation above deliberately abstracts from specific software frameworks; the EvoCoCo realization evaluated in this paper instantiates the source and target environments with PlatEMO and EvoX/PyTorch, respectively, as described below.

\begin{figure*}[!t]
\centering
\includegraphics[width=\textwidth]{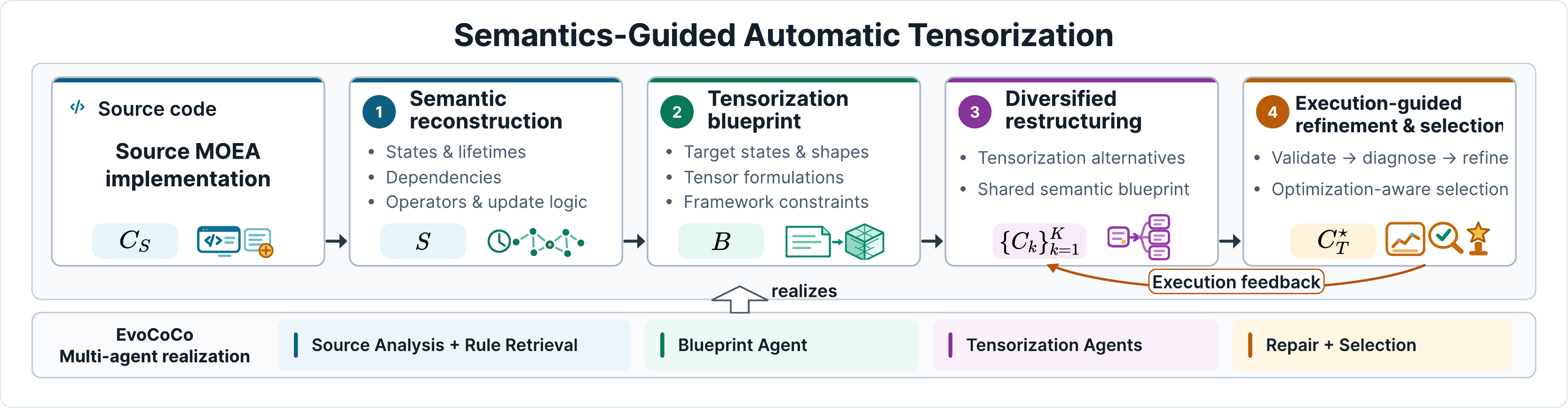}
\caption{Methodological overview of semantics-guided automatic tensorization for MOEAs. A source MOEA implementation is reconstructed into a structured semantic representation and a shared tensorization blueprint. Multiple branches then explore alternative computational realizations under common semantic constraints. Execution feedback supports refinement and optimization-aware selection of the final tensorized implementation.}
\label{fig:main_framework}
\end{figure*}

\subsection{Methodological Overview}

Semantics-guided tensorization separates source invariants from target-side design choices. As illustrated in Fig.~\ref{fig:main_framework}, automatic tensorization proceeds through four logical functions: semantic reconstruction, blueprint construction, diversified computational restructuring, and execution-guided refinement and selection. This decomposition defines the functional requirements of the transformation independently of any particular agent organization. The multi-agent framework introduced below provides the computational architecture that coordinates these functions.

Structured intermediate representations make this separation explicit through the information flow
\begin{equation}
C_S
\rightarrow S
\rightarrow \mathcal{R}_S
\rightarrow B
\rightarrow \{C_k\}_{k=1}^{K}
\rightarrow \mathcal{Z}
\rightarrow C_T^\star,
\end{equation}
where $S$ is the structured semantic representation reconstructed from the source implementation, $\mathcal{R}_S$ is the task-relevant subset retrieved from the rule base $\mathcal{R}$, and $B$ is the tensorization blueprint. Each restructuring strategy $s_k\in\mathcal{S}$ produces a candidate implementation $C_k$ under the same $S$, $\mathcal{R}_S$, and $B$. Candidates are then executed and, when necessary, refined using diagnostic feedback. The validated candidate set $\mathcal{Z}$ is hierarchically assessed to obtain the selected tensorized implementation $C_T^\star$.

The decomposition also determines where alternative design choices are permitted. The semantic representation $S$ records what the source implementation does, while the blueprint $B$ specifies how those requirements may be realized in the target framework. Candidate branches can therefore explore different tensor formulations without independently reinterpreting the source MOEA. Execution feedback then resolves target-side constraints that cannot be determined reliably before a candidate is run.

\subsection{EvoCoCo: Multi-Agent Framework}

\begin{figure*}[!t]
\centering
\includegraphics[width=0.94\textwidth]{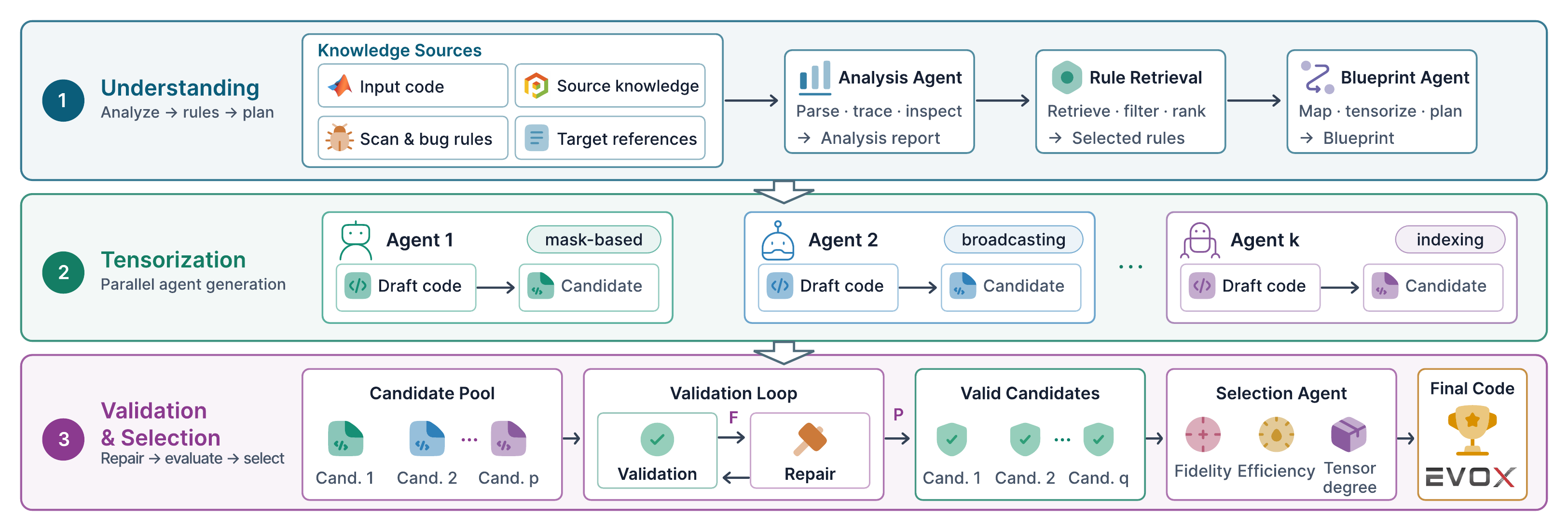}
\caption{Multi-agent architecture of EvoCoCo. Specialized agents assume distinct transformation responsibilities while sharing source analysis, retrieved rules, and the tensorization blueprint. Parallel Tensorization Agents explore alternative computational realizations. Execution feedback then drives repair, and the Selection Agent assesses the validated candidates.}
\label{fig:multi_agent_framework}
\end{figure*}

In the implementation evaluated in this work, EvoCoCo transforms PlatEMO source programs into EvoX/PyTorch targets. EvoCoCo operationalizes each methodological function through a specialized agent role. As illustrated in Fig.~\ref{fig:multi_agent_framework}, the Source Analysis Agent reconstructs $S$, the rule-retrieval role supplies contextual transformation knowledge, and the Blueprint Agent constructs $B$. Multiple Tensorization Agents instantiate alternative restructuring hypotheses under this shared blueprint. Execution failures provide diagnostic feedback to the Repair Agent, and the Selection Agent assesses validated candidates to obtain the final tensorized implementation.

Role specialization prevents each restructuring branch from solving the full transformation problem independently. Semantic interpretation is separated from target-code generation, planning establishes constraints shared by all branches, and execution diagnosis is deferred until target-side evidence becomes available. Shared intermediate representations coordinate these roles. Consequently, the framework can combine heterogeneous reasoning functions without requiring each branch to reinterpret the source program.

Algorithm~\ref{alg:evococo} makes this principle-to-architecture mapping explicit. Shared semantic constraints precede diversified restructuring, and execution-guided refinement closes the transformation loop.

\begin{algorithm}[!t]
\caption{EvoCoCo Multi-Agent Realization of Semantics-Guided Automatic Tensorization}
\label{alg:evococo}
\begin{algorithmic}[1]
\REQUIRE Source MOEA implementation $C_S$, rule base $\mathcal{R}$, restructuring strategies $\mathcal{S}=\{s_1,\ldots,s_K\}$, maximum refinement iterations $I$
\ENSURE Tensorized target implementation $C_T^\star$ or \textsc{Fail}

\STATE /* Semantic Reconstruction and Blueprint Construction */
\STATE $S \leftarrow \textsc{AnalyzeSource}(C_S)$
\STATE $\mathcal{R}_S \leftarrow \textsc{RetrieveRules}(S,\mathcal{R})$
\STATE $B \leftarrow \textsc{BuildBlueprint}(C_S,S,\mathcal{R}_S)$

\STATE /* Diversified Computational Restructuring */
\FOR{$k=1$ to $K$ \textbf{in parallel}}
    \STATE $C_k \leftarrow \textsc{Transform}(C_S,B,\mathcal{R}_S,s_k)$
\ENDFOR

\STATE /* Execution-Guided Refinement and Selection */
\FOR{$k=1$ to $K$ \textbf{in parallel}}
    \STATE $i \leftarrow 0$
    \STATE $(r_k,e_k,H_k,T_k) \leftarrow \textsc{Validate}(C_k,B)$
    \WHILE{$r_k=\textsc{Fail}$ and $i<I$}
        \STATE $C_k \leftarrow \textsc{Repair}(C_k,e_k,C_S,\mathcal{R}_S)$
        \STATE $i \leftarrow i+1$
        \STATE $(r_k,e_k,H_k,T_k) \leftarrow \textsc{Validate}(C_k,B)$
    \ENDWHILE
    \IF{$r_k=\textsc{Pass}$}
        \STATE $Q_k \leftarrow \textsc{AssessTensorization}(C_k,B)$
        \STATE $z_k \leftarrow (C_k,H_k,T_k,Q_k)$
    \ELSE
        \STATE $z_k \leftarrow \textsc{Invalid}$
    \ENDIF
\ENDFOR

\STATE $\mathcal{Z} \leftarrow \{z_k \mid z_k\neq\textsc{Invalid},\ k=1,\ldots,K\}$
\IF{$\mathcal{Z}=\emptyset$}
    \RETURN \textsc{Fail}
\ENDIF
\STATE $C_T^\star \leftarrow \textsc{HierarchicalSelect}(\mathcal{Z})$
\RETURN $C_T^\star$
\end{algorithmic}
\end{algorithm}

\subsection{Semantic Reconstruction and Tensorization Blueprint}

Semantic reconstruction establishes the source-level constraints required before target code can be generated. Source syntax does not explicitly identify which variables constitute persistent algorithmic state, which update relations define the optimization mechanism, or which loops express population-level computation. Consequently, direct line-by-line translation cannot provide the required source-level account. EvoCoCo therefore reconstructs a structured semantic representation $S$, retrieves the rules relevant to that representation, and constructs the shared tensorization blueprint $B$.

\subsubsection{Source Semantic Reconstruction}

The Source Analysis Agent reconstructs the structured semantic representation $S$ from $C_S$ without generating target code. The representation identifies the algorithm category, principal optimization mechanisms, execution workflow, external dependencies, and auxiliary functions. It records initialization, mating selection, variation, objective evaluation, environmental selection, archive maintenance, state updates, and termination when these components are present in the source implementation.

The representation separates persistent algorithmic states from temporary values. For each state, it records the semantic role, source-side dimensions, lifetime, and dependencies. Persistent states may include populations, objective values, ideal points, reference vectors, neighborhoods, archives, velocity states, and generation-dependent parameters. The analysis also records quantities such as population size $N$, number of objectives $M$, decision dimension $D$, and iteration budget $G$. Target tensor shapes and concrete implementation choices are deferred to the blueprint, which keeps source interpretation separate from target realization.

The analysis also identifies computational structures that constrain tensorization. Typical structures include individual-level loops, pairwise comparisons, dynamic lists, iterative front extraction, conditional population updates, and repeated subset operations. It separates population-level computations that admit reformulation through broadcasting, masking, batched indexing, or tensor sorting from procedures that contain genuine sequential dependencies. Source-specific conventions such as one-based indexing, MATLAB dimension semantics, implicit broadcasting, population-object slicing, persistent-state accumulation, modified dominance relations, nonstandard normalization, and custom tie-breaking are recorded explicitly because they can affect the identity of the implemented MOEA.

\subsubsection{Context-Aware Rule Retrieval}

Context-aware rule retrieval supplements reconstructed algorithmic semantics with target-framework knowledge. The rule base $\mathcal{R}$ contains target-framework requirements, recurring migration risks, and validated tensorization practices. The Rule Retriever selects a compact subset $\mathcal{R}_S$ according to the mechanisms and transformation risks recorded in $S$. Universal rules cover EvoX interfaces, state management, device and data-type consistency, MATLAB-to-PyTorch dimension conversion, numerical safeguards, unsupported operations, and unintended host-device synchronization. Contextual rules are activated only when corresponding structures occur in the source implementation. For example, crowding-distance methods require front-mask and subset-indexing rules, dominance-based methods require pairwise-comparison rules, and decomposition-based methods require weight-vector broadcasting and batched scalarization. Each retrieved rule therefore specifies an applicability condition, a known failure pattern, and a corresponding transformation requirement. The Rule Retriever filters rules whose applicability conditions match the structures recorded in $S$, ranks the retained rules by relevance to the identified mechanisms and transformation risks, and passes the resulting compact set as $\mathcal{R}_S$.

\subsubsection{Tensorization Blueprint Construction}

The tensorization blueprint $B$ converts reconstructed algorithmic semantics into explicit target-side requirements. Unlike $S$, which describes the source implementation, $B$ defines the admissible target realization. The Blueprint Agent constructs $B$ from $C_S$, $S$, and $\mathcal{R}_S$. It specifies the EvoX state architecture, execution workflow, tensor formulations, auxiliary-function contracts, and framework constraints shared by all candidate branches.

The blueprint specifies both state mapping and computational restructuring. It maps source variables to persistent EvoX states or temporary tensors and defines their target shapes, data types, initialization procedures, and update rules. It also maps the source workflow to the target lifecycle, including initialization, variation, environmental selection, state updates, and required helper functions. Operations on PlatEMO population objects are decomposed into synchronized updates of decision, objective, constraint, and auxiliary tensors. For computation-intensive structures, $B$ records population-level formulations such as broadcasting for pairwise relations, batched scalarization for decomposition metrics, Boolean masks for conditional selection, and tensor indices for subset operations. When a procedure contains an unavoidable sequential dependency, the blueprint may retain a bounded loop, but the population-level work inside each iteration remains tensorized.

The blueprint also fixes target-side constraints shared by all candidate branches. Candidate implementations must conform to the EvoX algorithm interface, maintain consistent devices and data types, update related states synchronously, and avoid unnecessary host-side computation or device synchronization. Individual-level population loops, NumPy fallbacks, repeated host-device transfers, and scalar extraction are disallowed unless the source analysis identifies an unavoidable sequential dependency. Because these constraints are fixed before candidate generation, branch diversity reflects alternative computational realizations rather than inconsistent interpretations of the source MOEA.

\subsection{Diversified Computational Restructuring}

Diversified restructuring treats tensorization as a constrained search over alternative target realizations. Given $C_S$, $B$, $\mathcal{R}_S$, and the strategy set $\mathcal{S}=\{s_1,\ldots,s_K\}$, multiple Tensorization Agents generate the candidate implementations $\{C_k\}_{k=1}^{K}$ in parallel. All branches share the same reconstructed algorithmic semantics, state architecture, framework constraints, and lifecycle requirements. Their differences therefore represent alternative restructuring hypotheses rather than independent interpretations of the source MOEA.

EvoCoCo instantiates this strategy space with six transformation biases. \emph{Broadcasting} targets pairwise and population-level relations through dimension expansion. \emph{Einsum Optimization} targets tensor contraction and batched aggregation. \emph{Masked Operations} replaces data-dependent branches with Boolean tensor computation. \emph{In-Place Efficiency} reduces avoidable intermediate storage through state updates that remain consistent with the source mechanism. \emph{Advanced Operations} uses high-level PyTorch operations for sorting, indexing, and gather or scatter patterns. \emph{Tensorized Iterative Selection} retains bounded control loops for genuine sequential dependencies and uses preallocated tensor buffers and tensor-based indexing within each iteration.

Each Tensorization Agent applies one strategy as its primary restructuring bias and produces a self-contained EvoX implementation with the required lifecycle methods and helper functions. A strategy may change data layout, indexing, or control structure, but it must satisfy the common source-level and framework constraints encoded by $B$. Because all branches share these constraints, the strategy space exposes multiple realizations of the same source mechanism. The execution stage can then resolve choices that cannot be ranked reliably from static generation alone. In this sense, EvoCoCo searches the candidate program space through a generate--evaluate--refine--select cycle rather than a single source-to-target mapping.

\subsection{Execution-Guided Refinement and Candidate Selection}

Execution feedback resolves target-side constraints that cannot be established reliably before a candidate is run. Tensor shapes, framework interfaces, device placement, numerical behavior, and state transitions can remain invalid even when a candidate is syntactically complete. Consequently, the final part of the methodology closes the transformation loop through execution, diagnostic feedback, refinement, and selection. Each branch is processed independently, so a failure in one computational realization does not prevent the remaining alternatives from being evaluated.

\subsubsection{Candidate Validation}

Candidate validation separates static program checks from runtime execution checks. Static validation examines program integrity, dependencies, target interfaces, and state declarations. Runtime validation determines whether the implementation initializes, executes, and updates its states correctly in EvoX. It also detects tensor, device, numerical, and control-flow failures that syntax alone cannot establish. A lightweight validation problem provides a common optimization signal during generation and refinement, while the complete benchmark-level evaluation is conducted separately. For candidate $C_k$, validation returns status $r_k$, diagnostic feedback $e_k$, performance-indicator result $H_k$, and execution time $T_k$.

\subsubsection{Execution-Guided Refinement}

The Repair Agent revises failed candidates without relaxing the source or framework constraints used for generation. It uses the diagnostic feedback $e_k$, the source implementation $C_S$, and the retrieved rules $\mathcal{R}_S$. Static refinement addresses local program and interface errors, while runtime refinement addresses tensor computation, state transitions, numerical behavior, and control flow. The repair step remains constrained by the source algorithmic semantics and blueprint, so correction of a target-side failure cannot replace the source MOEA with a nominally similar standard implementation.

Each revised candidate re-enters validation until it passes or reaches the maximum refinement budget $I$. Candidates that exhaust this budget are marked invalid. Consequently, execution feedback becomes part of the transformation procedure rather than a post hoc debugging step. This loop is required because source analysis alone cannot fully determine target-framework constraints and target-side executability.

\subsubsection{Optimization-Aware Candidate Selection}

Candidate selection uses a hierarchy of validated evidence rather than a fixed weighted score. Although candidates share the same blueprint, they can differ in observed optimization outcome, execution time, and degree of tensorization. For each validated candidate $C_k$, let $Q_k$ denote the code-level tensorization assessment produced by $\textsc{AssessTensorization}(C_k,B)$. EvoCoCo represents the candidate as $z_k=(C_k,H_k,T_k,Q_k)$ and applies hierarchical selection to the validated set $\mathcal{Z}$. The performance-indicator result $H_k$ provides the primary signal for grouping candidates with comparable observed optimization outcomes. In the current EvoCoCo realization, an LLM-based qualitative assessment receives the performance-indicator results $H_k$ of the validated candidates and returns the subset judged to have comparable observed optimization outcomes. Execution time $T_k$ and tensorization assessment $Q_k$ are then considered within this comparable group.

The tensorization assessment $Q_k$ checks whether principal population-level computations are expressed through tensor operations without avoidable host-side processing, synchronization, or individual-level iteration. This selection procedure is a design choice in the current EvoCoCo realization. The methodological requirement is only that optimization outcome be considered after executability has been established. The selected implementation is denoted by $C_T^\star$; if no candidate survives validation and refinement, EvoCoCo returns \textsc{Fail}.

\section{Benchmark Design and Evaluation}
\label{sec:benchmark}

The evaluation separates three properties that answer different validity questions. Migration reliability asks whether a source implementation can be converted into a usable target implementation. Optimization fidelity asks whether the selected implementation attains comparable optimization outcomes under matched settings. Computational scalability asks whether the restructuring exposes increasing runtime advantage as problem scale grows. These properties are evaluated separately because executability does not establish optimization fidelity, and acceleration is meaningful only for an implementation whose optimization outcomes satisfy the predefined criterion.

\subsection{Benchmark Construction}

The benchmark covers 48 MOEAs whose PlatEMO implementations expose heterogeneous computational structures. The selected MOEAs span dominance-based, decomposition-based, reference-guided, indicator-based, particle-swarm, archive-based, sparse, and large-scale paradigms. Their implementations differ in states, operators, control structures, helper functions, and update schedules. This diversity provides the testbed for evaluating semantics-guided restructuring. The detailed benchmark taxonomy and implementation provenance are provided in Section II-B of the Supplementary Document. For the experiments below, the generic notation is instantiated as $C_P \equiv C_S$ for a PlatEMO source implementation and $C_E^\star \equiv C_T^\star$ for the selected tensorized EvoX implementation.

Generation-time validation is isolated from benchmark-level evaluation. During automatic tensorization, each candidate implementation is tested on a lightweight validation problem under a fixed configuration, which provides diagnostic and performance signals for refinement and selection. Five independent migration attempts are performed for each source implementation. Algorithm~\ref{alg:evococo} describes branch-level selection within one conversion attempt; after the five attempt-level outputs are obtained, the converged outputs are compared using the same hierarchical evidence order, and the best converged attempt-level output is retained as the selected tensorized implementation $C_E^\star$. The resulting 48 implementations are then fixed for all subsequent fidelity and scaling experiments, with no further repair or reselection based on benchmark results. They are independently evaluated on the DTLZ~\cite{deb2005dtlz}, WFG~\cite{huband2006wfg}, LSMOP~\cite{cheng2017large}, and MaF~\cite{cheng2017maf} suites.

\subsection{Migration Reliability}

Migration reliability measures whether a source MOEA implementation can be transformed into a usable EvoX implementation under the prescribed validation setting. The measure contains three cumulative levels. A \emph{syntax pass} requires the generated implementation to be complete and syntactically valid. An \emph{execution pass} further requires it to complete the validation workflow without interface errors, runtime exceptions, invalid tensor states, persistent nonfinite values, or termination failures. A \emph{convergence pass} additionally requires the implementation to satisfy the predefined optimization criterion on the validation problem. A migration attempt is regarded as successful only when it reaches the convergence-pass level.

Let $N_{\mathrm{success}}$ denote the number of attempts that reach the convergence-pass level, and let $N_{\mathrm{total}}$ denote the total number of attempts. The migration success rate (MSR) is defined as
\begin{equation}
\mathrm{MSR}
=
\frac{N_{\mathrm{success}}}{N_{\mathrm{total}}}.
\end{equation}
The syntax-pass and execution-pass rates locate failures within the transformation process. Failed attempts are categorized by the earliest observed stage as syntax, execution, or convergence failures. Numerical failures and timeouts are recorded as execution-stage subtypes, while tensorization violations are recorded separately because they characterize the computational structure of an otherwise generated implementation rather than the migration outcome itself.

\subsection{Optimization Fidelity}

Optimization fidelity measures whether the selected tensorized implementation $C_E^\star$ attains optimization outcomes comparable to those of the corresponding PlatEMO implementation $C_P$ across the evaluated benchmark problems. Exact agreement is neither expected nor required because the two implementations may use different random-number generators, numerical libraries, and execution orders. Consequently, the comparison uses repeated independent runs under matched population sizes, objective and decision dimensions, and evaluation budgets.

For each MOEA--problem combination, $C_E^\star$ and $C_P$ are independently executed multiple times. Let $I_E$ and $I_P$ denote their mean IGD values, respectively. The absolute difference $D$ and relative increase $R$ are defined as
\begin{equation}
D = I_E - I_P,
\qquad
R = \frac{I_E - I_P}{I_P}.
\end{equation}
Two predefined tolerances, $\varepsilon_{\mathrm{abs}}=0.10$ and $\varepsilon_{\mathrm{rel}}=2.0$, account for differences in IGD scale across benchmark problems. The absolute tolerance provides an absolute-scale criterion when baseline IGD is small, while the relative tolerance provides a scale-aware criterion when IGD values are larger. Section~III-D of the Supplementary Document reports sensitivity to nearby tolerance settings.

The classification separates improvement from tolerated degradation under the predefined fidelity criterion. A result is labeled \emph{Improved} when $I_E \leq I_P$, which indicates a better observed optimization outcome under the evaluated setting. When $I_E>I_P$, the result is labeled \emph{Preserved} if either $D\leq\varepsilon_{\mathrm{abs}}$ or $R\leq\varepsilon_{\mathrm{rel}}$. Otherwise, it is labeled \emph{IGD Degradation}. Both \emph{Improved} and \emph{Preserved} count as passes under the predefined fidelity criterion.

Reference validity is determined independently of the tensorized implementation. A PlatEMO reference is valid only when all 21 runs complete and produce finite final-IGD values. A MOEA--problem combination is labeled \emph{Reference invalid} when the PlatEMO implementation encounters an execution error or timeout and fails to provide all 21 usable runs. Such cases are excluded from the corresponding coverage denominators, so reference failure is not interpreted as evidence against the tensorized implementation.

Let $N^{\mathrm{pass}}$ and $N^{\mathrm{valid}}$ denote the numbers of fidelity-passing and valid MOEA--problem comparisons, respectively. The overall fidelity coverage is
\begin{equation}
\rho_{\mathrm{overall}}
=
\frac{N^{\mathrm{pass}}}{N^{\mathrm{valid}}}.
\end{equation}
For MOEA $a$, let $N_a^{\mathrm{pass}}$ and $N_a^{\mathrm{valid}}$ denote the numbers of fidelity-passing and valid problem comparisons. Its fidelity coverage is
\begin{equation}
\rho_a
=
\frac{N_a^{\mathrm{pass}}}{N_a^{\mathrm{valid}}}.
\end{equation}
A MOEA satisfies the algorithm-level fidelity criterion when $\rho_a\geq80\%$. At the framework level, the predefined criterion is satisfied when $\rho_{\mathrm{overall}}\geq80\%$ and at least $80\%$ of the MOEAs satisfy the algorithm-level criterion.

\subsection{Computational Scalability}

Computational scalability measures how the runtime advantage of the selected tensorized implementations changes with problem scale. For the same MOEA and problem setting, let $R_P$ and $R_E$ denote the measured execution times of the PlatEMO and EvoX implementations, respectively, and let $G_P$ and $G_E$ denote their corresponding numbers of completed generations. Speedup is defined using the average execution time per generation as
\begin{equation}
\mathrm{Speedup}
=
\frac{R_P/G_P}{R_E/G_E}.
\end{equation}

The scaling protocol varies either population size or decision dimension while holding the remaining settings fixed. Initialization and warm-up are excluded from the measured execution time, and GPU operations are synchronized before and after timing. Measured aggregates include only pairs completed by both implementations. PlatEMO timeouts are shown as lower-bound speedups in the scale-wise distributions and reported separately in the exception counts. Out-of-memory (OOM) events and execution failures are also reported separately.

Speedup is interpreted jointly with migration reliability and optimization fidelity. The scaling study does not seek to establish a universal GPU advantage for every implementation. Instead, it tests whether computational restructuring exposes increasing population-level tensor parallelism as problem scale grows.

\section{Experimental Study}
\label{sec:experiments}

The five research questions (RQs) form a progressive evidence chain for the proposed methodology. RQ1 asks whether heterogeneous source implementations can be transformed reliably. RQ2 examines whether the selected tensorized implementations retain comparable optimization outcomes under the predefined fidelity criterion. RQ3 asks whether the resulting restructuring provides increasing runtime advantage as population size or decision dimension grows. RQ4 examines transfer to MATLAB implementations outside the main PlatEMO source set. RQ5 analyzes which components contribute to migration reliability.

The experiments use two hardware environments. The migration, transfer, and ablation experiments were conducted on a Windows 11 workstation with an NVIDIA GeForce RTX 5060 Ti GPU and an Intel Core Ultra 7 265K CPU. The optimization-fidelity and computational-scalability experiments were conducted on a server with an NVIDIA GeForce RTX 4090 GPU and an AMD EPYC 7543 CPU.

\begin{table*}[!t]
\centering
\caption{Migration results across the seven experimental conditions, with 240 independent conversion attempts per condition. IGD and runtime statistics are computed only over converged attempts.}
\label{tab:exp1-one-shot}
\scriptsize
\setlength{\tabcolsep}{3.2pt}
\begin{tabular}{l c c c c c c}
\toprule
Method & Syntax pass & Execution pass & Convergence pass & Algs. with $\geq 1$ pass & Median IGD $\downarrow$ & Mean runtime (s) $\downarrow$ \\
\midrule
GLM-5.1 one-shot
& 99.58\% & 72.92\% & 61.67\% & 45/48 & 0.07347 & 3.59 \\
DeepSeek V4 Pro one-shot
& \textbf{100.00\%} & 35.00\% & 27.08\% & 32/48 & 0.07826 & 2.81 \\
DeepSeek V4 Flash one-shot
& \textbf{100.00\%} & 26.25\% & 16.67\% & 24/48 & 0.07431 & 2.13 \\
Gemini 3 Flash one-shot
& \textbf{100.00\%} & 37.08\% & 26.67\% & 25/48 & 0.07304 & 8.71 \\
EvoCoCo (DeepSeek V4 Flash)
& 97.08\% & 79.58\% & 59.58\% & 47/48 & 0.08299 & 6.34 \\
EvoCoCo (DeepSeek V4 Pro)
& 99.17\% & 84.58\% & 65.42\% & 46/48 & 0.07928 & 2.99 \\
EvoCoCo (Gemini 3 Flash)
& \textbf{100.00\%} & \textbf{93.33\%} & \textbf{78.75\%} & \textbf{48/48} & 0.07477 & 3.86 \\
\bottomrule
\end{tabular}
\end{table*}

\subsection{RQ1: Migration Reliability}

RQ1 compares EvoCoCo with direct one-shot LLM translation in terms of migration reliability. The primary comparisons use matched model backends, so EvoCoCo and the one-shot baseline share the same underlying LLM. EvoCoCo includes parallel candidate generation, execution-guided refinement, and selection, whereas a one-shot attempt generates one target implementation directly. Consequently, the comparison measures realized workflow reliability rather than resource-normalized generation efficiency. Cross-backend results are reported separately because they also reflect differences in model capability.

\subsubsection{Experimental Settings}

The migration benchmark contains the PlatEMO implementations of 48 MOEAs. Seven conditions are evaluated: four one-shot conditions using GLM-5.1, DeepSeek V4 Pro, DeepSeek V4 Flash, and Gemini 3 Flash, and three EvoCoCo conditions using Gemini 3 Flash, DeepSeek V4 Pro, and DeepSeek V4 Flash. The reasoning-effort setting is left at the provider default for GLM-5.1, set to \texttt{minimal} for all Gemini 3 Flash conditions, and set to \texttt{low} for all DeepSeek V4 conditions.

Each condition uses five independent conversion attempts for every source implementation, which yields 240 attempts per condition. A one-shot attempt generates one output directly. In contrast, each EvoCoCo attempt constructs six strategy-guided candidates and applies validation, execution-guided repair, and branch-level selection. Migration metrics are computed at the attempt level from the final output. For EvoCoCo with Gemini 3 Flash, after each attempt has returned its branch-selected output, the converged attempt-level outputs are compared using the same hierarchical evidence order, and the best one is retained for each MOEA in the subsequent fidelity and scaling experiments.

Each output is evaluated on DTLZ2 with population size $N=100$, $M=3$ objectives, decision dimension $D=12$, and 100 generations. An attempt passes the convergence test if it completes without persistent nonfinite values and achieves a final IGD below 0.25 within the 60-s time limit. Conditional IGD and runtime statistics are computed only over converged attempts.

\subsubsection{Comparison Results}

EvoCoCo with Gemini 3 Flash attains the highest migration reliability among the seven evaluated conditions. Its execution and convergence pass rates reach 93.33\% and 78.75\%, respectively, and all 48 benchmark MOEAs have at least one converged implementation. Among the one-shot conditions, GLM-5.1 performs best, with an execution pass rate of 72.92\%, a convergence pass rate of 61.67\%, and coverage of 45 benchmark MOEAs. Relative to this strongest one-shot condition, EvoCoCo with Gemini 3 Flash increases the execution and convergence pass rates by 20.41 and 17.08 percentage points, respectively, and extends coverage to all 48 benchmark MOEAs.

\begin{figure}[!t]
\centering
\includegraphics[width=\columnwidth]{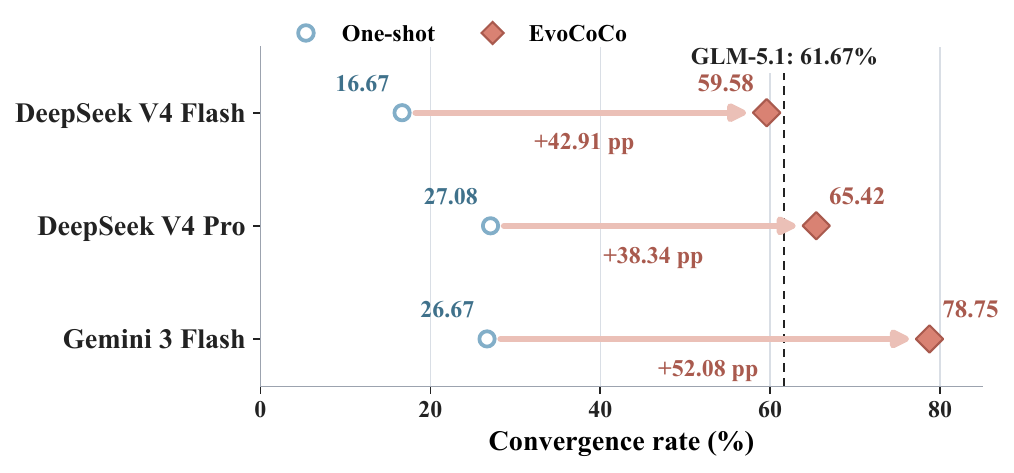}
\caption{Convergence improvements of EvoCoCo over one-shot translation under matched model backends. Each condition contains 240 independent conversion attempts. Labels show convergence pass rates, arrows indicate percentage-point improvements, and the dashed line denotes the best one-shot result.}
\label{fig:exp1-convergence-uplift}
\end{figure}

The matched-backend comparisons reduce confounding from model choice when assessing differences associated with the EvoCoCo workflow. As shown in Fig.~\ref{fig:exp1-convergence-uplift}, EvoCoCo increases the convergence pass rate by 52.08 percentage points over Gemini 3 Flash one-shot, by 38.34 points over DeepSeek V4 Pro one-shot, and by 42.91 points over DeepSeek V4 Flash one-shot. At the MOEA level, EvoCoCo produces more converged attempts than the corresponding one-shot method on 38, 35, and 43 of the 48 benchmark MOEAs for the three matched backends, respectively. Thus, the complete EvoCoCo workflow attains higher migration reliability than direct one-shot translation under all three matched-backend settings. Detailed migration results are provided in Section II of the Supplementary Document.

Model choice still affects migration performance within EvoCoCo. Gemini 3 Flash produces 189 converged attempts, compared with 157 for DeepSeek V4 Pro and 143 for DeepSeek V4 Flash. However, the differences are not uniform across MOEAs. DeepSeek V4 Pro, for example, produces more converged attempts overall than DeepSeek V4 Flash but covers one fewer benchmark MOEA. These results indicate that backend capability affects individual conversions, whereas the matched-backend comparisons more directly reflect differences associated with the EvoCoCo workflow.

\subsection{RQ2: Optimization Fidelity}

RQ2 evaluates whether the selected tensorized implementations retain comparable optimization outcomes after successful migration. The analysis reports problem-level fidelity and coverage across the 48 benchmark MOEAs.

\subsubsection{Experimental Settings}
The 48 selected tensorized implementations from the primary migration experiment are evaluated on DTLZ1--DTLZ7, WFG1--WFG9, LSMOP1--LSMOP9, and MaF1--MaF15. This design yields 1,920 MOEA--problem combinations.

For each combination, EvoX and PlatEMO are independently executed 21 times under matched experimental settings. The population size is set to $N=100$, and the number of objectives is fixed at three. Both implementations use the same decision dimension, variable bounds, and a budget of 100 generations.

Final IGD is the primary performance metric. Each combination is evaluated using the arithmetic mean of the final IGD values from the 21 runs. Outcomes are classified as Improved, Preserved, or IGD Degradation according to the combined absolute and relative tolerance rule defined in Section~\ref{sec:benchmark}. A combination is labeled Reference invalid if PlatEMO does not provide the required number of usable reference runs, and such combinations are excluded from the effective denominators. Framework-level fidelity is then assessed using the overall and algorithm-level coverage criteria defined in Section~\ref{sec:benchmark}.

\subsubsection{Comparison Results}
\begin{table}[!t]
\centering
\caption{Optimization fidelity across the DTLZ, WFG, LSMOP, and MaF benchmark suites.}
\label{tab:fidelity-suite-main}
\scriptsize
\setlength{\tabcolsep}{3.4pt}
\begin{tabular}{lrrrrrr}
\toprule
Suite & Valid & Improved & Preserved & IGD degr. & Passes & Coverage \\
\midrule
DTLZ    & 332   & 141 & 138 & 53  & 279   & 84.0\% \\
WFG     & 429   & 161 & 253 & 15  & 414   & 96.5\% \\
LSMOP   & 427   & 149 & 213 & 65  & 362   & 84.8\% \\
MaF     & 716   & 273 & 352 & 91  & 625   & 87.3\% \\
\midrule
Overall & 1,904 & 724 & 956 & 224 & 1,680 & 88.2\% \\
\bottomrule
\end{tabular}
\end{table}

\begin{figure}[!t]
\centering
\includegraphics[width=\columnwidth]{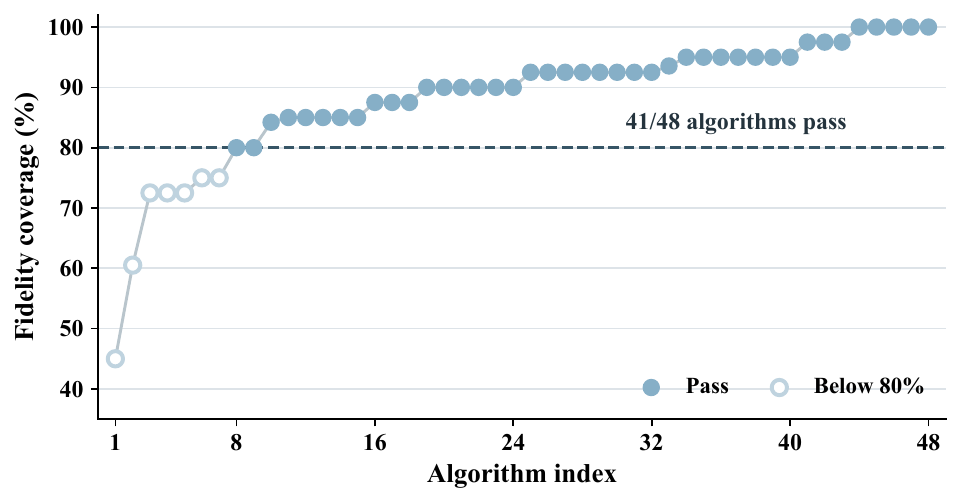}
\caption{Optimization-fidelity coverage of the 48 tensorized implementations, ordered by increasing coverage. Filled and open markers indicate algorithms above and below the 80\% threshold, respectively.}
\label{fig:fidelity-algorithm-coverage}
\end{figure}

Overall fidelity coverage reaches 88.2\% across the valid benchmark comparisons. Among the 1,920 MOEA--problem combinations, 16 PlatEMO references are unavailable because of execution errors or incomplete reference runs. The remaining 1,904 comparisons include 724 Improved cases and 956 Preserved cases, which yields 1,680 passes under the predefined fidelity criterion. The other 224 comparisons show IGD degradation.

Fidelity coverage remains above 80\% in all four benchmark families. WFG achieves the highest aggregate coverage at 96.5\% (414/429), followed by MaF at 87.3\% (625/716), LSMOP at 84.8\% (362/427), and DTLZ at 84.0\% (279/332). Consequently, the overall result is not dominated by a single benchmark suite. The differences among the four families also indicate that optimization fidelity varies with problem characteristics.

Algorithm-level coverage shows a similar pattern. As shown in Fig.~\ref{fig:fidelity-algorithm-coverage}, 41 of the 48 tensorized implementations achieve at least 80\% fidelity coverage over their valid reference problems. Thus, 85.4\% of the benchmark MOEAs satisfy the algorithm-level criterion. The seven MOEAs below this criterion are MaOEA-CSS (18/40), OSP-NSDE (23/38), BCE-MOEA-D (29/40), NSGA-II-SDR (29/40), TS-SparseEA (29/40), MOEA-D-PaS (30/40), and TS-NSGA-II (30/40).

Taken together, the overall and algorithm-level results satisfy the two predefined 80\% fidelity thresholds. They therefore provide empirical support for comparable optimization outcomes across the evaluated problems and MOEAs under the stated criterion. However, they do not imply trajectory-level or element-wise equivalence between the two implementations.

\begin{figure*}[!t]
\centering
\includegraphics[width=0.92\textwidth]{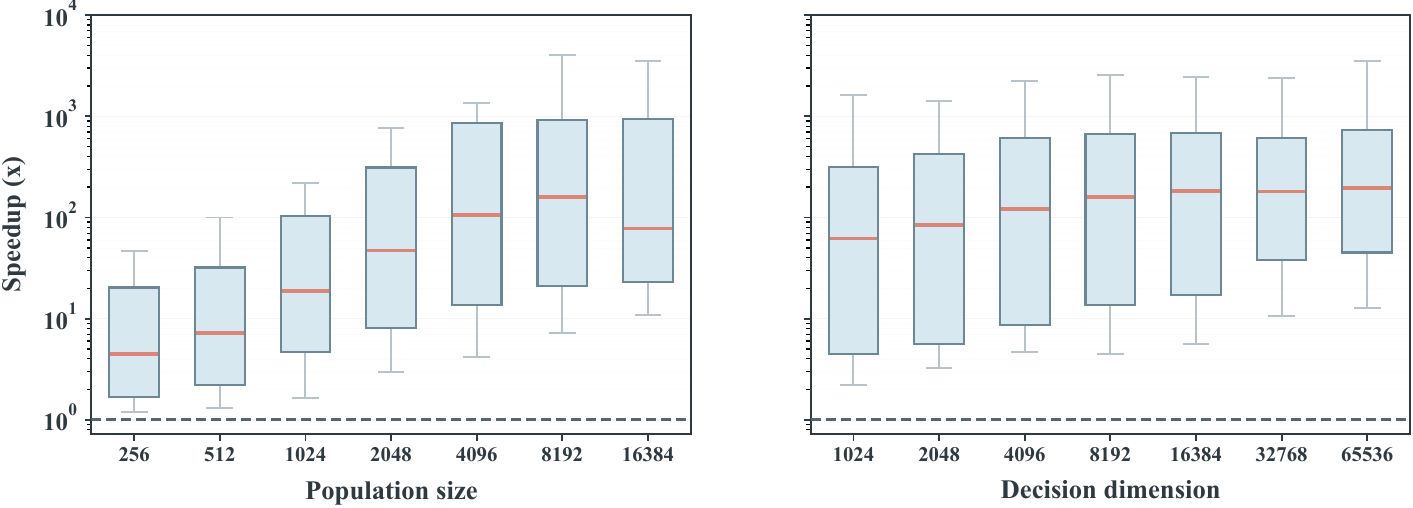}
\caption{Distribution of speedups across the 48 tensorized implementations as population size (left) or decision dimension (right) increases. Boxes show the interquartile range, center lines show medians, and whiskers show the 10th--90th percentiles. The dashed line denotes $1\times$ speedup.}
\label{fig:representative_scaling}
\end{figure*}

\subsection{RQ3: Computational Scalability}

RQ3 evaluates whether computational restructuring produces increasing runtime advantage as problem scale grows. Population size and decision dimension are varied separately because they stress different population-level operations and memory patterns in the tensorized implementations.

\subsubsection{Experimental Settings}

Runtime scalability is evaluated on DTLZ3 with three objectives. For population scaling, the population size is successively doubled from 256 to 16384 while the decision dimension remains fixed at $D=12$. For dimension scaling, the population size remains fixed at 1000 while the decision dimension is successively doubled from 1024 to 65536. EvoX executes the tensorized implementations on the GPU, and PlatEMO serves as the CPU baseline.

The EvoX workflow step is compiled using \texttt{torch.compile}. Initialization and compilation costs are excluded from runtime measurements. Runtime is measured in milliseconds per generation and averaged over 100 generations. A three-hour runtime limit is imposed for 100 generations, corresponding to 108000 ms/gen. When PlatEMO reaches this limit, the corresponding speedup is reported as a lower bound. Insufficient-memory cases and execution errors are reported separately.

\subsubsection{Comparison Results}

Dimension scaling yields larger aggregate runtime gains than population scaling. As shown in Table~\ref{tab:scaling_summary}, the median measured speedup is $22.6\times$ under population scaling and $80.2\times$ under decision-dimension scaling, while the corresponding geometric means are $29.9\times$ and $71.5\times$. The interquartile ranges are $4.5$--$163.5\times$ and $8.4$--$404.9\times$, respectively. These statistics indicate a higher central tendency under dimension scaling together with substantial heterogeneity across MOEAs and scale settings.

The scale-wise distributions show that the typical runtime advantage generally increases with computational scale, although heterogeneity remains across implementations. PlatEMO timeout cases with valid EvoX runtimes are included in Fig.~\ref{fig:representative_scaling} as lower-bound speedups. The median reported speedup under population scaling rises from $4.5\times$ at $N=256$ to $77.5\times$ at $N=16384$. Under decision-dimension scaling, the corresponding median rises from $62.1\times$ at $D=1024$ to $194.8\times$ at $D=65536$. The widening interquartile ranges at several scales further show that the benefit varies with the computational structure of the underlying MOEA implementation.

\begin{table}[!t]
\centering
\caption{Aggregate runtime scaling of the 48 tensorized implementations. Median, geometric mean, and interquartile range (IQR; 25th--75th percentile) are computed only for cases completed by both PlatEMO and EvoX. Timeout and exceptional cases are reported separately.}
\label{tab:scaling_summary}
\scriptsize
\setlength{\tabcolsep}{3pt}
\begin{tabular}{lcccc}
\toprule
Scaling axis & Measured pairs & Median & Geomean & IQR \\
\midrule
Population size & 289 & $22.6\times$ & $29.9\times$ & $4.5$--$163.5\times$ \\
Decision dimension & 276 & $80.2\times$ & $71.5\times$ & $8.4$--$404.9\times$ \\
\bottomrule
\end{tabular}
\end{table}

\begin{figure*}[!t]
\centering
\subfloat[MOEA/D-DE ($N$)]{\includegraphics[width=0.235\textwidth]{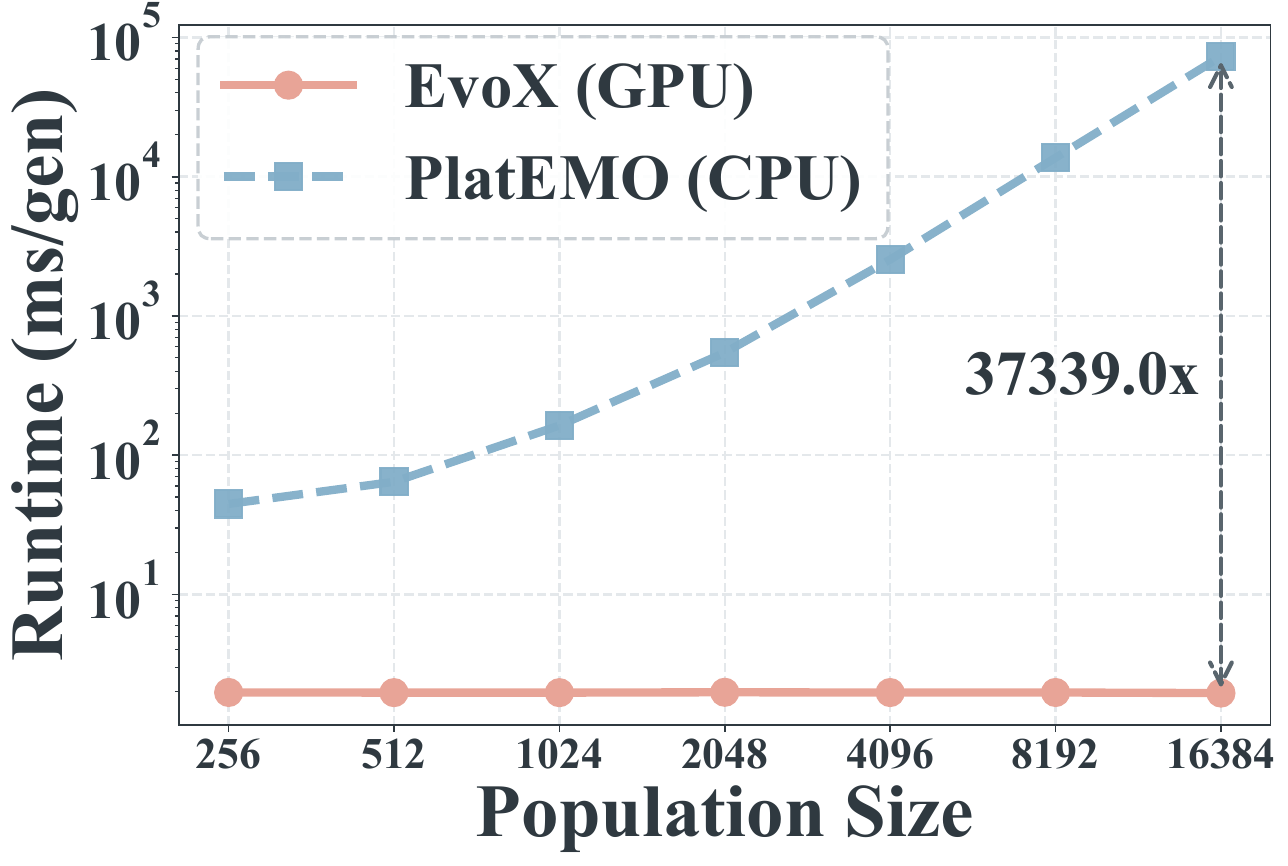}}
\hfill
\subfloat[WASF-GA ($N$)]{\includegraphics[width=0.235\textwidth]{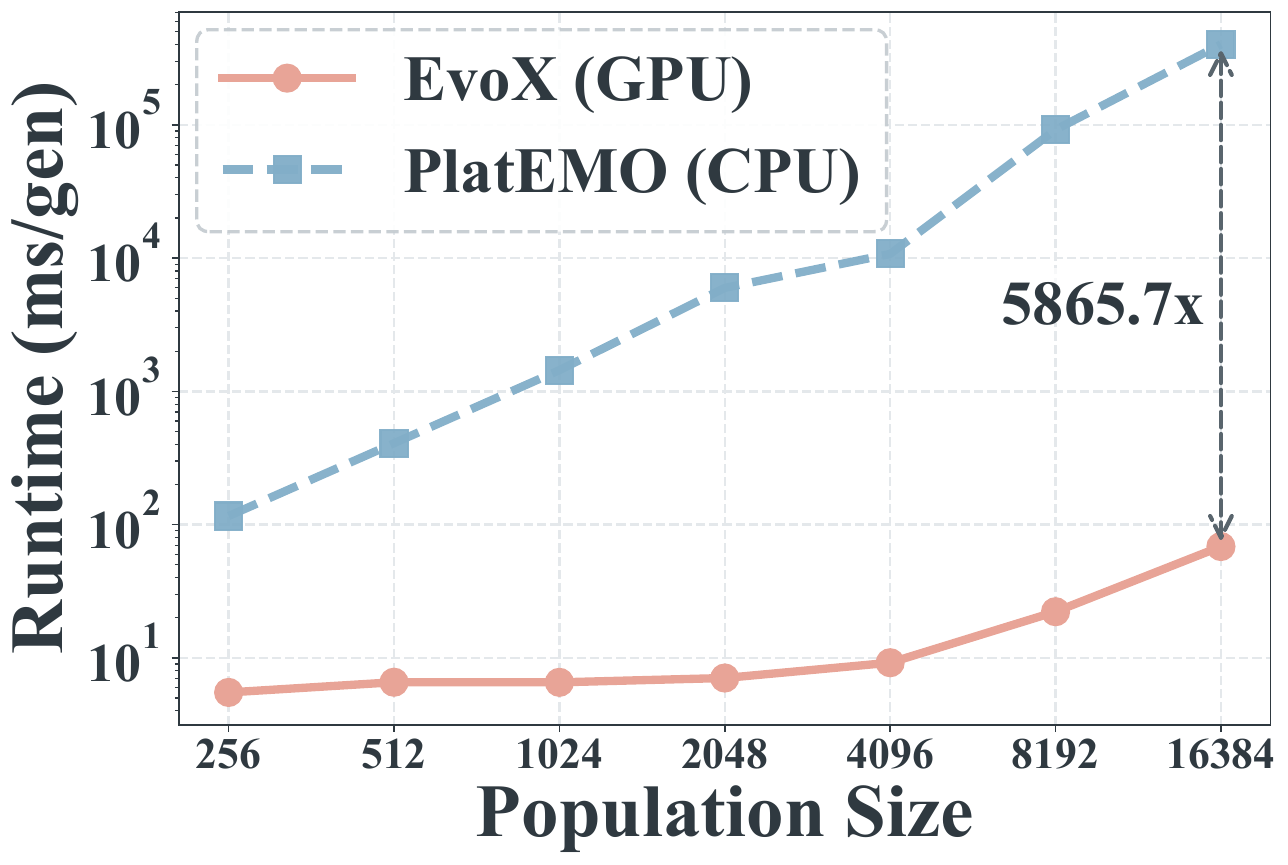}}
\hfill
\subfloat[PESA-II ($D$)]{\includegraphics[width=0.235\textwidth]{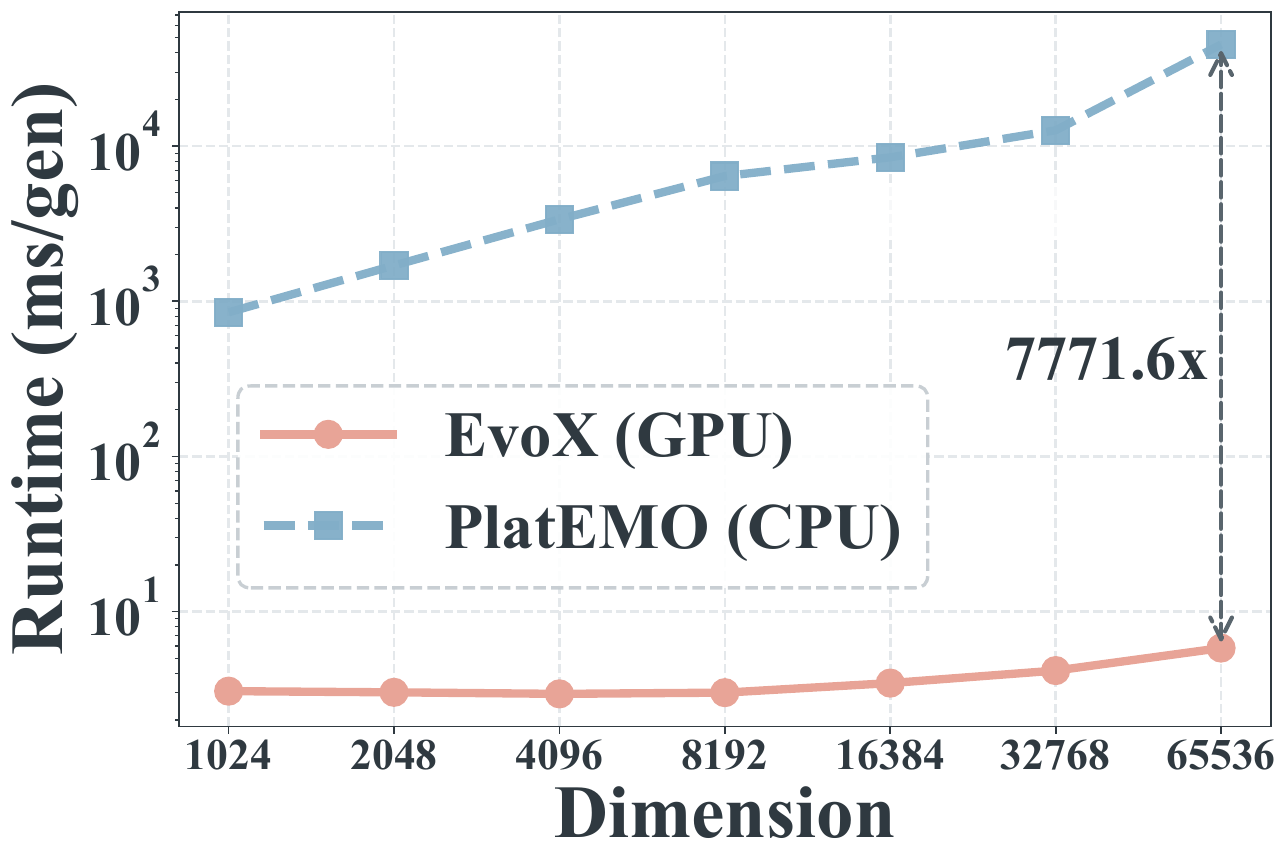}}
\hfill
\subfloat[LSMOF ($D$)]{\includegraphics[width=0.235\textwidth]{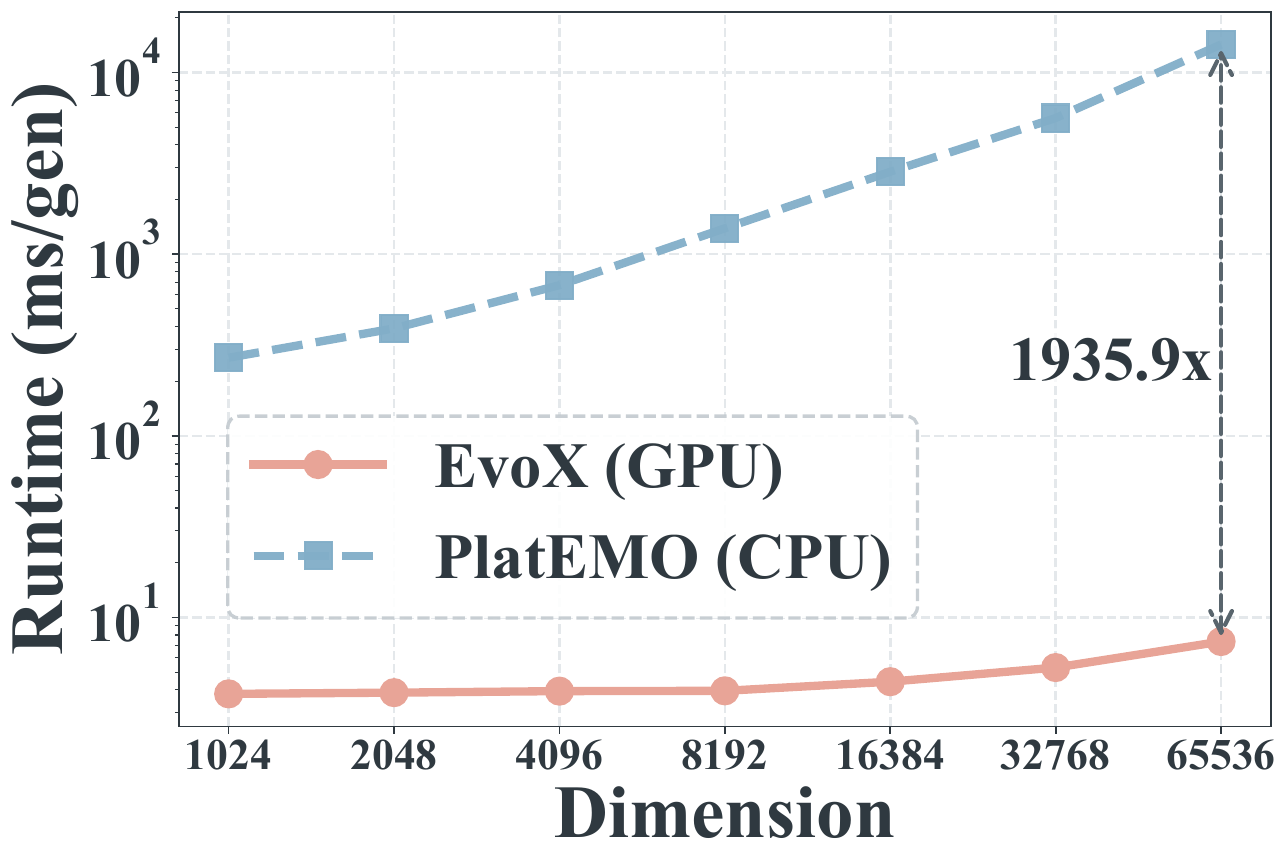}}
\caption{Runtime scaling of four representative tensorized MOEAs. (a) MOEA/D-DE and (b) WASF-GA are evaluated by successively doubling the population size. (c) PESA-II and (d) LSMOF are evaluated by successively doubling the decision dimension. Annotations indicate the speedups at the largest evaluated scale.}
\label{fig:selected_scaling_trajectories}
\end{figure*}

Representative scaling trajectories illustrate the upper range of the observed runtime gains without characterizing their typical magnitude. As shown in Fig.~\ref{fig:selected_scaling_trajectories}, MOEA/D-DE and WASF-GA achieve speedups of $37339.0\times$ and $5865.7\times$, respectively, at $N=16384$. PESA-II and LSMOF achieve $7771.6\times$ and $1935.9\times$, respectively, at $D=65536$. Among all completed runs, MOEA/D-DE has the largest measured speedup under population scaling, while DM-MOEA reaches the largest measured speedup of $23467.4\times$ under decision-dimension scaling at $D=8192$. For PlatEMO timeout cases, the largest reported lower bounds are $29900.3\times$ for MaOEA-CSS at $N=1024$ and $105365.9\times$ at $D=2048$ and $D=4096$. These values describe extreme cases, whereas Table~\ref{tab:scaling_summary} provides the distributional summary used to characterize typical acceleration.

Boundary cases further characterize the scaling regime. At the largest settings, some PlatEMO runs reach the three-hour runtime limit or encounter execution failures, while some EvoX runs encounter GPU out-of-memory conditions. Execution failures and OOM events are reported separately. Table~\ref{tab:scaling_summary} summarizes measured speedups only for settings completed by both implementations.

Together with the fidelity results, the scaling experiments show that the selected tensorized implementations can gain runtime advantage as population size or decision dimension grows while satisfying the predefined optimization-fidelity criterion in the corresponding benchmark evaluation. Detailed scaling results are provided in Section IV of the Supplementary Document.

\subsection{RQ4: Transfer to External MOEA Implementations}

RQ4 evaluates transfer beyond the PlatEMO implementations used in the main benchmark. The ten source programs are drawn from external or legacy MATLAB and Octave code bases with different coding conventions, helper functions, and program structures, while EvoX remains the target framework.

\subsubsection{Experimental Settings}

The transfer set includes implementations of ten MOEAs: DEMO, DEMO-IBEA, DEMO-PBEA, ISDE+, LSMaODE, PAR-DEMO-IND, PAR-DEMO-NDS, RTEA, R-DEMO, and Two-Arch2. LSMaODE is based on an earlier version of PlatEMO and is not directly compatible with the current version, while Two-Arch2 uses an independent implementation. EvoCoCo and the one-shot baseline each perform five independent migration attempts for every MOEA, which yields 50 attempts per method.

During migration, each generated implementation undergoes a lightweight 50-generation run for execution validation. This run is used only for generation and repair and is excluded from the final evaluation. For the final benchmark, complete implementations are evaluated on DTLZ2 with population size $N=100$, three objectives, decision dimension $D=12$, and 100 generations. An executable implementation is considered converged if its final IGD is below 0.25.

\subsubsection{Comparison Results}
\begin{table}[!b]
\centering
\caption{Migration performance on ten algorithms beyond the main PlatEMO setting, with 50 attempts per method.}
\label{tab:non_platemo_translation}
\scriptsize
\setlength{\tabcolsep}{4pt}
\begin{tabular}{lcc}
\toprule
Metric & EvoCoCo & One-shot \\
\midrule
Syntax pass & 100.00\% (50/50) & 70.00\% (35/50) \\
Execution pass & 96.00\% (48/50) & 34.00\% (17/50) \\
Converged & 62.00\% (31/50) & 18.00\% (9/50) \\
\bottomrule
\end{tabular}
\end{table}

EvoCoCo attains higher migration reliability than one-shot generation on the external and legacy source set. As shown in Table~\ref{tab:non_platemo_translation}, EvoCoCo achieves syntax and execution pass rates of 100.00\% and 96.00\%, compared with 70.00\% and 34.00\% for the one-shot baseline. Of the 50 EvoCoCo attempts, 31 satisfy the convergence threshold, which yields a convergence rate of 62.00\%; only 9 of the 50 one-shot attempts converge, corresponding to 18.00\%. At the MOEA level, EvoCoCo produces at least one converged result for nine of the ten MOEAs, compared with two for one-shot generation.

Failure analysis shows that the largest difference appears before optimization-level evaluation. For one-shot generation, 15 attempts fail the syntax check and another 18 fail during execution, which leaves 17 executable attempts. In contrast, all EvoCoCo attempts pass the syntax check and only two fail during execution. Manual inspection shows that many syntactically invalid one-shot outputs are incomplete and terminate within expressions, imports, function calls, or formatted strings. Among the 48 executable EvoCoCo results, 31 satisfy the convergence criterion, while the remaining 17 execute successfully but do not reach the specified IGD threshold.

The remaining transfer failures occur primarily after successful execution. This pattern indicates that optimization-level validity is a more restrictive transfer requirement than syntactic or runtime validity. Thus, the external-source study supports transfer of the staged transformation process while identifying optimization validity as the main remaining limitation. Detailed results are provided in Section V of the Supplementary Document.

\subsection{RQ5: Component Contributions}

RQ5 evaluates how the major EvoCoCo components contribute to migration reliability. The tested components are blueprint construction, execution-guided refinement, context-aware rule retrieval, and multi-branch generation and selection.

\subsubsection{Experimental Settings}

The ablation study uses a diagnostic subset of 12 MOEAs: BCE-IBEA, NSGA-II-SDR, AGE-MOEA, GrEA, SMPSO, SparseEA, MOEA-D-AWA, MOEA-D-PaS, DM-MOEA, SIBEA, Two-Arch2, and WOF. The subset spans consistently successful cases, convergence-challenging cases, decomposition-based MOEAs, runtime-sensitive cases, and implementations with complex program structures.

Each source implementation is migrated five times under each variant, which yields 60 attempts per variant and 300 attempts in total. All variants use Gemini 3 Flash and follow the same DTLZ2 validation setting and convergence criterion as the primary migration experiment. Full EvoCoCo serves as the reference configuration. The four ablated variants are w/o Blueprint Agent, w/o Repair Agent, w/o Rule Retrieval, and w/o Multi-Branch. The first three remove the named component, while w/o Multi-Branch uses one tensorization branch instead of generating and selecting among multiple branches.

\subsubsection{Comparison Results}

\begin{table*}[!t]
\centering
\caption{Ablation results on a diagnostic subset of 12 algorithms, with 60 attempts evaluated for each variant. Average IGD is computed over converged attempts only.}
\label{tab:ablation_subset_main}
\scriptsize
\setlength{\tabcolsep}{4.2pt}
\begin{tabular}{l l c c c c}
\toprule
Variant & Removed component & Syntax pass & Exec. pass & Converged & Avg. IGD \\
\midrule
w/o Blueprint Agent & Blueprint Agent & \textbf{100.0\%} & 95.0\% & 78.3\% & \textbf{0.0814} \\
w/o Repair Agent & Repair Agent & \textbf{100.0\%} & 60.0\% & 41.7\% & 0.1005 \\
w/o Rule Retrieval & rule retrieval & \textbf{100.0\%} & 93.3\% & 75.0\% & 0.0993 \\
w/o Multi-Branch & Multi-branch generation and selection & \textbf{100.0\%} & 68.3\% & 40.0\% & 0.1036 \\
Full EvoCoCo & None & \textbf{100.0\%} & \textbf{98.3\%} & \textbf{83.3\%} & 0.0944 \\
\bottomrule
\end{tabular}
\end{table*}

Execution-guided repair has the largest effect on execution reliability in the ablation study. Full EvoCoCo achieves an execution pass rate of 98.3\% and a convergence rate of 83.3\%, while all variants maintain a 100.0\% syntax pass rate. Removing the Repair Agent reduces the execution pass rate to 60.0\% and the convergence rate to 41.7\%. The corresponding failures are mainly associated with EvoX API misuse, unavailable APIs, and tensor-shape mismatches. This failure profile is consistent with the role of runtime feedback in resolving errors that code generation alone does not expose.

Multi-branch generation also contributes to migration reliability. Using one branch reduces the execution pass rate to 68.3\% and the convergence rate to 40.0\%. Thus, alternative tensorization strategies increase the likelihood of obtaining an executable and converged implementation. Removing rule retrieval or Blueprint Agent produces smaller reductions in execution and convergence rates. Their failure records contain more dependency, interface, and initialization-related errors, which indicates that these components primarily support code construction and adaptation to the EvoX workflow.

Conditional IGD must be interpreted together with convergence rate because it is computed only over converged attempts. Although w/o Blueprint Agent achieves the lowest average IGD of 0.0814, its convergence rate is 78.3\%, compared with 83.3\% for Full EvoCoCo. Taken together, the ablation results assign distinct roles to the four components: the Blueprint Agent and rule retrieval support code construction, the Repair Agent improves execution reliability, and the multi-branch mechanism increases the likelihood of obtaining a converged tensorized implementation. Detailed ablation results are provided in Section VI of the Supplementary Document.

\subsection{Discussion}
\label{subsec:discussion}

Migration results show that automatic tensorization is constrained more by target-side executability than by code completion alone. Most failures occur after syntactically valid code has been produced, when the target implementation must satisfy framework requirements, tensor-state consistency, and optimization requirements simultaneously. The ablation results support the same interpretation. Execution-guided refinement and diversified restructuring contribute directly to reliability, while blueprint construction and rule retrieval constrain how the source implementation is reconstructed and realized.

Beyond role specialization, EvoCoCo also introduces an evolution-like mechanism into program restructuring. Candidate implementations are generated as alternative realizations of a shared semantic blueprint, exposed to execution feedback, refined when necessary, and selected according to target-side evidence. This process is evolutionary in structure, although EvoCoCo is not formulated as a conventional evolutionary algorithm.

Optimization fidelity defines a deeper boundary between tensorization and reimplementation. The source and target programs can differ in data layout, execution order, and control structure while still encoding the same principal operators and persistent states. However, small deviations in normalization, reference association, environmental selection, or state updates can accumulate differently across problem landscapes. Lower coverage on DTLZ1, DTLZ6, LSMOP6, LSMOP7, MaF3, MaF6, and MaF9 illustrates this sensitivity. The fixed DTLZ2 validation setting cannot expose every problem-dependent deviation before final evaluation. Consequently, the fidelity results indicate empirical outcome agreement under the predefined criterion rather than trajectory-level equivalence.

Computational scaling also reveals a runtime-memory trade-off. As population size or decision dimension grows, the selected EvoX implementations increasingly benefit from population-level GPU execution, while several PlatEMO runs reach timeout or execution limits. However, the scaling experiments also record EvoX out-of-memory cases at the largest settings. Thus, computational restructuring can exchange a CPU runtime bottleneck for GPU memory pressure because broadcasting and batched intermediate tensors may increase peak memory use. More memory-efficient restructuring strategies are required when intermediate tensor materialization becomes the dominant cost.

These observations define the scope of the claims supported by the present methodology and multi-agent framework. The experiments establish feasibility for heterogeneous PlatEMO-to-EvoX transformation and provide additional evidence on a limited set of external and legacy MATLAB implementations. They do not establish framework-independent generality, resource-normalized superiority over direct translation, formal program equivalence, or exact stochastic trajectory equivalence. Within this scope, the results support a constrained program-restructuring view of automatic tensorization: the target computation may change considerably, while reconstructed algorithmic states, operators, dependencies, and update relations bound the admissible changes. Future work will extend the methodology to additional source and target environments and develop validation and restructuring strategies for different MOEA structures and memory regimes.

\section{Conclusion}

This work formulates automatic tensorization for multiobjective evolutionary algorithms as semantics-guided computational restructuring and develops EvoCoCo as a multi-agent framework that realizes this formulation across heterogeneous MOEA implementations. EvoCoCo maps the methodological decomposition to specialized roles for semantic analysis, rule retrieval, blueprint construction, diversified restructuring, execution-guided repair, and candidate selection. A shared tensorization blueprint coordinates these roles and constrains alternative computational realizations.

The experiments on 48 MOEAs show that the framework can produce executable tensorized implementations across heterogeneous algorithmic structures, satisfy the predefined optimization-fidelity criterion in most valid comparisons, and expose increasing GPU acceleration as computational scale grows. External-source and ablation studies further examine transfer and component roles. Together, these results support automatic tensorization for MOEAs as a structured transformation methodology in which implementation-level computation may change considerably while remaining constrained by the underlying MOEA.

The current evidence remains specific to the evaluated source and target environments. It does not imply exact stochastic equivalence, framework-independent generality, or resource-normalized superiority over direct translation. Future work will extend the methodology and multi-agent framework to additional programming environments, strengthen optimization-aware validation, and develop memory-efficient restructuring strategies for larger computational scales.

\bibliographystyle{IEEEtran}
\bibliography{evococo_refs}

\clearpage
\onecolumn
\setcounter{section}{0}
\setcounter{subsection}{0}
\setcounter{subsubsection}{0}
\setcounter{table}{0}
\setcounter{figure}{0}
\setcounter{equation}{0}
\setcounter{footnote}{0}
\renewcommand{\thesection}{\Roman{section}}
\renewcommand{\thetable}{S.\Roman{table}}
\renewcommand{\thefigure}{S.\arabic{figure}}
\renewcommand{\theHsection}{supp.\arabic{section}}
\renewcommand{\theHsubsection}{supp.\arabic{section}.\arabic{subsection}}
\renewcommand{\theHsubsubsection}{supp.\arabic{section}.\arabic{subsection}.\arabic{subsubsection}}
\renewcommand{\theHtable}{supp.\arabic{table}}
\renewcommand{\theHfigure}{supp.\arabic{figure}}
\renewcommand{\theHequation}{supp.\arabic{equation}}
\renewcommand{\theHfootnote}{supp.\arabic{footnote}}

\begin{center}
{\LARGE\bfseries Semantics-Guided Automatic Tensorization for Multiobjective Evolutionary Algorithms:\\[0.3em]
A Multi-Agent Framework\par}
\vspace{1.0em}
{\large\emph{Supplementary Document}\par}
\end{center}
\vspace{1.0em}
\section{Overview}

This supplementary document provides detailed evidence for the five research questions (RQs) concerning the multiobjective evolutionary algorithm (MOEA) implementations transformed by Evolutionary Code Conversion (EvoCoCo). Section~II reports migration reliability across seven conversion conditions, including complete attempt-level outcomes, failure stages, backend comparisons, token use, benchmark taxonomy, and provenance. Section~III reports optimization-fidelity results across DTLZ, WFG, LSMOP, and MaF. Section~IV reports population- and dimension-scaling results, including complete curves and exception records. Section~V reports transfer to external and legacy MATLAB/Octave implementations. Section~VI reports the component-ablation results. All PlatEMO source and reference implementations used in the main benchmark are from PlatEMO v4.14.

The experiments use the same two hardware environments reported in the main paper. The migration-reliability, external-transfer, and ablation experiments were conducted locally under Windows 11 on a workstation equipped with an NVIDIA GeForce RTX 5060 Ti graphics processing unit (GPU) with 16 GB of GPU memory and an Intel Core Ultra 7 265K central processing unit (CPU). The optimization-fidelity and computational-scalability experiments were conducted on a server equipped with an NVIDIA GeForce RTX 4090 GPU with 24 GB of GPU memory and an AMD EPYC 7543 CPU. This supplementary document reports the complete evidence behind the compact main-paper results and introduces no additional post hoc repair or reselection of the 48 tensorized implementations used in the fidelity and scaling studies.

\begin{table}[H]
\centering
\caption{Navigation of the supplementary document and the corresponding main-paper questions.}
\label{tab:supp_navigation}
\small
\setlength{\tabcolsep}{5pt}
\begin{tabular}{p{0.12\textwidth} p{0.50\textwidth} p{0.28\textwidth}}
\toprule
Section & Detailed content & Main-paper question \\
\midrule
II & Migration outcomes, failure stages, source taxonomy, token use, backend comparisons & RQ1: Migration reliability \\
III & Suite-, problem-, and MOEA-level fidelity results, invalid references, and tolerance sensitivity & RQ2: Optimization fidelity \\
IV & Aggregate and complete population/dimension scaling results, timeouts, errors, and out-of-memory cases & RQ3: Computational scalability \\
V & Matched migration results for external and legacy MATLAB/Octave implementations & RQ4: Transfer \\
VI & Blueprint, repair, rule-retrieval, and multi-branch ablations & RQ5: Component contributions \\
\bottomrule
\end{tabular}
\end{table}
\section{Detailed Migration-Reliability Results}

\subsection{Detailed Migration Configuration and Batch Inclusion}

The migration benchmark evaluates automatic tensorization from heterogeneous PlatEMO implementations to EvoX. The attempt-level experimental unit is the final EvoX implementation returned by one conversion attempt. Each of the seven conditions contains five independent attempts for each of 48 source MOEA implementations, which yields $48\times5=240$ attempt-level implementations. The reasoning-effort setting is left at the provider default for GLM-5.1, set to \texttt{minimal} for both Gemini 3 Flash conditions, and set to \texttt{low} for all DeepSeek V4 conditions.

The two generation modes differ in transformation structure. One-shot generation returns one candidate directly from the supplied source context. EvoCoCo instead performs source analysis, rule retrieval, blueprint construction, generation through six strategy-guided branches, validation, execution-guided repair, and final selection. Within each EvoCoCo attempt, the six branch candidates are reduced to one attempt-level output by the hierarchical selection procedure described in the main paper. The branch strategies emphasize broadcasting, einsum optimization, masked operations, in-place efficiency, advanced tensor operations, and tensorized iterative selection.

All generated implementations are validated on DTLZ2 with $N=100$, $M=3$, $D=12$, and 100 generations, subject to a 60-s time limit. A candidate is converged when it executes without persistent NaNs and its final inverted generational distance (IGD) is below 0.25. Conditional IGD uses converged records with finite final IGD, and conditional runtime uses converged records with nonnegative execution time. For the downstream fidelity and scaling studies, the five attempt-level outputs produced by EvoCoCo with Gemini 3 Flash for each MOEA are compared using the same hierarchical evidence order, and the best converged attempt-level output is fixed before downstream benchmark evaluation.

\begin{table}[H]
\centering
\caption{Migration-reliability configurations for the seven experimental conditions, including model reasoning-effort settings and retained generation metadata. Each condition comprises 240 independent conversion attempts.}
\label{tab:exp1_batch_inclusion}
\scriptsize
\setlength{\tabcolsep}{3.2pt}
\begin{tabular}{l c p{0.44\textwidth} c}
\toprule
Method & Branches & Reasoning effort and execution metadata & Records \\
\midrule
GLM-5.1 one-shot & 1 & Provider default; detailed reasoning metadata not retained & 240 \\
DeepSeek V4 Pro one-shot & 1 & \texttt{low} & 240 \\
DeepSeek V4 Flash one-shot & 1 & \texttt{low} & 240 \\
Gemini 3 Flash one-shot & 1 & \texttt{minimal} & 240 \\
EvoCoCo (DeepSeek V4 Flash) & 6 & \texttt{low} & 240 \\
EvoCoCo (DeepSeek V4 Pro) & 6 & \texttt{low}; five repeats concurrent and at most 30 branches active & 240 \\
EvoCoCo (Gemini 3 Flash) & 6 & \texttt{minimal} & 240 \\
\bottomrule
\end{tabular}
\end{table}

\subsection{Benchmark Composition, Taxonomy, and Algorithm Provenance}

The benchmark contains PlatEMO implementations of 48 source MOEAs. For a compact and mutually exclusive taxonomy, each algorithm is assigned to one of four classes according to its primary environmental-selection or population-update mechanism: dominance-based, decomposition-based, indicator-based, or specialized-operator-based. Hybrid algorithms are assigned to the mechanism that most directly determines survival or population updating, while the detailed mechanism column records important secondary components and target problem settings. Table~\ref{tab:algorithm_provenance} reports this taxonomy together with the original publication associated with every source algorithm.

\begingroup
\scriptsize
\setlength{\tabcolsep}{2.4pt}
\renewcommand{\arraystretch}{1.06}
\begin{longtable}{p{0.14\textwidth} p{0.14\textwidth} p{0.53\textwidth} p{0.09\textwidth}}
\multicolumn{4}{c}{\label{tab:algorithm_provenance}\normalfont\footnotesize TABLE~\thetable}\\[-0.2ex]
\multicolumn{4}{c}{\normalfont\footnotesize\scshape Taxonomy, Primary Mechanisms, and Source Publications of the 48 Benchmark Algorithms.}\\[0.5ex]
\toprule
Class & Algorithm & Detailed mechanism & Reference \\
\midrule
\endfirsthead
\multicolumn{4}{c}{\normalfont\footnotesize TABLE~\thetable\ (Continued)}\\[-0.2ex]
\multicolumn{4}{c}{\normalfont\footnotesize\scshape Taxonomy, Primary Mechanisms, and Source Publications of the 48 Benchmark Algorithms.}\\[0.5ex]
\toprule
Class & Algorithm & Detailed mechanism & Reference \\
\midrule
\endhead
\midrule
\multicolumn{4}{r}{Continued on next page}\\
\endfoot
\bottomrule
\endlastfoot
Dominance-based & AGE-MOEA & Combines nondominated sorting with adaptive $L_p$-geometry estimates for proximity--diversity survival scores. & \cite{AGEMOEA2019} \\
Dominance-based & BiGE & Maps proximity and crowding to a bi-goal space and applies nondominated sorting in that space. & \cite{BiGE2015} \\
Dominance-based & e-MOEA & Uses epsilon-dominance for steady-state replacement and maintenance of a bounded external archive. & \cite{eMOEA2003} \\
Dominance-based & GDE3 & Combines differential-evolution variation with Pareto dominance and crowding-based truncation. & \cite{GDE32005} \\
Dominance-based & GrEA & Uses nondominated sorting followed by grid rank, grid crowding, and grid-dominance criteria. & \cite{GrEA2013} \\
Dominance-based & KnEA & Uses nondominated fronts, adaptive knee-point identification, and distance-based last-front truncation. & \cite{KnEA2015} \\
Dominance-based & NSBiDiCo & Combines nondominated sorting with bidirectional differential coevolution for population updating. & \cite{NSBiDiCo2023} \\
Dominance-based & NSGA-II-SDR & Replaces standard Pareto comparison with a strengthened dominance relation encoding convergence and diversity. & \cite{NSGAIISDR2019} \\
Dominance-based & PESA-II & Maintains a nondominated archive and selects sparsely occupied objective-space hyperboxes. & \cite{PESAII2001} \\
Dominance-based & SPEA-R & Uses strength-Pareto fitness and reference directions to balance convergence and population spread. & \cite{SPEAR2017} \\
Dominance-based & t-DEA & Uses theta-dominance within reference-vector clusters to rank many-objective solutions. & \cite{tDEA2016} \\
\addlinespace
Decomposition-based & BCE-MOEA-D & Couples MOEA/D subproblems with a Pareto bi-criterion to balance convergence and diversity. & \cite{BCEIBEA2016} \\
Decomposition-based & EFR-RR & Ranks solutions by ensemble aggregation functions and restricts ranking through weight-vector niches. & \cite{EFRRR2016} \\
Decomposition-based & GWASF-GA & Uses multiple global weighting achievement scalarizing functions to approximate the complete Pareto front. & \cite{GWASFGA2017} \\
Decomposition-based & MOEA-D-AWA & Adapts weight vectors online when the population distribution becomes imbalanced. & \cite{MOEADAWA2014} \\
Decomposition-based & MOEA-D-DCWV & Controls the distribution of weight vectors to better match the current Pareto-front geometry. & \cite{MOEADDCWV2019} \\
Decomposition-based & MOEA/D-DE & Optimizes weighted subproblems through neighborhood replacement and differential-evolution variation. & \cite{MOEADDE2009} \\
Decomposition-based & MOEA-D-DRA & Dynamically allocates search effort among subproblems according to their recent utility. & \cite{MOEADDRA2009} \\
Decomposition-based & MOEA-D-DU & Combines aggregation-based convergence with perpendicular-distance diversity updating. & \cite{EFRRR2016} \\
Decomposition-based & MOEA-D-DYTS & Selects variation operators for decomposed subproblems through dynamic Thompson sampling. & \cite{MOEADDYTS2020} \\
Decomposition-based & MOEA-D-FRRMAB & Uses fitness-rate-rank credit and a multi-armed bandit to select differential-evolution operators. & \cite{MOEADFRRMAB2014} \\
Decomposition-based & MOEA-D-PaS & Learns Pareto-adaptive scalarizing methods for Pareto fronts with different geometries. & \cite{MOEADPaS2016} \\
Decomposition-based & MOEA-D-URAW & Combines random initial weights with online adaptive weight addition and deletion. & \cite{MOEADURAW2019} \\
Decomposition-based & tDEA-CPBI & Uses reference-vector decomposition with a constrained penalty-boundary-intersection criterion. & \cite{tDEACPBI2023} \\
Decomposition-based & WASF-GA & Uses achievement scalarizing functions centered on preference reference points. & \cite{WASFGA2015} \\
\addlinespace
Indicator-based & BCE-IBEA & Uses indicator fitness inside bi-criterion environmental selection to balance Pareto and non-Pareto criteria. & \cite{BCEIBEA2016} \\
Indicator-based & PREA & Defines pairwise indicator fitness, identifies promising regions, and iteratively removes crowded candidates. & \cite{PREA2021} \\
Indicator-based & SIBEA & Uses a binary weighted-hypervolume indicator for Pareto-compliant fitness assignment. & \cite{SIBEA2007} \\
\addlinespace
Specialized operator & CLIA & Combines cascade clustering with incremental reference-point learning to adapt environmental-selection niches. & \cite{CLIA2019} \\
Specialized operator & CMOEA-MS & Balances objective progress and constraint satisfaction using constraint-aware fitness and density estimation. & \cite{CMOEAMS2022} \\
Specialized operator & CMOPSO & Uses competitive swarm learning, leader competition, and particle-velocity updates. & \cite{CMOPSO2018} \\
Specialized operator & CoMMEA & Coevolves convergence- and diversity-oriented populations for multimodal Pareto sets. & \cite{CoMMEA2023} \\
Specialized operator & DM-MOEA & Uses dual models to respond to changes and exploit sparse variables in dynamic large-scale problems. & \cite{DMMOEA2025} \\
Specialized operator & LSMOF & Reformulates a high-dimensional decision problem into lower-dimensional search components. & \cite{LSMOF2019} \\
Specialized operator & MaOEA-CSS & Coordinates complementary convergence- and diversity-oriented selection strategies. & \cite{MaOEACSS2017} \\
Specialized operator & OSP-NSDE & Uses objective-space prediction to guide offspring generation within a nondominated-sorting DE framework. & \cite{OSPNSDE2019} \\
Specialized operator & PICEA-g & Coevolves candidate solutions and goal vectors, rewarding solutions that satisfy relatively rare goals. & \cite{PICEAg2013} \\
Specialized operator & SMPSO & Uses speed-constrained particle updates, mutation, and a nondominated external archive. & \cite{SMPSO2009} \\
Specialized operator & S-NSGA-II & Adds sparse-problem-specific crossover and mutation operators to the NSGA-II framework. & \cite{SNSGAII2024} \\
Specialized operator & SparseEA & Scores decision-variable activation patterns and applies sparsity-oriented variation. & \cite{SparseEA2020} \\
Specialized operator & SparseEA2 & Refines variable scoring and sparse variation to improve convergence and scalability. & \cite{SparseEA22023} \\
Specialized operator & SSCEA & Segments the search space into subspaces and coevolves subpopulations for convergence and diversity. & \cite{SSCEA2023} \\
Specialized operator & TELSO & Uses two-layer encoding and frequent-itemset learning for sparse large-scale swarm search. & \cite{TELSO2024} \\
Specialized operator & TS-NSGA-II & Switches between stage-wise convergence and diversity selection mechanisms during many-objective search. & \cite{TSNSGAII2022} \\
Specialized operator & TS-SparseEA & Separates sparse-pattern discovery and solution refinement into two evolutionary stages. & \cite{TSSparseEA2022} \\
Specialized operator & Two\_Arch2 & Maintains separate convergence and diversity archives with restricted mating and archive-specific updates. & \cite{TwoArch22015} \\
Specialized operator & VaEA & Uses vector-angle niching and angle-based elimination to preserve directional diversity. & \cite{VaEA2017} \\
Specialized operator & WOF & Reformulates large-scale problems through weight-based decision-variable transformation. & \cite{WOF2018} \\
\end{longtable}
\endgroup

The taxonomy contains 11 dominance-based, 14 decomposition-based, 3 indicator-based, and 20 specialized-operator-based algorithms. The 48 algorithm labels correspond to 46 unique publications. BCE-IBEA and BCE-MOEA-D originate from the same bi-criterion evolution study, while EFR-RR and MOEA-D-DU originate from the same decomposition-based many-objective study.

\subsection{Failure-Stage Decomposition}

The four outcome categories in Table~\ref{tab:exp1_failure_stages} are mutually exclusive and sum to 240 within each condition. An execution failure is syntactically valid code that does not complete execution; a convergence failure is executable code that does not meet the final-IGD criterion.

\begin{table}[H]
\centering
\caption{Migration results across the seven experimental conditions, classified by the earliest unsuccessful evaluation stage. Counts and percentages are computed over 240 attempts per condition.}
\label{tab:exp1_failure_stages}
\scriptsize
\setlength{\tabcolsep}{3.2pt}
\begin{tabular}{l c c c c}
\toprule
Method & Syntax failure & Execution failure & Convergence failure & Converged \\
\midrule
GLM-5.1 one-shot & 1 (0.42\%) & 64 (26.67\%) & 27 (11.25\%) & 148 (61.67\%) \\
DeepSeek V4 Pro one-shot & 0 & 156 (65.00\%) & 19 (7.92\%) & 65 (27.08\%) \\
Gemini 3 Flash one-shot & 0 & 151 (62.92\%) & 25 (10.42\%) & 64 (26.67\%) \\
DeepSeek V4 Flash one-shot & 0 & 177 (73.75\%) & 23 (9.58\%) & 40 (16.67\%) \\
EvoCoCo (DeepSeek V4 Flash) & 7 (2.92\%) & 42 (17.50\%) & 48 (20.00\%) & 143 (59.58\%) \\
EvoCoCo (DeepSeek V4 Pro) & 2 (0.83\%) & 35 (14.58\%) & 46 (19.17\%) & 157 (65.42\%) \\
EvoCoCo (Gemini 3 Flash) & 0 & 16 (6.67\%) & 35 (14.58\%) & 189 (78.75\%) \\
\bottomrule
\end{tabular}
\end{table}

\begin{figure}[H]
\centering
\includegraphics[width=0.65\textwidth]{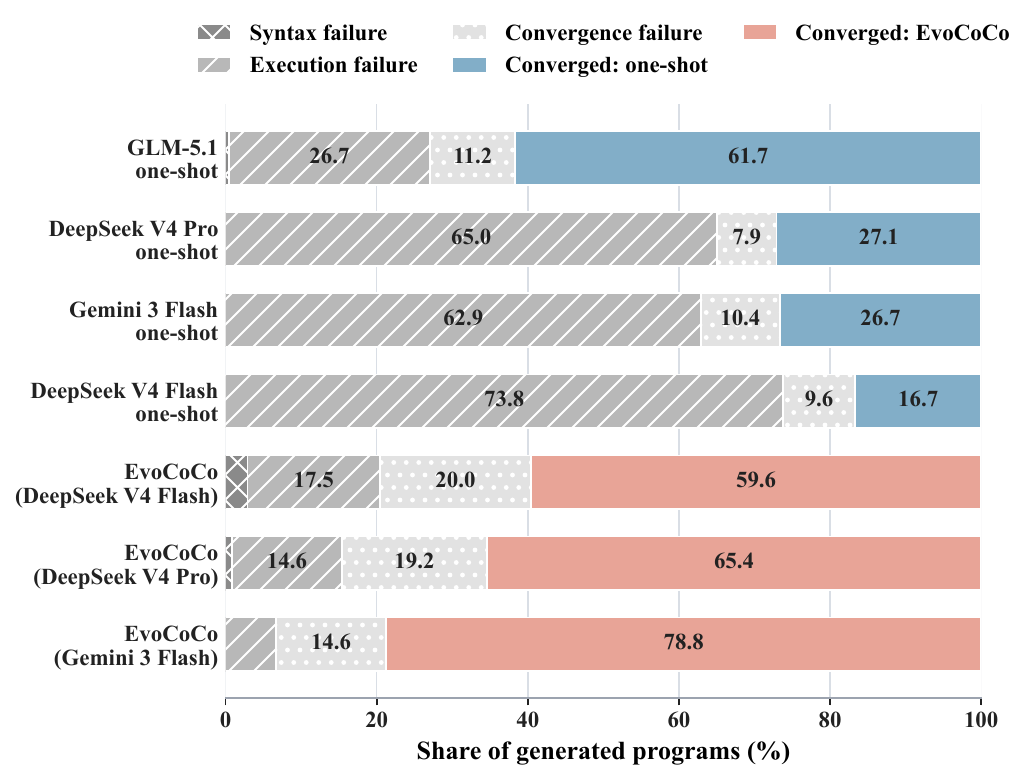}
\caption{Stage-wise distribution of migration results across the seven experimental conditions. Each condition comprises 240 conversion attempts. Patterned segments denote unsuccessful stages, while blue and coral denote converged one-shot and EvoCoCo attempts, respectively.}
\label{fig:exp1_failure_composition}
\end{figure}

Execution-stage failures account for 62.92\%--73.75\% of the three backend-matched one-shot conditions, compared with 6.67\%--17.50\% under EvoCoCo. More EvoCoCo candidates therefore reach optimization evaluation. The larger convergence-failure segment under some EvoCoCo conditions should be interpreted together with the substantial expansion of the converged segment, rather than as standalone evidence of reduced optimization fidelity.

\subsection{Algorithm-Level Stability and Paired Comparisons}

Table~\ref{tab:exp1_stability} reports how many algorithms achieve exactly zero through five converged attempts. EvoCoCo with Gemini 3 Flash is the only condition with no 0/5 algorithm and reaches 5/5 convergence on 19 algorithms.

\begin{table}[H]
\centering
\caption{Algorithm-level distribution of convergence counts across five attempts per condition. Each row accounts for all 48 benchmark algorithms.}
\label{tab:exp1_stability}
\scriptsize
\setlength{\tabcolsep}{5pt}
\begin{tabular}{l c c c c c c}
\toprule
Method & 0/5 & 1/5 & 2/5 & 3/5 & 4/5 & 5/5 \\
\midrule
GLM-5.1 one-shot & 3 & 4 & 10 & 10 & 11 & 10 \\
DeepSeek V4 Pro one-shot & 16 & 17 & 4 & 4 & 7 & 0 \\
Gemini 3 Flash one-shot & 23 & 10 & 4 & 2 & 5 & 4 \\
DeepSeek V4 Flash one-shot & 24 & 14 & 6 & 2 & 2 & 0 \\
EvoCoCo (DeepSeek V4 Flash) & 1 & 7 & 6 & 17 & 12 & 5 \\
EvoCoCo (DeepSeek V4 Pro) & 2 & 6 & 6 & 9 & 13 & 12 \\
EvoCoCo (Gemini 3 Flash) & \textbf{0} & 1 & 4 & 11 & 13 & \textbf{19} \\
\bottomrule
\end{tabular}
\end{table}

An algorithm is a win when the left method has more converged attempts than the right method, a loss when it has fewer, and a tie when the counts are equal. The exploratory two-sided sign test discards ties and treats algorithms as paired units. Holm adjustment is applied across the eight comparisons in Table~\ref{tab:exp1_paired_tests}.

\begin{table}[H]
\centering
\caption{Paired algorithm-level comparisons of convergence counts across five attempts per algorithm. Win/tie/loss counts are accompanied by two-sided sign-test $p$-values before and after Holm correction.}
\label{tab:exp1_paired_tests}
\scriptsize
\setlength{\tabcolsep}{2.6pt}
\begin{tabular}{p{0.30\textwidth} p{0.25\textwidth} c r r r}
\toprule
Left method & Right method & W/T/L & Added conv. & Raw $p$ & Holm $p$ \\
\midrule
EvoCoCo (Gemini 3 Flash) & Gemini 3 Flash one-shot & 38/3/7 & +125 & $3.12\times10^{-6}$ & $1.87\times10^{-5}$ \\
EvoCoCo (DeepSeek V4 Pro) & DeepSeek V4 Pro one-shot & 35/9/4 & +92 & $3.35\times10^{-7}$ & $2.35\times10^{-6}$ \\
EvoCoCo (DeepSeek V4 Flash) & DeepSeek V4 Flash one-shot & 43/3/2 & +103 & $5.89\times10^{-11}$ & $4.71\times10^{-10}$ \\
EvoCoCo (Gemini 3 Flash) & GLM-5.1 one-shot & 27/9/12 & +41 & 0.0237 & 0.1185 \\
EvoCoCo (DeepSeek V4 Pro) & GLM-5.1 one-shot & 16/16/16 & +9 & 1.0000 & 1.0000 \\
EvoCoCo (DeepSeek V4 Flash) & GLM-5.1 one-shot & 18/12/18 & $-5$ & 1.0000 & 1.0000 \\
EvoCoCo (Gemini 3 Flash) & EvoCoCo (DeepSeek V4 Pro) & 24/11/13 & +32 & 0.0989 & 0.3955 \\
EvoCoCo (DeepSeek V4 Pro) & EvoCoCo (DeepSeek V4 Flash) & 21/14/13 & +14 & 0.2295 & 0.6884 \\
\bottomrule
\end{tabular}
\begin{flushleft}
\footnotesize The tests quantify directional consistency at the algorithm level. They do not establish a causal pipeline effect because model versions, reasoning behavior, concurrency, and worker settings were not fully controlled across batches.
\end{flushleft}
\end{table}

\subsection{Conditional Quality, Runtime, and Token Use}

IGD and benchmark runtime in Table~\ref{tab:exp1_conditional_metrics} are computed over converged runs only. They describe the quality and execution time of the successful subset, not end-to-end migration reliability.

\begin{table}[H]
\centering
\caption{Optimization quality and runtime among converged migration attempts. IGD and runtime statistics are computed only over converged attempts.}
\label{tab:exp1_conditional_metrics}
\scriptsize
\setlength{\tabcolsep}{3.0pt}
\begin{tabular}{l r r r r r}
\toprule
Method & Conv. & Mean IGD & Median IGD & Mean time (s) & Median time (s) \\
\midrule
GLM-5.1 one-shot & 148 & \textbf{0.07817} & 0.07347 & 3.59 & 1.76 \\
DeepSeek V4 Pro one-shot & 65 & 0.08242 & 0.07826 & 2.81 & 1.16 \\
Gemini 3 Flash one-shot & 64 & 0.09500 & \textbf{0.07304} & 8.71 & 4.28 \\
DeepSeek V4 Flash one-shot & 40 & 0.08276 & 0.07431 & \textbf{2.13} & \textbf{0.59} \\
EvoCoCo (DeepSeek V4 Flash) & 143 & 0.09849 & 0.08299 & 6.34 & 1.38 \\
EvoCoCo (DeepSeek V4 Pro) & 157 & 0.09412 & 0.07928 & 2.99 & 0.67 \\
EvoCoCo (Gemini 3 Flash) & 189 & 0.08611 & 0.07477 & 3.86 & 1.63 \\
\bottomrule
\end{tabular}
\end{table}

Token use measures large language model (LLM) generation resources and is separate from benchmark execution time. Provider-reported reasoning tokens are included within completion tokens rather than added again. Because provider prices, cache accounting, and source completeness differ across batches, monetary cost is not compared here.

\begin{table}[H]
\centering
\caption{LLM token consumption across the seven experimental conditions, reported in total, per conversion attempt, and per converged attempt.}
\label{tab:exp1_tokens}
\scriptsize
\setlength{\tabcolsep}{3.0pt}
\begin{tabular}{l r r r}
\toprule
Method & Total tokens & Per attempt & Per converged result \\
\midrule
GLM-5.1 one-shot & 11,308,504 & 47.1k & 76.4k \\
DeepSeek V4 Pro one-shot & 3,692,224 & 15.4k & 56.8k \\
Gemini 3 Flash one-shot & \textbf{1,811,231} & \textbf{7.5k} & \textbf{28.3k} \\
DeepSeek V4 Flash one-shot & 3,876,583 & 16.2k & 96.9k \\
EvoCoCo (DeepSeek V4 Flash) & 101,165,726 & 421.5k & 707.5k \\
EvoCoCo (DeepSeek V4 Pro) & 103,274,934 & 430.3k & 657.8k \\
EvoCoCo (Gemini 3 Flash) & 53,725,242 & 223.9k & 284.3k \\
\bottomrule
\end{tabular}
\end{table}

The one-shot conditions use fewer tokens, whereas EvoCoCo attains higher reliability and coverage. For example, EvoCoCo with Gemini 3 Flash uses about ten times more tokens per converged result than its one-shot counterpart but expands convergence from 64 to 189 runs and coverage from 25 to 48 algorithms. Token efficiency therefore exposes a resource--reliability trade-off rather than a standalone migration-quality ranking.

\subsection{Algorithm-Level Outcomes and Backend Heterogeneity}

\begin{figure}[H]
\centering
\includegraphics[width=0.8\textwidth]{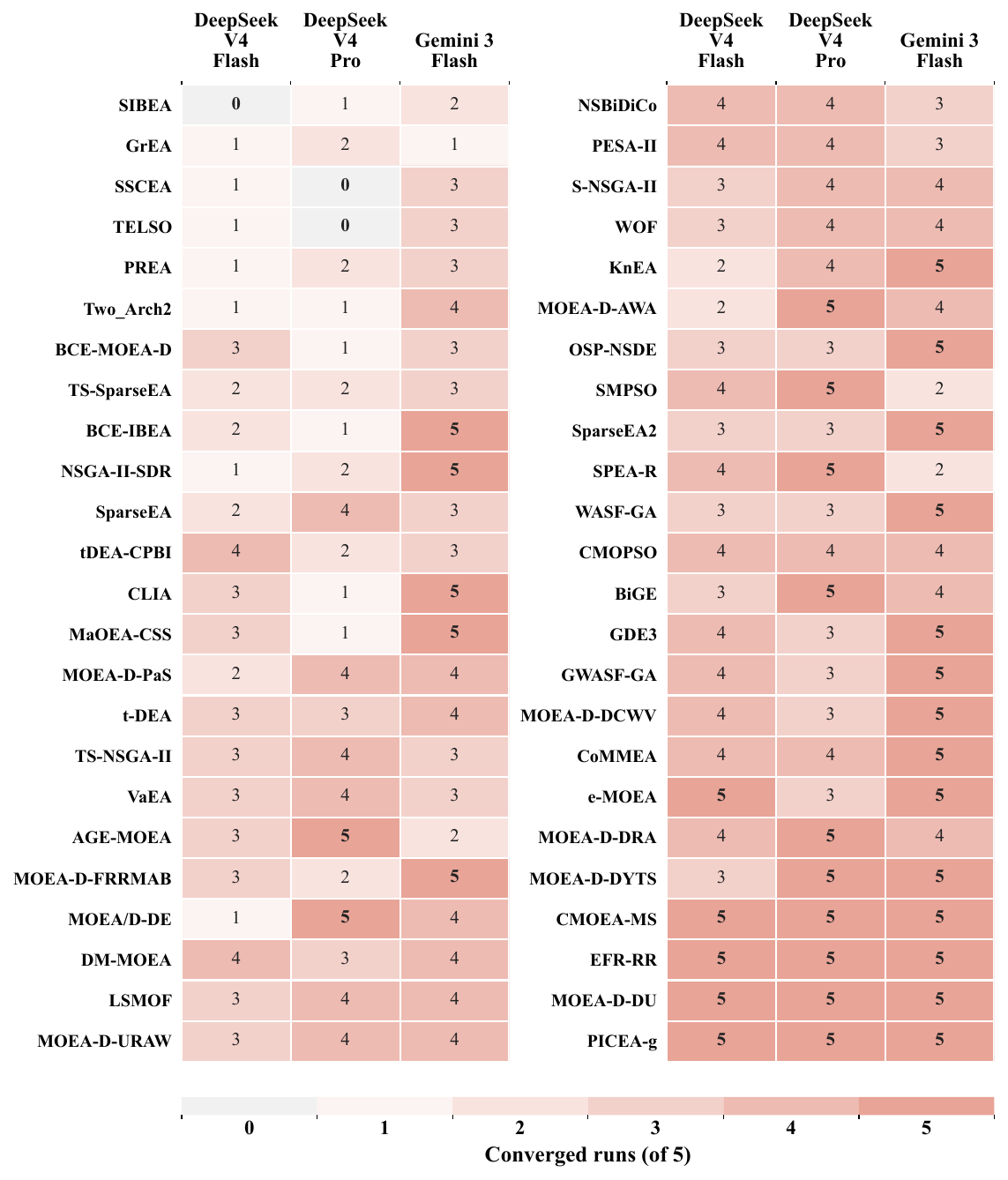}
\caption{Backend-specific convergence performance of EvoCoCo across the 48 benchmark algorithms. Each cell reports the number of converged attempts out of five; algorithms are ordered by total convergence across the three model backends.}
\label{fig:exp1_backend_heatmap}
\end{figure}

Figure~\ref{fig:exp1_backend_heatmap} orders the 48 algorithms by their combined number of converged runs under the three EvoCoCo backends. The backend effect is algorithm dependent rather than a uniform shift. SIBEA, GrEA, SSCEA, and TELSO remain difficult across backends, while CMOEA-MS, EFR-RR, MOEA-D-DU, and PICEA-g converge in all 15 attempts. Gemini 3 Flash avoids every 0/5 result; DeepSeek V4 Pro has no convergence on SSCEA and TELSO, and DeepSeek V4 Flash has none on SIBEA.

The complete algorithm-level execution/convergence matrix is reported in Table~\ref{tab:exp1_algorithm_matrix}. Each entry is $E/C$, where $E$ and $C$ are the numbers of executable and converged implementations among five attempts.

\begingroup
\scriptsize
\setlength{\tabcolsep}{2.3pt}
\begin{longtable}{l c c c c c c c}
\multicolumn{8}{c}{\label{tab:exp1_algorithm_matrix}\normalfont\footnotesize TABLE~\thetable}\\[-0.2ex]
\multicolumn{8}{c}{\normalfont\footnotesize\scshape Algorithm-Level Migration Results Across the Seven Experimental Conditions.}\\[0.5ex]
\toprule
Algorithm & \shortstack{GLM-5.1\\one-shot} & \shortstack{DeepSeek V4\\Pro one-shot} & \shortstack{Gemini 3 Flash\\one-shot} & \shortstack{DeepSeek V4\\Flash one-shot} & \shortstack{EvoCoCo\\(DeepSeek V4 Flash)} & \shortstack{EvoCoCo\\(DeepSeek V4 Pro)} & \shortstack{EvoCoCo\\(Gemini 3 Flash)} \\
\midrule
\endfirsthead
\multicolumn{8}{c}{\normalfont\footnotesize TABLE~\thetable\ (Continued)}\\[-0.2ex]
\multicolumn{8}{c}{\normalfont\footnotesize\scshape Algorithm-Level Migration Results Across the Seven Experimental Conditions.}\\[0.5ex]
\toprule
Algorithm & \shortstack{GLM-5.1\\one-shot} & \shortstack{DeepSeek V4\\Pro one-shot} & \shortstack{Gemini 3 Flash\\one-shot} & \shortstack{DeepSeek V4\\Flash one-shot} & \shortstack{EvoCoCo\\(DeepSeek V4 Flash)} & \shortstack{EvoCoCo\\(DeepSeek V4 Pro)} & \shortstack{EvoCoCo\\(Gemini 3 Flash)} \\
\midrule
\endhead
\midrule
\multicolumn{8}{r}{Continued on next page}\\
\endfoot
\bottomrule
\endlastfoot
AGE-MOEA & 2/2 & 4/4 & 1/1 & 1/1 & 3/3 & 5/5 & 4/2 \\
BCE-IBEA & 5/3 & 0/0 & 2/0 & 0/0 & 5/2 & 4/1 & 5/5 \\
BCE-MOEA-D & 4/4 & 0/0 & 0/0 & 0/0 & 3/3 & 2/1 & 4/3 \\
BiGE & 5/5 & 2/1 & 1/1 & 2/1 & 3/3 & 5/5 & 5/4 \\
CLIA & 2/2 & 1/1 & 0/0 & 0/0 & 5/3 & 2/1 & 5/5 \\
CMOEA-MS & 4/3 & 1/1 & 5/2 & 1/1 & 5/5 & 5/5 & 5/5 \\
CMOPSO & 4/4 & 4/4 & 1/1 & 2/2 & 4/4 & 4/4 & 4/4 \\
CoMMEA & 5/4 & 0/0 & 1/0 & 1/1 & 5/4 & 5/4 & 5/5 \\
DM-MOEA & 4/4 & 0/0 & 3/3 & 1/0 & 5/4 & 4/3 & 4/4 \\
EFR-RR & 5/4 & 1/1 & 0/0 & 2/1 & 5/5 & 5/5 & 5/5 \\
e-MOEA & 4/4 & 2/2 & 0/0 & 1/1 & 5/5 & 5/3 & 5/5 \\
GDE3 & 4/4 & 5/4 & 1/1 & 4/4 & 4/4 & 5/3 & 5/5 \\
GrEA & 4/3 & 1/1 & 3/2 & 2/0 & 1/1 & 4/2 & 5/1 \\
GWASF-GA & 3/3 & 4/4 & 2/1 & 4/2 & 5/4 & 4/3 & 5/5 \\
KnEA & 2/1 & 1/1 & 5/5 & 0/0 & 2/2 & 5/4 & 5/5 \\
LSMOF & 4/4 & 1/1 & 1/1 & 0/0 & 3/3 & 4/4 & 4/4 \\
MaOEA-CSS & 5/5 & 4/4 & 5/5 & 4/4 & 3/3 & 5/1 & 5/5 \\
MOEA-D-AWA & 2/2 & 0/0 & 0/0 & 0/0 & 4/2 & 5/5 & 4/4 \\
MOEA-D-DCWV & 1/1 & 1/1 & 0/0 & 0/0 & 5/4 & 5/3 & 5/5 \\
MOEA/D-DE & 3/3 & 3/3 & 0/0 & 2/1 & 5/1 & 5/5 & 5/4 \\
MOEA-D-DRA & 2/2 & 3/2 & 0/0 & 1/1 & 4/4 & 5/5 & 4/4 \\
MOEA-D-DU & 3/3 & 0/0 & 3/3 & 0/0 & 5/5 & 5/5 & 5/5 \\
MOEA-D-DYTS & 5/5 & 2/1 & 0/0 & 1/1 & 5/3 & 5/5 & 5/5 \\
MOEA-D-FRRMAB & 2/2 & 0/0 & 2/2 & 0/0 & 3/3 & 4/2 & 5/5 \\
MOEA-D-PaS & 2/2 & 2/1 & 0/0 & 0/0 & 5/2 & 4/4 & 5/4 \\
MOEA-D-URAW & 4/4 & 1/1 & 0/0 & 0/0 & 3/3 & 5/4 & 4/4 \\
NSBiDiCo & 2/2 & 1/1 & 0/0 & 1/1 & 4/4 & 4/4 & 4/3 \\
NSGA-II-SDR & 5/1 & 1/0 & 4/0 & 2/0 & 3/1 & 3/2 & 5/5 \\
OSP-NSDE & 3/3 & 4/3 & 0/0 & 1/1 & 3/3 & 3/3 & 5/5 \\
PESA-II & 4/4 & 4/4 & 5/5 & 1/1 & 5/4 & 4/4 & 4/3 \\
PICEA-g & 3/3 & 3/3 & 1/1 & 3/3 & 5/5 & 5/5 & 5/5 \\
PREA & 5/5 & 4/4 & 4/4 & 2/1 & 5/1 & 5/2 & 5/3 \\
SIBEA & 2/2 & 1/1 & 3/0 & 2/2 & 0/0 & 1/1 & 2/2 \\
SMPSO & 5/5 & 1/1 & 5/5 & 3/3 & 5/4 & 5/5 & 5/2 \\
S-NSGA-II & 5/5 & 2/2 & 4/4 & 2/2 & 5/3 & 5/4 & 5/4 \\
SparseEA & 5/5 & 2/1 & 1/1 & 0/0 & 3/2 & 5/4 & 5/3 \\
SparseEA2 & 3/3 & 0/0 & 1/1 & 0/0 & 5/3 & 3/3 & 5/5 \\
SPEA-R & 3/3 & 1/0 & 4/1 & 1/0 & 5/4 & 5/5 & 4/2 \\
SSCEA & 2/0 & 0/0 & 0/0 & 2/0 & 3/1 & 4/0 & 4/3 \\
t-DEA & 4/4 & 3/3 & 0/0 & 0/0 & 3/3 & 3/3 & 5/4 \\
tDEA-CPBI & 5/5 & 1/1 & 0/0 & 2/2 & 4/4 & 2/2 & 5/3 \\
TELSO & 3/0 & 2/0 & 0/0 & 0/0 & 5/1 & 4/0 & 5/3 \\
TS-NSGA-II & 5/2 & 0/0 & 4/4 & 1/0 & 4/3 & 5/4 & 5/3 \\
TS-SparseEA & 4/2 & 1/0 & 5/4 & 0/0 & 3/2 & 4/2 & 5/3 \\
Two\_Arch2 & 4/1 & 2/1 & 1/0 & 3/0 & 5/1 & 5/1 & 4/4 \\
VaEA & 5/5 & 4/2 & 4/4 & 3/1 & 4/3 & 4/4 & 5/3 \\
WASF-GA & 3/0 & 4/0 & 5/0 & 3/0 & 4/3 & 4/3 & 5/5 \\
WOF & 5/5 & 0/0 & 2/2 & 2/2 & 3/3 & 4/4 & 5/4 \\
\end{longtable}
\endgroup

\subsection{Failure Analysis}

Across the three EvoCoCo conditions, the numbers of non-converged candidates are 97 for DeepSeek V4 Flash, 83 for DeepSeek V4 Pro, and 51 for Gemini 3 Flash. All 129 convergence-stage failures produce a finite final IGD above 0.25. These cases therefore represent optimization-behavior failures rather than missing application programming interface (API) or runtime failures. The raw backend ranking is consistent for both executability and convergence, although algorithm-level contrasts remain substantial.

The DeepSeek V4 Flash batch retains a single-worker retry with exception traces. Serial retry recovers 23 of the 42 raw execution failures and eight converged runs. This recovery shows that part of the raw failure count is sensitive to concurrent execution. The remaining 19 reproducible execution failures are summarized in Table~\ref{tab:exp1_concrete_failures}. This retry is a sensitivity analysis and is not used in the main comparison because the other backends were not subjected to the same policy.

\begin{table}[H]
\centering
\caption{Reproducible execution-failure categories for the 19 DeepSeek V4 Flash attempts that remained unsuccessful under single-worker serial verification.}
\label{tab:exp1_concrete_failures}
\scriptsize
\setlength{\tabcolsep}{3.5pt}
\begin{tabular}{p{0.42\textwidth} c p{0.42\textwidth}}
\toprule
Failure type & Count & Representative cause \\
\midrule
Unsupported Boolean \texttt{argmax} & 5 & \texttt{torch.argmax} called on a Boolean tensor \\
GPU index/device assertion & 3 & Invalid scatter, gather, or index values \\
Indexed-assignment dtype mismatch & 3 & Incompatible integer or floating tensor dtypes \\
Tensor shape mismatch & 2 & Operations combine incompatible dimensions \\
Selection-API argument mismatch & 2 & Integer passed where a tensor list is required \\
Tuple return treated as tensor & 2 & Tuple used with tensor operations or methods \\
Missing API/import & 1 & \texttt{hypervolume} unavailable from installed package \\
Singular linear system & 1 & \texttt{torch.linalg.solve} receives a singular matrix \\
\bottomrule
\end{tabular}
\end{table}

These errors separate into API-contract failures, tensor-contract failures, and numerical-robustness failures. The first two groups account for 18 of the 19 reproducible failures. This distribution indicates that API awareness and tensor-contract validation are more important for these cases than additional syntax-only checks. The DeepSeek V4 Pro and Gemini 3 Flash batches do not retain comparable per-run stderr traces, so their execution failures cannot be assigned defensibly to the same concrete categories.

\FloatBarrier
\section{Detailed Optimization-Fidelity Results}

This section reports the optimization-fidelity results for the 48 tensorized implementations on DTLZ1--DTLZ7, WFG1--WFG9, LSMOP1--LSMOP9, and MaF1--MaF15. For every algorithm--problem combination, EvoX and PlatEMO are each evaluated in 21 independent runs with population size $N=100$ and a budget of 100 generations. The two frameworks use the same problem definitions, objective and decision dimensions, variable bounds, population sizes, and generation budgets. Final IGD is the primary endpoint, and classification uses the arithmetic mean of the 21 final IGD values.

A combination is \emph{Improved} when EvoX is no worse than PlatEMO and \emph{Preserved} when its positive degradation satisfies either an absolute IGD tolerance of 0.10 or a relative IGD tolerance of 200\%. Other finite comparisons are labeled \emph{IGD degradation}. A combination is labeled \emph{Reference invalid} when the PlatEMO implementation does not provide the required number of usable reference runs.

Among the 1,920 algorithm--problem combinations, 16 PlatEMO references are invalid. The remaining 1,904 valid comparisons contain 724 Improved, 956 Preserved, and 224 IGD-degradation outcomes. Thus, 1,680 valid comparisons satisfy the fidelity criterion, which yields an overall fidelity coverage of 88.2\%. At the algorithm level, 41 of the 48 tensorized implementations achieve at least 80\% fidelity coverage over their valid reference problems. Therefore, 85.4\% of the benchmark MOEAs satisfy the algorithm-level coverage criterion.

\subsection{Problem Configuration}

All recorded EvoX and PlatEMO problem groups use three objectives. MaF8--MaF9 therefore use the experiment's explicit $M=3$ setting rather than the PlatEMO class default of 10. The confirmed DTLZ7 setting is $D=22$, matching $D=M+19$ for $M=3$. The decision dimensions and variable bounds used in the matched runs are listed in Table~\ref{tab:fidelity-problem-configuration}.

\begin{table}[H]
\centering
\caption{Matched problem configurations for the optimization-fidelity experiment, including decision dimensions and variable bounds for each benchmark suite.}
\label{tab:fidelity-problem-configuration}
\scriptsize
\setlength{\tabcolsep}{5pt}
\begin{tabular}{lrrll}
\toprule
Problems & $M$ & $D$ & Lower bound & Upper bound \\
\midrule
DTLZ1 & 3 & 7 & $0$ & $1$ \\
DTLZ2--DTLZ6 & 3 & 12 & $0$ & $1$ \\
DTLZ7 & 3 & 22 & $0$ & $1$ \\
WFG1--WFG9 & 3 & 12 & $0$ & $[2,4,\ldots,24]$ \\
LSMOP1--LSMOP9 & 3 & 300 & $0$ & $[1,1,10,\ldots,10]$ \\
MaF1--MaF6 & 3 & 12 & $0$ & $1$ \\
MaF7 & 3 & 22 & $0$ & $1$ \\
MaF8--MaF9 & 3 & 2 & $[-10^4,-10^4]$ & $[10^4,10^4]$ \\
MaF10--MaF12 & 3 & 12 & $0$ & $[2,4,\ldots,24]$ \\
MaF13 & 3 & 5 & $[0,0,-2,-2,-2]$ & $[1,1,2,2,2]$ \\
MaF14--MaF15 & 3 & 60 & $0$ & $[1,1,10,\ldots,10]$ \\
\bottomrule
\end{tabular}
\end{table}

\subsection{Suite-Level Results}

\begin{table}[H]
\centering
\caption{Overall and suite-level optimization-fidelity results across the DTLZ, WFG, LSMOP, and MaF benchmark suites. Coverage denotes the proportion of valid PlatEMO references classified as Improved or Preserved.}
\label{tab:fidelity-suite-summary}
\scriptsize
\setlength{\tabcolsep}{3.2pt}
\begin{tabular}{lrrrrrrrr}
\toprule
Suite & All & Valid & Improved & Preserved & IGD degr. & Ref. invalid & Passes & Coverage \\
\midrule
DTLZ & 336 & 332 & 141 & 138 & 53 & 4 & 279 & 84.0\% \\
WFG & 432 & 429 & 161 & 253 & 15 & 3 & 414 & 96.5\% \\
LSMOP & 432 & 427 & 149 & 213 & 65 & 5 & 362 & 84.8\% \\
MaF & 720 & 716 & 273 & 352 & 91 & 4 & 625 & 87.3\% \\
\textbf{Overall} & \textbf{1920} & \textbf{1904} & \textbf{724} & \textbf{956} & \textbf{224} & \textbf{16} & \textbf{1680} & \textbf{88.2\%} \\
\bottomrule
\end{tabular}
\end{table}

\begin{figure}[H]
\centering
\includegraphics[width=\textwidth]{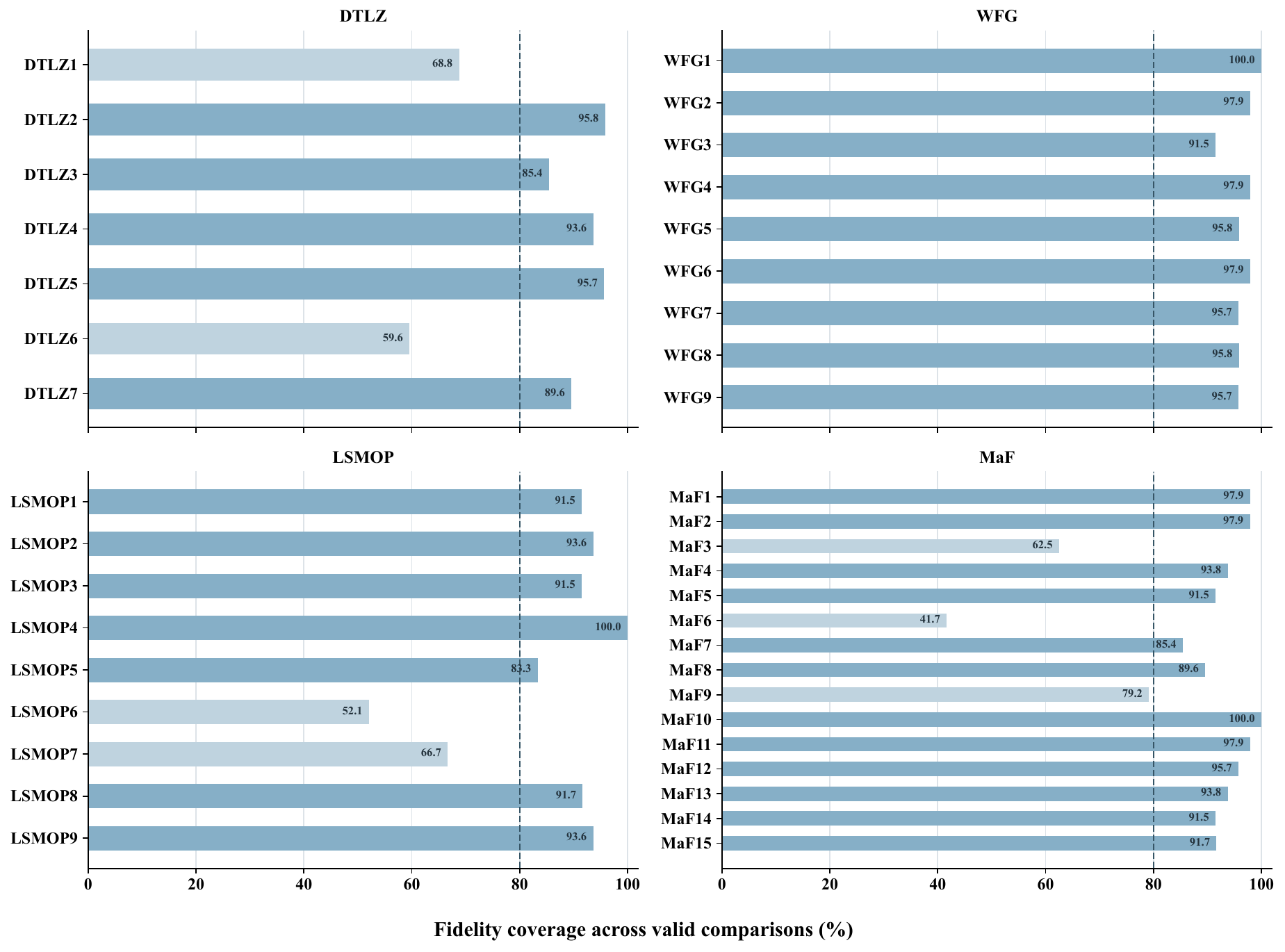}
\caption{Problem-level optimization-fidelity coverage across the DTLZ, WFG, LSMOP, and MaF benchmark suites. Dark-blue and pale-blue bars indicate coverage above and below the 80\% reference level, respectively; Reference-invalid cases are excluded.}
\label{fig:supp-fidelity-problems}
\end{figure}

Figure~\ref{fig:supp-fidelity-problems} summarizes problem-level coverage, while the four outcome matrices below retain every algorithm--problem classification and each algorithm's suite-specific pass count. These views provide the complete categorical evidence without repeating it in separate problem- and algorithm-level summary tables.
\subsection{Invalid PlatEMO References}

Among the 16 reference-invalid combinations, nine are caused by PlatEMO timeouts for SIBEA on DTLZ5, DTLZ6, WFG3, WFG7, WFG9, LSMOP2, LSMOP4, MaF2, and MaF12. The remaining seven arise from PlatEMO execution errors: OSP-NSDE on DTLZ4 and MaF5, CLIA on DTLZ5 and LSMOP9, and TELSO on LSMOP1, LSMOP3, and MaF14. These cases are reported for completeness but are excluded from all effective coverage denominators.

\subsection{Tolerance Sensitivity}

\begin{figure}[H]
\centering
\includegraphics[width=0.7\textwidth]{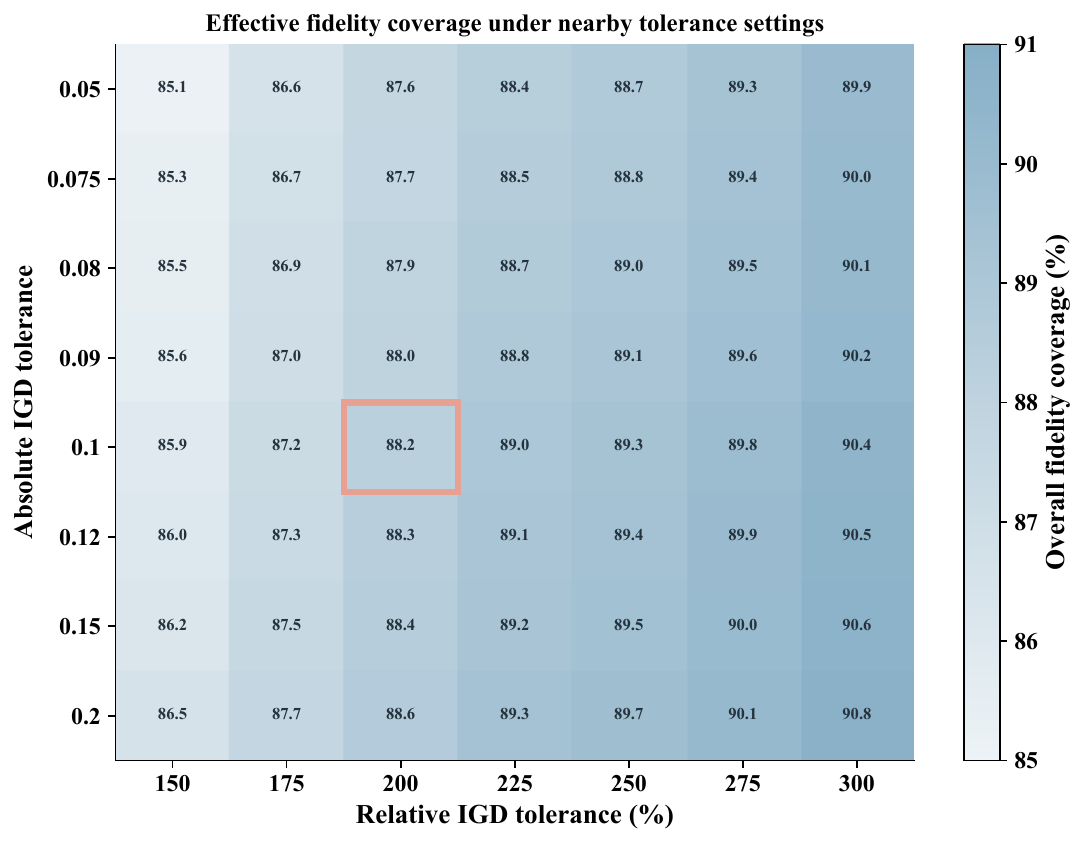}
\caption{Sensitivity of overall optimization-fidelity coverage to the absolute and relative IGD tolerances. The outlined cell denotes the selected setting of 0.10 and 200\%, respectively.}
\label{fig:supp-fidelity-sensitivity}
\end{figure}

The selected rule lies inside a stable neighborhood: the overall coverage changes gradually as either tolerance is varied, rather than jumping at the selected cell. The chosen setting therefore provides a transparent practical threshold without relying on a narrow, data-specific discontinuity.

\subsection{Representative Convergence Trajectories}

Each convergence curve reports the checkpoint-wise arithmetic mean over 21 independent runs, with a band of plus or minus one sample standard deviation. The recorded trajectories contain 101 IGD checkpoints but no checkpoint-level timestamps. The displayed time coordinate is therefore obtained by mapping the normalized checkpoint index linearly to the median total runtime of the corresponding algorithm--problem combination and should be interpreted as an approximate wall-clock scale rather than instrumented per-generation timing.

\begin{figure}[p]
\centering
\subfloat{\includegraphics[width=0.235\textwidth]{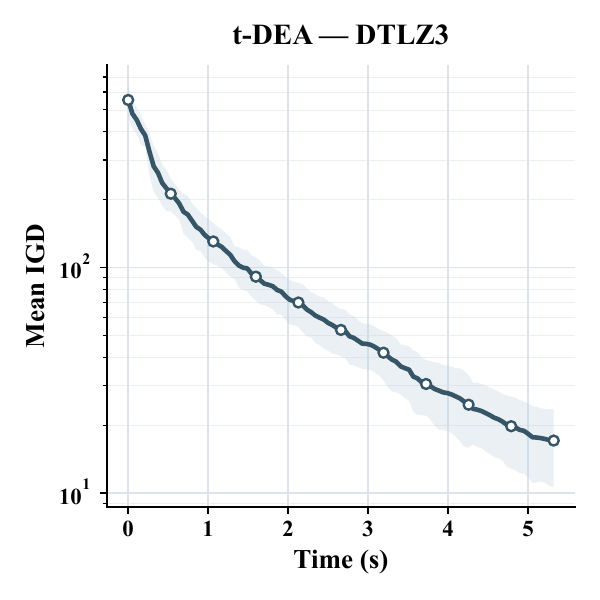}}\hfill
\subfloat{\includegraphics[width=0.235\textwidth]{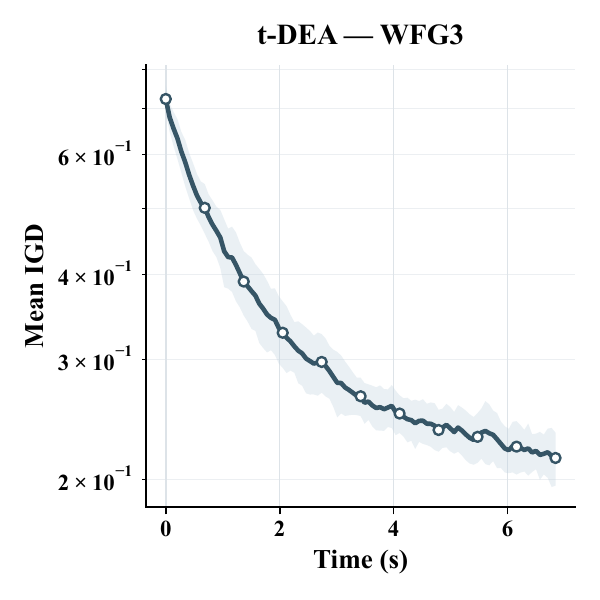}}\hfill
\subfloat{\includegraphics[width=0.235\textwidth]{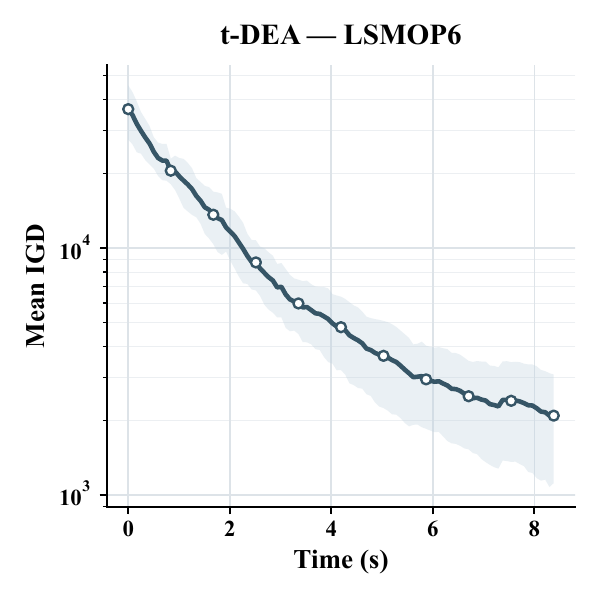}}\hfill
\subfloat{\includegraphics[width=0.235\textwidth]{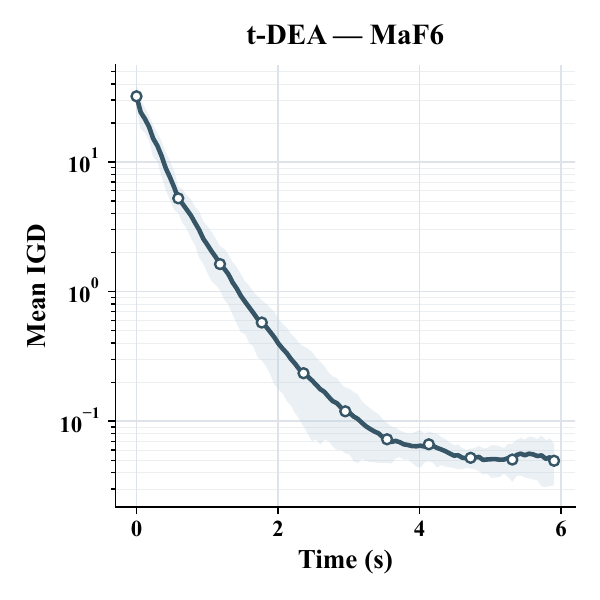}}\\[1mm]
\subfloat{\includegraphics[width=0.235\textwidth]{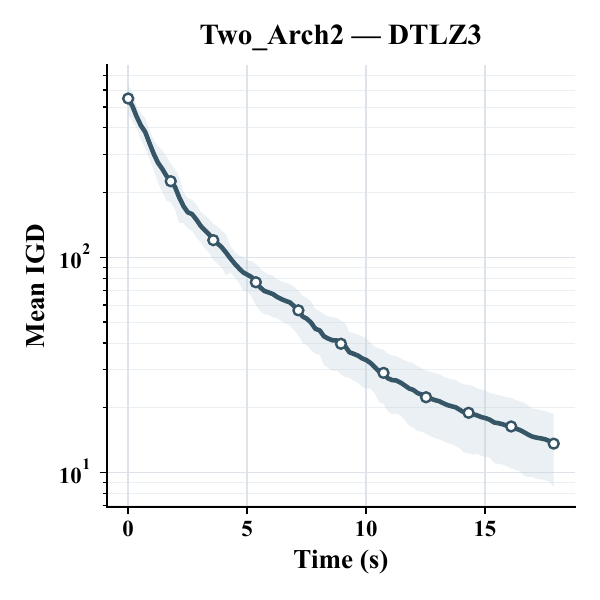}}\hfill
\subfloat{\includegraphics[width=0.235\textwidth]{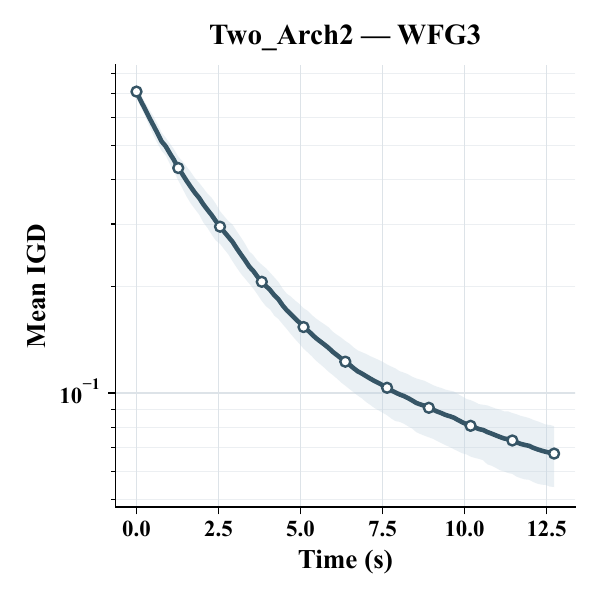}}\hfill
\subfloat{\includegraphics[width=0.235\textwidth]{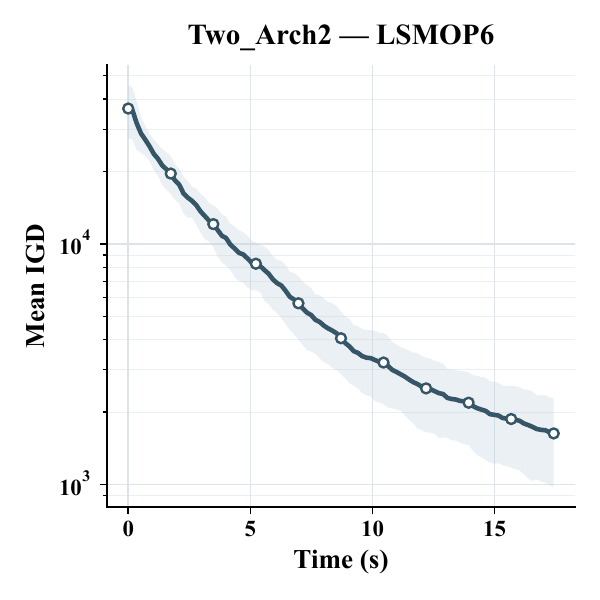}}\hfill
\subfloat{\includegraphics[width=0.235\textwidth]{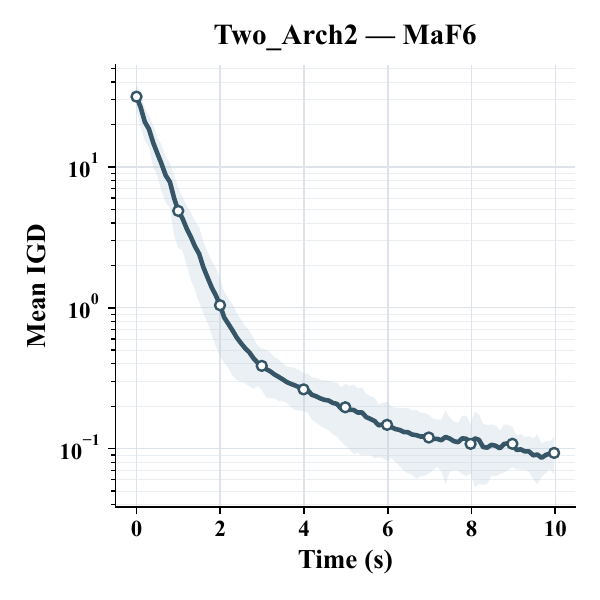}}\\[1mm]
\subfloat{\includegraphics[width=0.235\textwidth]{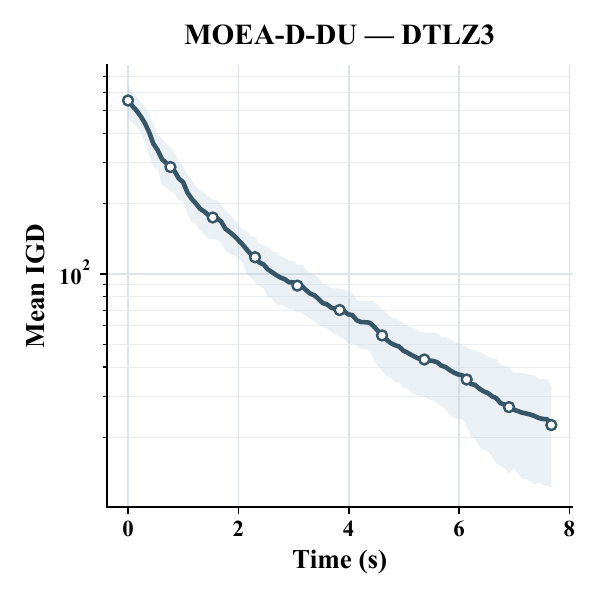}}\hfill
\subfloat{\includegraphics[width=0.235\textwidth]{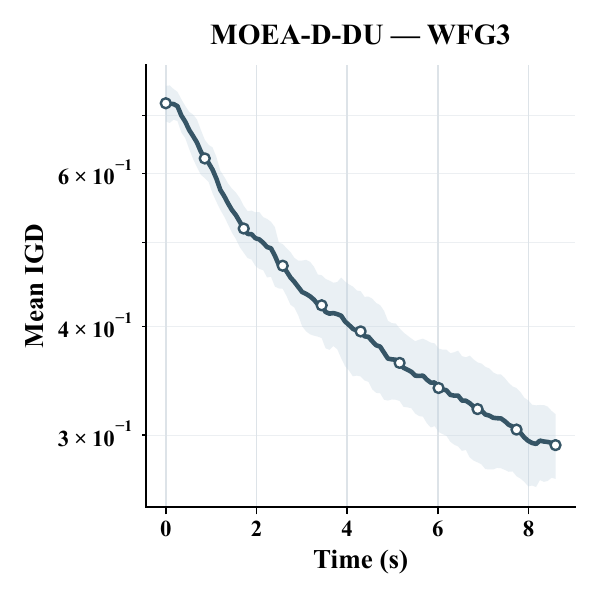}}\hfill
\subfloat{\includegraphics[width=0.235\textwidth]{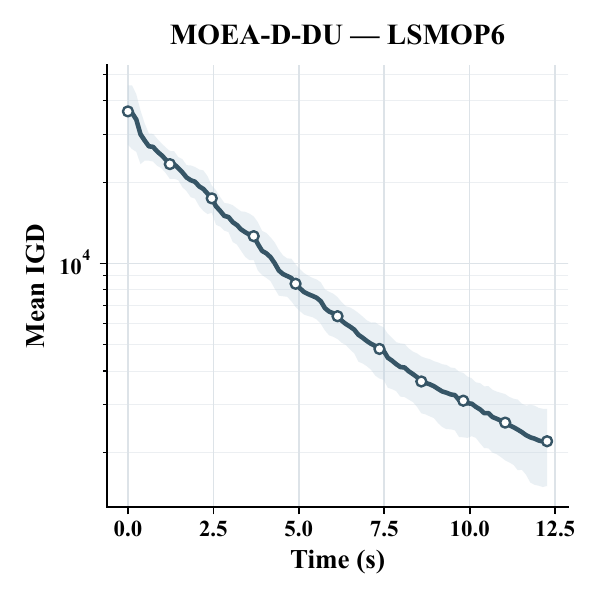}}\hfill
\subfloat{\includegraphics[width=0.235\textwidth]{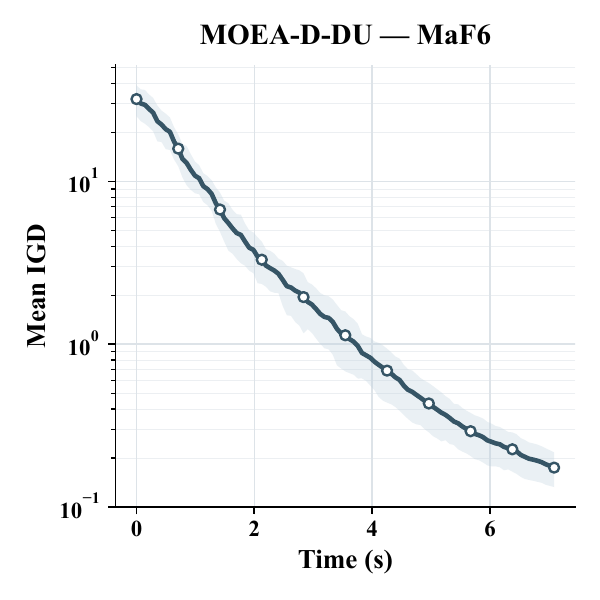}}\\[1mm]
\subfloat{\includegraphics[width=0.235\textwidth]{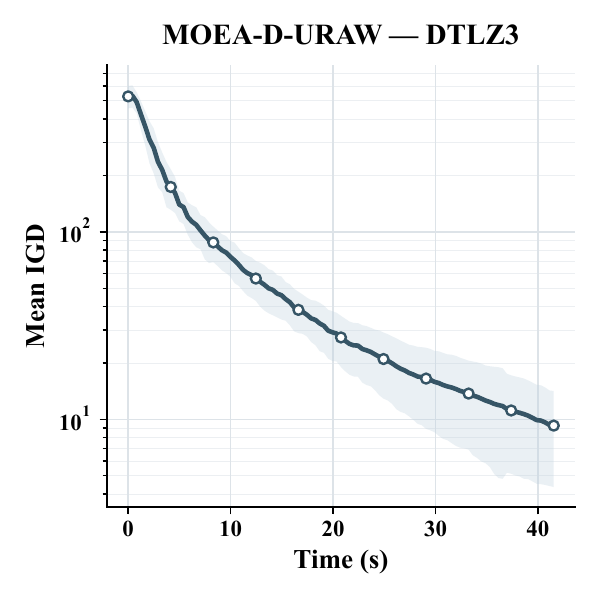}}\hfill
\subfloat{\includegraphics[width=0.235\textwidth]{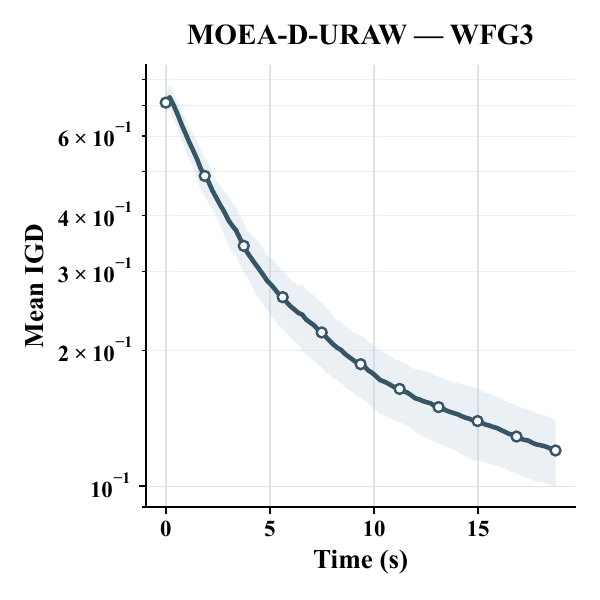}}\hfill
\subfloat{\includegraphics[width=0.235\textwidth]{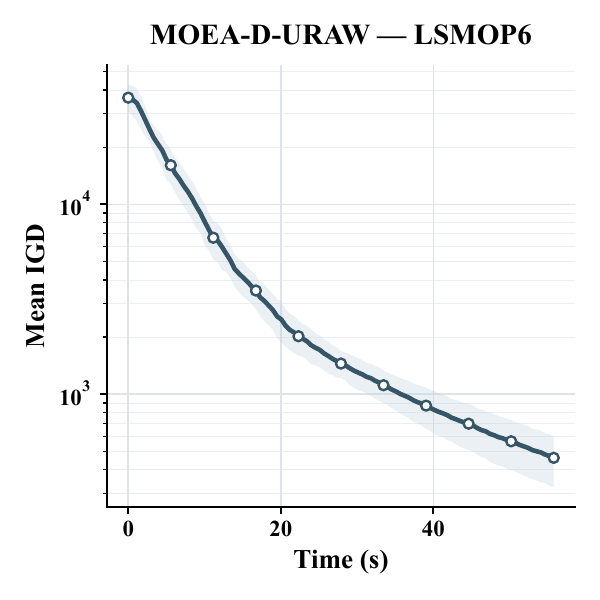}}\hfill
\subfloat{\includegraphics[width=0.235\textwidth]{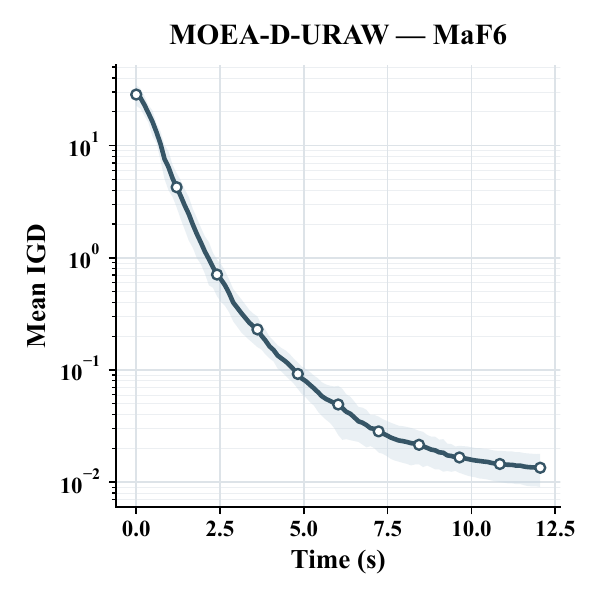}}
\caption{Representative convergence trajectories for four algorithms with at least 95\% effective optimization-fidelity coverage. Rows correspond to t-DEA, Two\_Arch2, MOEA-D-DU, and MOEA-D-URAW; columns correspond to DTLZ3, WFG3, LSMOP6, and MaF6. All displayed combinations are classified as Improved or Preserved. The horizontal axes are approximate wall-clock scales obtained by mapping checkpoint index to the median total runtime.}
\label{fig:supp-fidelity-convergence-95}
\end{figure}

Figure~\ref{fig:supp-fidelity-convergence-95} presents 16 representative trajectories selected from combinations classified as \emph{Improved} or \emph{Preserved}. The four algorithms each have at least 95\% overall effective coverage and contribute one trajectory from every benchmark suite. Combinations classified as IGD degradation are omitted so that the figure focuses on fidelity-preserving convergence behavior.

\FloatBarrier

\clearpage
\begingroup
\subsection{Complete Algorithm--Problem Outcome Matrices}

The five portrait matrices below retain all 1,920 algorithm--problem classifications while avoiding 40 repetitions of the same table structure. DTLZ, WFG, and LSMOP are each shown in one matrix; the wider MaF suite is divided into two column blocks with the same algorithm order. Each cell applies the arithmetic-mean rule defined above. The final three columns in the second MaF block report full-suite passed comparisons, valid references, and coverage. The 16 X cases and their exclusion reasons are described above.

\noindent\textbf{Cell legend:} \statI\ Improved; \statP\ Preserved; \statD\ IGD degradation; \statX\ Reference invalid. Letter codes remain visible in grayscale and do not rely on color alone.

\scriptsize
\setlength{\tabcolsep}{4.0pt}
\renewcommand{\arraystretch}{0.96}
\begin{longtable}{l*{7}{c} r r r}
\multicolumn{11}{c}{\label{tab:fidelity-status-dtlz}\normalfont\footnotesize TABLE~\thetable}\\[-0.2ex]
\multicolumn{11}{c}{\normalfont\footnotesize\scshape Optimization-Fidelity Outcome Matrix for the DTLZ Suite.}\\[0.5ex]
\toprule
Algorithm & DTLZ1 & DTLZ2 & DTLZ3 & DTLZ4 & DTLZ5 & DTLZ6 & DTLZ7 & Pass & Valid & Coverage \\
\midrule
\endfirsthead
\multicolumn{11}{c}{\normalfont\footnotesize TABLE~\thetable\ (Continued)}\\[-0.2ex]
\multicolumn{11}{c}{\normalfont\footnotesize\scshape Optimization-Fidelity Outcome Matrix for the DTLZ Suite.}\\[0.5ex]
\toprule
Algorithm & DTLZ1 & DTLZ2 & DTLZ3 & DTLZ4 & DTLZ5 & DTLZ6 & DTLZ7 & Pass & Valid & Coverage \\
\midrule
\endhead
\midrule
\multicolumn{11}{r}{Continued on next page}\\
\endfoot
\bottomrule
\endlastfoot
AGE-MOEA & \statD & \statP & \statP & \statD & \statP & \statP & \statP & 5 & 7 & 71.4\% \\
BCE-IBEA & \statD & \statP & \statP & \statP & \statP & \statP & \statI & 6 & 7 & 85.7\% \\
BCE-MOEA-D & \statD & \statP & \statD & \statP & \statP & \statD & \statP & 4 & 7 & 57.1\% \\
BiGE & \statD & \statI & \statP & \statI & \statP & \statI & \statP & 6 & 7 & 85.7\% \\
CLIA & \statP & \statP & \statP & \statI & \statX & \statD & \statP & 5 & 6 & 83.3\% \\
CMOEA-MS & \statD & \statP & \statP & \statI & \statI & \statD & \statP & 5 & 7 & 71.4\% \\
CMOPSO & \statI & \statP & \statI & \statP & \statP & \statI & \statI & 7 & 7 & 100.0\% \\
CoMMEA & \statI & \statI & \statI & \statI & \statI & \statI & \statI & 7 & 7 & 100.0\% \\
DM-MOEA & \statD & \statI & \statP & \statI & \statI & \statP & \statP & 6 & 7 & 85.7\% \\
e-MOEA & \statI & \statP & \statI & \statI & \statI & \statP & \statI & 7 & 7 & 100.0\% \\
EFR-RR & \statI & \statI & \statI & \statI & \statP & \statP & \statP & 7 & 7 & 100.0\% \\
GDE3 & \statI & \statI & \statP & \statI & \statI & \statI & \statI & 7 & 7 & 100.0\% \\
GrEA & \statP & \statP & \statP & \statI & \statP & \statP & \statP & 7 & 7 & 100.0\% \\
GWASF-GA & \statD & \statI & \statD & \statI & \statP & \statD & \statP & 4 & 7 & 57.1\% \\
KnEA & \statD & \statI & \statP & \statI & \statI & \statP & \statD & 5 & 7 & 71.4\% \\
LSMOF & \statI & \statI & \statI & \statI & \statI & \statI & \statI & 7 & 7 & 100.0\% \\
MaOEA-CSS & \statP & \statD & \statI & \statD & \statP & \statD & \statD & 3 & 7 & 42.9\% \\
MOEA-D-AWA & \statD & \statP & \statP & \statI & \statP & \statD & \statP & 5 & 7 & 71.4\% \\
MOEA-D-DCWV & \statP & \statP & \statI & \statP & \statP & \statD & \statI & 6 & 7 & 85.7\% \\
MOEA/D-DE & \statP & \statP & \statP & \statD & \statP & \statD & \statP & 5 & 7 & 71.4\% \\
MOEA-D-DRA & \statI & \statI & \statI & \statI & \statP & \statI & \statP & 7 & 7 & 100.0\% \\
MOEA-D-DU & \statP & \statP & \statP & \statP & \statP & \statP & \statI & 7 & 7 & 100.0\% \\
MOEA-D-DYTS & \statI & \statI & \statP & \statI & \statP & \statP & \statI & 7 & 7 & 100.0\% \\
MOEA-D-FRRMAB & \statI & \statI & \statI & \statP & \statP & \statD & \statP & 6 & 7 & 85.7\% \\
MOEA-D-PaS & \statI & \statD & \statI & \statP & \statD & \statD & \statI & 4 & 7 & 57.1\% \\
MOEA-D-URAW & \statI & \statP & \statP & \statI & \statP & \statI & \statP & 7 & 7 & 100.0\% \\
NSBiDiCo & \statI & \statI & \statI & \statI & \statI & \statI & \statI & 7 & 7 & 100.0\% \\
NSGA-II-SDR & \statD & \statI & \statD & \statI & \statP & \statD & \statD & 3 & 7 & 42.9\% \\
OSP-NSDE & \statD & \statI & \statD & \statX & \statP & \statD & \statI & 3 & 6 & 50.0\% \\
PESA-II & \statP & \statP & \statI & \statP & \statP & \statP & \statI & 7 & 7 & 100.0\% \\
PICEA-g & \statD & \statP & \statP & \statP & \statP & \statD & \statP & 5 & 7 & 71.4\% \\
PREA & \statP & \statP & \statI & \statI & \statP & \statI & \statP & 7 & 7 & 100.0\% \\
S-NSGA-II & \statI & \statP & \statI & \statI & \statI & \statI & \statI & 7 & 7 & 100.0\% \\
SIBEA & \statP & \statP & \statP & \statP & \statX & \statX & \statP & 5 & 5 & 100.0\% \\
SMPSO & \statD & \statI & \statD & \statP & \statP & \statI & \statI & 5 & 7 & 71.4\% \\
SparseEA & \statI & \statI & \statI & \statI & \statI & \statP & \statI & 7 & 7 & 100.0\% \\
SparseEA2 & \statP & \statI & \statI & \statI & \statI & \statP & \statI & 7 & 7 & 100.0\% \\
SPEA-R & \statP & \statI & \statI & \statI & \statI & \statD & \statP & 6 & 7 & 85.7\% \\
SSCEA & \statD & \statP & \statP & \statI & \statP & \statD & \statI & 5 & 7 & 71.4\% \\
t-DEA & \statP & \statP & \statP & \statI & \statP & \statD & \statP & 6 & 7 & 85.7\% \\
tDEA-CPBI & \statI & \statP & \statI & \statP & \statP & \statD & \statI & 6 & 7 & 85.7\% \\
TELSO & \statI & \statI & \statP & \statI & \statI & \statI & \statI & 7 & 7 & 100.0\% \\
TS-NSGA-II & \statD & \statP & \statD & \statP & \statD & \statD & \statD & 2 & 7 & 28.6\% \\
TS-SparseEA & \statP & \statI & \statP & \statI & \statP & \statD & \statD & 5 & 7 & 71.4\% \\
Two\_Arch2 & \statP & \statP & \statP & \statI & \statP & \statI & \statI & 7 & 7 & 100.0\% \\
VaEA & \statP & \statP & \statI & \statP & \statP & \statP & \statP & 7 & 7 & 100.0\% \\
WASF-GA & \statP & \statI & \statD & \statI & \statI & \statI & \statI & 6 & 7 & 85.7\% \\
WOF & \statI & \statP & \statI & \statI & \statI & \statI & \statI & 7 & 7 & 100.0\% \\
\end{longtable}
\endgroup

\clearpage
\begingroup
\scriptsize
\setlength{\tabcolsep}{2.8pt}
\renewcommand{\arraystretch}{0.96}
\begin{longtable}{l*{9}{c} r r r}
\multicolumn{13}{c}{\label{tab:fidelity-status-wfg}\normalfont\footnotesize TABLE~\thetable}\\[-0.2ex]
\multicolumn{13}{c}{\normalfont\footnotesize\scshape Optimization-Fidelity Outcome Matrix for the WFG Suite.}\\[0.5ex]
\toprule
Algorithm & WFG1 & WFG2 & WFG3 & WFG4 & WFG5 & WFG6 & WFG7 & WFG8 & WFG9 & Pass & Valid & Coverage \\
\midrule
\endfirsthead
\multicolumn{13}{c}{\normalfont\footnotesize TABLE~\thetable\ (Continued)}\\[-0.2ex]
\multicolumn{13}{c}{\normalfont\footnotesize\scshape Optimization-Fidelity Outcome Matrix for the WFG Suite.}\\[0.5ex]
\toprule
Algorithm & WFG1 & WFG2 & WFG3 & WFG4 & WFG5 & WFG6 & WFG7 & WFG8 & WFG9 & Pass & Valid & Coverage \\
\midrule
\endhead
\midrule
\multicolumn{13}{r}{Continued on next page}\\
\endfoot
\bottomrule
\endlastfoot
AGE-MOEA & \statP & \statP & \statD & \statP & \statP & \statP & \statD & \statP & \statP & 7 & 9 & 77.8\% \\
BCE-IBEA & \statP & \statP & \statP & \statP & \statP & \statP & \statP & \statP & \statP & 9 & 9 & 100.0\% \\
BCE-MOEA-D & \statP & \statP & \statP & \statP & \statP & \statP & \statP & \statP & \statP & 9 & 9 & 100.0\% \\
BiGE & \statI & \statI & \statP & \statI & \statI & \statI & \statP & \statI & \statP & 9 & 9 & 100.0\% \\
CLIA & \statP & \statP & \statP & \statP & \statP & \statP & \statP & \statP & \statP & 9 & 9 & 100.0\% \\
CMOEA-MS & \statI & \statI & \statI & \statI & \statI & \statI & \statI & \statP & \statI & 9 & 9 & 100.0\% \\
CMOPSO & \statP & \statP & \statP & \statP & \statP & \statP & \statP & \statP & \statP & 9 & 9 & 100.0\% \\
CoMMEA & \statI & \statI & \statD & \statI & \statP & \statI & \statI & \statI & \statI & 8 & 9 & 88.9\% \\
DM-MOEA & \statI & \statP & \statI & \statI & \statI & \statI & \statI & \statI & \statI & 9 & 9 & 100.0\% \\
e-MOEA & \statI & \statP & \statP & \statP & \statP & \statP & \statP & \statP & \statP & 9 & 9 & 100.0\% \\
EFR-RR & \statI & \statI & \statP & \statI & \statI & \statI & \statI & \statP & \statI & 9 & 9 & 100.0\% \\
GDE3 & \statI & \statI & \statI & \statI & \statI & \statI & \statI & \statI & \statI & 9 & 9 & 100.0\% \\
GrEA & \statP & \statP & \statP & \statP & \statP & \statP & \statP & \statP & \statP & 9 & 9 & 100.0\% \\
GWASF-GA & \statP & \statP & \statP & \statP & \statP & \statP & \statP & \statI & \statP & 9 & 9 & 100.0\% \\
KnEA & \statP & \statP & \statP & \statP & \statP & \statI & \statP & \statP & \statP & 9 & 9 & 100.0\% \\
LSMOF & \statI & \statI & \statI & \statI & \statI & \statI & \statI & \statI & \statI & 9 & 9 & 100.0\% \\
MaOEA-CSS & \statP & \statD & \statD & \statD & \statD & \statD & \statD & \statD & \statD & 1 & 9 & 11.1\% \\
MOEA-D-AWA & \statP & \statP & \statP & \statP & \statP & \statP & \statP & \statP & \statP & 9 & 9 & 100.0\% \\
MOEA-D-DCWV & \statP & \statP & \statP & \statP & \statP & \statP & \statP & \statP & \statP & 9 & 9 & 100.0\% \\
MOEA/D-DE & \statI & \statP & \statP & \statP & \statP & \statP & \statP & \statI & \statP & 9 & 9 & 100.0\% \\
MOEA-D-DRA & \statI & \statP & \statI & \statP & \statP & \statP & \statI & \statI & \statI & 9 & 9 & 100.0\% \\
MOEA-D-DU & \statI & \statP & \statI & \statI & \statI & \statP & \statP & \statP & \statP & 9 & 9 & 100.0\% \\
MOEA-D-DYTS & \statI & \statP & \statI & \statP & \statP & \statP & \statI & \statI & \statI & 9 & 9 & 100.0\% \\
MOEA-D-FRRMAB & \statI & \statP & \statI & \statP & \statI & \statI & \statP & \statP & \statP & 9 & 9 & 100.0\% \\
MOEA-D-PaS & \statI & \statP & \statD & \statP & \statD & \statP & \statP & \statD & \statD & 5 & 9 & 55.6\% \\
MOEA-D-URAW & \statP & \statP & \statP & \statP & \statP & \statP & \statP & \statP & \statP & 9 & 9 & 100.0\% \\
NSBiDiCo & \statI & \statI & \statI & \statI & \statI & \statI & \statI & \statI & \statI & 9 & 9 & 100.0\% \\
NSGA-II-SDR & \statP & \statP & \statP & \statI & \statP & \statP & \statP & \statP & \statP & 9 & 9 & 100.0\% \\
OSP-NSDE & \statP & \statI & \statI & \statI & \statI & \statI & \statI & \statI & \statI & 9 & 9 & 100.0\% \\
PESA-II & \statP & \statP & \statP & \statP & \statP & \statP & \statP & \statP & \statP & 9 & 9 & 100.0\% \\
PICEA-g & \statP & \statP & \statP & \statP & \statP & \statP & \statP & \statP & \statP & 9 & 9 & 100.0\% \\
PREA & \statP & \statP & \statP & \statI & \statI & \statP & \statI & \statP & \statI & 9 & 9 & 100.0\% \\
S-NSGA-II & \statI & \statI & \statI & \statP & \statI & \statI & \statI & \statI & \statI & 9 & 9 & 100.0\% \\
SIBEA & \statI & \statI & \statX & \statI & \statP & \statP & \statX & \statP & \statX & 6 & 6 & 100.0\% \\
SMPSO & \statP & \statP & \statP & \statP & \statP & \statP & \statI & \statP & \statP & 9 & 9 & 100.0\% \\
SparseEA & \statI & \statI & \statP & \statP & \statI & \statP & \statP & \statI & \statI & 9 & 9 & 100.0\% \\
SparseEA2 & \statI & \statP & \statI & \statI & \statI & \statP & \statI & \statI & \statI & 9 & 9 & 100.0\% \\
SPEA-R & \statI & \statP & \statP & \statI & \statP & \statI & \statP & \statP & \statP & 9 & 9 & 100.0\% \\
SSCEA & \statI & \statP & \statI & \statP & \statP & \statP & \statI & \statP & \statI & 9 & 9 & 100.0\% \\
t-DEA & \statP & \statP & \statP & \statP & \statP & \statP & \statP & \statP & \statP & 9 & 9 & 100.0\% \\
tDEA-CPBI & \statP & \statP & \statP & \statP & \statP & \statP & \statP & \statP & \statP & 9 & 9 & 100.0\% \\
TELSO & \statP & \statI & \statI & \statI & \statI & \statI & \statI & \statI & \statI & 9 & 9 & 100.0\% \\
TS-NSGA-II & \statP & \statP & \statP & \statP & \statP & \statP & \statP & \statP & \statP & 9 & 9 & 100.0\% \\
TS-SparseEA & \statP & \statP & \statP & \statP & \statP & \statI & \statP & \statI & \statP & 9 & 9 & 100.0\% \\
Two\_Arch2 & \statP & \statP & \statI & \statP & \statP & \statP & \statP & \statP & \statP & 9 & 9 & 100.0\% \\
VaEA & \statP & \statP & \statP & \statI & \statI & \statI & \statI & \statP & \statI & 9 & 9 & 100.0\% \\
WASF-GA & \statI & \statI & \statI & \statI & \statI & \statI & \statI & \statI & \statI & 9 & 9 & 100.0\% \\
WOF & \statI & \statP & \statI & \statP & \statP & \statI & \statI & \statI & \statP & 9 & 9 & 100.0\% \\
\end{longtable}
\endgroup

\clearpage
\begingroup
\scriptsize
\setlength{\tabcolsep}{2.8pt}
\renewcommand{\arraystretch}{0.96}
\begin{longtable}{l*{9}{c} r r r}
\multicolumn{13}{c}{\label{tab:fidelity-status-lsmop}\normalfont\footnotesize TABLE~\thetable}\\[-0.2ex]
\multicolumn{13}{c}{\normalfont\footnotesize\scshape Optimization-Fidelity Outcome Matrix for the LSMOP Suite.}\\[0.5ex]
\toprule
Algorithm & LSMOP1 & LSMOP2 & LSMOP3 & LSMOP4 & LSMOP5 & LSMOP6 & LSMOP7 & LSMOP8 & LSMOP9 & Pass & Valid & Coverage \\
\midrule
\endfirsthead
\multicolumn{13}{c}{\normalfont\footnotesize TABLE~\thetable\ (Continued)}\\[-0.2ex]
\multicolumn{13}{c}{\normalfont\footnotesize\scshape Optimization-Fidelity Outcome Matrix for the LSMOP Suite.}\\[0.5ex]
\toprule
Algorithm & LSMOP1 & LSMOP2 & LSMOP3 & LSMOP4 & LSMOP5 & LSMOP6 & LSMOP7 & LSMOP8 & LSMOP9 & Pass & Valid & Coverage \\
\midrule
\endhead
\midrule
\multicolumn{13}{r}{Continued on next page}\\
\endfoot
\bottomrule
\endlastfoot
AGE-MOEA & \statP & \statP & \statP & \statP & \statP & \statD & \statD & \statI & \statI & 7 & 9 & 77.8\% \\
BCE-IBEA & \statP & \statP & \statP & \statP & \statP & \statD & \statP & \statP & \statI & 8 & 9 & 88.9\% \\
BCE-MOEA-D & \statP & \statP & \statP & \statP & \statD & \statD & \statD & \statI & \statP & 6 & 9 & 66.7\% \\
BiGE & \statP & \statP & \statP & \statP & \statP & \statP & \statD & \statP & \statP & 8 & 9 & 88.9\% \\
CLIA & \statP & \statP & \statP & \statP & \statD & \statP & \statD & \statP & \statX & 6 & 8 & 75.0\% \\
CMOEA-MS & \statI & \statP & \statP & \statP & \statP & \statI & \statI & \statI & \statP & 9 & 9 & 100.0\% \\
CMOPSO & \statP & \statP & \statP & \statP & \statP & \statD & \statD & \statD & \statI & 6 & 9 & 66.7\% \\
CoMMEA & \statI & \statP & \statI & \statP & \statI & \statD & \statI & \statI & \statI & 8 & 9 & 88.9\% \\
DM-MOEA & \statP & \statP & \statI & \statP & \statI & \statI & \statI & \statI & \statI & 9 & 9 & 100.0\% \\
e-MOEA & \statP & \statP & \statP & \statI & \statI & \statP & \statD & \statP & \statP & 8 & 9 & 88.9\% \\
EFR-RR & \statI & \statI & \statI & \statI & \statI & \statP & \statP & \statP & \statP & 9 & 9 & 100.0\% \\
GDE3 & \statP & \statP & \statP & \statP & \statP & \statD & \statP & \statI & \statP & 8 & 9 & 88.9\% \\
GrEA & \statP & \statP & \statP & \statP & \statP & \statD & \statI & \statD & \statP & 7 & 9 & 77.8\% \\
GWASF-GA & \statI & \statI & \statI & \statI & \statI & \statP & \statP & \statI & \statP & 9 & 9 & 100.0\% \\
KnEA & \statP & \statI & \statP & \statI & \statP & \statD & \statP & \statP & \statP & 8 & 9 & 88.9\% \\
LSMOF & \statI & \statP & \statI & \statP & \statD & \statD & \statP & \statP & \statI & 7 & 9 & 77.8\% \\
MaOEA-CSS & \statI & \statD & \statI & \statP & \statI & \statI & \statP & \statP & \statI & 8 & 9 & 88.9\% \\
MOEA-D-AWA & \statI & \statP & \statI & \statP & \statP & \statD & \statP & \statI & \statP & 8 & 9 & 88.9\% \\
MOEA-D-DCWV & \statI & \statP & \statI & \statP & \statI & \statP & \statI & \statP & \statI & 9 & 9 & 100.0\% \\
MOEA/D-DE & \statI & \statP & \statI & \statP & \statI & \statI & \statI & \statP & \statP & 9 & 9 & 100.0\% \\
MOEA-D-DRA & \statI & \statI & \statI & \statI & \statI & \statI & \statI & \statI & \statI & 9 & 9 & 100.0\% \\
MOEA-D-DU & \statP & \statP & \statP & \statP & \statP & \statP & \statP & \statP & \statI & 9 & 9 & 100.0\% \\
MOEA-D-DYTS & \statD & \statP & \statD & \statP & \statD & \statD & \statP & \statP & \statP & 5 & 9 & 55.6\% \\
MOEA-D-FRRMAB & \statI & \statP & \statI & \statP & \statI & \statI & \statI & \statP & \statP & 9 & 9 & 100.0\% \\
MOEA-D-PaS & \statP & \statD & \statI & \statP & \statI & \statI & \statI & \statI & \statI & 8 & 9 & 88.9\% \\
MOEA-D-URAW & \statI & \statP & \statP & \statP & \statI & \statI & \statI & \statI & \statI & 9 & 9 & 100.0\% \\
NSBiDiCo & \statP & \statI & \statI & \statI & \statP & \statP & \statP & \statI & \statI & 9 & 9 & 100.0\% \\
NSGA-II-SDR & \statP & \statP & \statP & \statP & \statP & \statD & \statD & \statP & \statP & 7 & 9 & 77.8\% \\
OSP-NSDE & \statD & \statI & \statD & \statI & \statD & \statD & \statD & \statD & \statD & 2 & 9 & 22.2\% \\
PESA-II & \statP & \statP & \statP & \statP & \statP & \statD & \statD & \statP & \statI & 7 & 9 & 77.8\% \\
PICEA-g & \statP & \statP & \statI & \statP & \statP & \statI & \statD & \statD & \statP & 7 & 9 & 77.8\% \\
PREA & \statP & \statP & \statP & \statP & \statP & \statD & \statD & \statP & \statP & 7 & 9 & 77.8\% \\
S-NSGA-II & \statP & \statP & \statP & \statI & \statI & \statI & \statI & \statI & \statI & 9 & 9 & 100.0\% \\
SIBEA & \statI & \statX & \statI & \statX & \statP & \statD & \statI & \statI & \statP & 6 & 7 & 85.7\% \\
SMPSO & \statI & \statP & \statI & \statP & \statI & \statI & \statD & \statP & \statI & 8 & 9 & 88.9\% \\
SparseEA & \statD & \statP & \statD & \statP & \statD & \statD & \statP & \statI & \statI & 5 & 9 & 55.6\% \\
SparseEA2 & \statP & \statP & \statP & \statI & \statP & \statD & \statI & \statI & \statI & 8 & 9 & 88.9\% \\
SPEA-R & \statP & \statD & \statI & \statP & \statP & \statD & \statD & \statP & \statI & 6 & 9 & 66.7\% \\
SSCEA & \statI & \statI & \statI & \statI & \statI & \statI & \statI & \statP & \statI & 9 & 9 & 100.0\% \\
t-DEA & \statP & \statP & \statP & \statP & \statP & \statP & \statP & \statP & \statP & 9 & 9 & 100.0\% \\
tDEA-CPBI & \statI & \statP & \statI & \statP & \statI & \statI & \statD & \statP & \statP & 8 & 9 & 88.9\% \\
TELSO & \statX & \statI & \statX & \statI & \statP & \statP & \statI & \statI & \statI & 7 & 7 & 100.0\% \\
TS-NSGA-II & \statP & \statP & \statP & \statP & \statP & \statD & \statD & \statP & \statP & 7 & 9 & 77.8\% \\
TS-SparseEA & \statD & \statP & \statD & \statI & \statD & \statD & \statP & \statP & \statD & 4 & 9 & 44.4\% \\
Two\_Arch2 & \statI & \statP & \statI & \statI & \statP & \statI & \statI & \statP & \statI & 9 & 9 & 100.0\% \\
VaEA & \statI & \statP & \statI & \statP & \statI & \statI & \statD & \statP & \statP & 8 & 9 & 88.9\% \\
WASF-GA & \statP & \statI & \statP & \statI & \statP & \statD & \statP & \statP & \statD & 7 & 9 & 77.8\% \\
WOF & \statP & \statP & \statP & \statP & \statD & \statD & \statP & \statP & \statP & 7 & 9 & 77.8\% \\
\end{longtable}
\endgroup

\clearpage
\begingroup
\scriptsize
\setlength{\tabcolsep}{4.0pt}
\renewcommand{\arraystretch}{0.96}
\begin{longtable}{l*{8}{c}}
\multicolumn{9}{c}{\label{tab:fidelity-status-maf}\normalfont\footnotesize TABLE~\thetable}\\[-0.2ex]
\multicolumn{9}{c}{\normalfont\footnotesize\scshape Optimization-Fidelity Outcome Matrix for the MaF Suite (MaF1--MaF8; Part I of II).}\\[0.5ex]
\toprule
Algorithm & MaF1 & MaF2 & MaF3 & MaF4 & MaF5 & MaF6 & MaF7 & MaF8 \\
\midrule
\endfirsthead
\multicolumn{9}{c}{\normalfont\footnotesize TABLE~\thetable\ (Continued)}\\[-0.2ex]
\multicolumn{9}{c}{\normalfont\footnotesize\scshape Optimization-Fidelity Outcome Matrix for the MaF Suite (MaF1--MaF8; Part I of II).}\\[0.5ex]
\toprule
Algorithm & MaF1 & MaF2 & MaF3 & MaF4 & MaF5 & MaF6 & MaF7 & MaF8 \\
\midrule
\endhead
\midrule
\multicolumn{9}{r}{Continued on next page}\\
\endfoot
\bottomrule
\endlastfoot
AGE-MOEA & \statP & \statP & \statD & \statP & \statD & \statP & \statP & \statP \\
BCE-IBEA & \statP & \statP & \statI & \statP & \statI & \statD & \statP & \statP \\
BCE-MOEA-D & \statP & \statP & \statD & \statP & \statP & \statD & \statP & \statP \\
BiGE & \statI & \statP & \statD & \statI & \statI & \statD & \statP & \statI \\
CLIA & \statP & \statP & \statD & \statP & \statI & \statP & \statP & \statP \\
CMOEA-MS & \statP & \statI & \statD & \statP & \statI & \statD & \statD & \statI \\
CMOPSO & \statP & \statI & \statI & \statP & \statP & \statP & \statI & \statP \\
CoMMEA & \statI & \statI & \statI & \statI & \statD & \statD & \statI & \statP \\
DM-MOEA & \statI & \statI & \statI & \statI & \statI & \statD & \statP & \statI \\
e-MOEA & \statP & \statP & \statI & \statP & \statI & \statP & \statI & \statP \\
EFR-RR & \statI & \statP & \statI & \statI & \statI & \statP & \statP & \statI \\
GDE3 & \statI & \statI & \statI & \statP & \statP & \statD & \statI & \statI \\
GrEA & \statP & \statP & \statP & \statP & \statP & \statD & \statP & \statP \\
GWASF-GA & \statI & \statI & \statD & \statP & \statI & \statD & \statD & \statD \\
KnEA & \statP & \statP & \statD & \statP & \statI & \statD & \statD & \statI \\
LSMOF & \statI & \statI & \statI & \statP & \statI & \statI & \statI & \statP \\
MaOEA-CSS & \statD & \statD & \statP & \statP & \statD & \statP & \statD & \statD \\
MOEA-D-AWA & \statP & \statP & \statD & \statP & \statI & \statP & \statP & \statP \\
MOEA-D-DCWV & \statP & \statP & \statD & \statP & \statP & \statD & \statI & \statP \\
MOEA/D-DE & \statP & \statP & \statP & \statP & \statD & \statP & \statP & \statP \\
MOEA-D-DRA & \statI & \statI & \statI & \statI & \statI & \statP & \statP & \statI \\
MOEA-D-DU & \statP & \statI & \statP & \statP & \statI & \statP & \statI & \statI \\
MOEA-D-DYTS & \statI & \statI & \statI & \statP & \statI & \statP & \statI & \statP \\
MOEA-D-FRRMAB & \statI & \statI & \statI & \statI & \statP & \statP & \statP & \statI \\
MOEA-D-PaS & \statP & \statP & \statI & \statP & \statP & \statD & \statI & \statI \\
MOEA-D-URAW & \statP & \statP & \statI & \statI & \statI & \statP & \statP & \statP \\
NSBiDiCo & \statI & \statI & \statI & \statI & \statI & \statD & \statI & \statD \\
NSGA-II-SDR & \statI & \statP & \statD & \statD & \statI & \statD & \statD & \statP \\
OSP-NSDE & \statI & \statI & \statD & \statD & \statX & \statD & \statI & \statP \\
PESA-II & \statP & \statP & \statP & \statI & \statP & \statD & \statI & \statP \\
PICEA-g & \statP & \statP & \statD & \statP & \statP & \statP & \statP & \statP \\
PREA & \statI & \statP & \statP & \statP & \statI & \statD & \statP & \statP \\
S-NSGA-II & \statI & \statI & \statI & \statI & \statP & \statD & \statP & \statI \\
SIBEA & \statP & \statX & \statP & \statP & \statI & \statD & \statP & \statI \\
SMPSO & \statI & \statI & \statD & \statD & \statP & \statD & \statP & \statP \\
SparseEA & \statP & \statI & \statI & \statI & \statI & \statD & \statI & \statD \\
SparseEA2 & \statI & \statI & \statI & \statI & \statI & \statD & \statP & \statP \\
SPEA-R & \statP & \statP & \statP & \statI & \statP & \statD & \statP & \statI \\
SSCEA & \statP & \statP & \statD & \statP & \statI & \statP & \statI & \statI \\
t-DEA & \statP & \statP & \statD & \statP & \statI & \statP & \statP & \statI \\
tDEA-CPBI & \statI & \statP & \statP & \statI & \statI & \statD & \statI & \statI \\
TELSO & \statI & \statI & \statI & \statP & \statI & \statI & \statI & \statI \\
TS-NSGA-II & \statP & \statP & \statD & \statP & \statP & \statD & \statD & \statI \\
TS-SparseEA & \statP & \statI & \statI & \statI & \statP & \statD & \statD & \statI \\
Two\_Arch2 & \statP & \statP & \statD & \statP & \statI & \statP & \statI & \statI \\
VaEA & \statP & \statI & \statI & \statI & \statP & \statD & \statI & \statP \\
WASF-GA & \statI & \statI & \statD & \statP & \statI & \statP & \statI & \statD \\
WOF & \statI & \statP & \statI & \statI & \statI & \statD & \statI & \statI \\
\end{longtable}

\clearpage
\scriptsize
\setlength{\tabcolsep}{4.0pt}
\renewcommand{\arraystretch}{0.96}
\begin{longtable}{l*{7}{c} r r r}
\multicolumn{11}{c}{\label{tab:fidelity-status-maf-2}\normalfont\footnotesize TABLE~\thetable}\\[-0.2ex]
\multicolumn{11}{c}{\normalfont\footnotesize\scshape Optimization-Fidelity Outcome Matrix for the MaF Suite (MaF9--MaF15; Part II of II).}\\[0.5ex]
\toprule
Algorithm & MaF9 & MaF10 & MaF11 & MaF12 & MaF13 & MaF14 & MaF15 & Pass & Valid & Coverage \\
\midrule
\endfirsthead
\multicolumn{11}{c}{\normalfont\footnotesize TABLE~\thetable\ (Continued)}\\[-0.2ex]
\multicolumn{11}{c}{\normalfont\footnotesize\scshape Optimization-Fidelity Outcome Matrix for the MaF Suite (MaF9--MaF15; Part II of II).}\\[0.5ex]
\toprule
Algorithm & MaF9 & MaF10 & MaF11 & MaF12 & MaF13 & MaF14 & MaF15 & Pass & Valid & Coverage \\
\midrule
\endhead
\midrule
\multicolumn{11}{r}{Continued on next page}\\
\endfoot
\bottomrule
\endlastfoot
AGE-MOEA & \statP & \statP & \statP & \statP & \statP & \statP & \statP & 13 & 15 & 86.7\% \\
BCE-IBEA & \statD & \statP & \statP & \statP & \statP & \statP & \statP & 13 & 15 & 86.7\% \\
BCE-MOEA-D & \statD & \statP & \statP & \statP & \statP & \statD & \statD & 10 & 15 & 66.7\% \\
BiGE & \statP & \statI & \statI & \statI & \statP & \statP & \statP & 13 & 15 & 86.7\% \\
CLIA & \statP & \statP & \statP & \statP & \statP & \statD & \statD & 12 & 15 & 80.0\% \\
CMOEA-MS & \statI & \statI & \statI & \statI & \statP & \statP & \statP & 12 & 15 & 80.0\% \\
CMOPSO & \statI & \statP & \statP & \statP & \statP & \statP & \statP & 15 & 15 & 100.0\% \\
CoMMEA & \statI & \statI & \statI & \statI & \statI & \statI & \statI & 13 & 15 & 86.7\% \\
DM-MOEA & \statI & \statI & \statP & \statI & \statP & \statI & \statP & 14 & 15 & 93.3\% \\
e-MOEA & \statP & \statI & \statP & \statP & \statI & \statP & \statI & 15 & 15 & 100.0\% \\
EFR-RR & \statI & \statI & \statI & \statI & \statP & \statP & \statP & 15 & 15 & 100.0\% \\
GDE3 & \statD & \statI & \statI & \statI & \statI & \statI & \statI & 13 & 15 & 86.7\% \\
GrEA & \statD & \statP & \statP & \statP & \statP & \statP & \statD & 12 & 15 & 80.0\% \\
GWASF-GA & \statD & \statP & \statP & \statP & \statI & \statI & \statI & 10 & 15 & 66.7\% \\
KnEA & \statI & \statP & \statP & \statP & \statI & \statP & \statP & 12 & 15 & 80.0\% \\
LSMOF & \statI & \statI & \statI & \statI & \statI & \statI & \statP & 15 & 15 & 100.0\% \\
MaOEA-CSS & \statD & \statP & \statD & \statD & \statD & \statI & \statP & 6 & 15 & 40.0\% \\
MOEA-D-AWA & \statP & \statP & \statP & \statI & \statP & \statP & \statP & 14 & 15 & 93.3\% \\
MOEA-D-DCWV & \statP & \statP & \statP & \statP & \statP & \statI & \statP & 13 & 15 & 86.7\% \\
MOEA/D-DE & \statP & \statI & \statP & \statP & \statP & \statI & \statP & 14 & 15 & 93.3\% \\
MOEA-D-DRA & \statP & \statI & \statP & \statP & \statI & \statI & \statI & 15 & 15 & 100.0\% \\
MOEA-D-DU & \statP & \statI & \statP & \statI & \statP & \statP & \statP & 15 & 15 & 100.0\% \\
MOEA-D-DYTS & \statP & \statI & \statP & \statI & \statI & \statP & \statP & 15 & 15 & 100.0\% \\
MOEA-D-FRRMAB & \statI & \statI & \statP & \statP & \statI & \statI & \statP & 15 & 15 & 100.0\% \\
MOEA-D-PaS & \statI & \statI & \statP & \statD & \statP & \statI & \statP & 13 & 15 & 86.7\% \\
MOEA-D-URAW & \statI & \statP & \statP & \statP & \statP & \statI & \statP & 15 & 15 & 100.0\% \\
NSBiDiCo & \statI & \statI & \statI & \statI & \statI & \statI & \statI & 13 & 15 & 86.7\% \\
NSGA-II-SDR & \statD & \statP & \statP & \statP & \statP & \statP & \statP & 10 & 15 & 66.7\% \\
OSP-NSDE & \statD & \statP & \statI & \statI & \statI & \statD & \statP & 9 & 14 & 64.3\% \\
PESA-II & \statD & \statP & \statP & \statP & \statP & \statP & \statP & 13 & 15 & 86.7\% \\
PICEA-g & \statP & \statP & \statP & \statP & \statD & \statP & \statP & 13 & 15 & 86.7\% \\
PREA & \statP & \statP & \statP & \statI & \statP & \statP & \statP & 14 & 15 & 93.3\% \\
S-NSGA-II & \statI & \statI & \statI & \statP & \statI & \statI & \statI & 14 & 15 & 93.3\% \\
SIBEA & \statP & \statI & \statP & \statX & \statP & \statP & \statP & 12 & 13 & 92.3\% \\
SMPSO & \statI & \statP & \statP & \statP & \statP & \statI & \statI & 12 & 15 & 80.0\% \\
SparseEA & \statP & \statI & \statI & \statI & \statP & \statI & \statP & 13 & 15 & 86.7\% \\
SparseEA2 & \statP & \statI & \statP & \statI & \statI & \statP & \statP & 14 & 15 & 93.3\% \\
SPEA-R & \statI & \statI & \statP & \statP & \statP & \statI & \statP & 14 & 15 & 93.3\% \\
SSCEA & \statI & \statI & \statP & \statI & \statP & \statI & \statI & 14 & 15 & 93.3\% \\
t-DEA & \statI & \statP & \statP & \statP & \statP & \statP & \statP & 14 & 15 & 93.3\% \\
tDEA-CPBI & \statP & \statP & \statP & \statP & \statP & \statI & \statI & 14 & 15 & 93.3\% \\
TELSO & \statP & \statP & \statI & \statI & \statI & \statX & \statI & 14 & 14 & 100.0\% \\
TS-NSGA-II & \statP & \statP & \statP & \statP & \statP & \statP & \statP & 12 & 15 & 80.0\% \\
TS-SparseEA & \statI & \statP & \statP & \statP & \statI & \statD & \statD & 11 & 15 & 73.3\% \\
Two\_Arch2 & \statP & \statP & \statP & \statP & \statD & \statI & \statI & 13 & 15 & 86.7\% \\
VaEA & \statP & \statP & \statP & \statI & \statP & \statI & \statI & 14 & 15 & 93.3\% \\
WASF-GA & \statD & \statI & \statI & \statI & \statI & \statP & \statP & 12 & 15 & 80.0\% \\
WOF & \statI & \statI & \statP & \statP & \statP & \statP & \statP & 14 & 15 & 93.3\% \\
\end{longtable}
\endgroup

\FloatBarrier
\section{Supplementary Scaling Results}

This section reports the complete scaling results for the 48 tensorized implementations. Runtime scalability is evaluated on DTLZ3 with three objectives. EvoX denotes the tensorized GPU implementation and PlatEMO denotes the MATLAB CPU baseline. Population scaling varies $N$ over $\{256,512,1024,2048,4096,8192,16384\}$ while fixing the decision dimension at $D=12$. Dimension scaling fixes $N=1000$ and varies $D$ over $\{1024,2048,4096,8192,16384,32768,65536\}$. The EvoX workflow step is compiled using \texttt{torch.compile}, and initialization and compilation costs are excluded. Runtime is reported in milliseconds per generation using the average over 100 generations. A three-hour PlatEMO timeout corresponds to 108000 ms/gen. When the corresponding EvoX run completes, the timeout-derived speedup is reported as a lower bound rather than a measured value. EvoX OOM events terminate without a valid runtime and are reported separately.

\subsection{Aggregate Results and Selected Cases}

Across all completed pairs, population scaling contains 289 measured comparisons and dimension scaling contains 276. The corresponding median measured speedups are $22.6\times$ and $80.2\times$, the geometric means are $29.9\times$ and $71.5\times$, and the interquartile ranges are $4.5$--$163.5\times$ and $8.4$--$404.9\times$, respectively.

PlatEMO reaches the time limit in 30 population-scaling and 42 dimension-scaling cases. Among these, 26 and 34 cases, respectively, have valid EvoX runtimes and contribute lower-bound speedups to the scale-wise distributions. The remaining 4 and 8 cases coincide with EvoX OOM events and therefore have no speedup value. Across all settings, EvoX encounters 6 population-scaling and 11 dimension-scaling OOM events. PlatEMO encounters 15 and 17 execution errors, respectively, together with one additional OOM event under dimension scaling.

\begin{figure}[H]
\centering
\includegraphics[width=0.6\textwidth]{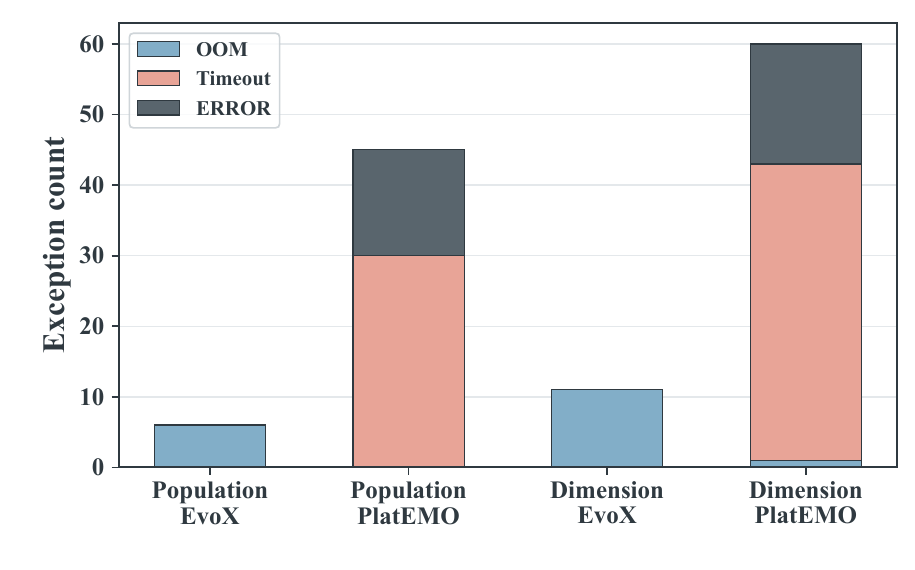}
\caption{Runtime-scaling exception counts by implementation and scaling axis, separated into timeout, execution-error, and out-of-memory cases.}
\label{fig:supp_scaling_exception_summary}
\end{figure}

Extreme scaling cases are retained for completeness rather than used to characterize typical acceleration. Among completed pairs, the largest measured speedups are $37339.0\times$ for MOEA/D-DE at $N=16384$ and $23467.4\times$ for DM-MOEA at $D=8192$. Among PlatEMO timeout cases with valid EvoX runtimes, the largest reported lower bounds are $29900.3\times$ for MaOEA-CSS at $N=1024$ and $105365.9\times$ for MaOEA-CSS at both $D=2048$ and $D=4096$. These extrema are shown in the complete algorithm-level curves below, while the median and interquartile range characterize the central distribution. Fig.~\ref{fig:supp_scaling_exception_summary} summarizes the exception counts, and Table~\ref{tab:supp_scaling_exceptions} provides the corresponding records.

\subsection{Complete Curves and Exception Records}

\begin{figure}[p]
\centering
\subfloat[AGE-MOEA: varying $N$]{\includegraphics[width=0.22\textwidth]{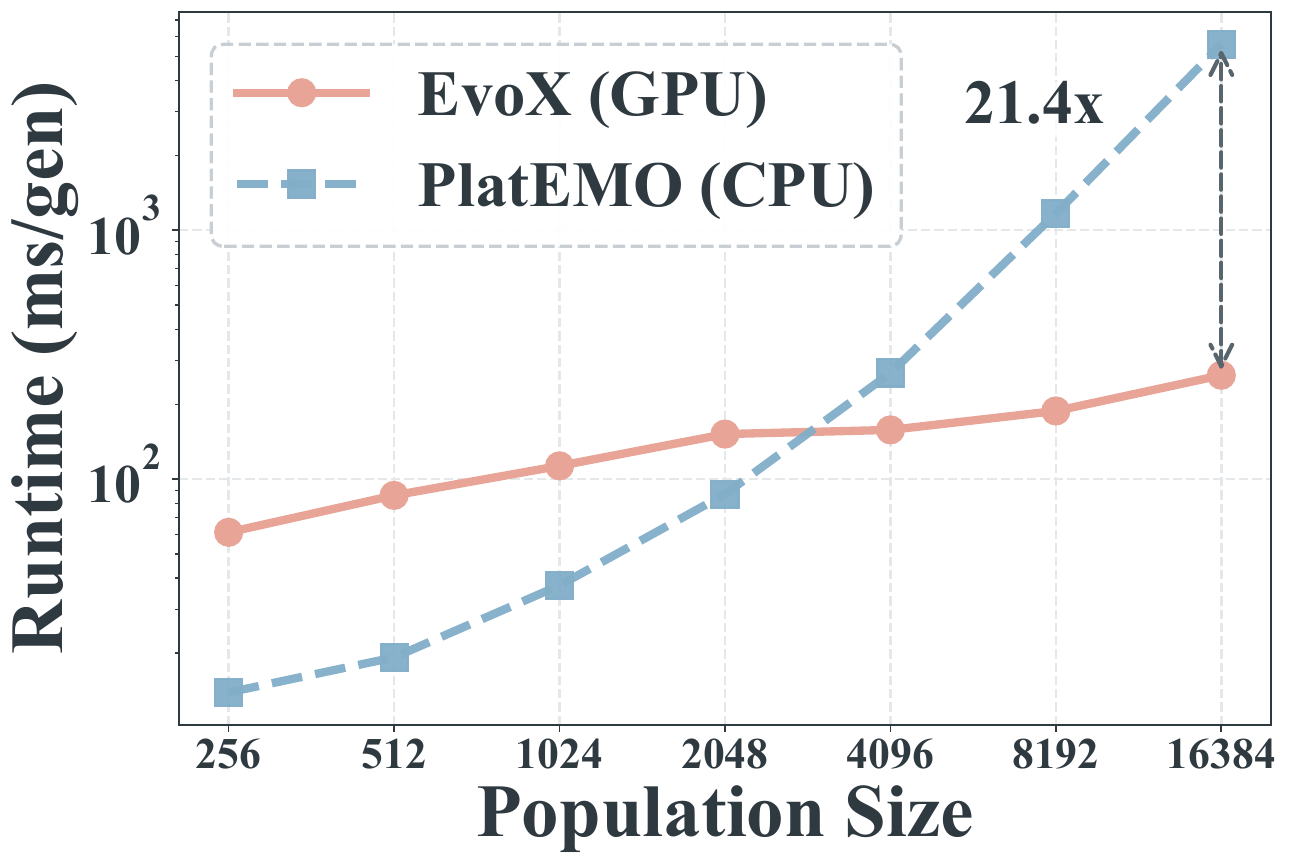}}
\hfill
\subfloat[AGE-MOEA: varying $D$]{\includegraphics[width=0.22\textwidth]{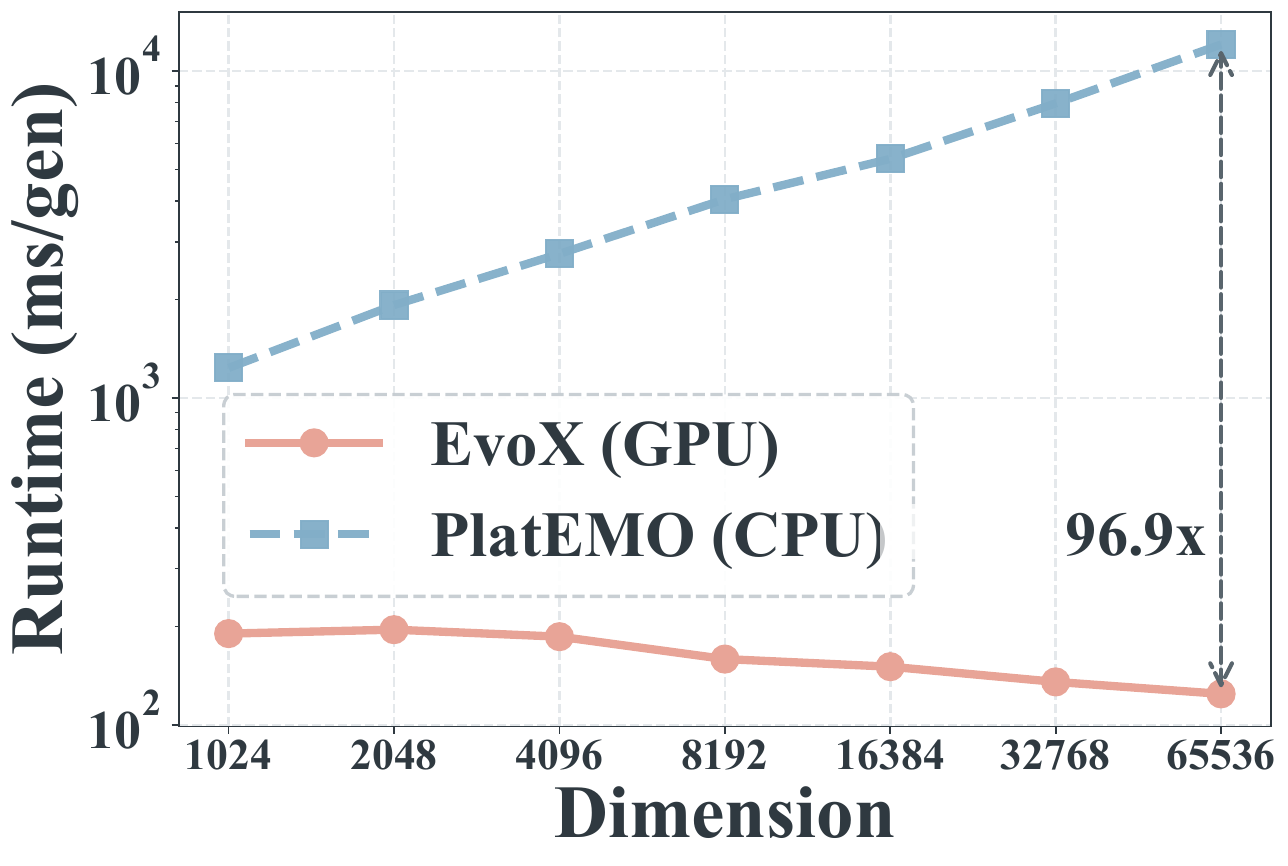}}
\hfill
\subfloat[BCE-IBEA: varying $N$]{\includegraphics[width=0.22\textwidth]{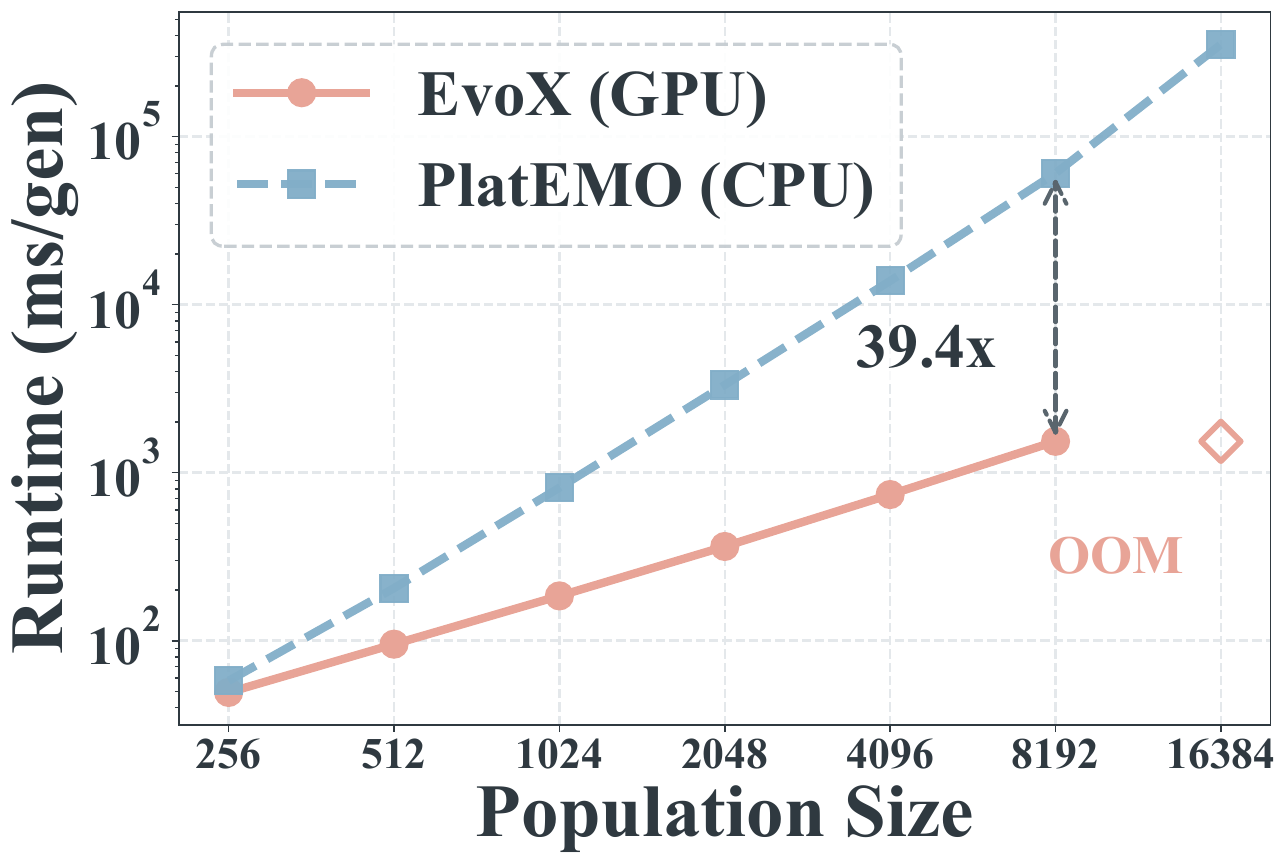}}
\hfill
\subfloat[BCE-IBEA: varying $D$]{\includegraphics[width=0.22\textwidth]{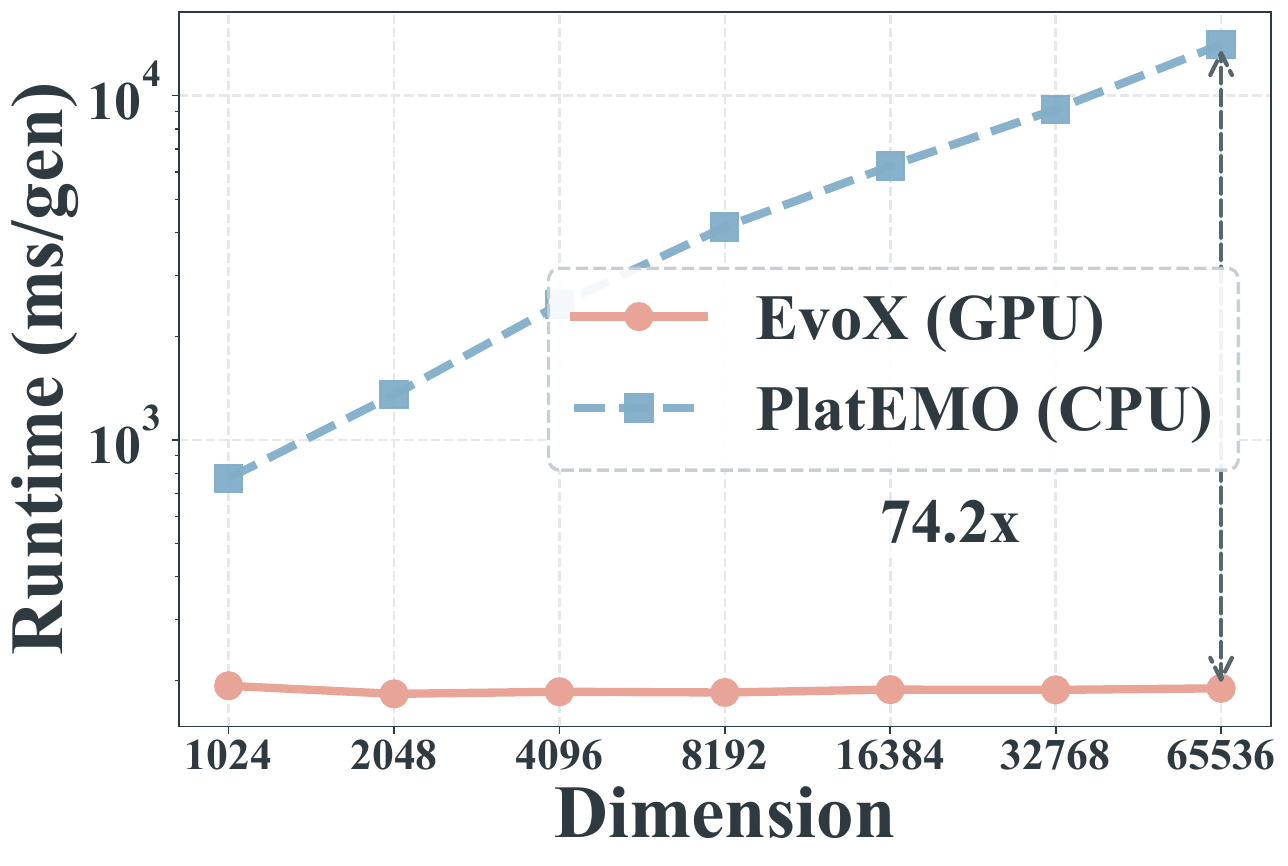}}
\\[-1mm]
\subfloat[BCE-MOEA-D: varying $N$]{\includegraphics[width=0.22\textwidth]{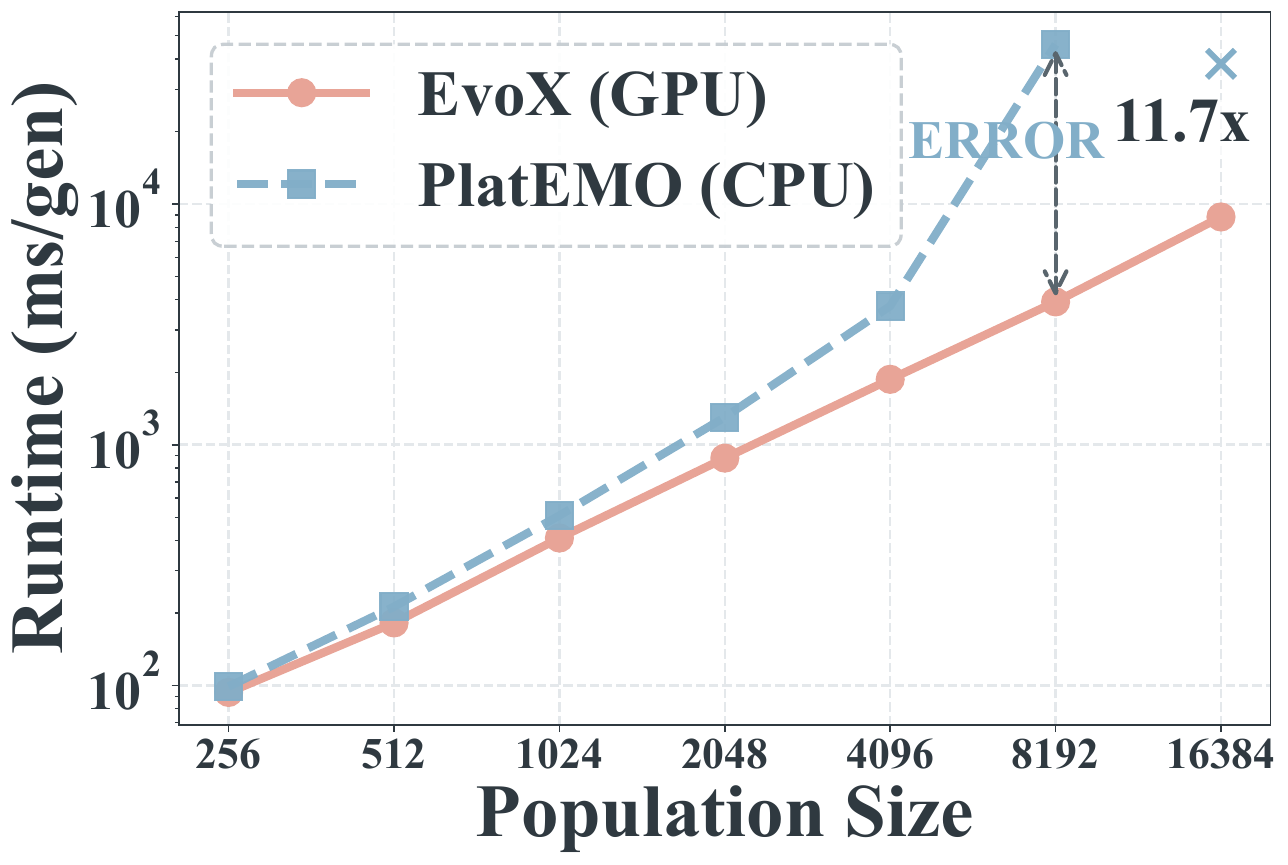}}
\hfill
\subfloat[BCE-MOEA-D: varying $D$]{\includegraphics[width=0.22\textwidth]{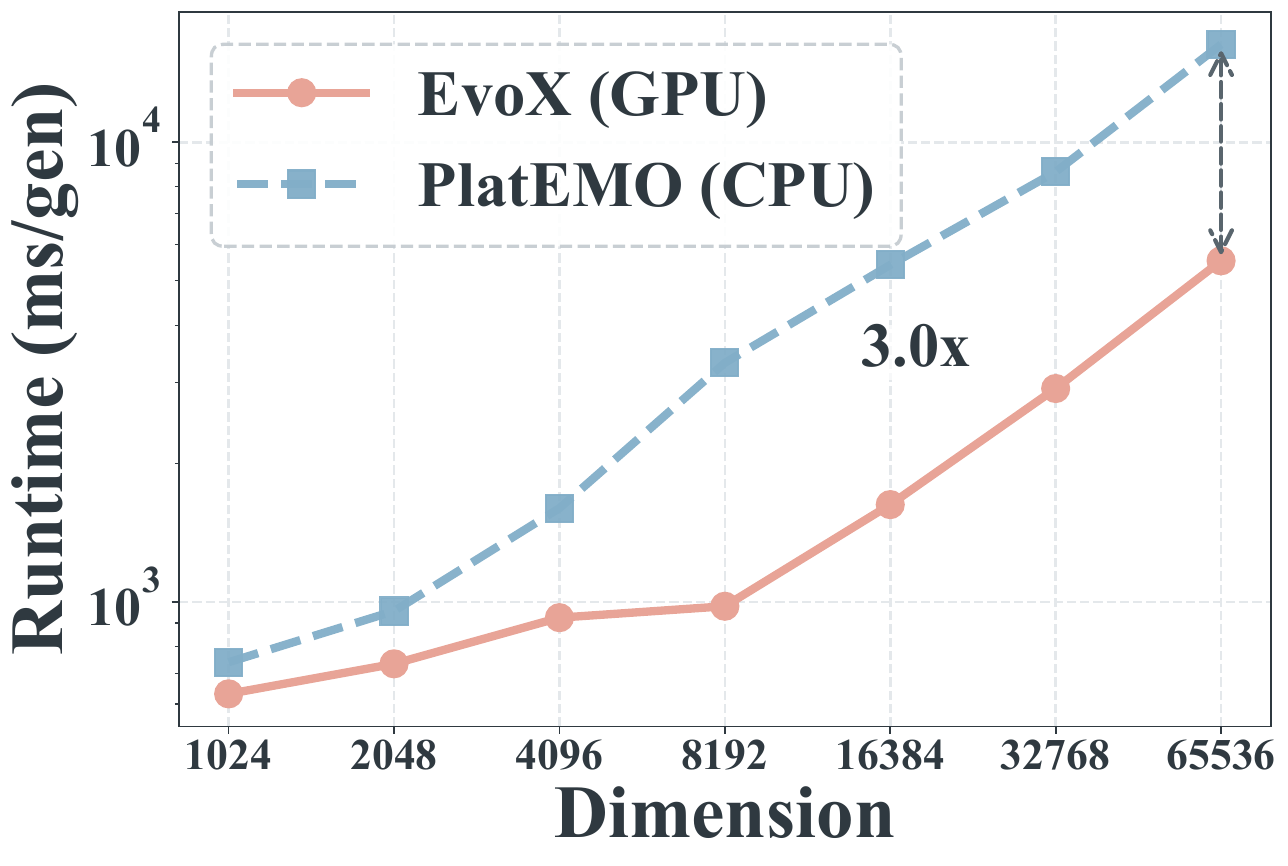}}
\hfill
\subfloat[BiGE: varying $N$]{\includegraphics[width=0.22\textwidth]{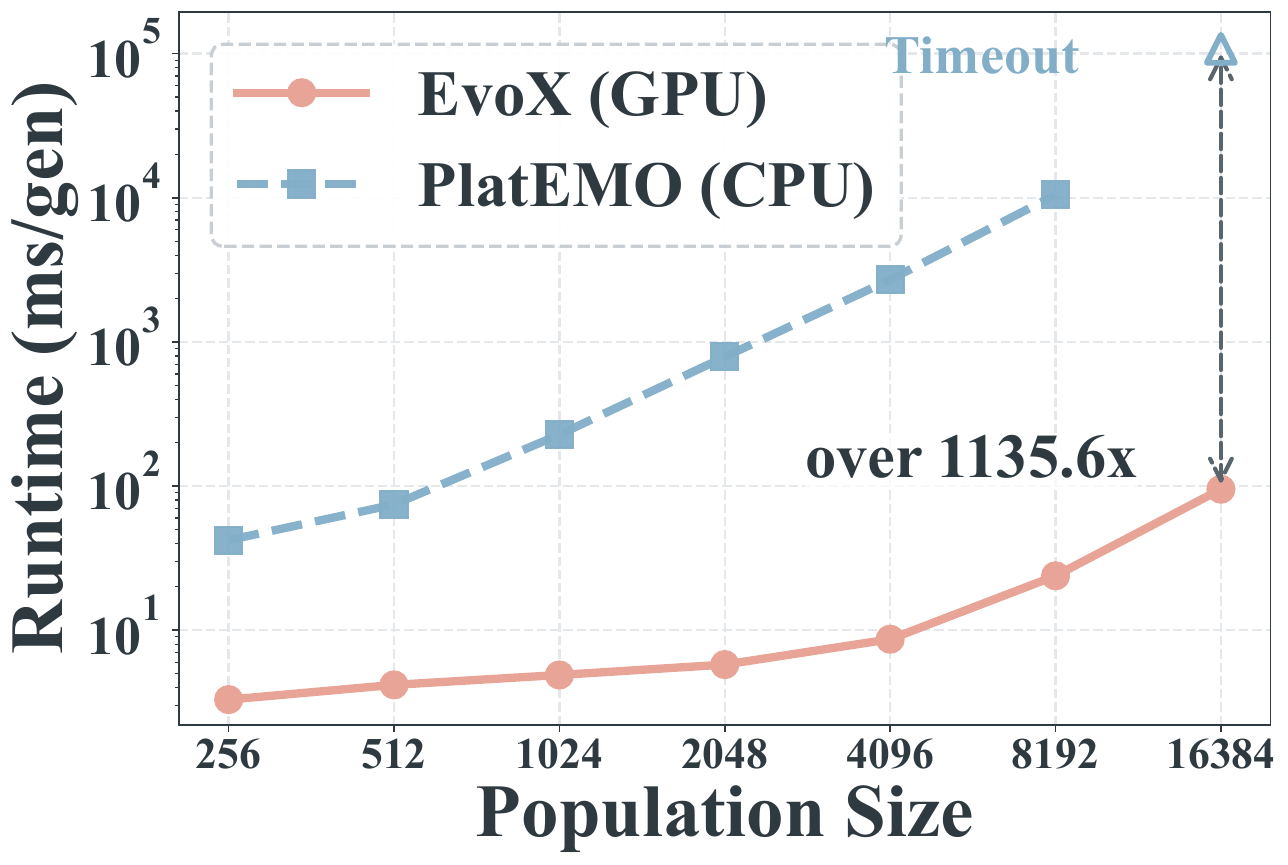}}
\hfill
\subfloat[BiGE: varying $D$]{\includegraphics[width=0.22\textwidth]{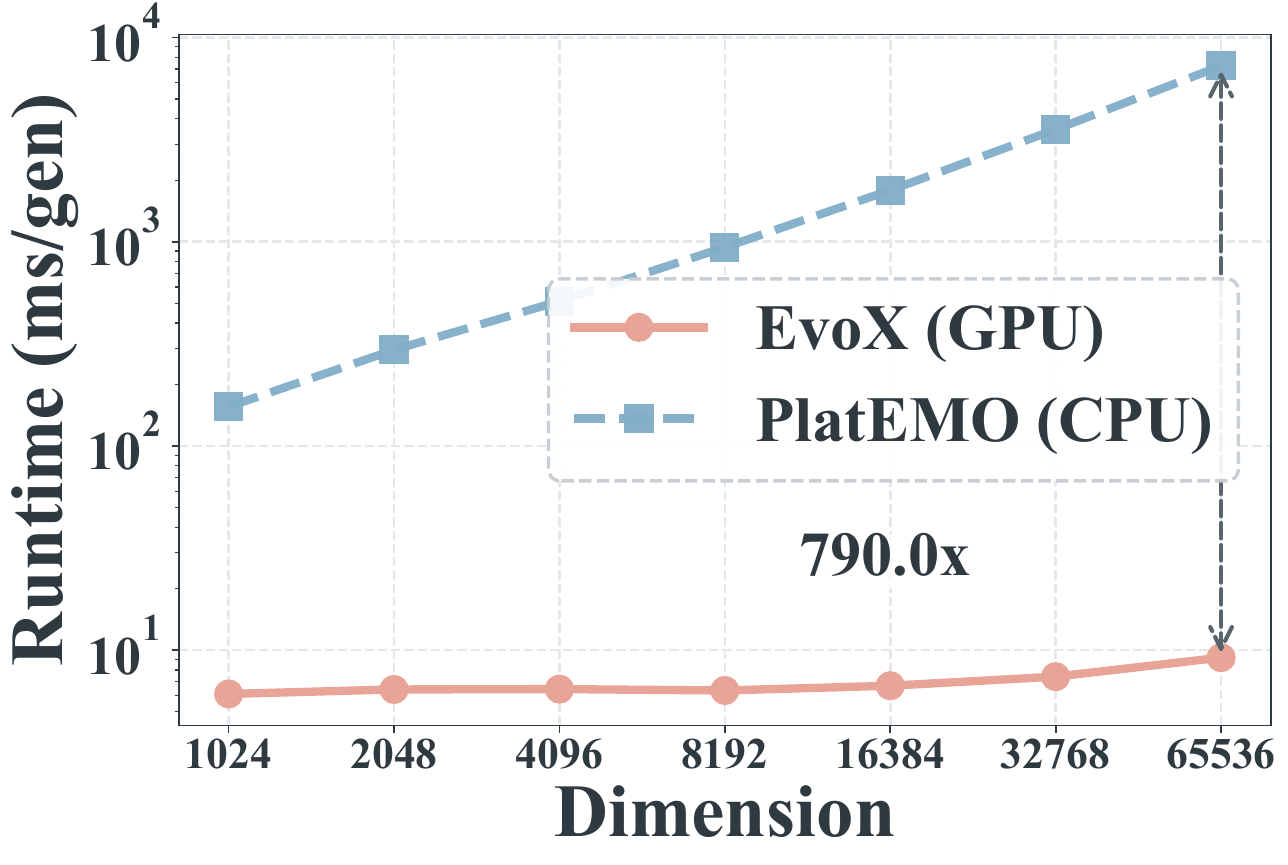}}
\\[-1mm]
\subfloat[CLIA: varying $N$]{\includegraphics[width=0.22\textwidth]{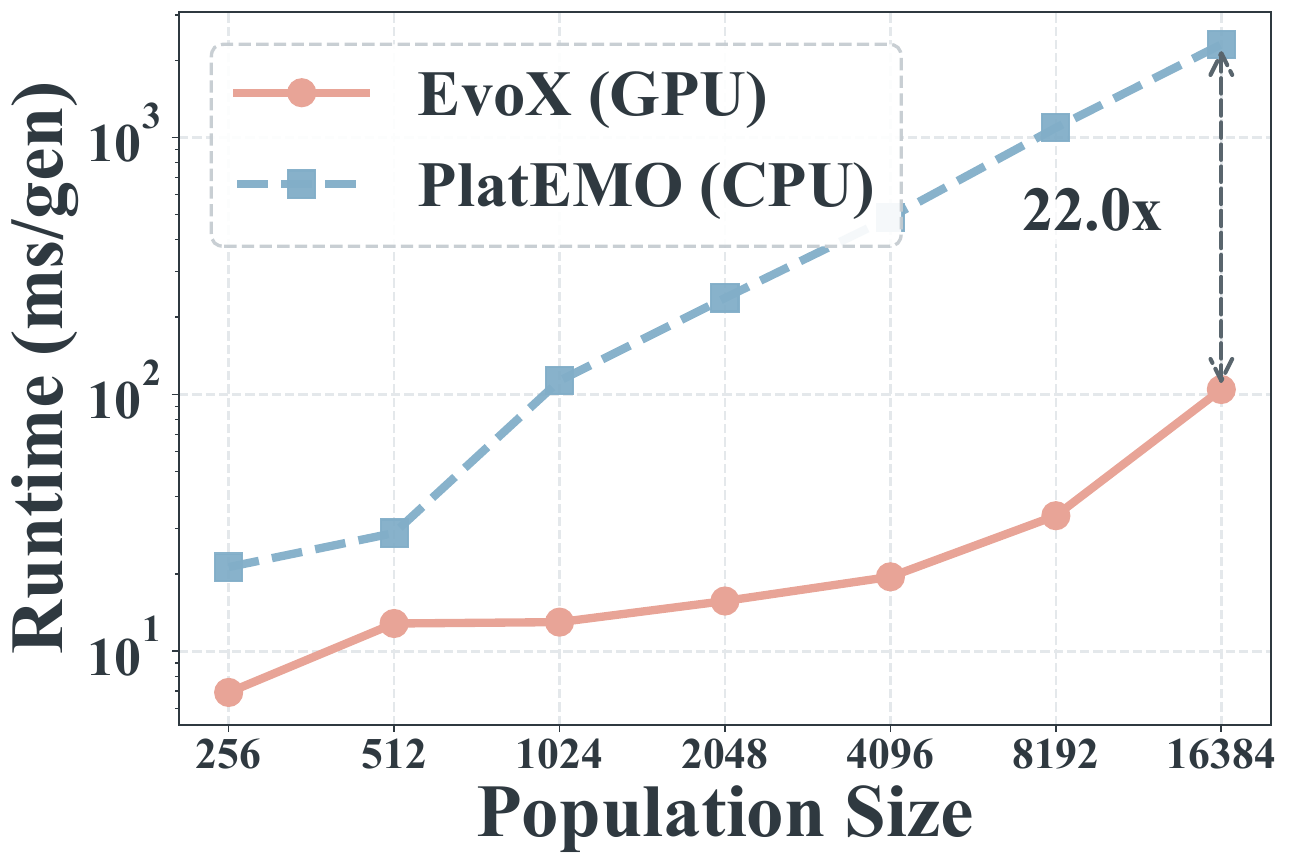}}
\hfill
\subfloat[CLIA: varying $D$]{\includegraphics[width=0.22\textwidth]{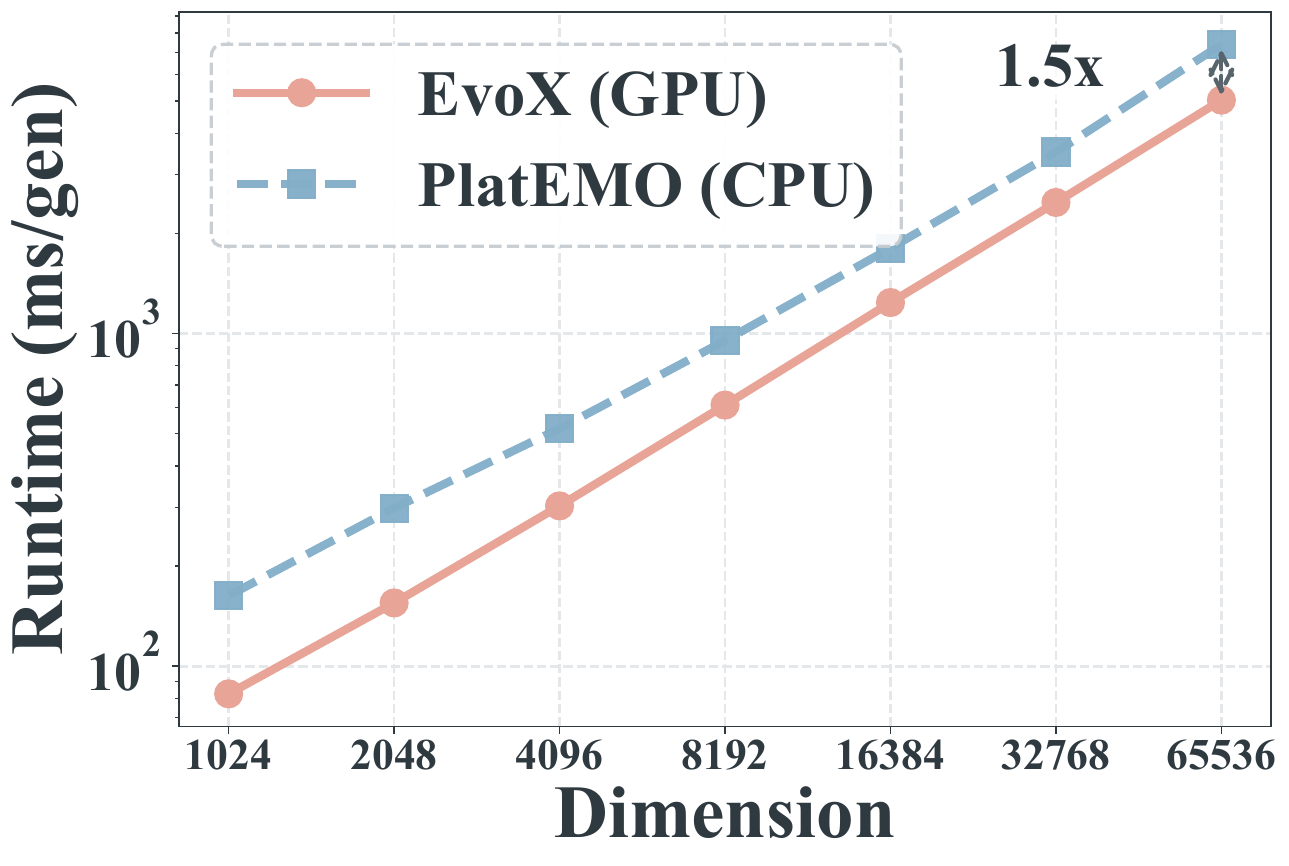}}
\hfill
\subfloat[CMOEA-MS: varying $N$]{\includegraphics[width=0.22\textwidth]{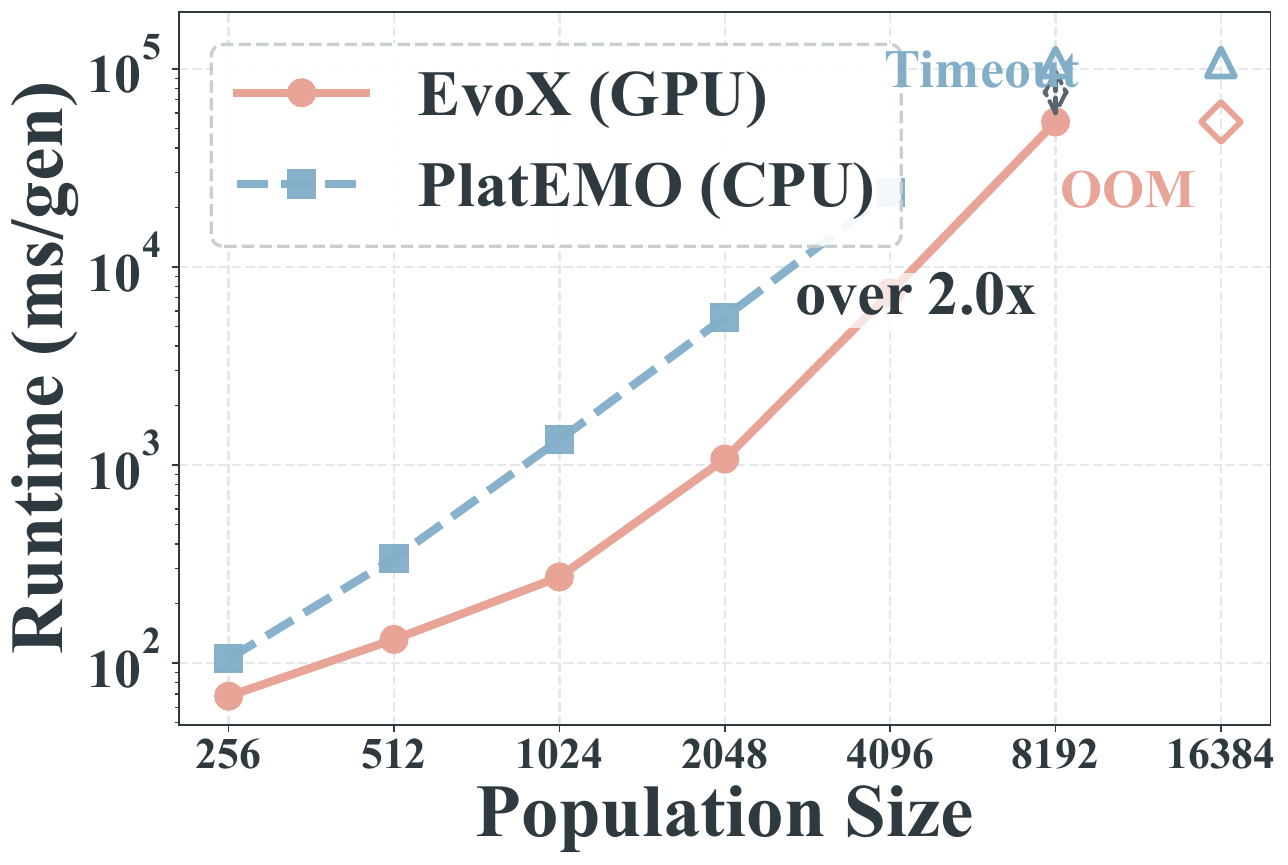}}
\hfill
\subfloat[CMOEA-MS: varying $D$]{\includegraphics[width=0.22\textwidth]{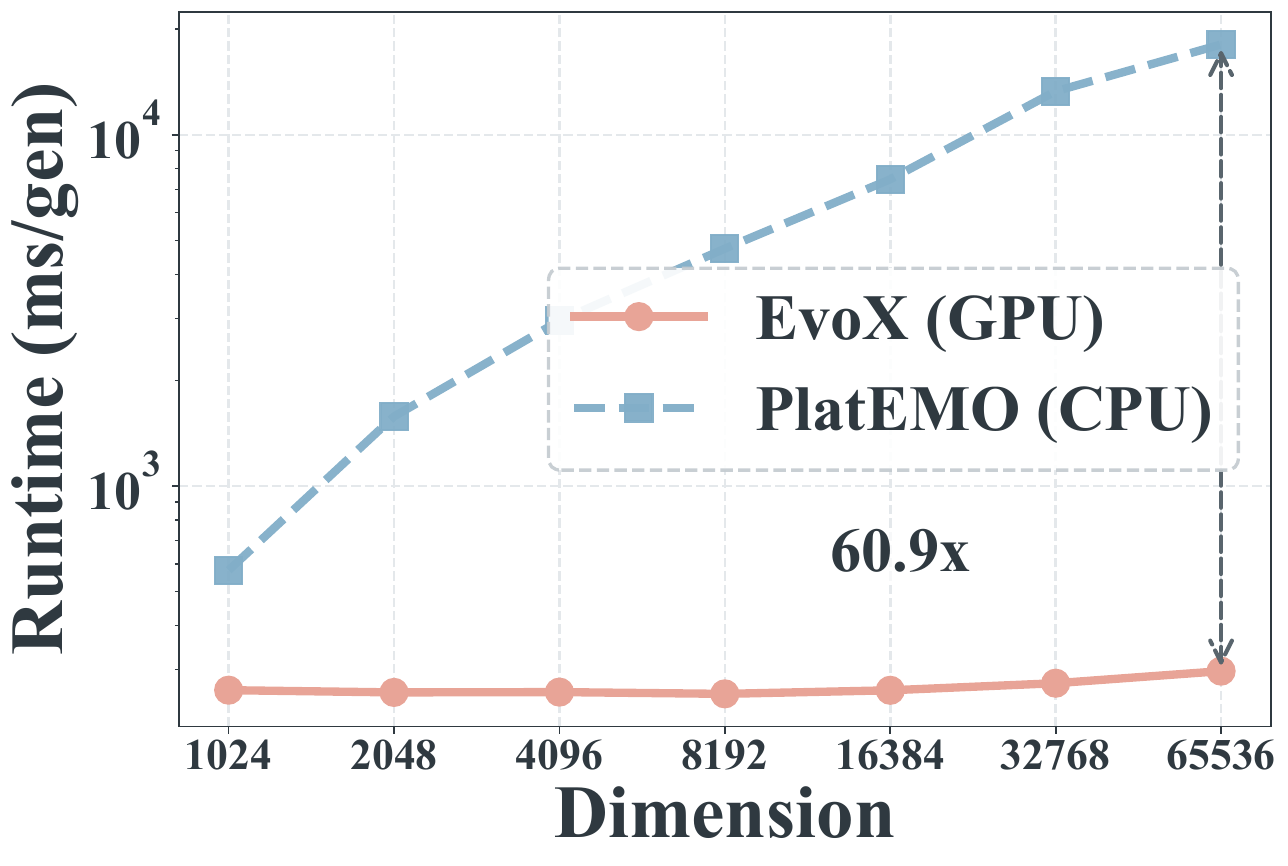}}
\\[-1mm]
\subfloat[CMOPSO: varying $N$]{\includegraphics[width=0.22\textwidth]{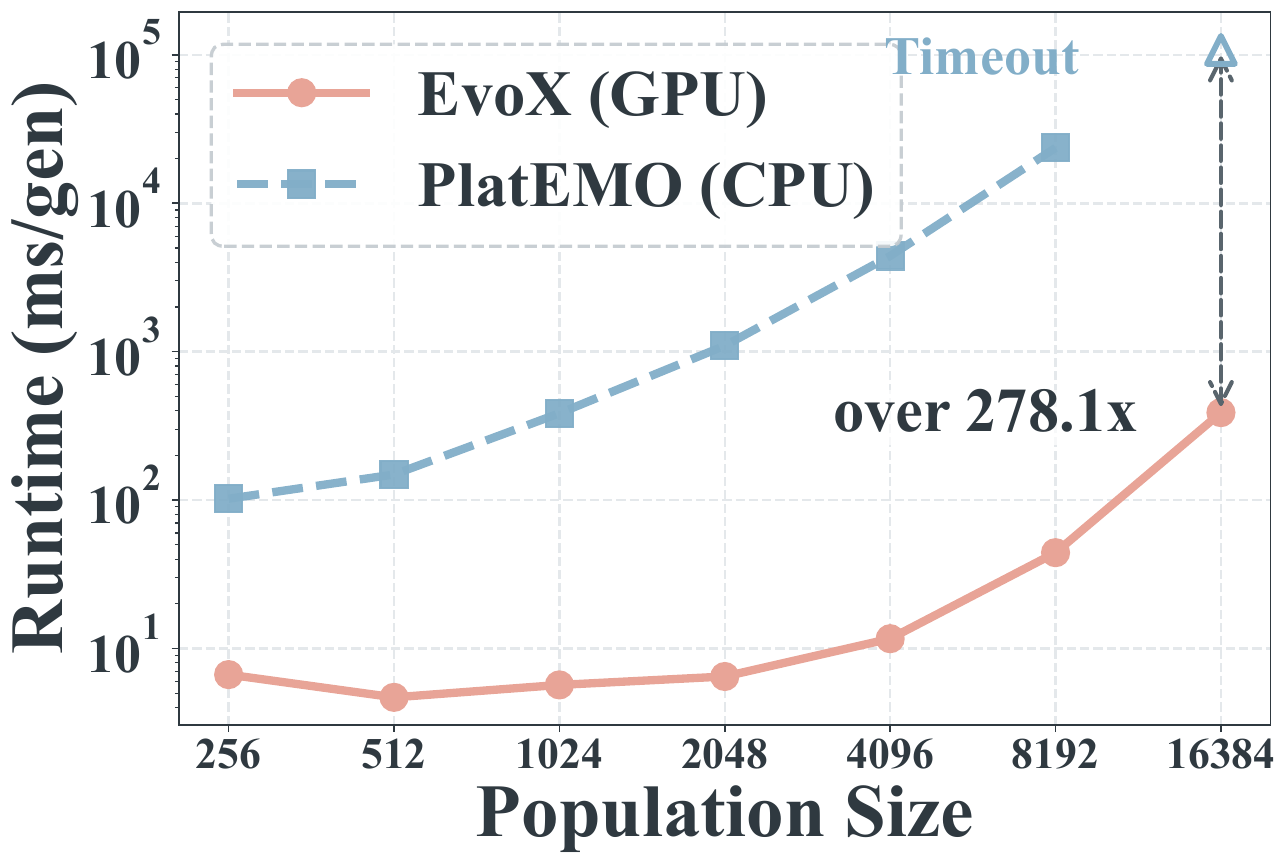}}
\hfill
\subfloat[CMOPSO: varying $D$]{\includegraphics[width=0.22\textwidth]{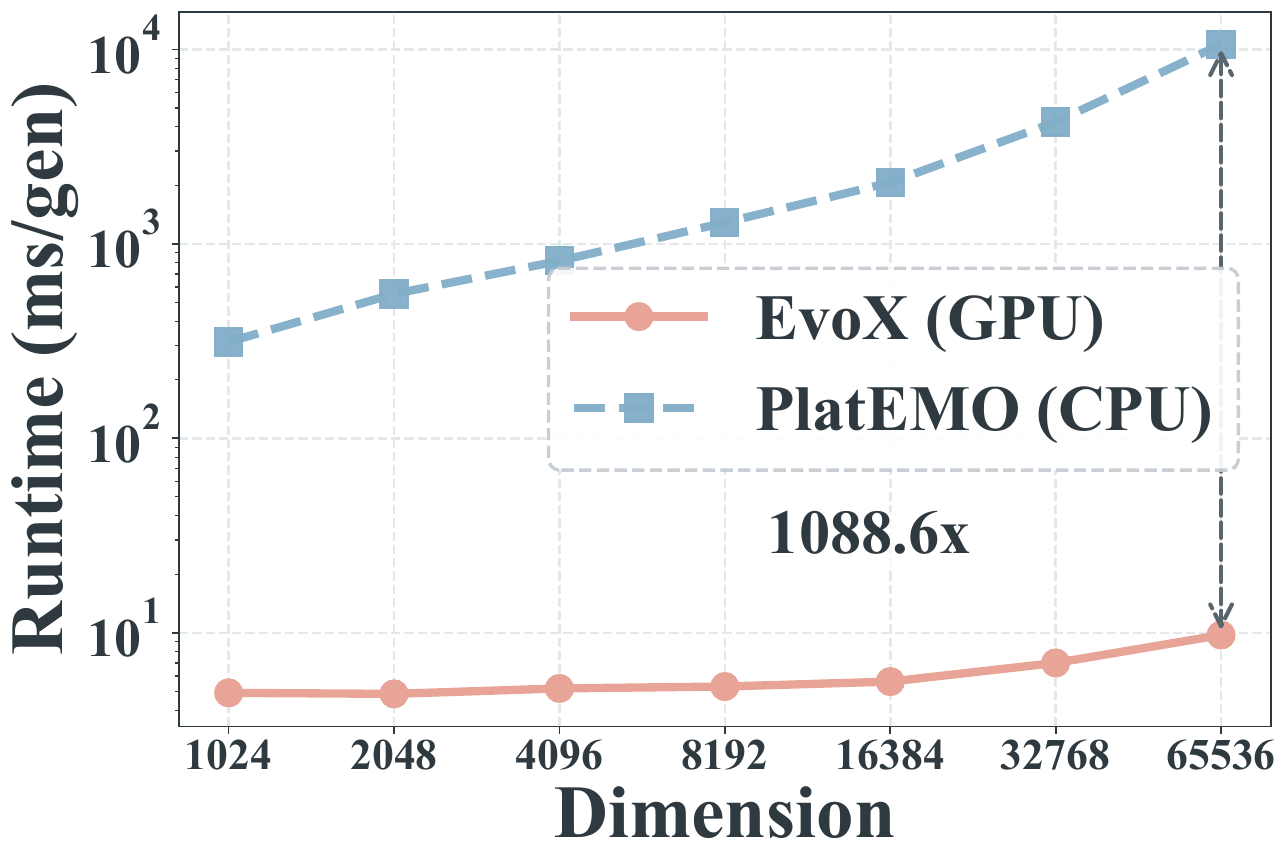}}
\hfill
\subfloat[CoMMEA: varying $N$]{\includegraphics[width=0.22\textwidth]{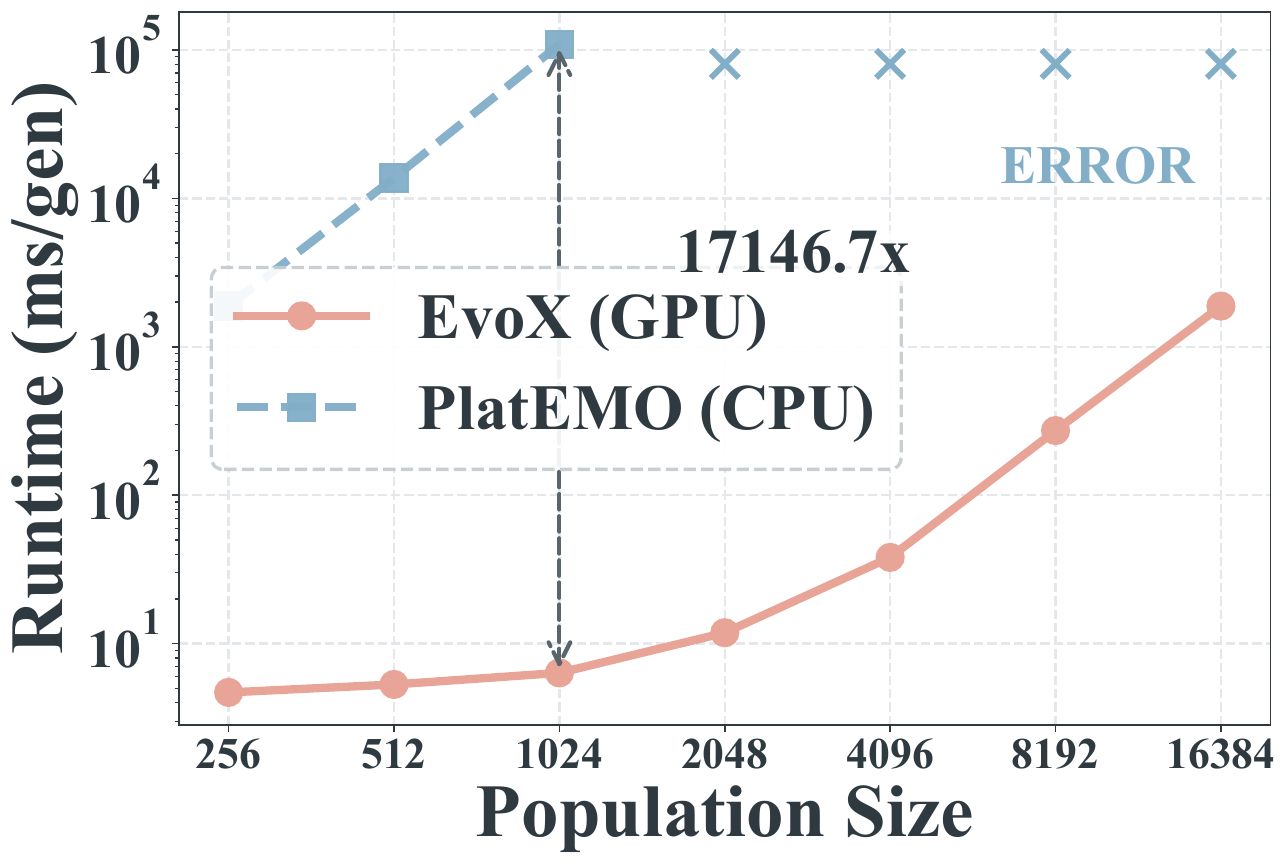}}
\hfill
\subfloat[CoMMEA: varying $D$]{\includegraphics[width=0.22\textwidth]{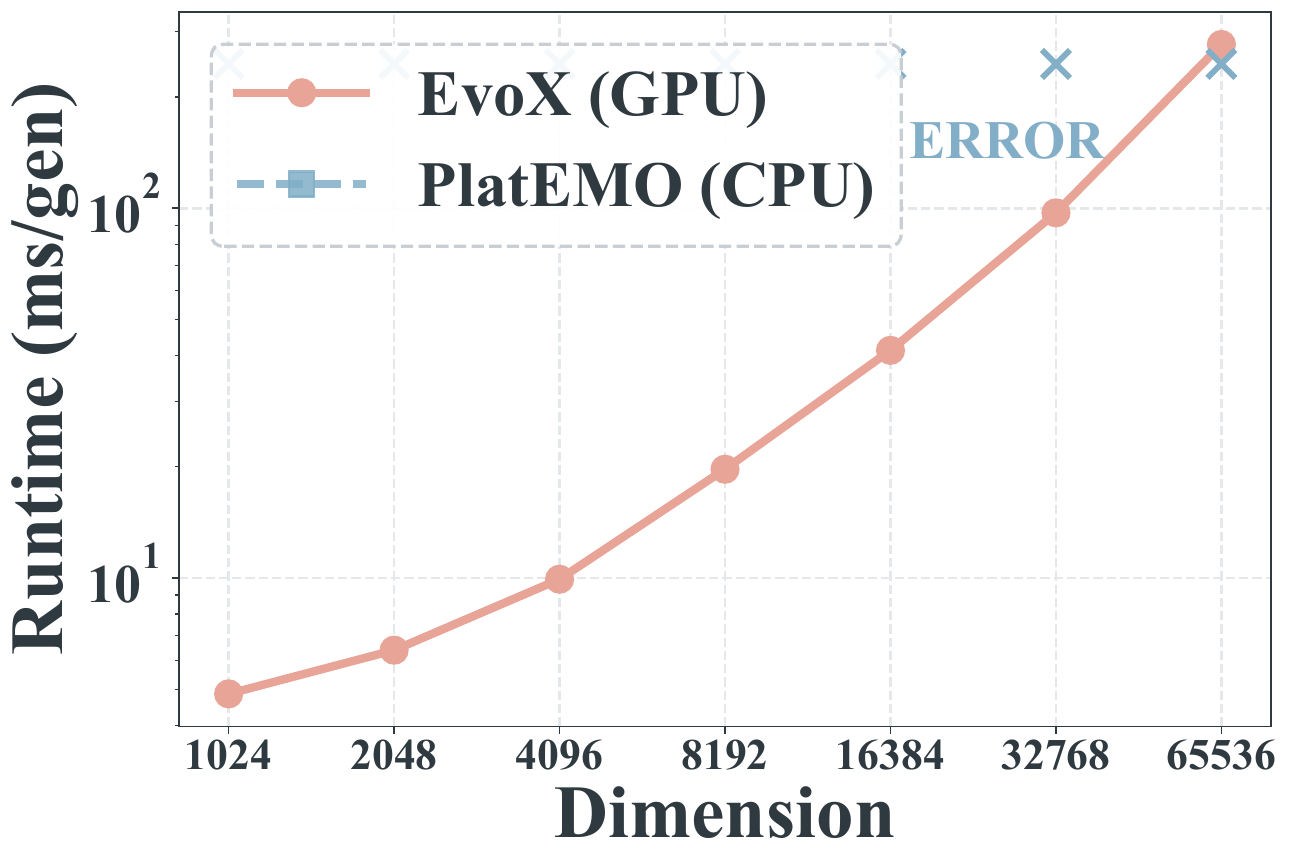}}
\\[-1mm]
\subfloat[DM-MOEA: varying $N$]{\includegraphics[width=0.22\textwidth]{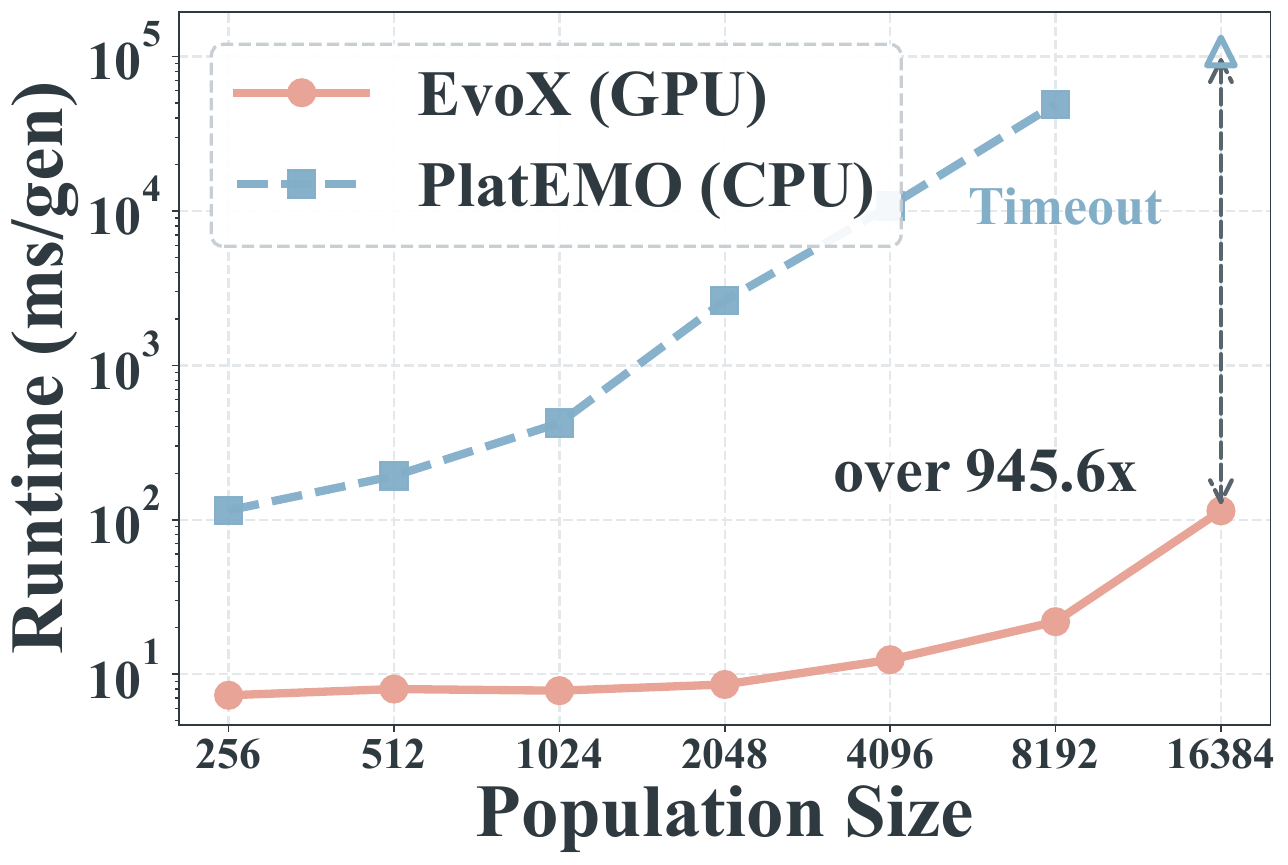}}
\hfill
\subfloat[DM-MOEA: varying $D$]{\includegraphics[width=0.22\textwidth]{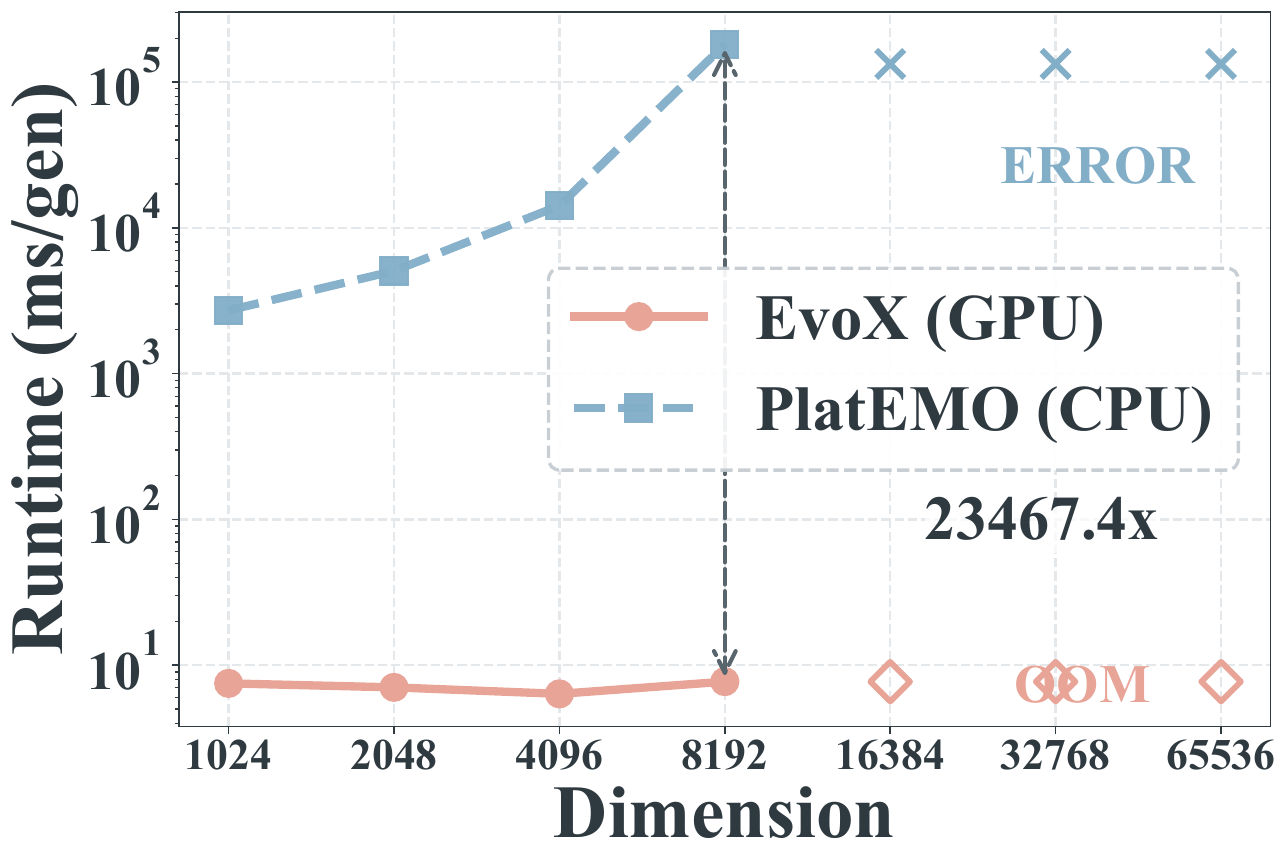}}
\hfill
\subfloat[EFR-RR: varying $N$]{\includegraphics[width=0.22\textwidth]{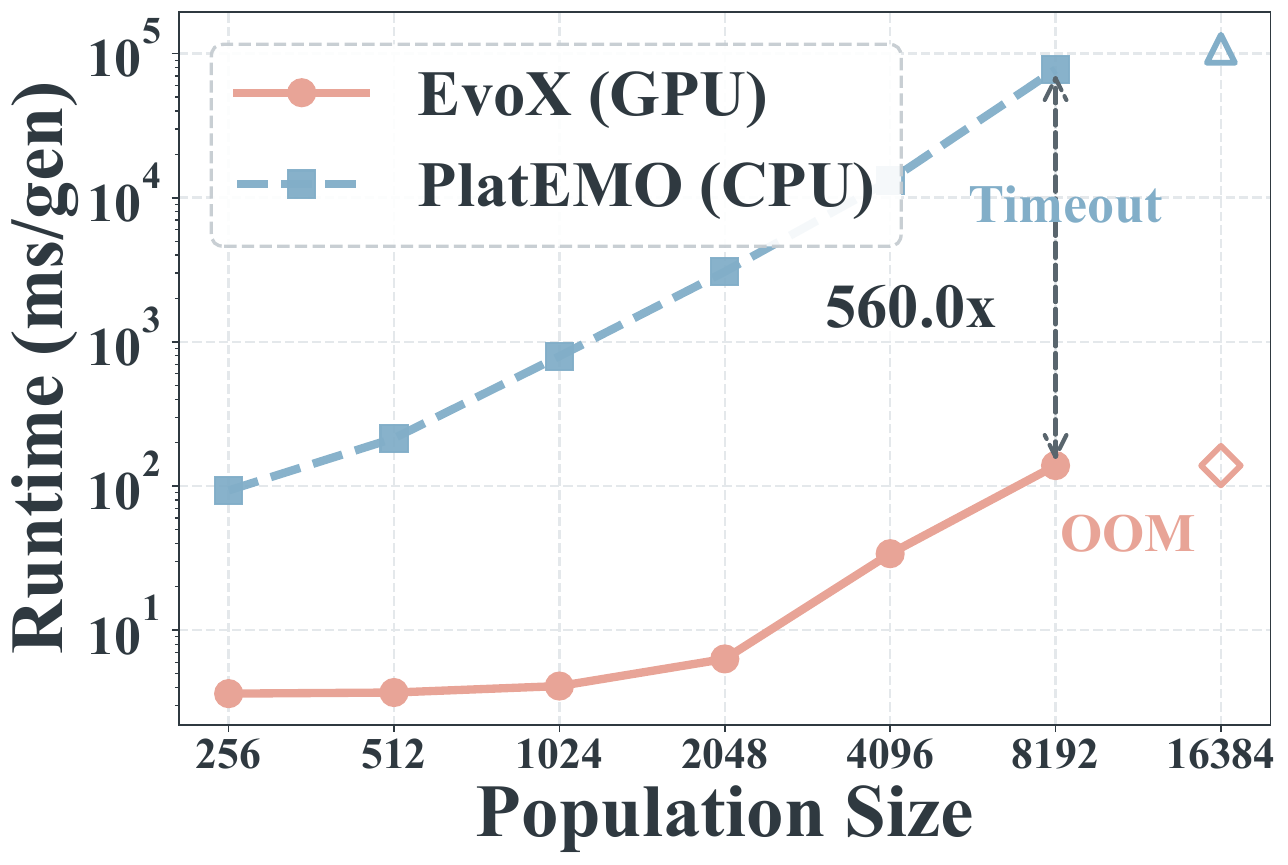}}
\hfill
\subfloat[EFR-RR: varying $D$]{\includegraphics[width=0.22\textwidth]{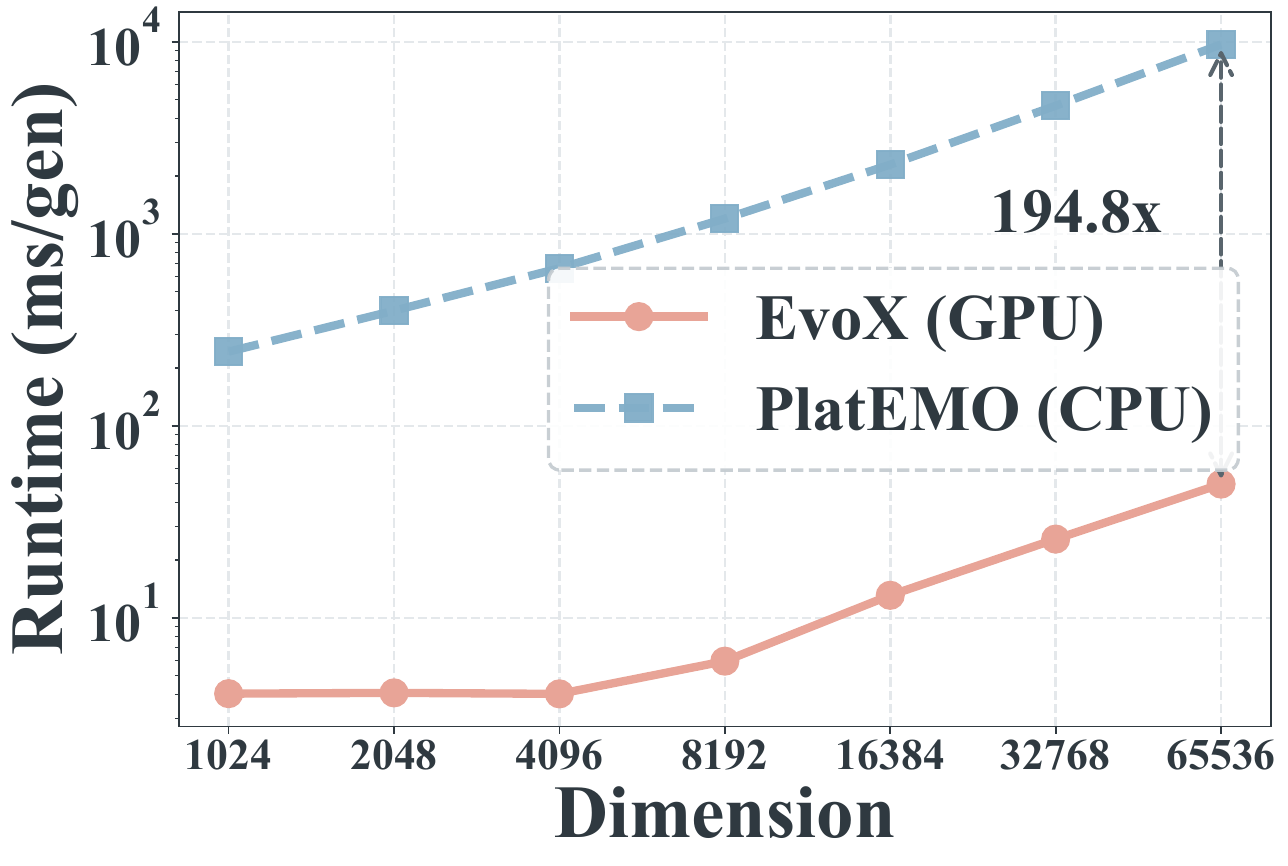}}
\caption{Complete population-size ($N$) and decision-dimension ($D$) scaling results for AGE-MOEA through EFR-RR (Part I of V). Adjacent panels report the two scaling axes for each algorithm.}
\label{fig:supp_scaling_curves_01}
\end{figure}

\begin{figure}[p]
\centering
\subfloat[e-MOEA: varying $N$]{\includegraphics[width=0.22\textwidth]{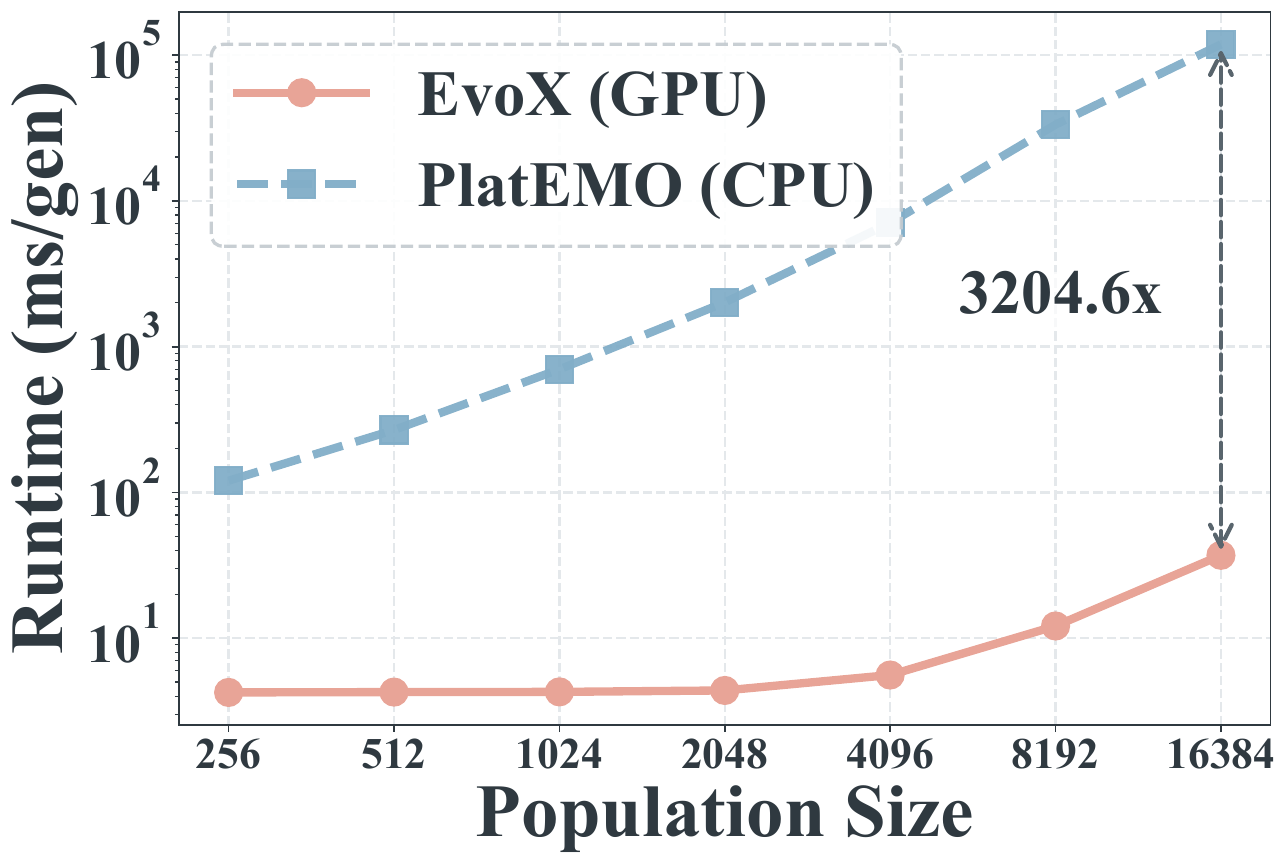}}
\hfill
\subfloat[e-MOEA: varying $D$]{\includegraphics[width=0.22\textwidth]{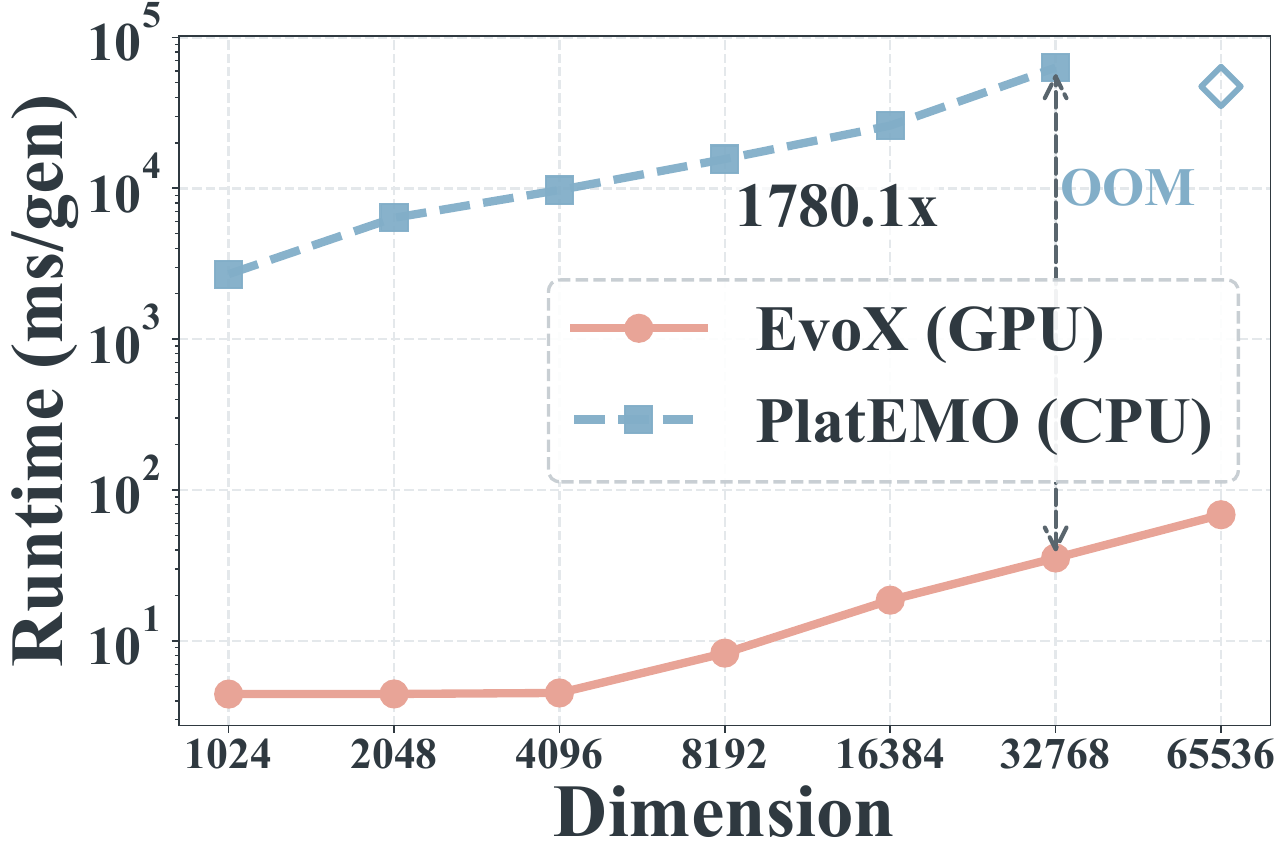}}
\hfill
\subfloat[GDE3: varying $N$]{\includegraphics[width=0.22\textwidth]{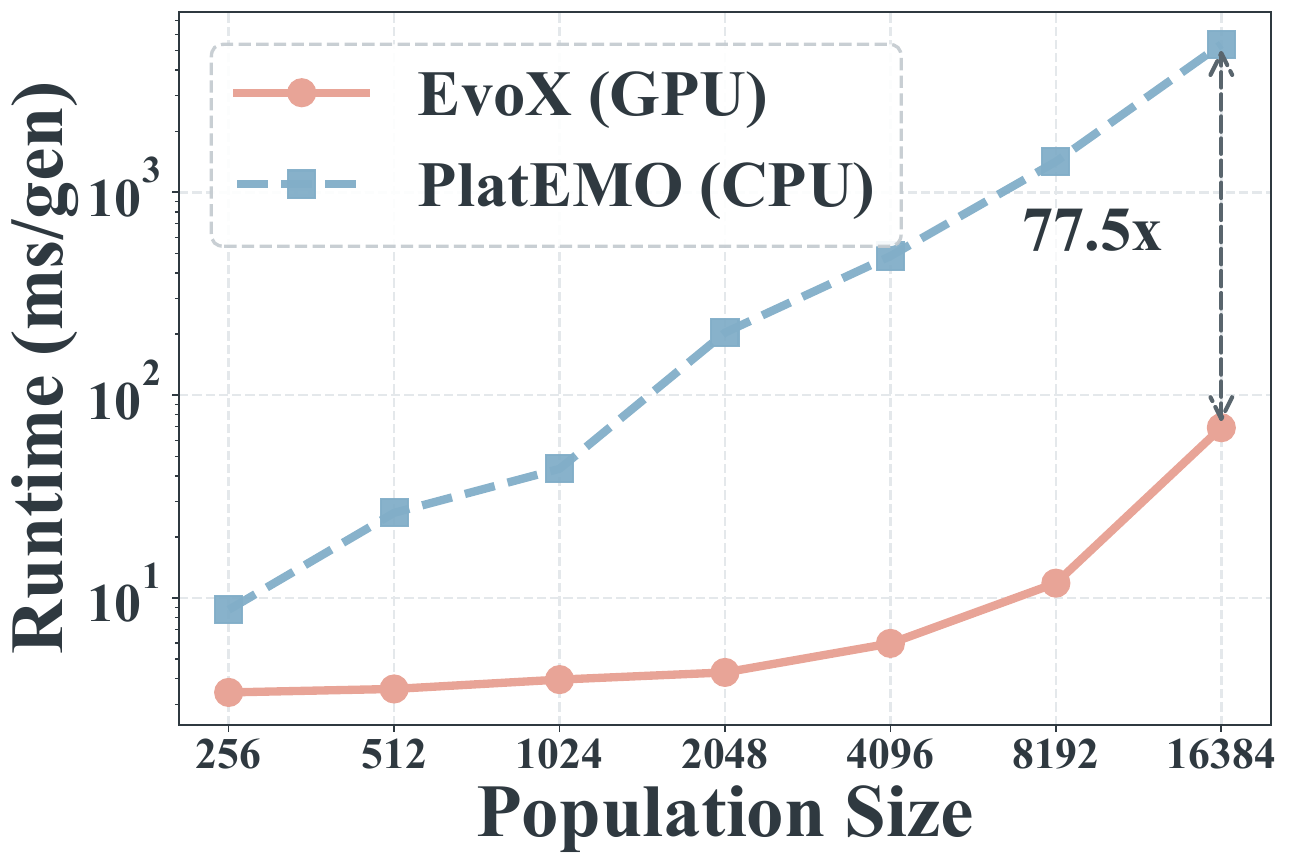}}
\hfill
\subfloat[GDE3: varying $D$]{\includegraphics[width=0.22\textwidth]{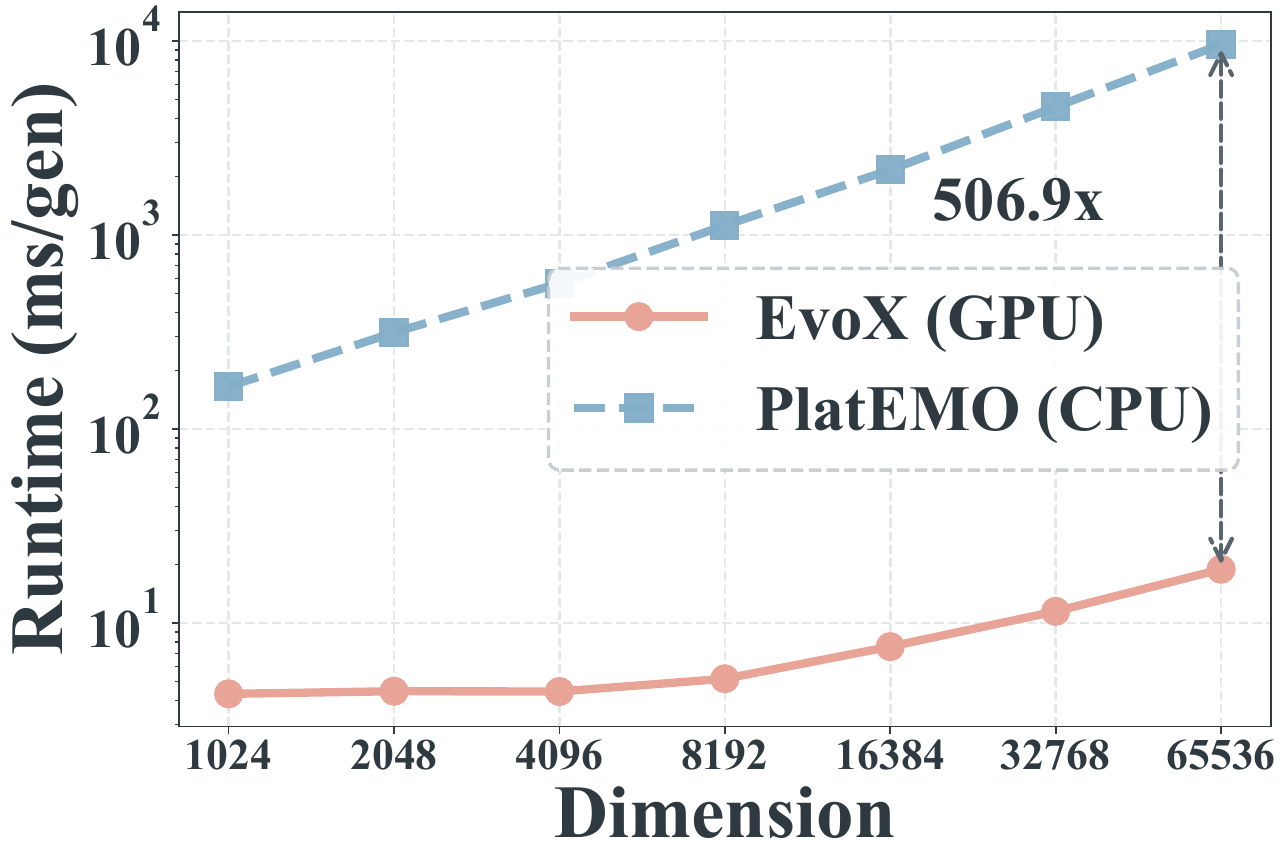}}
\\[-1mm]
\subfloat[GrEA: varying $N$]{\includegraphics[width=0.22\textwidth]{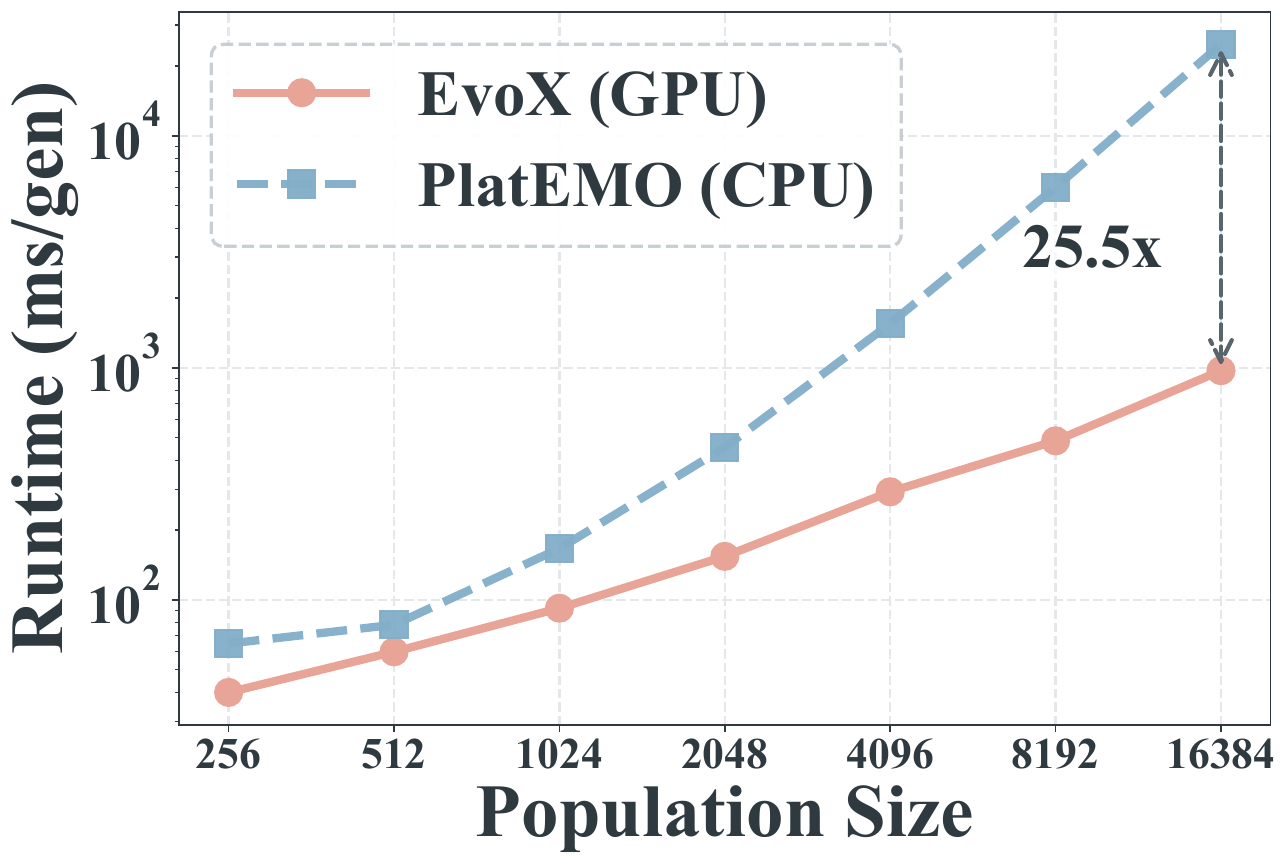}}
\hfill
\subfloat[GrEA: varying $D$]{\includegraphics[width=0.22\textwidth]{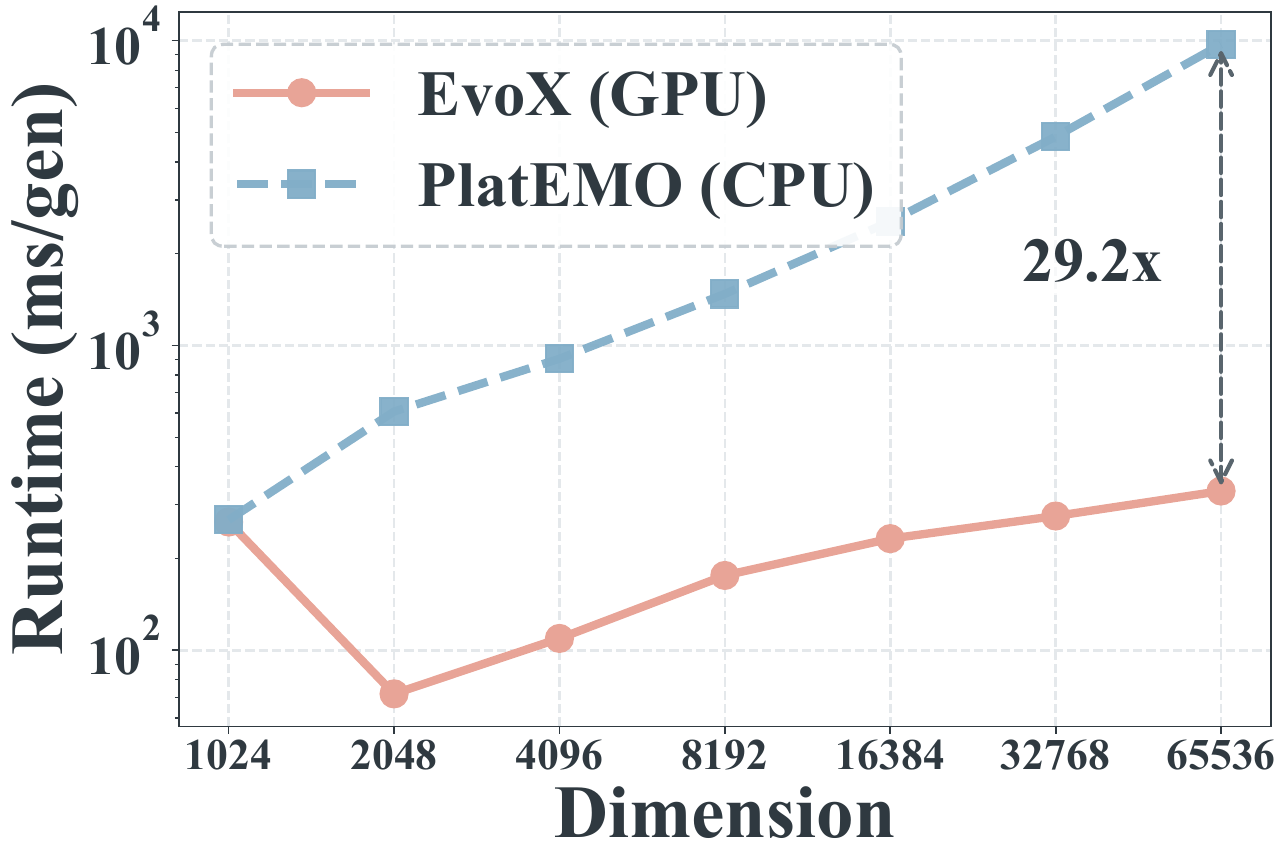}}
\hfill
\subfloat[GWASF-GA: varying $N$]{\includegraphics[width=0.22\textwidth]{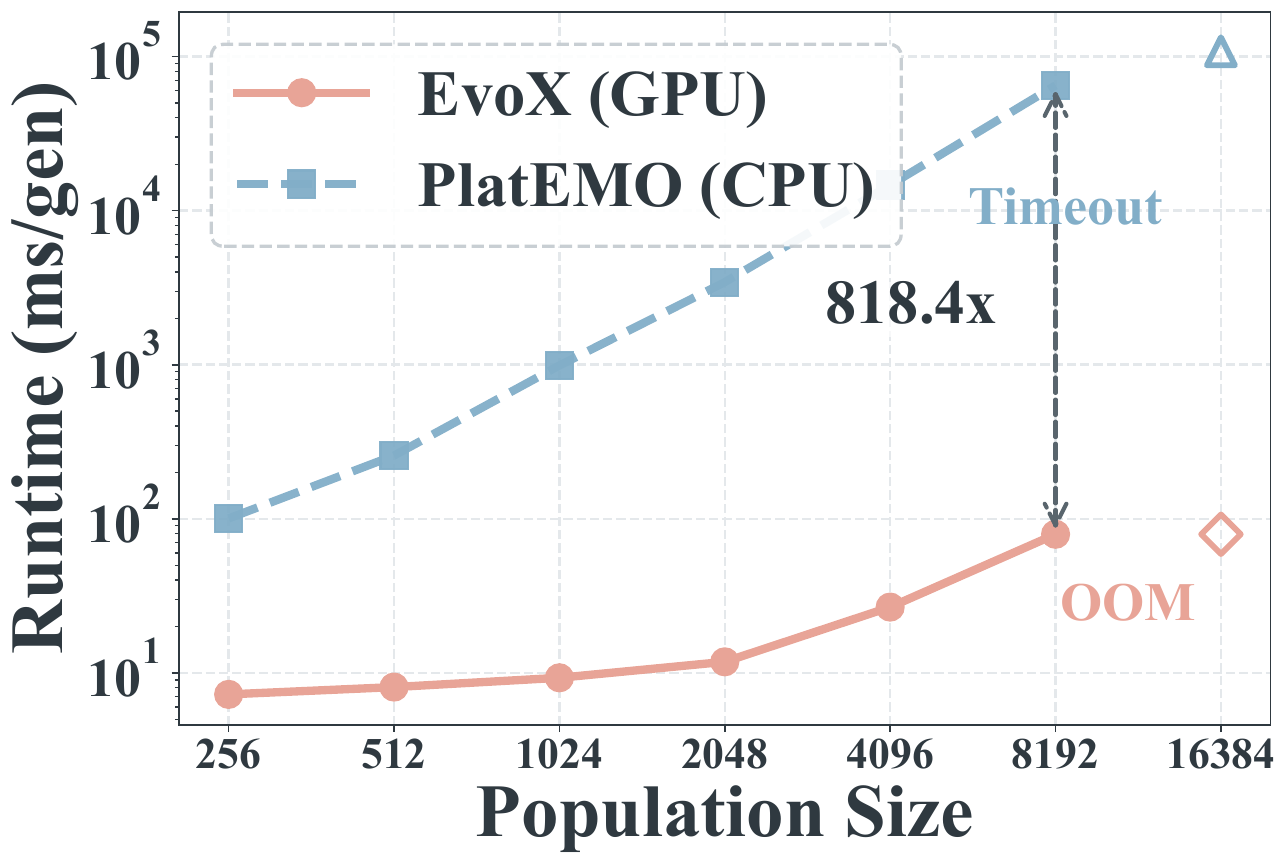}}
\hfill
\subfloat[GWASF-GA: varying $D$]{\includegraphics[width=0.22\textwidth]{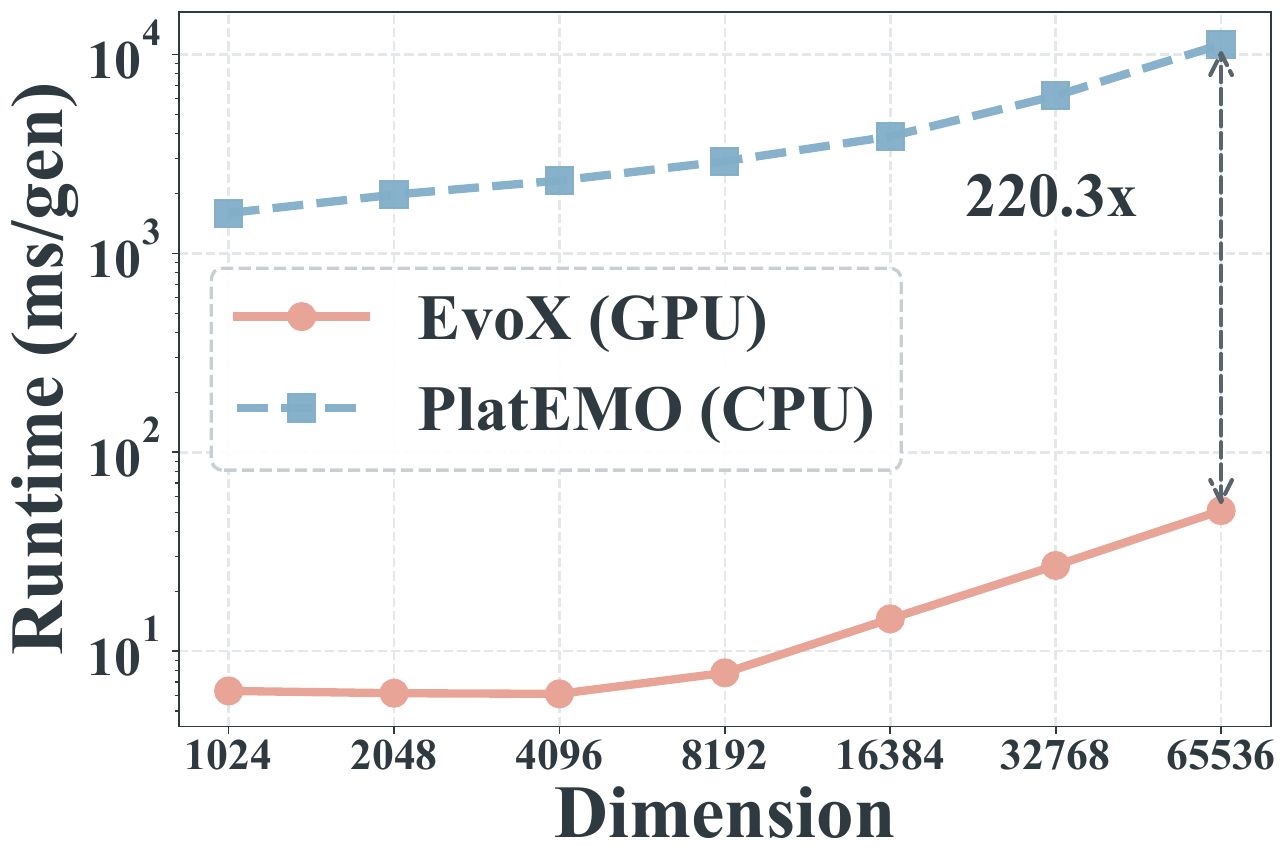}}
\\[-1mm]
\subfloat[KnEA: varying $N$]{\includegraphics[width=0.22\textwidth]{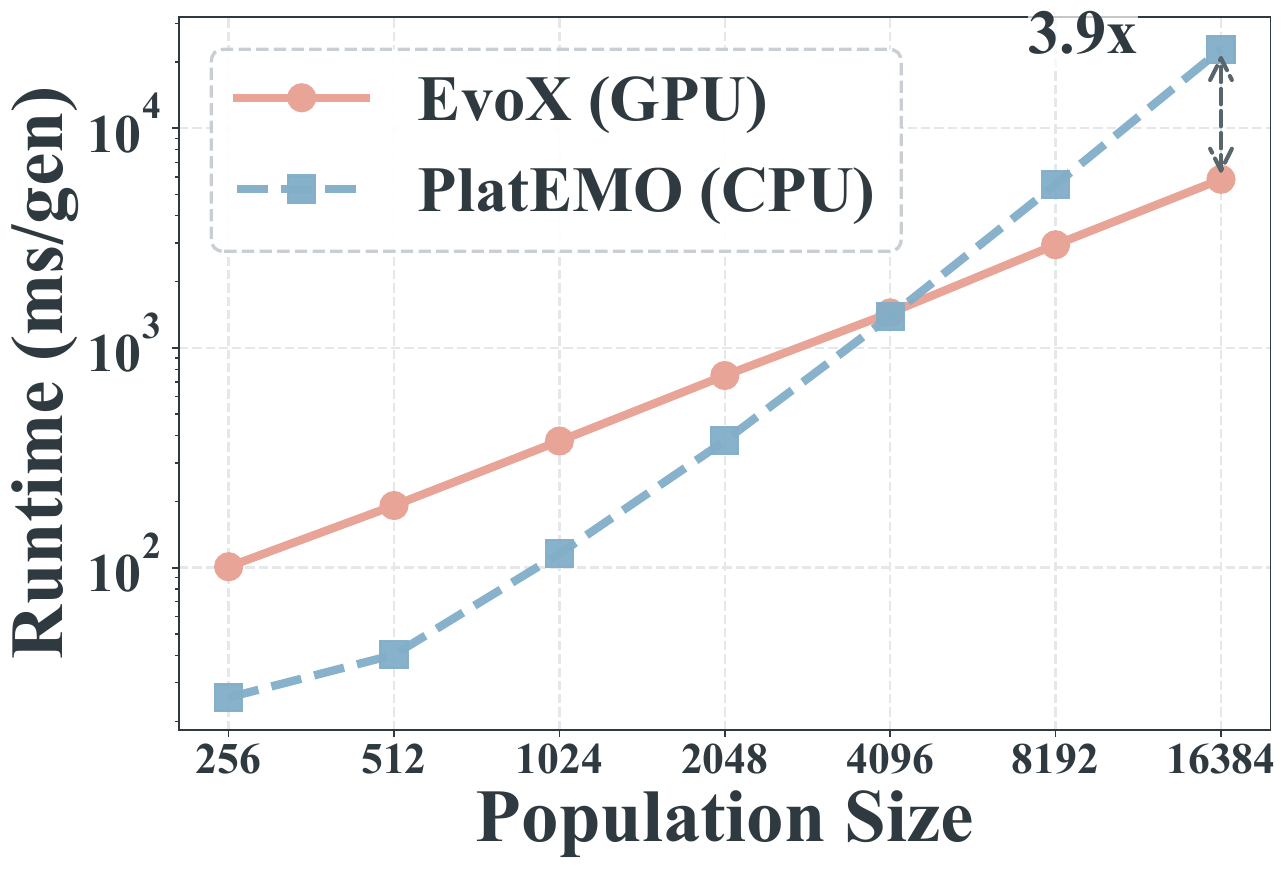}}
\hfill
\subfloat[KnEA: varying $D$]{\includegraphics[width=0.22\textwidth]{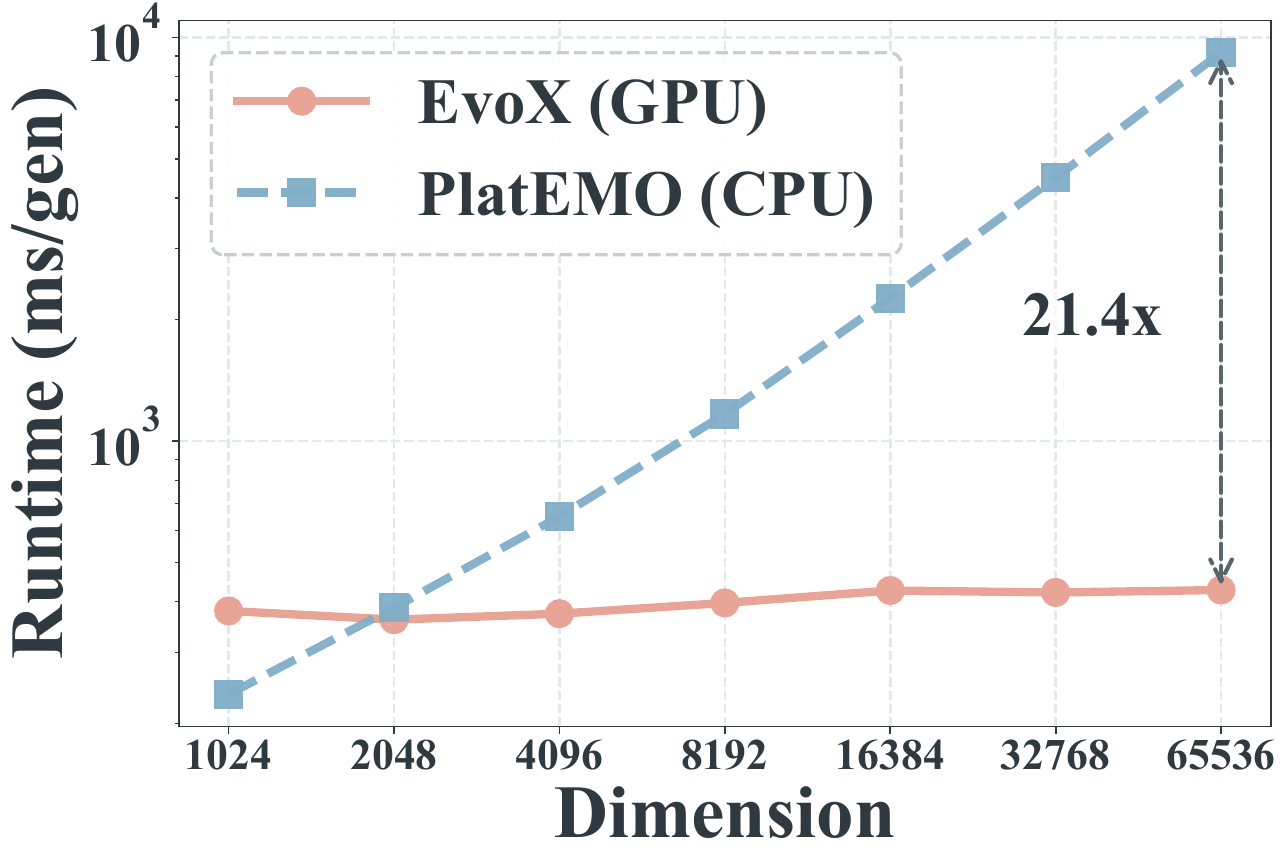}}
\hfill
\subfloat[LSMOF: varying $N$]{\includegraphics[width=0.22\textwidth]{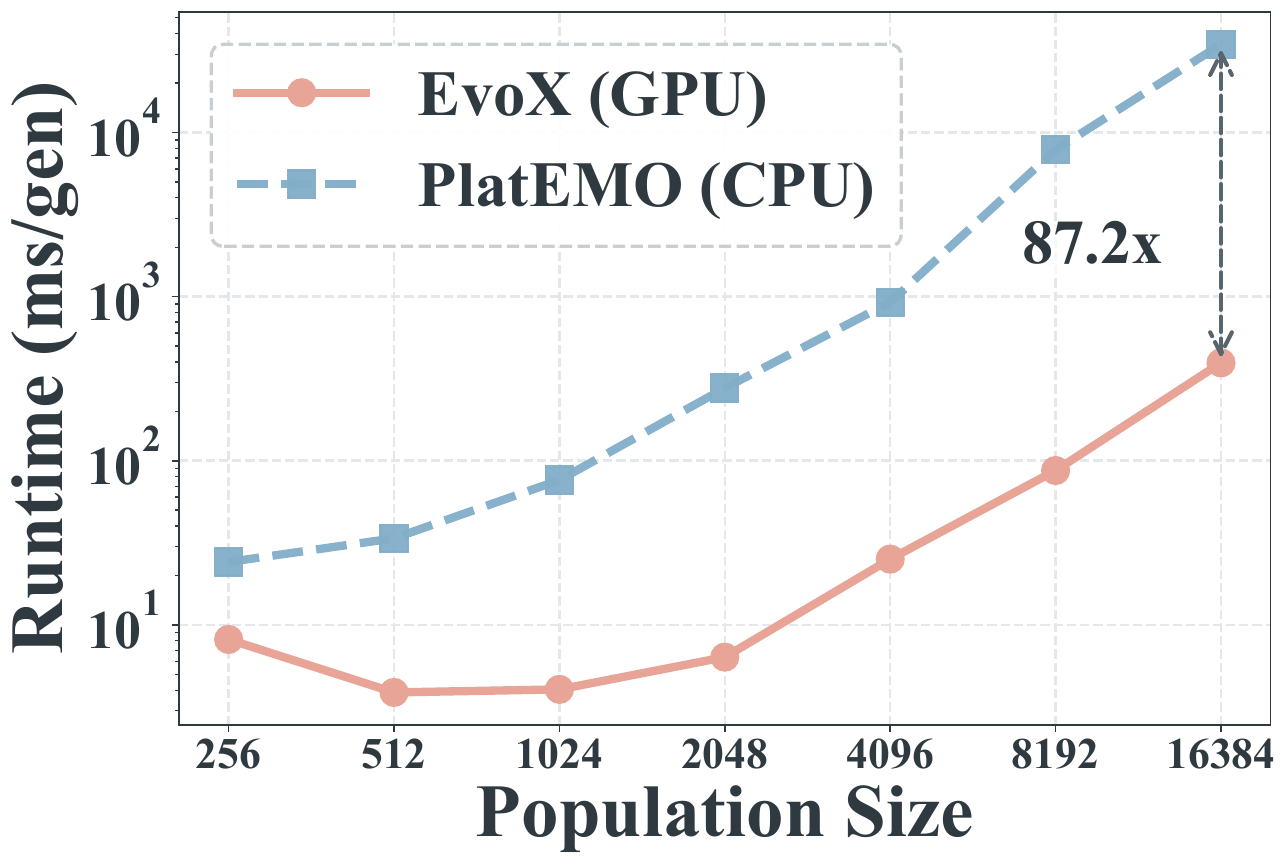}}
\hfill
\subfloat[LSMOF: varying $D$]{\includegraphics[width=0.22\textwidth]{figures/exp3_scaling/curves/scaling_dim_comparison_LSMOF_dim.pdf}}
\\[-1mm]
\subfloat[MaOEA-CSS: varying $N$]{\includegraphics[width=0.22\textwidth]{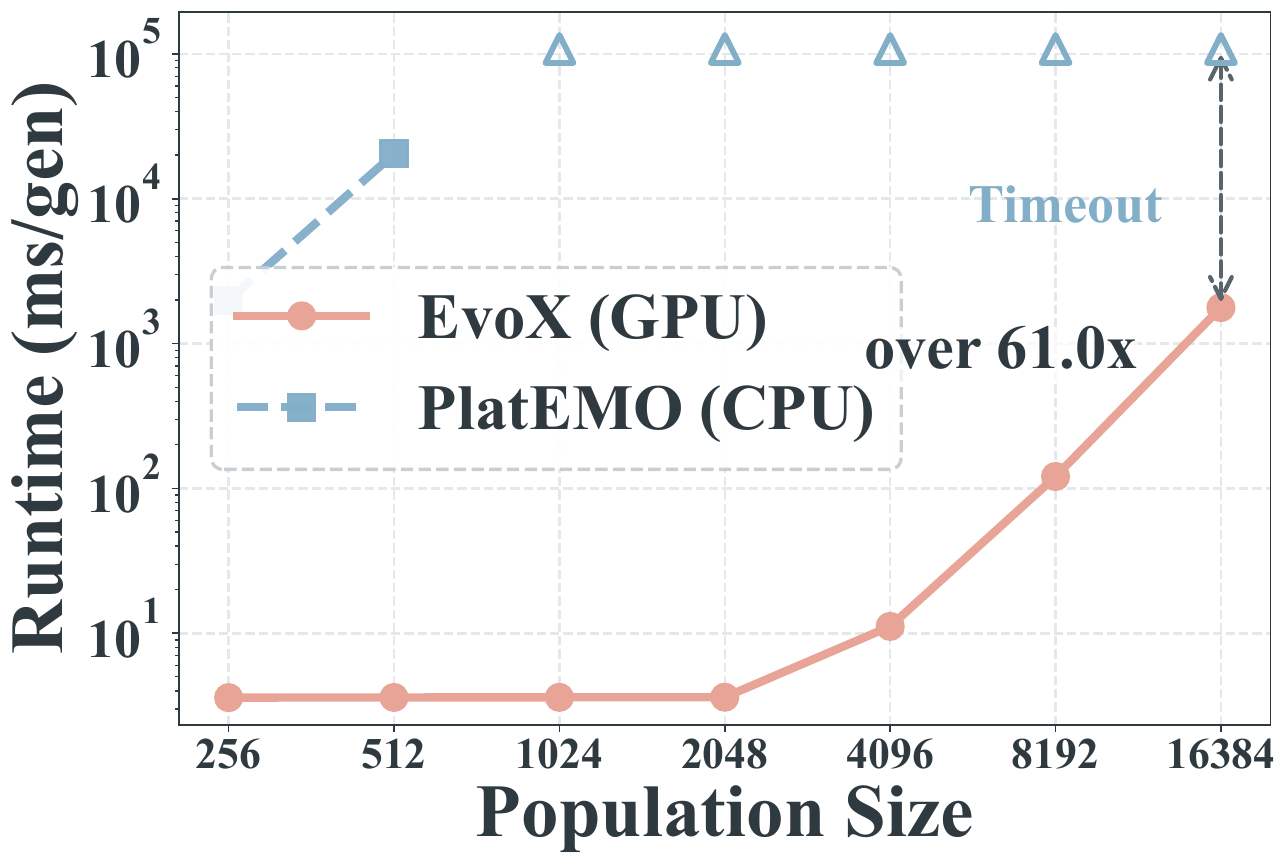}}
\hfill
\subfloat[MaOEA-CSS: varying $D$]{\includegraphics[width=0.22\textwidth]{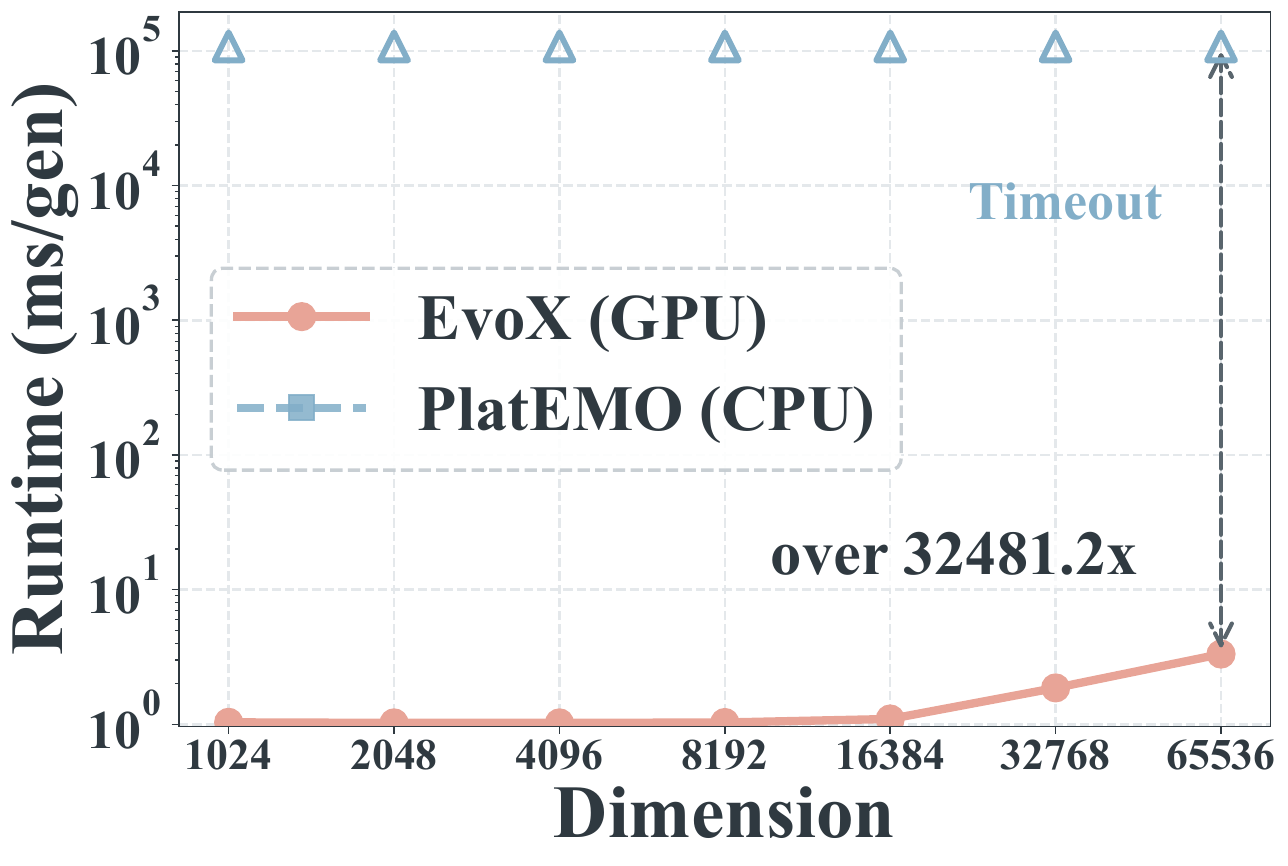}}
\hfill
\subfloat[MOEA-D-AWA: varying $N$]{\includegraphics[width=0.22\textwidth]{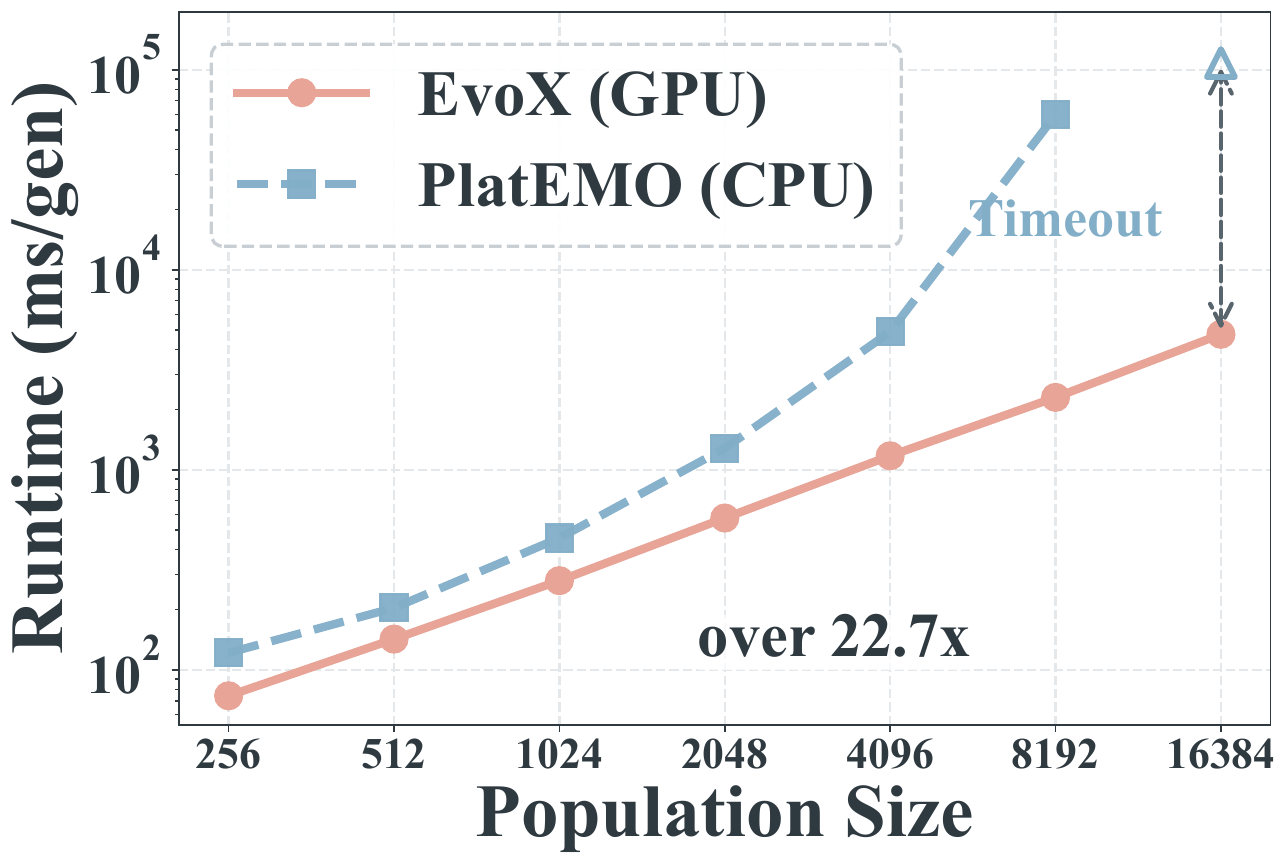}}
\hfill
\subfloat[MOEA-D-AWA: varying $D$]{\includegraphics[width=0.22\textwidth]{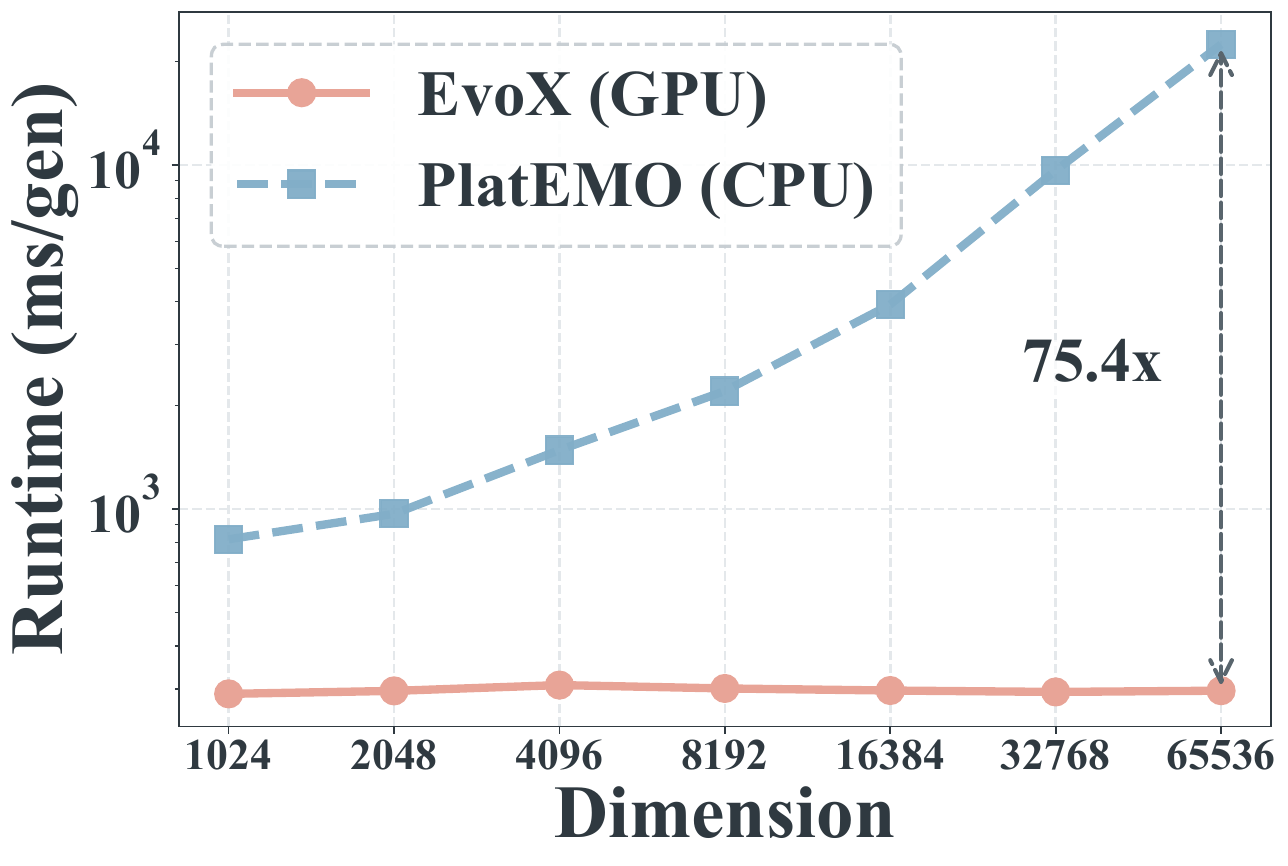}}
\\[-1mm]
\subfloat[MOEA-D-DCWV: varying $N$]{\includegraphics[width=0.22\textwidth]{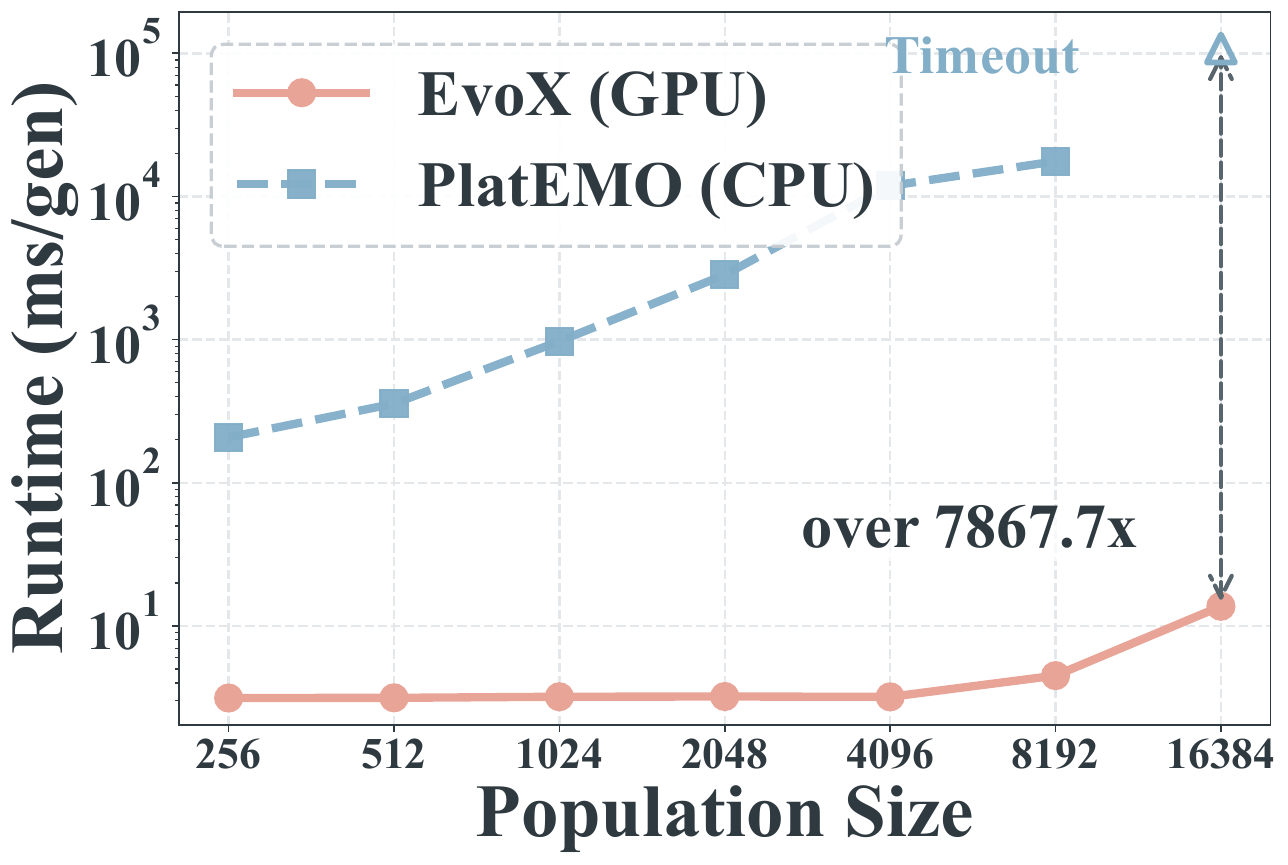}}
\hfill
\subfloat[MOEA-D-DCWV: varying $D$]{\includegraphics[width=0.22\textwidth]{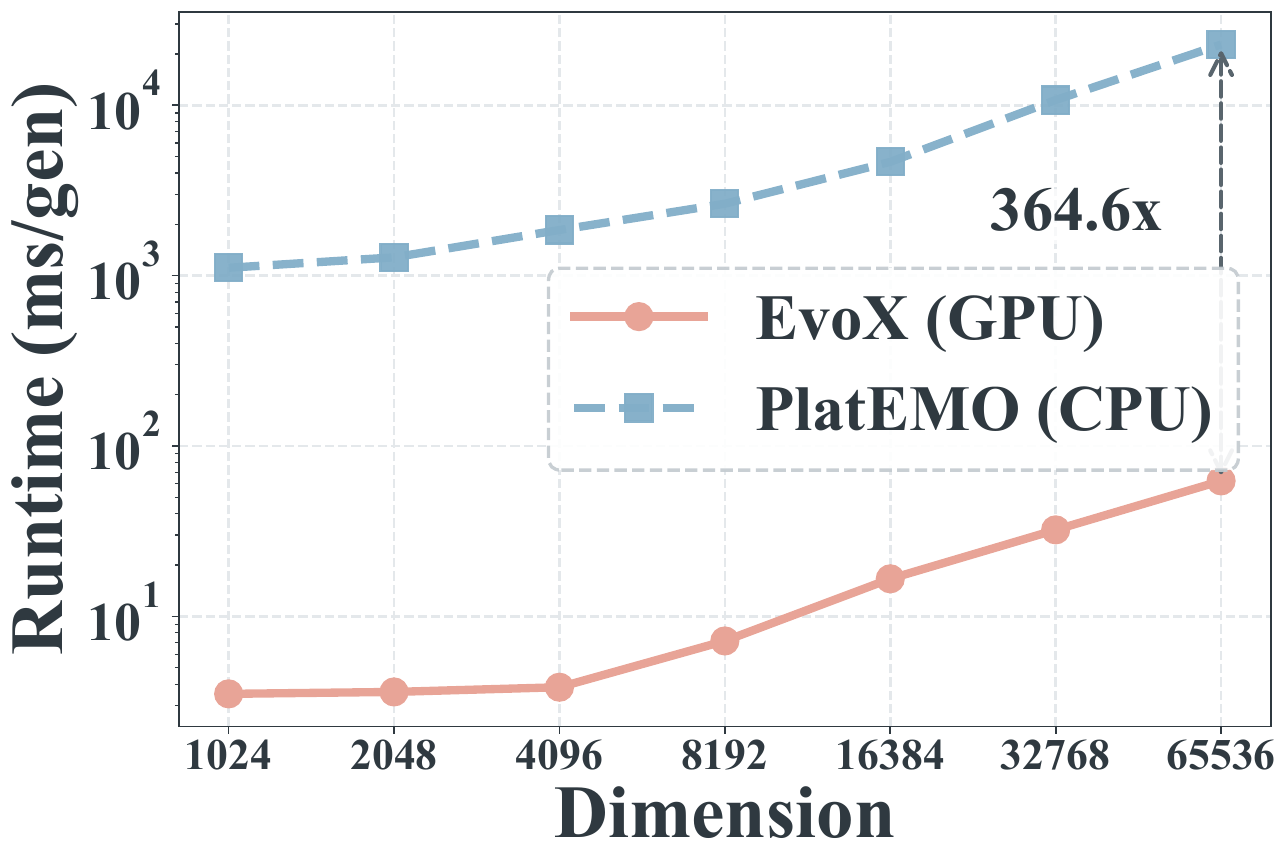}}
\hfill
\subfloat[MOEA/D-DE: varying $N$]{\includegraphics[width=0.22\textwidth]{figures/exp3_scaling/curves/scaling_pop_comparison_MOEA-D-DE_pop.pdf}}
\hfill
\subfloat[MOEA/D-DE: varying $D$]{\includegraphics[width=0.22\textwidth]{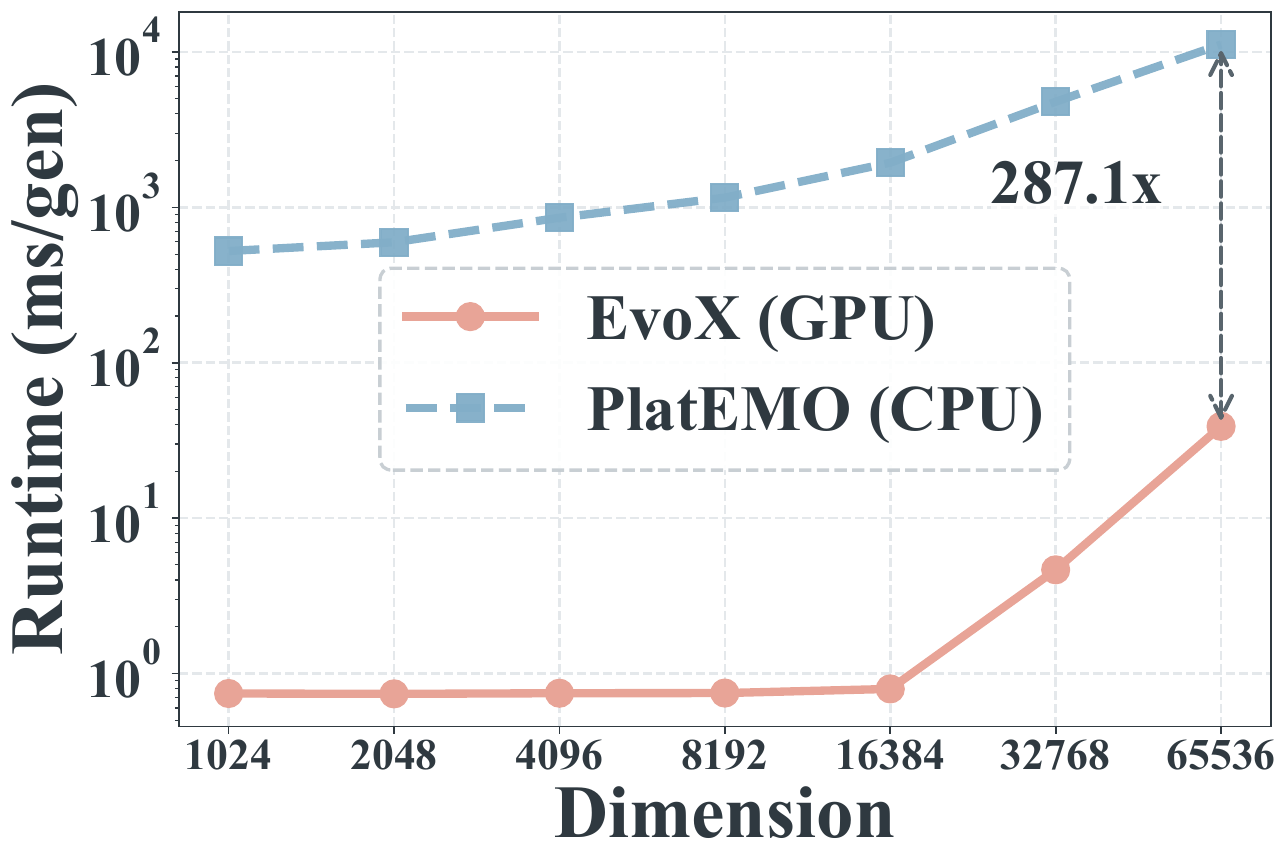}}
\caption{Complete population-size ($N$) and decision-dimension ($D$) scaling results for e-MOEA through MOEA/D-DE (Part II of V). Adjacent panels report the two scaling axes for each algorithm.}
\label{fig:supp_scaling_curves_02}
\end{figure}

\begin{figure}[p]
\centering
\subfloat[MOEA-D-DRA: varying $N$]{\includegraphics[width=0.22\textwidth]{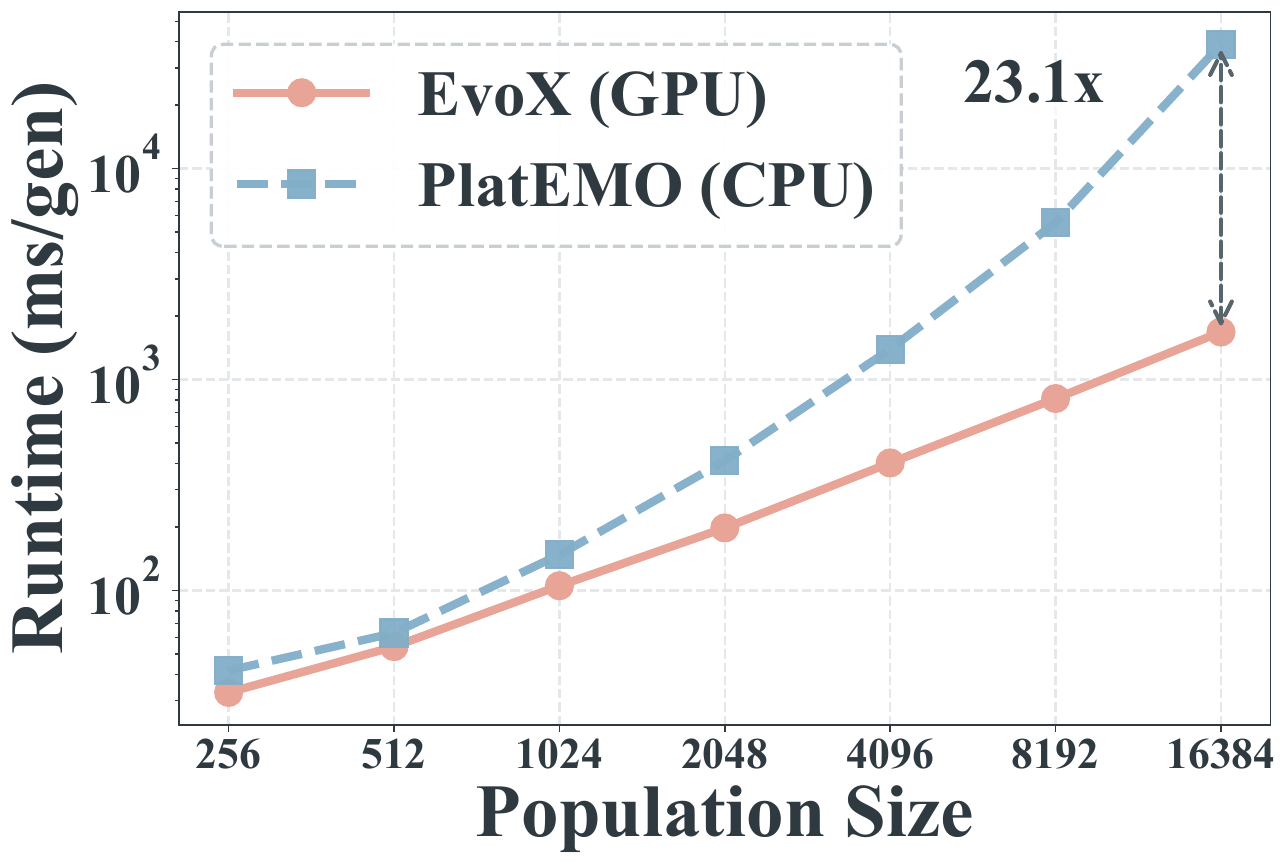}}
\hfill
\subfloat[MOEA-D-DRA: varying $D$]{\includegraphics[width=0.22\textwidth]{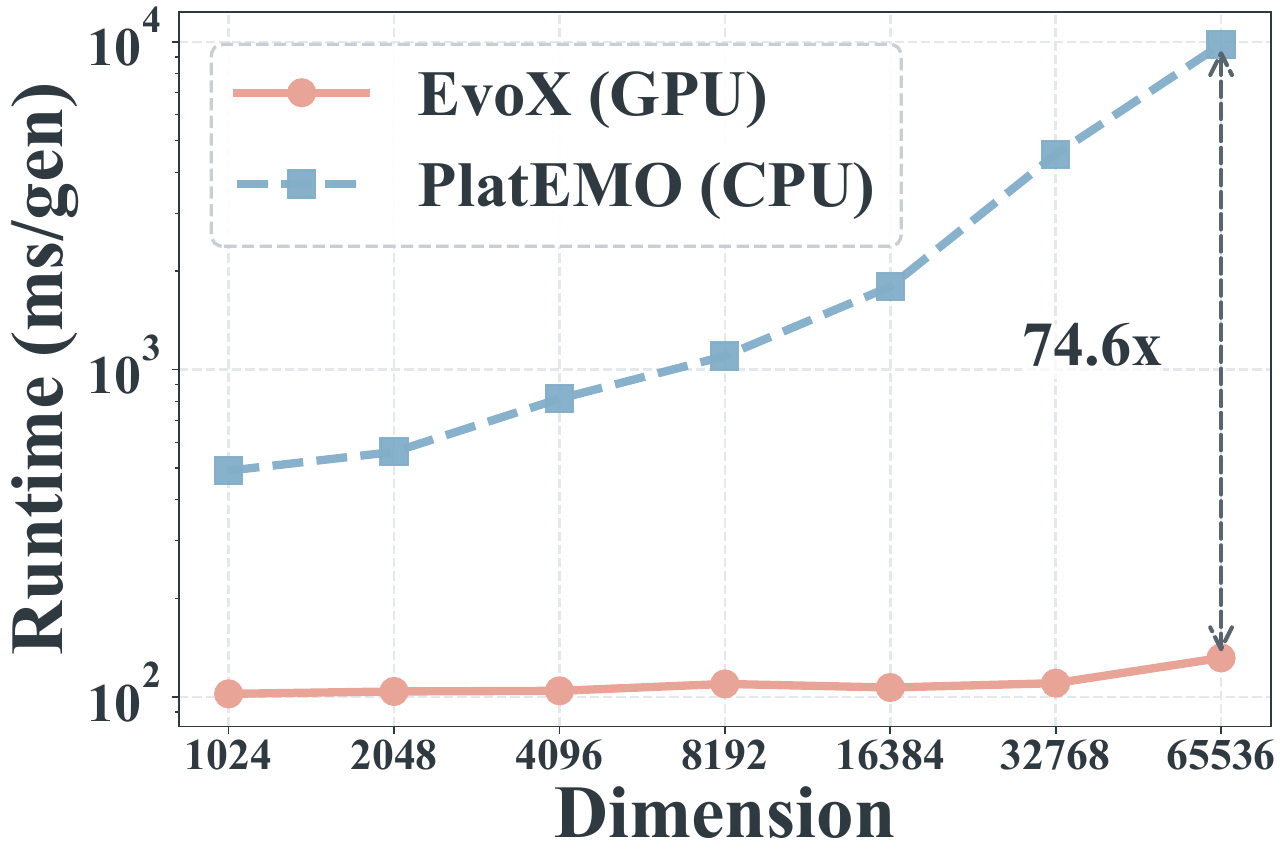}}
\hfill
\subfloat[MOEA-D-DU: varying $N$]{\includegraphics[width=0.22\textwidth]{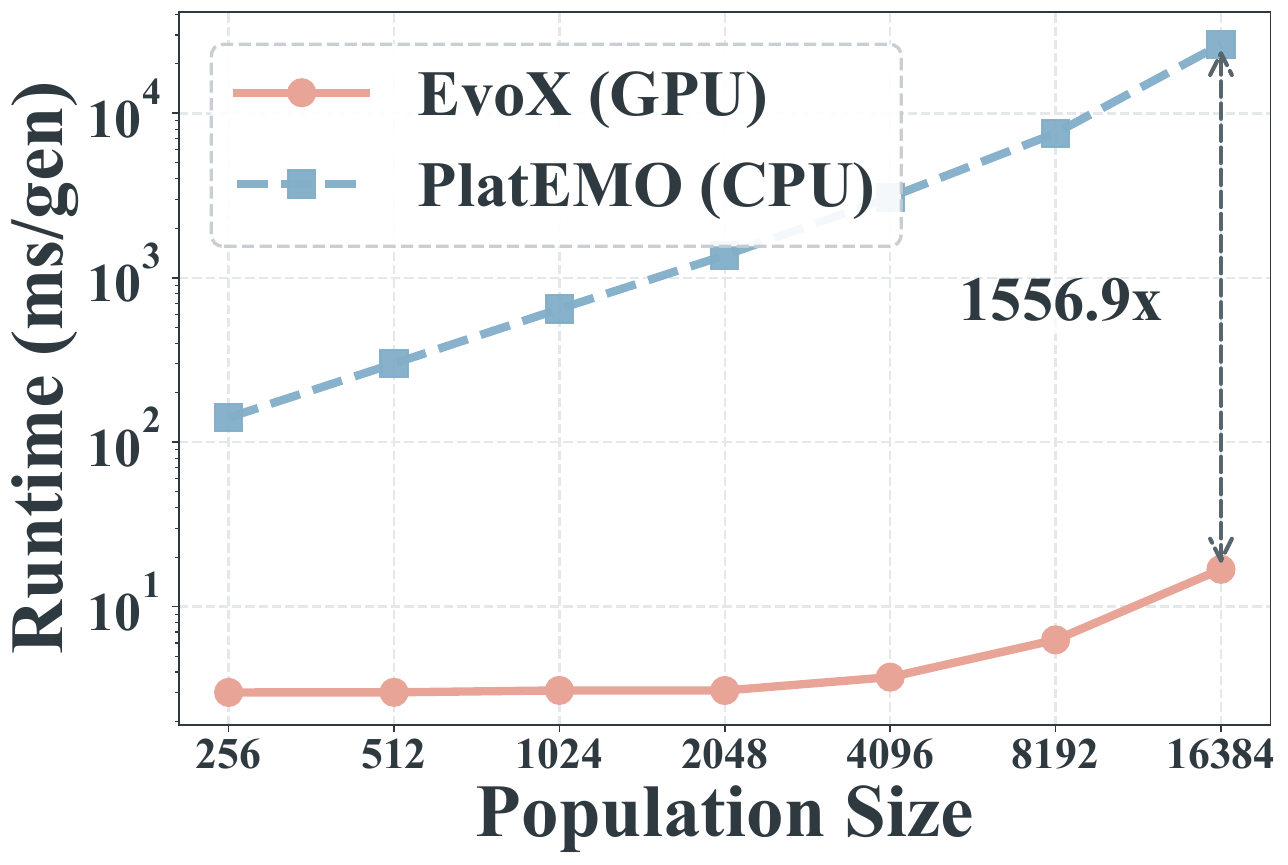}}
\hfill
\subfloat[MOEA-D-DU: varying $D$]{\includegraphics[width=0.22\textwidth]{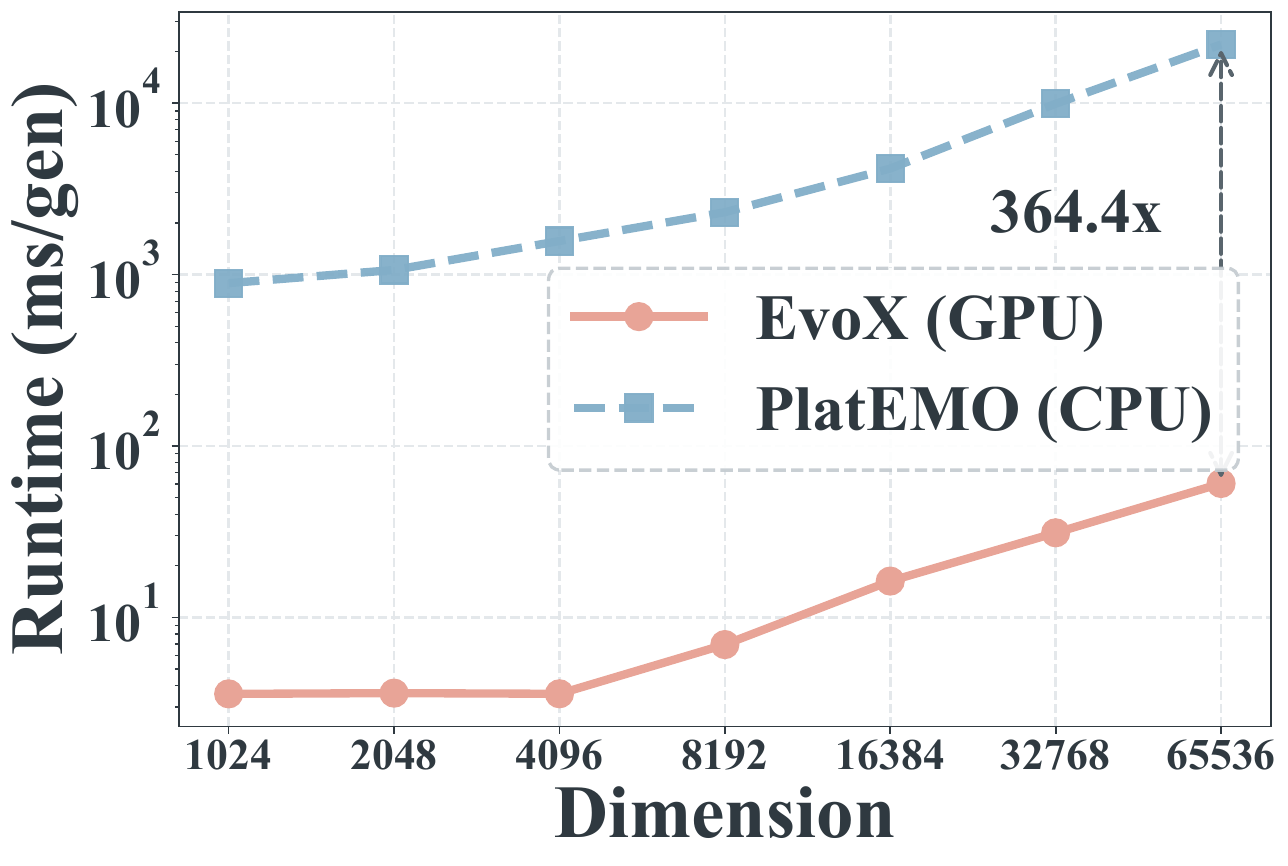}}
\\[-1mm]
\subfloat[MOEA-D-DYTS: varying $N$]{\includegraphics[width=0.22\textwidth]{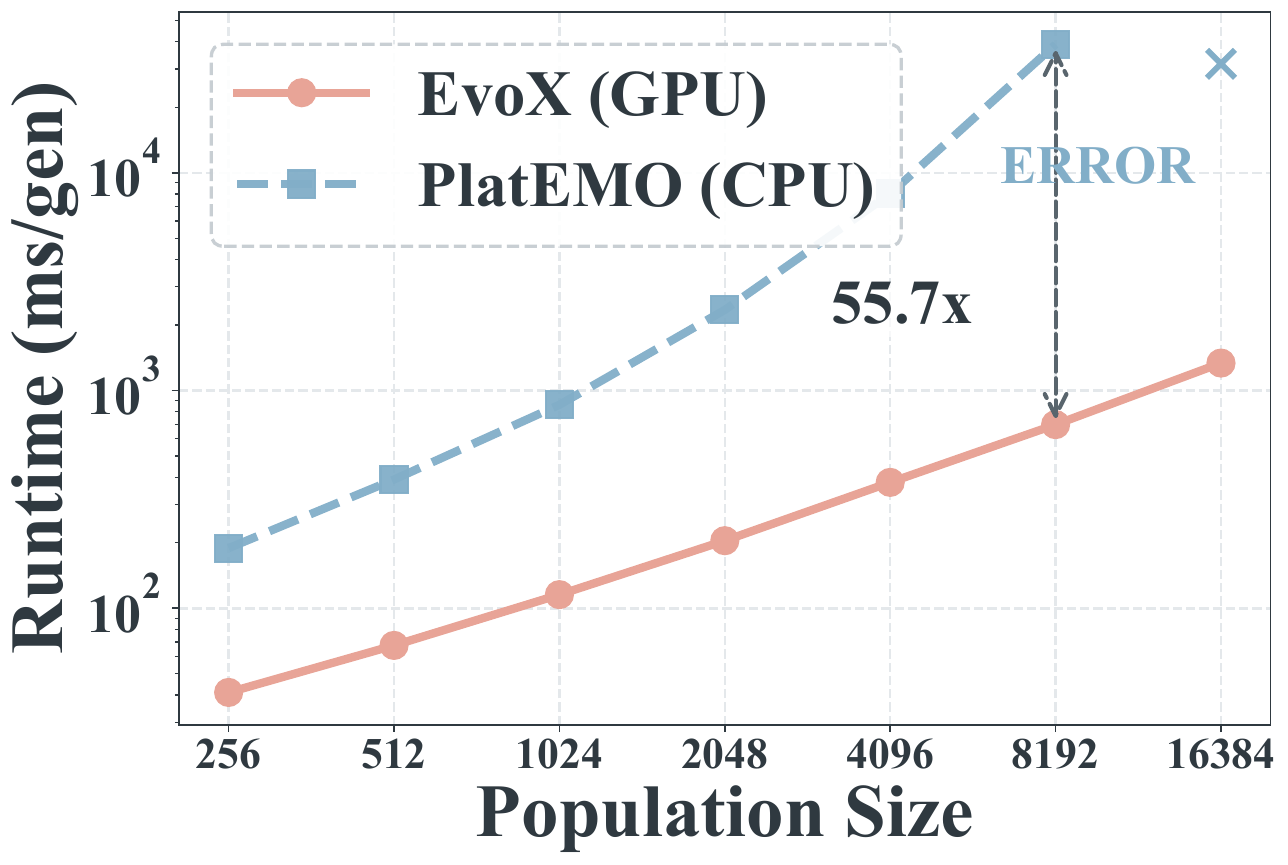}}
\hfill
\subfloat[MOEA-D-DYTS: varying $D$]{\includegraphics[width=0.22\textwidth]{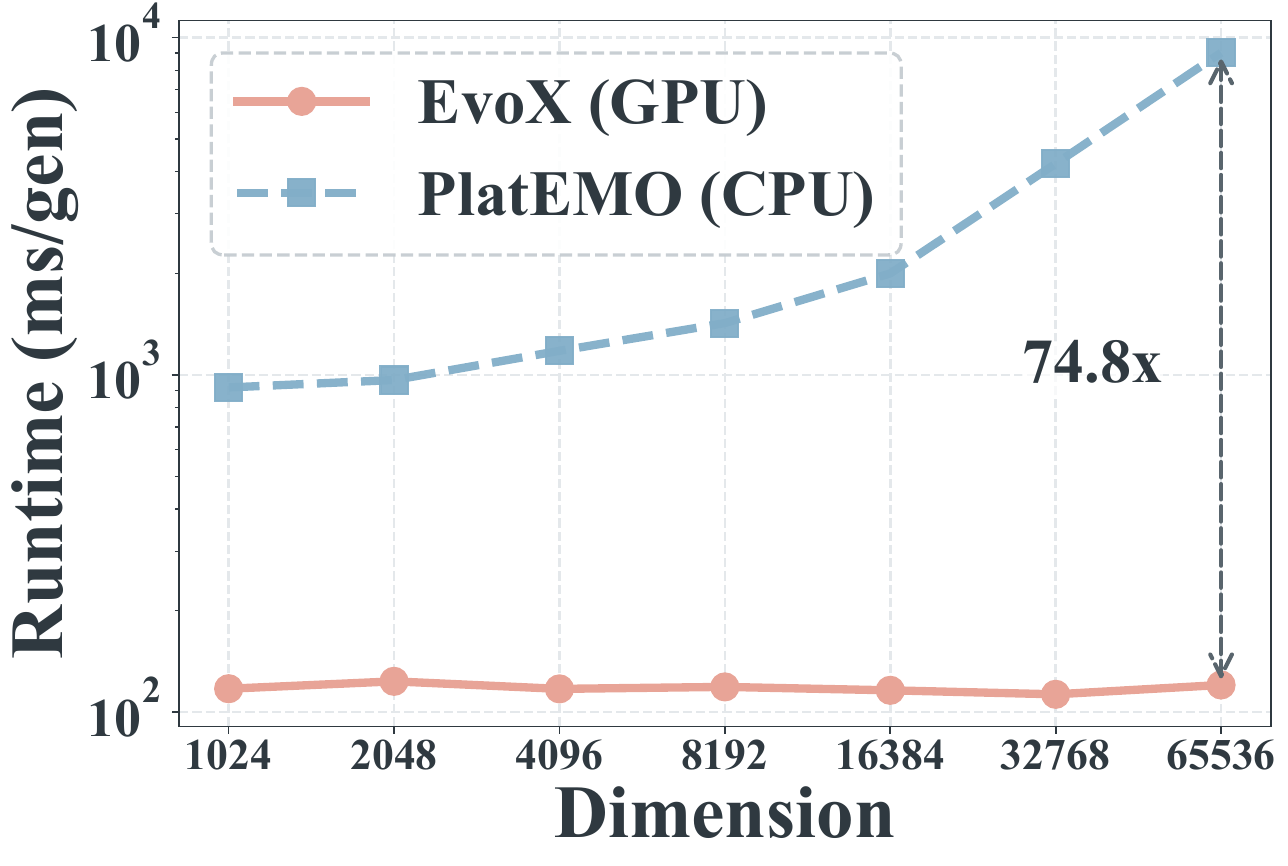}}
\hfill
\subfloat[MOEA-D-FRRMAB: varying $N$]{\includegraphics[width=0.22\textwidth]{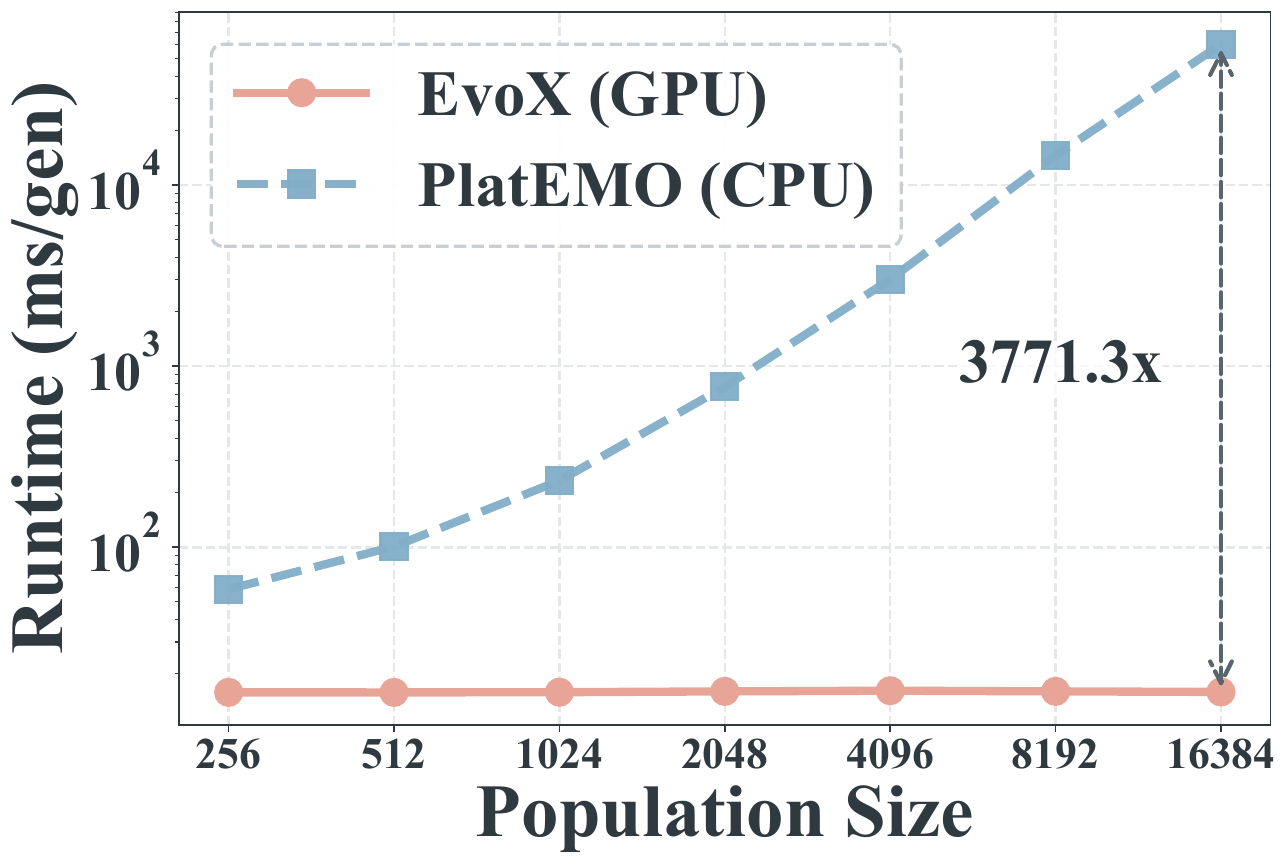}}
\hfill
\subfloat[MOEA-D-FRRMAB: varying $D$]{\includegraphics[width=0.22\textwidth]{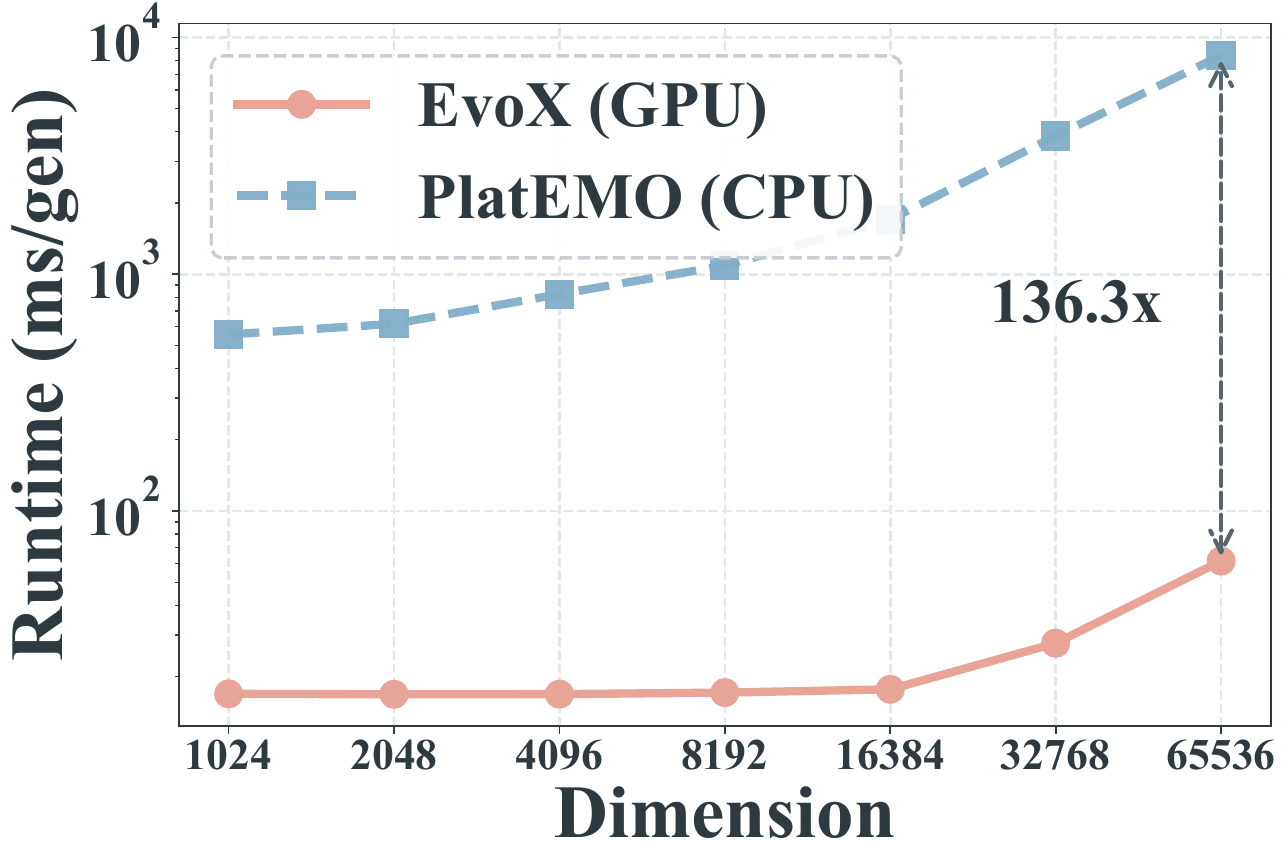}}
\\[-1mm]
\subfloat[MOEA-D-PaS: varying $N$]{\includegraphics[width=0.22\textwidth]{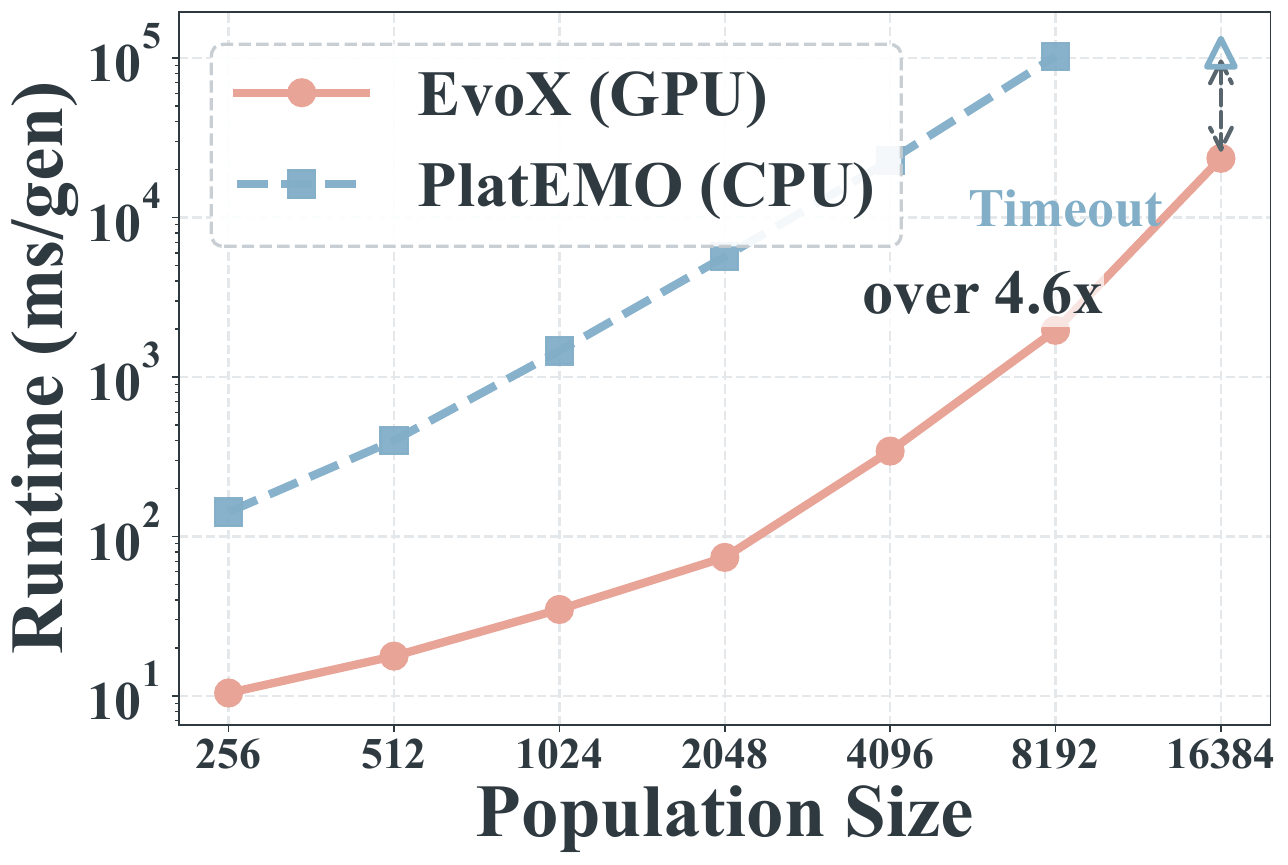}}
\hfill
\subfloat[MOEA-D-PaS: varying $D$]{\includegraphics[width=0.22\textwidth]{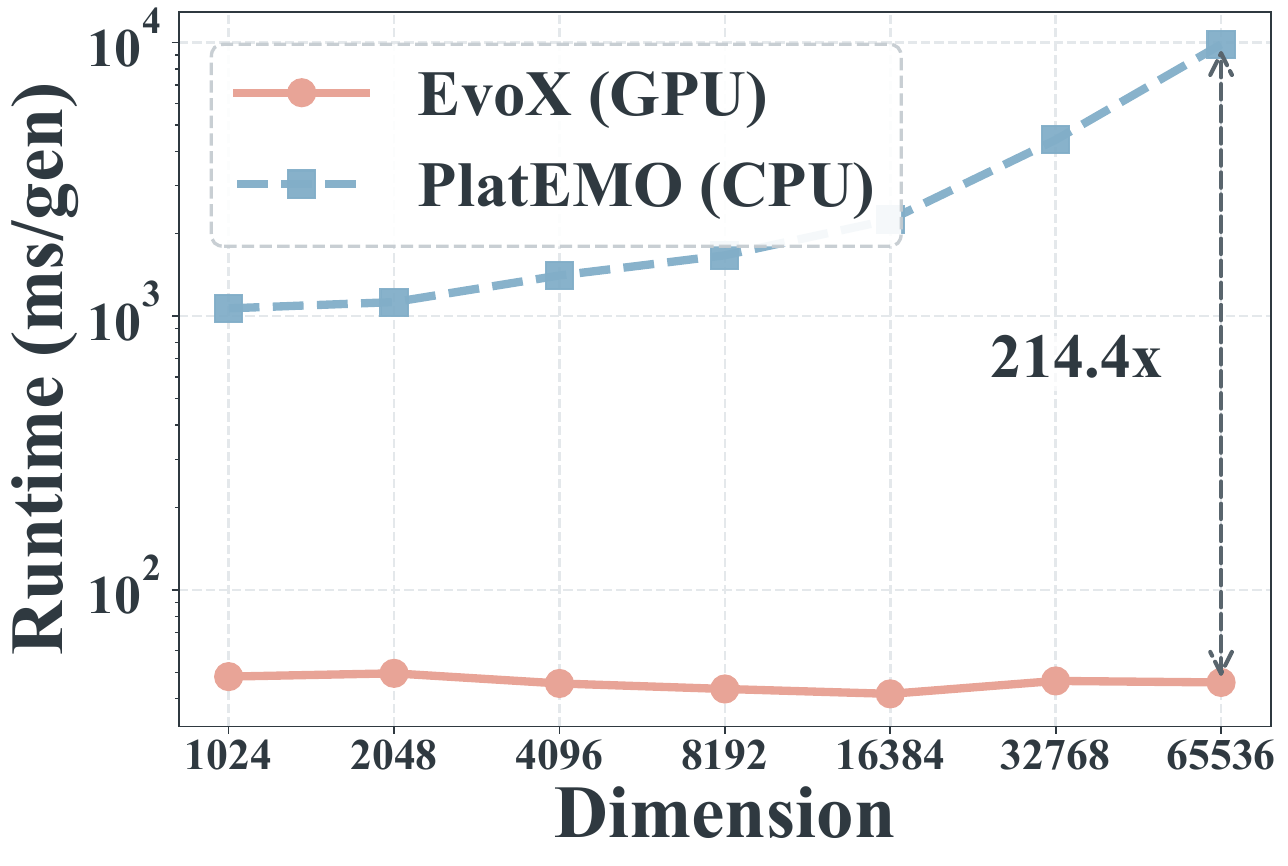}}
\hfill
\subfloat[MOEA-D-URAW: varying $N$]{\includegraphics[width=0.22\textwidth]{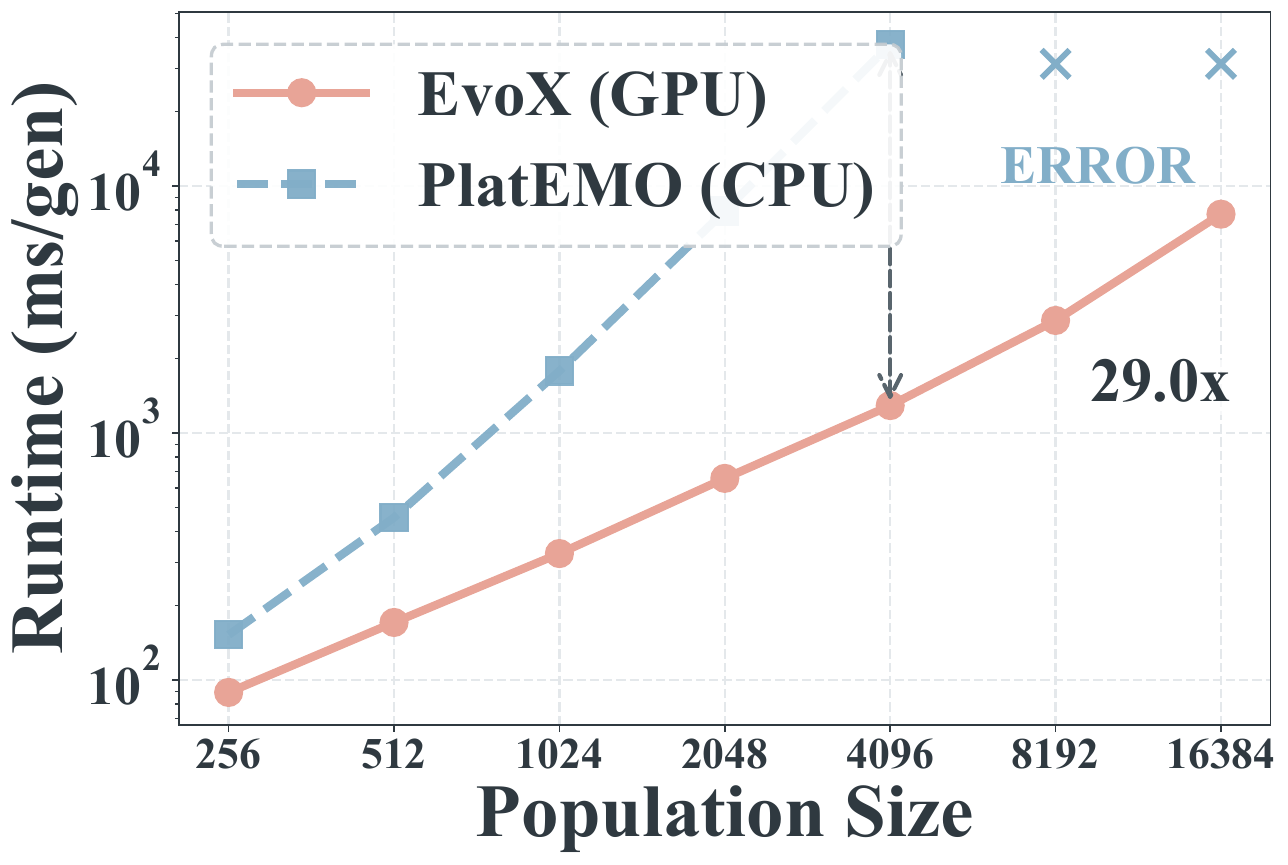}}
\hfill
\subfloat[MOEA-D-URAW: varying $D$]{\includegraphics[width=0.22\textwidth]{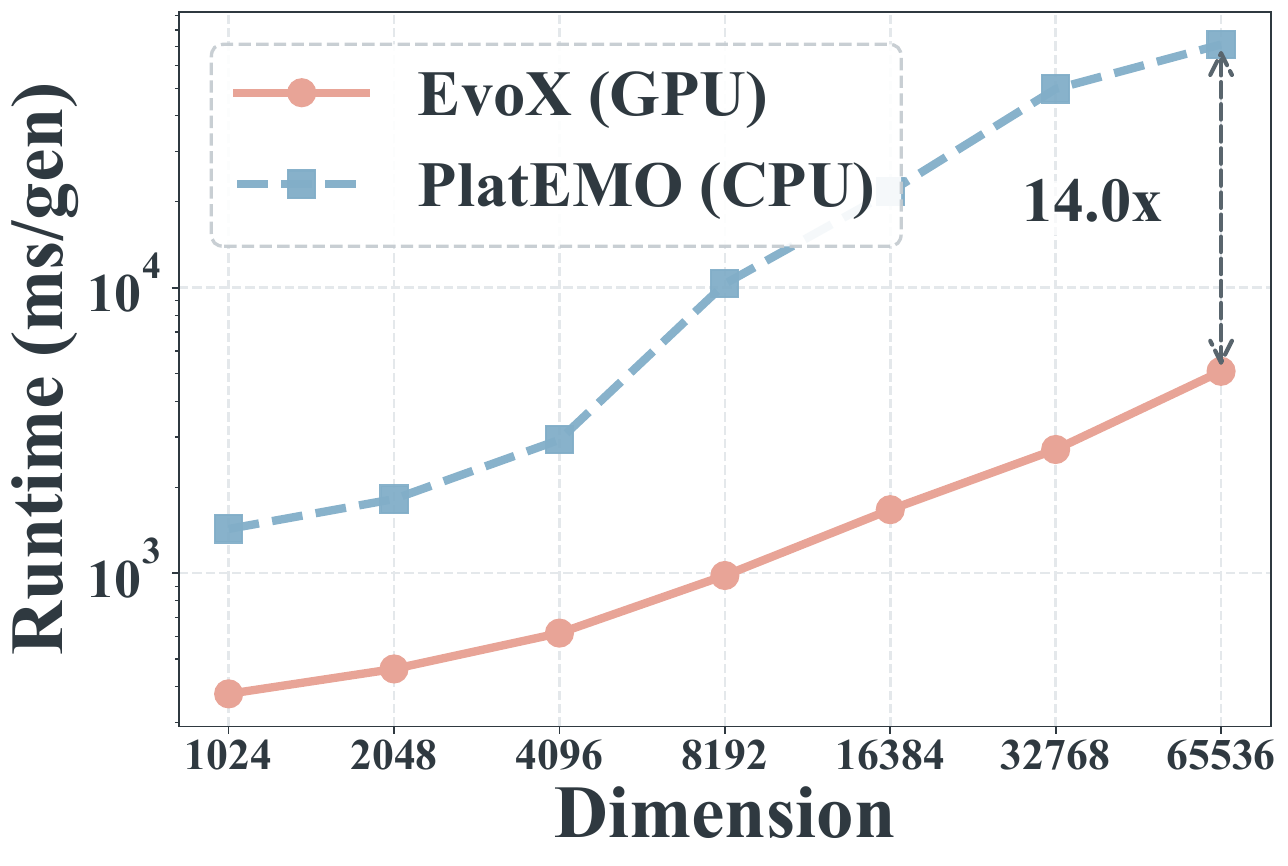}}
\\[-1mm]
\subfloat[NSBiDiCo: varying $N$]{\includegraphics[width=0.22\textwidth]{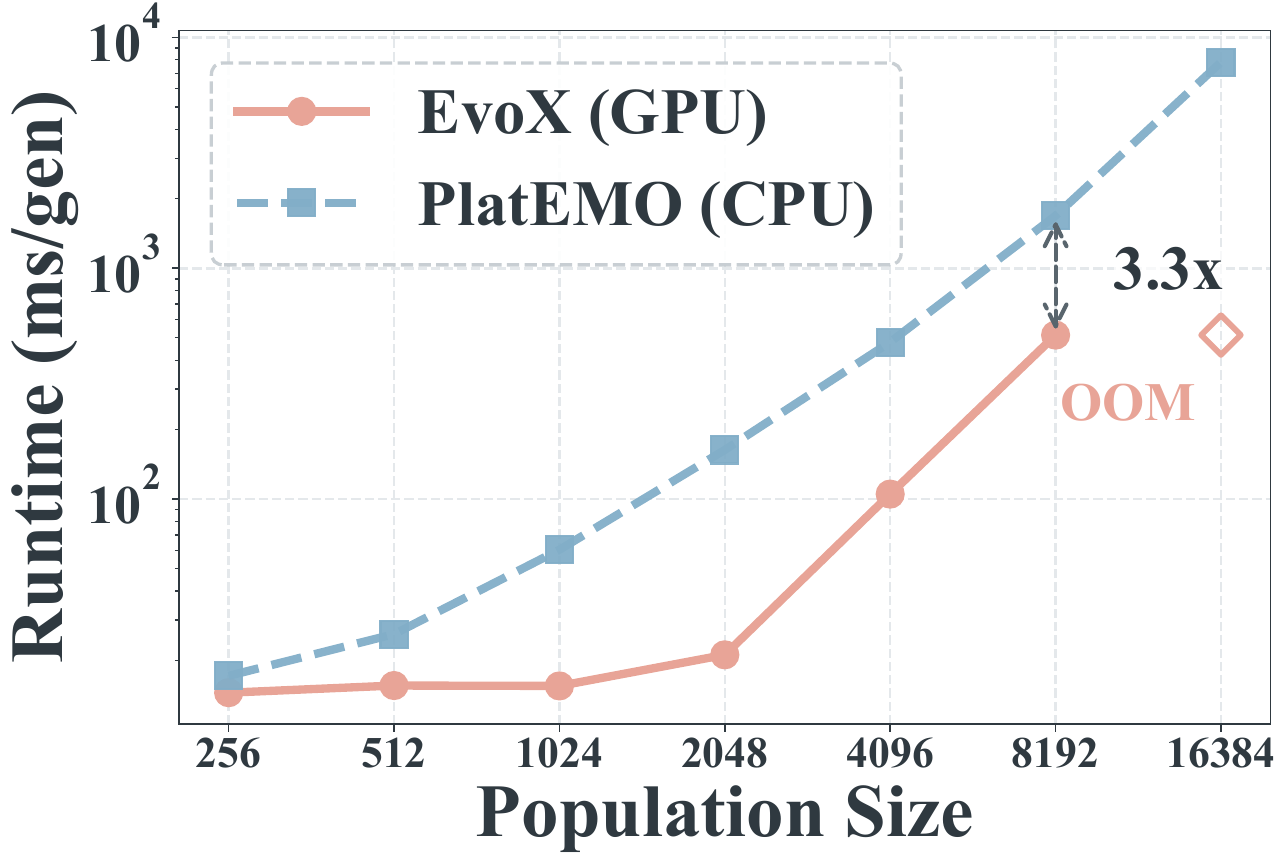}}
\hfill
\subfloat[NSBiDiCo: varying $D$]{\includegraphics[width=0.22\textwidth]{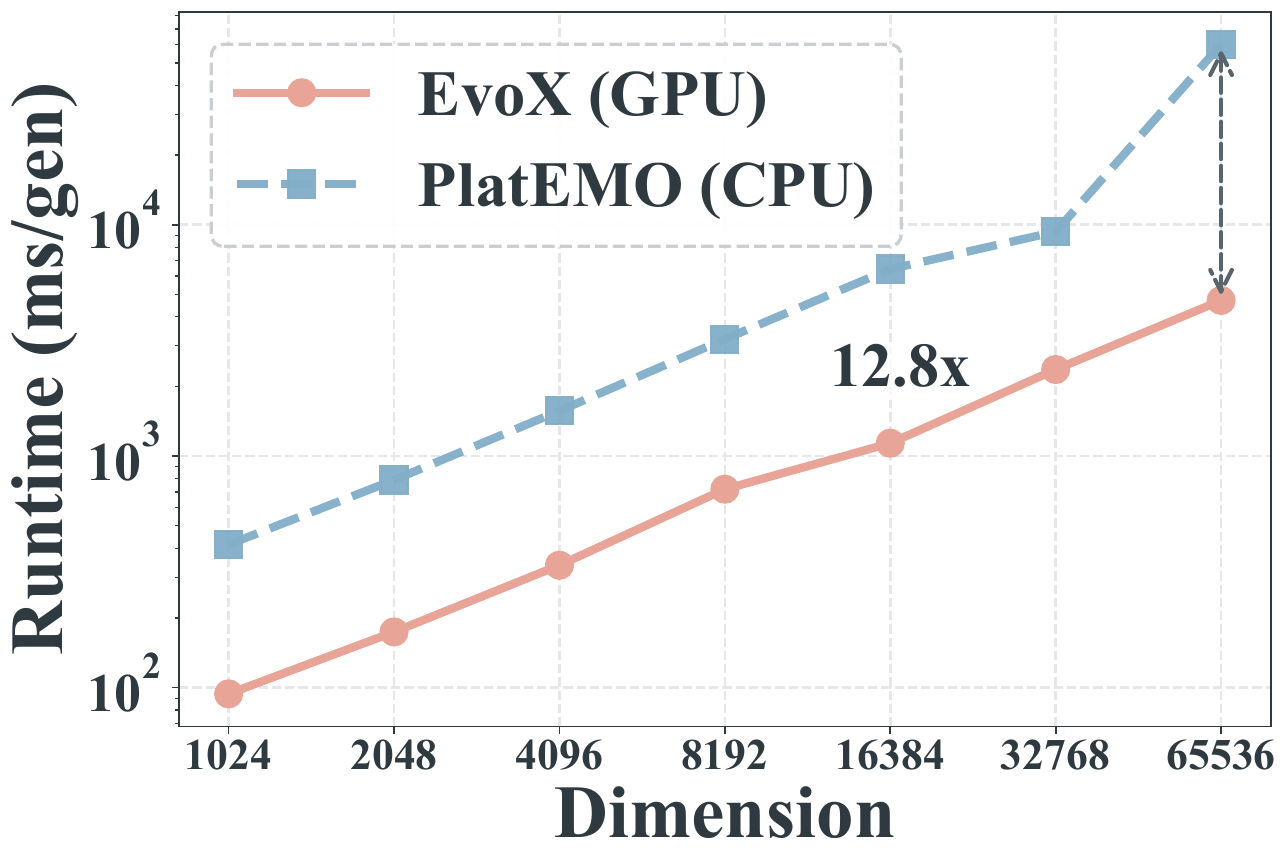}}
\hfill
\subfloat[NSGA-II-SDR: varying $N$]{\includegraphics[width=0.22\textwidth]{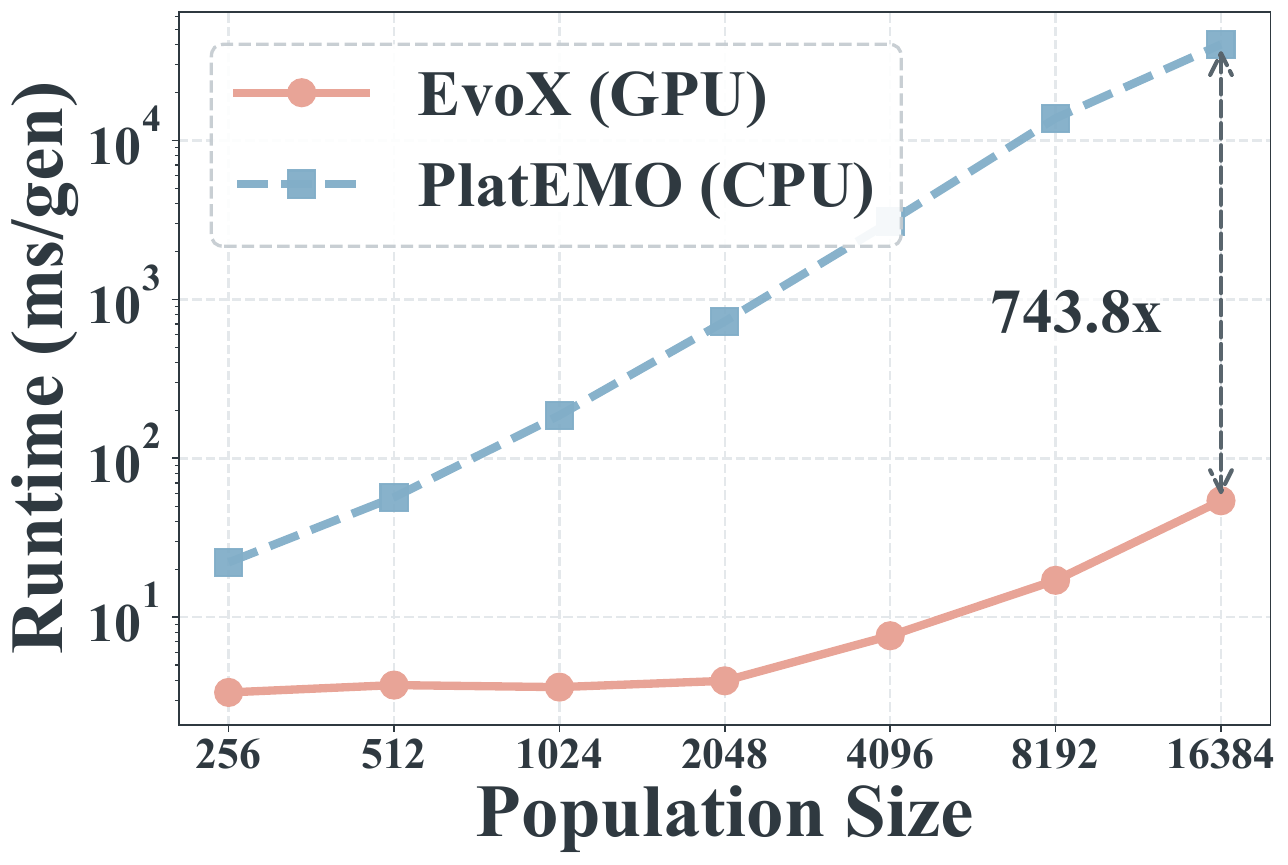}}
\hfill
\subfloat[NSGA-II-SDR: varying $D$]{\includegraphics[width=0.22\textwidth]{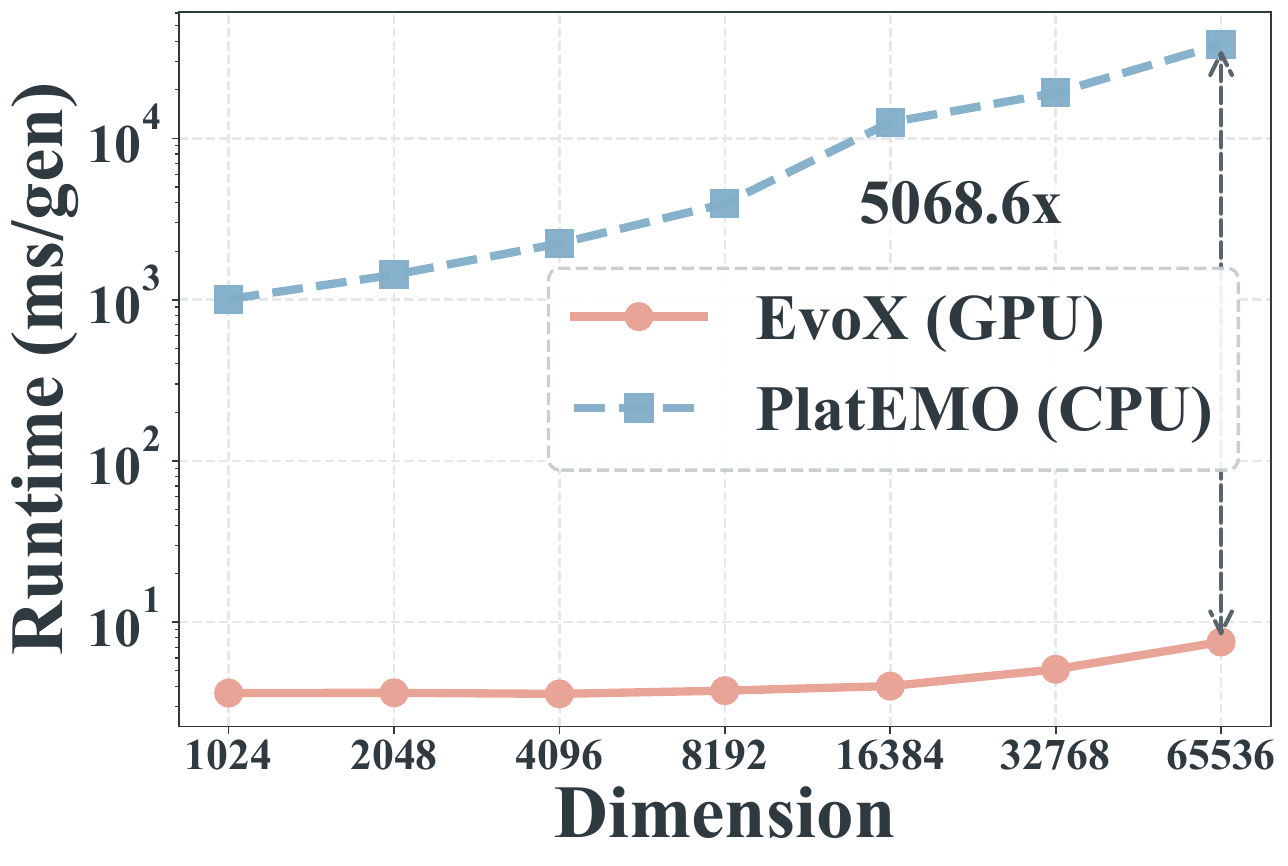}}
\\[-1mm]
\subfloat[OSP-NSDE: varying $N$]{\includegraphics[width=0.22\textwidth]{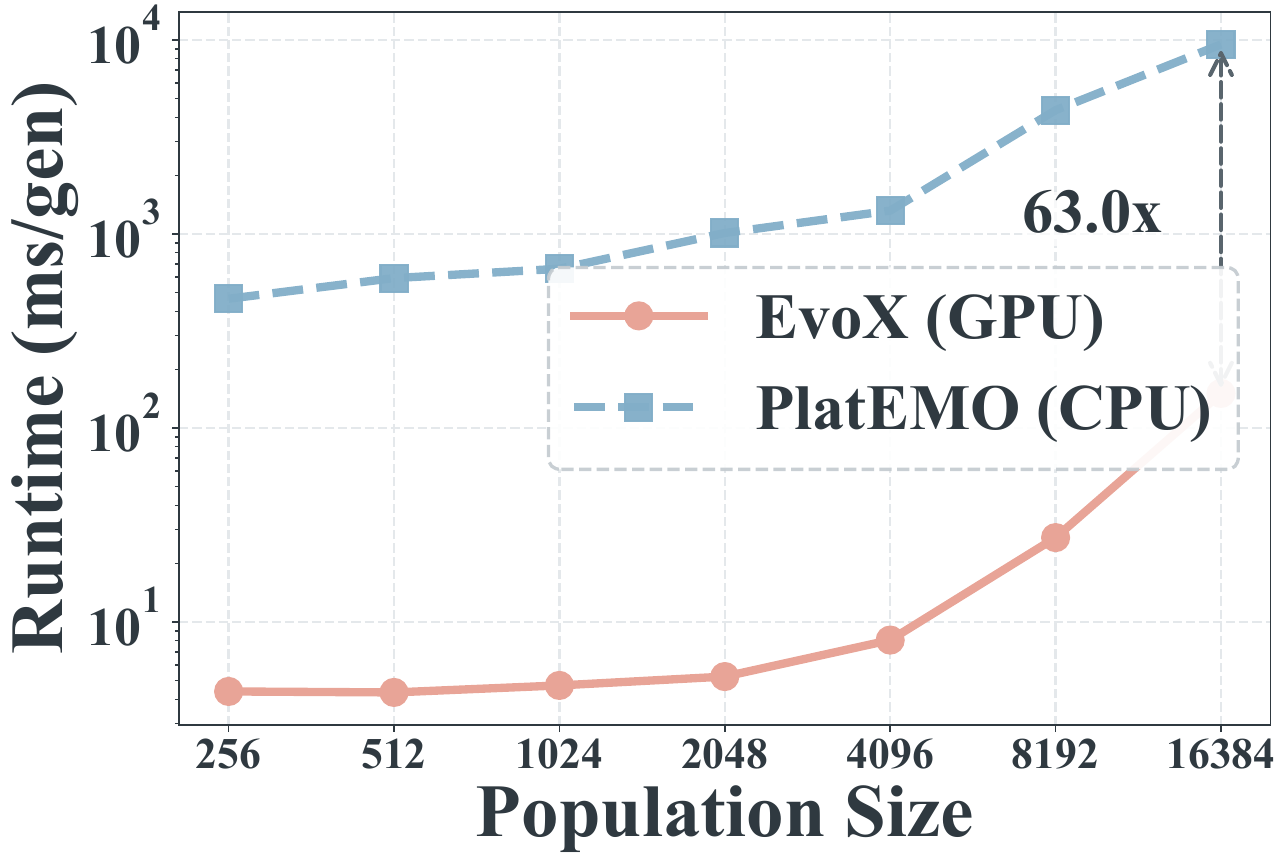}}
\hfill
\subfloat[OSP-NSDE: varying $D$]{\includegraphics[width=0.22\textwidth]{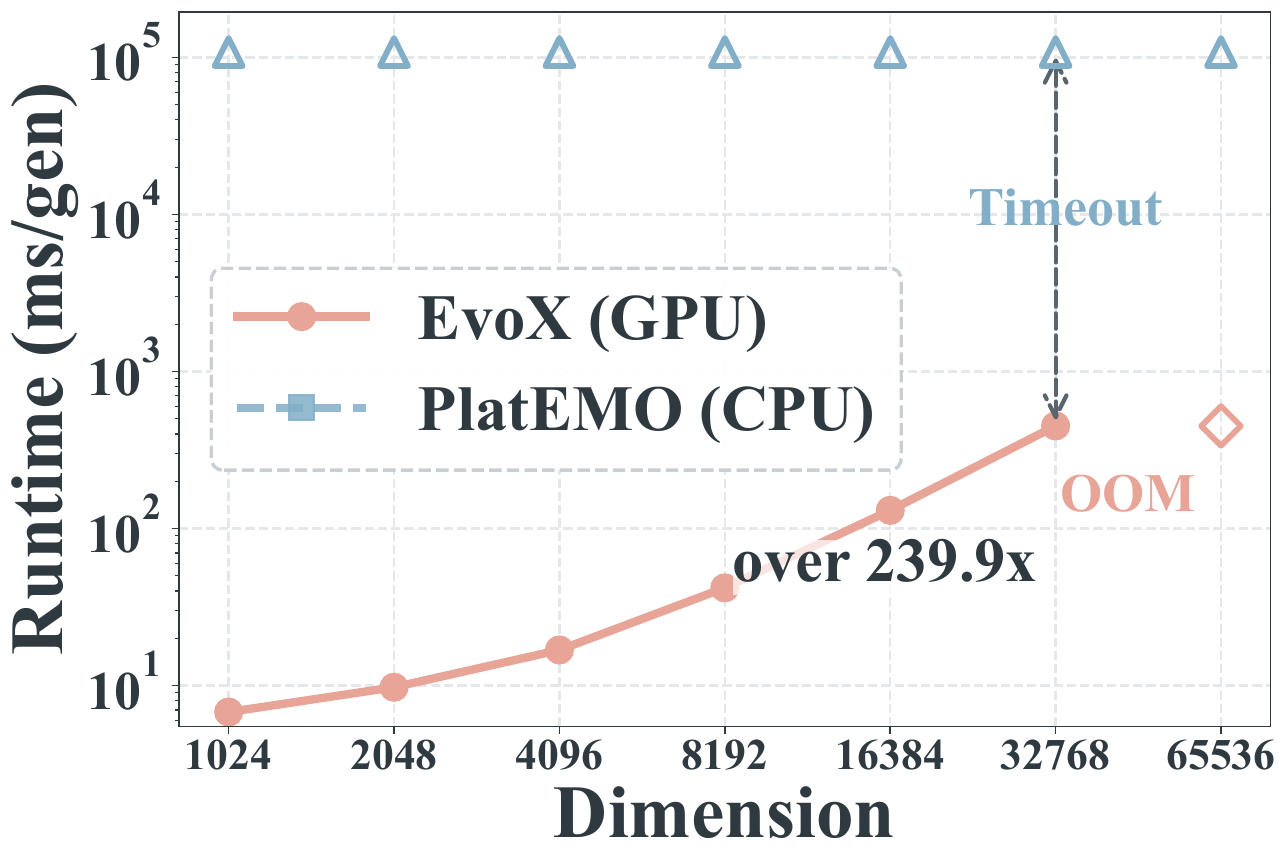}}
\hfill
\subfloat[PESA-II: varying $N$]{\includegraphics[width=0.22\textwidth]{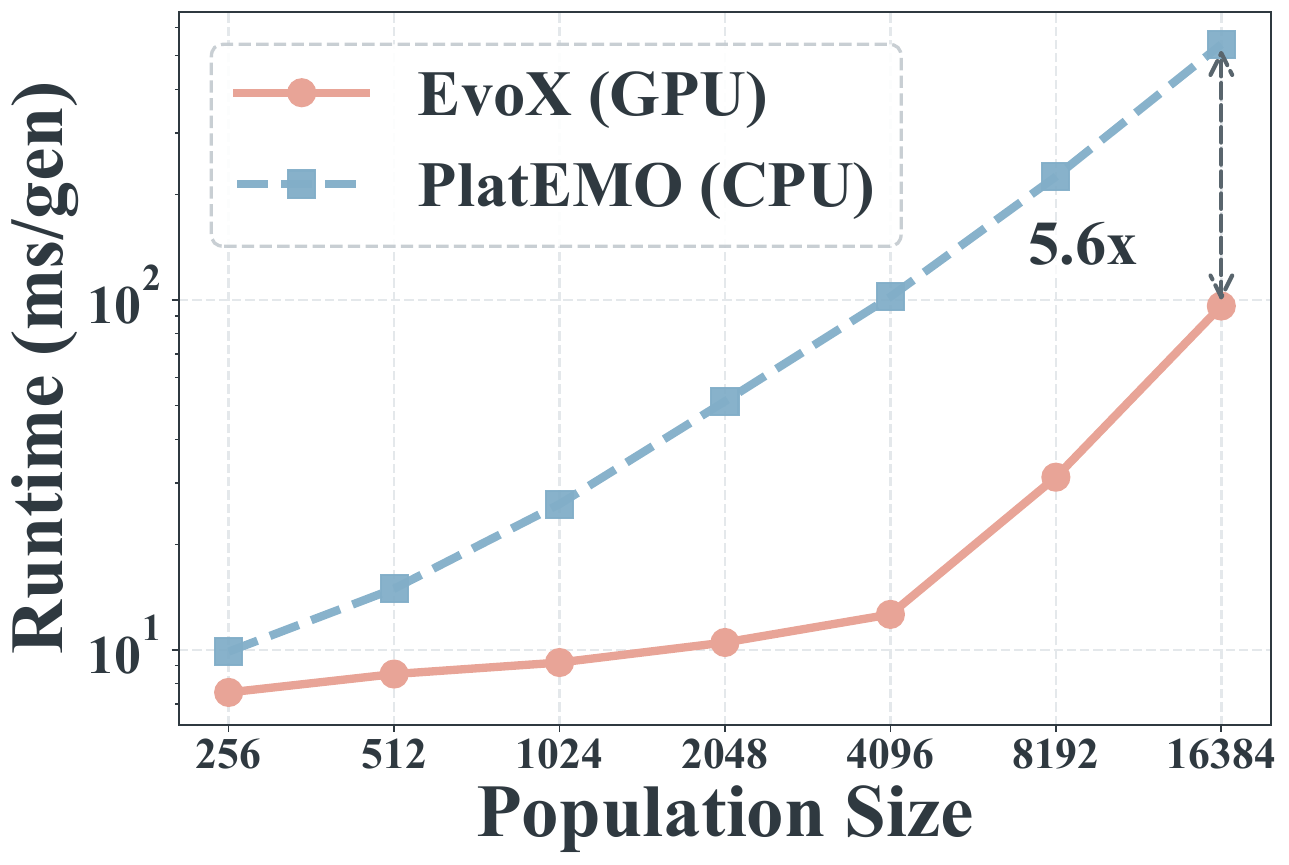}}
\hfill
\subfloat[PESA-II: varying $D$]{\includegraphics[width=0.22\textwidth]{figures/exp3_scaling/curves/scaling_dim_comparison_PESA-II_dim.pdf}}
\caption{Complete population-size ($N$) and decision-dimension ($D$) scaling results for MOEA-D-DRA through PESA-II (Part III of V). Adjacent panels report the two scaling axes for each algorithm.}
\label{fig:supp_scaling_curves_03}
\end{figure}

\begin{figure}[p]
\centering
\subfloat[PICEA-g: varying $N$]{\includegraphics[width=0.22\textwidth]{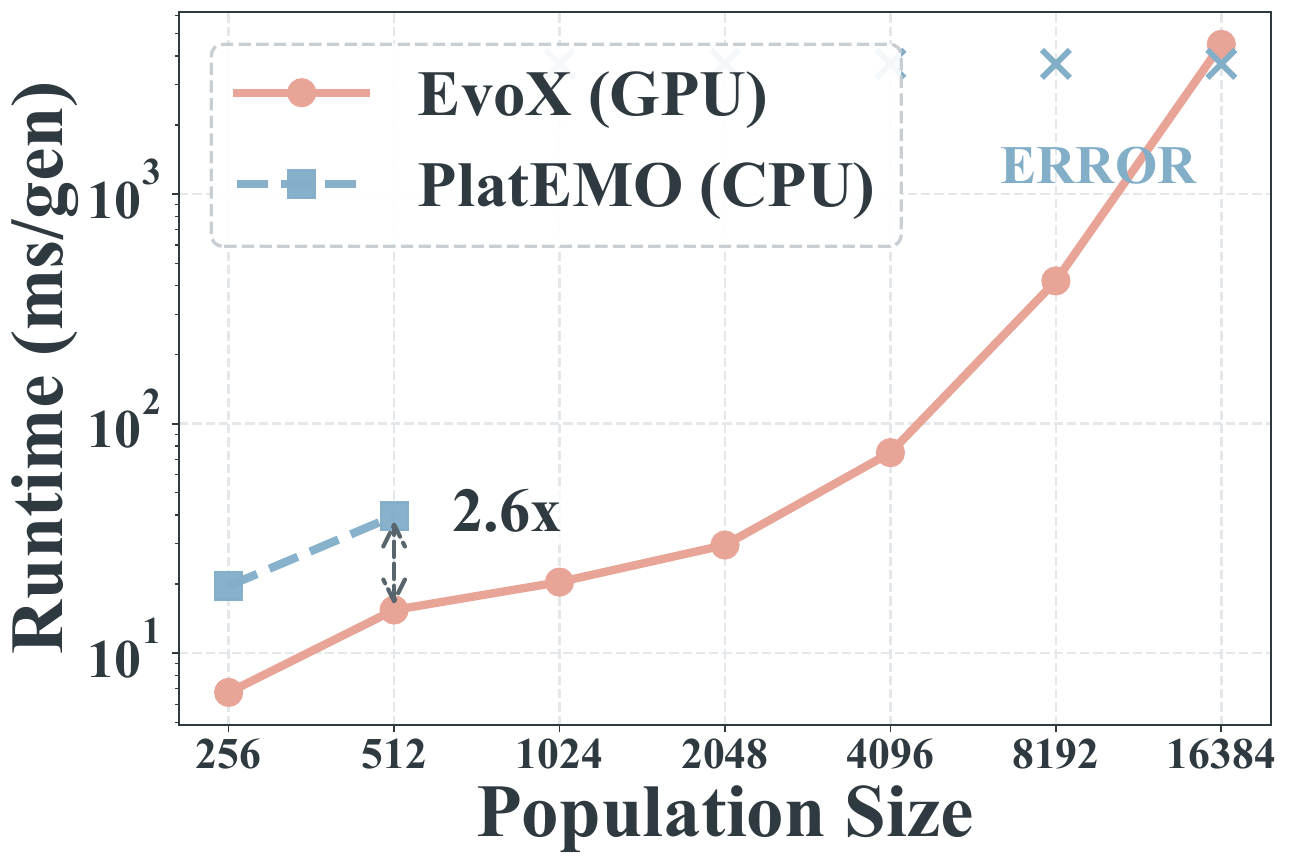}}
\hfill
\subfloat[PICEA-g: varying $D$]{\includegraphics[width=0.22\textwidth]{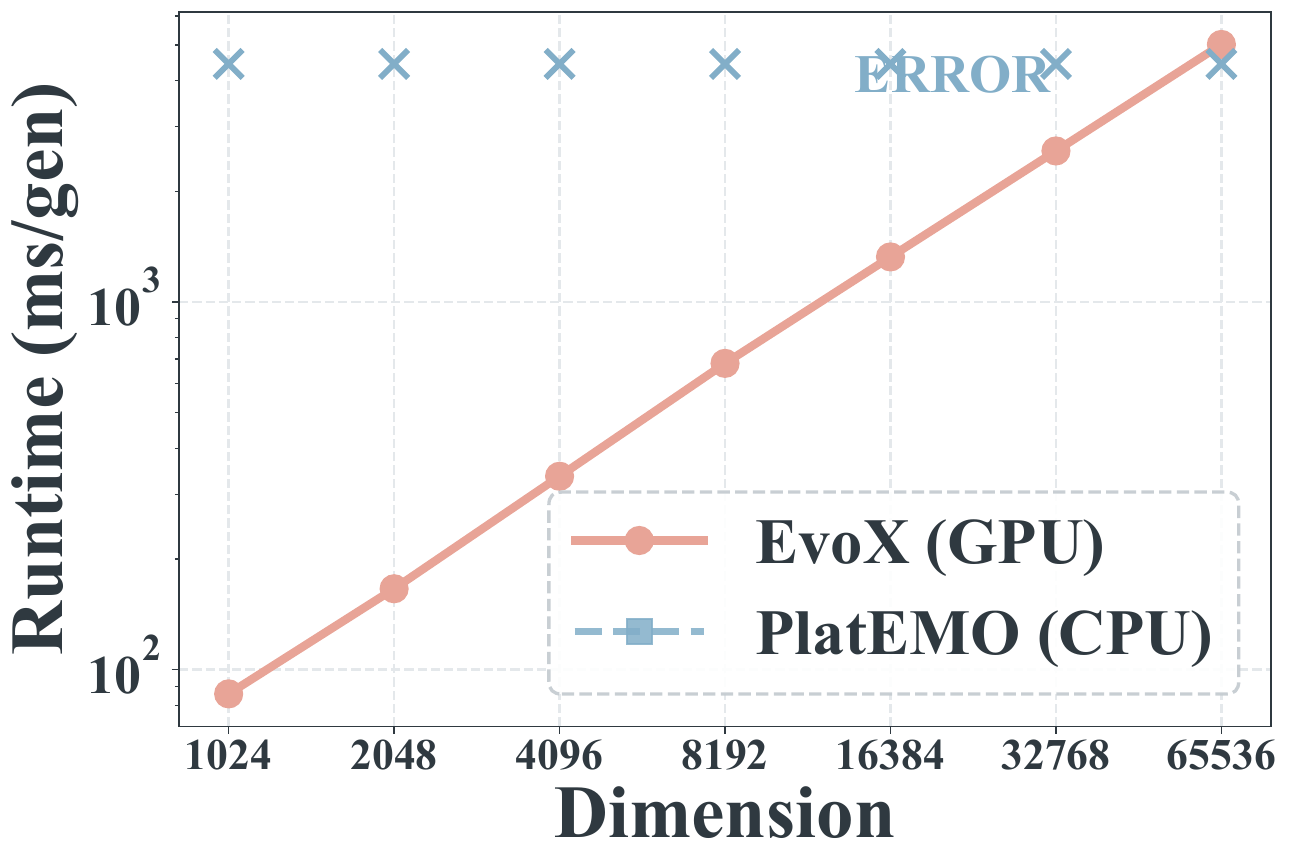}}
\hfill
\subfloat[PREA: varying $N$]{\includegraphics[width=0.22\textwidth]{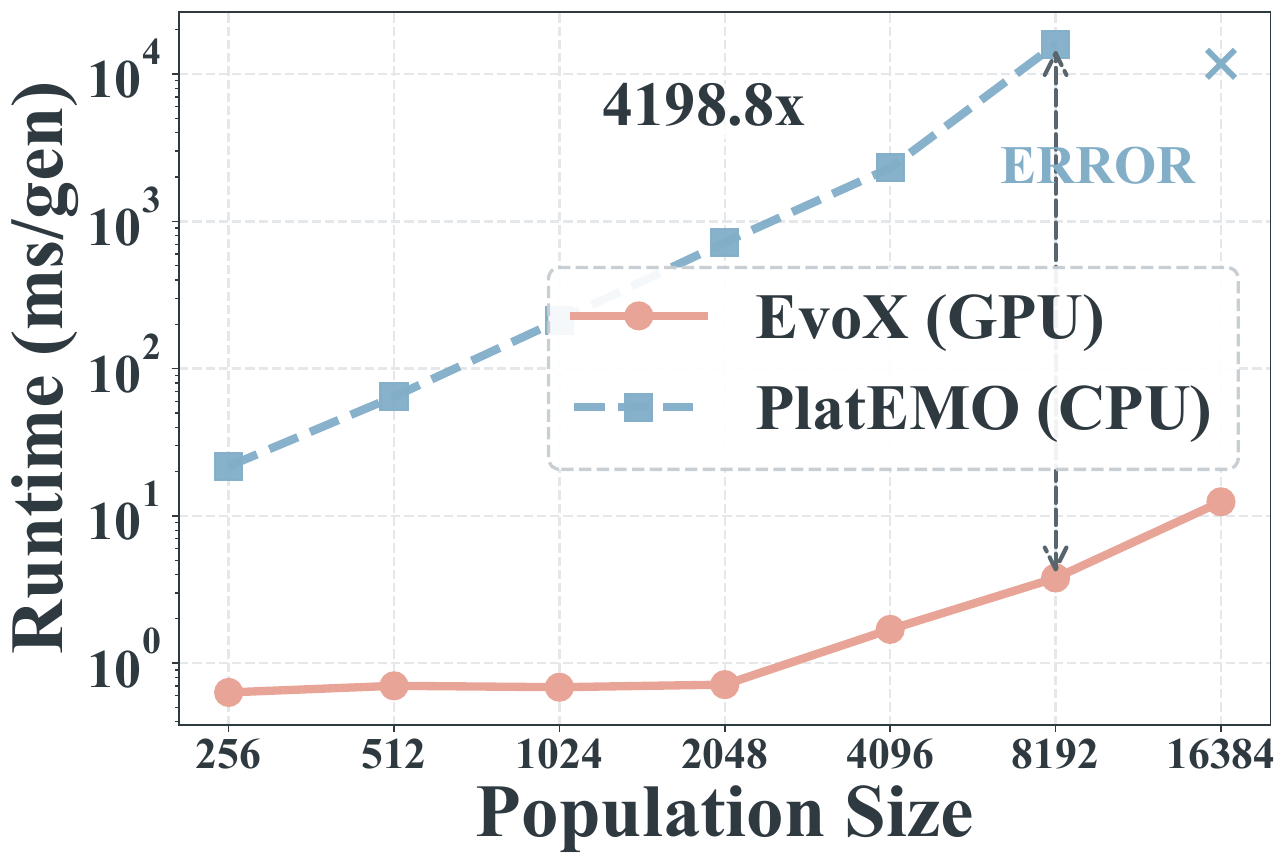}}
\hfill
\subfloat[PREA: varying $D$]{\includegraphics[width=0.22\textwidth]{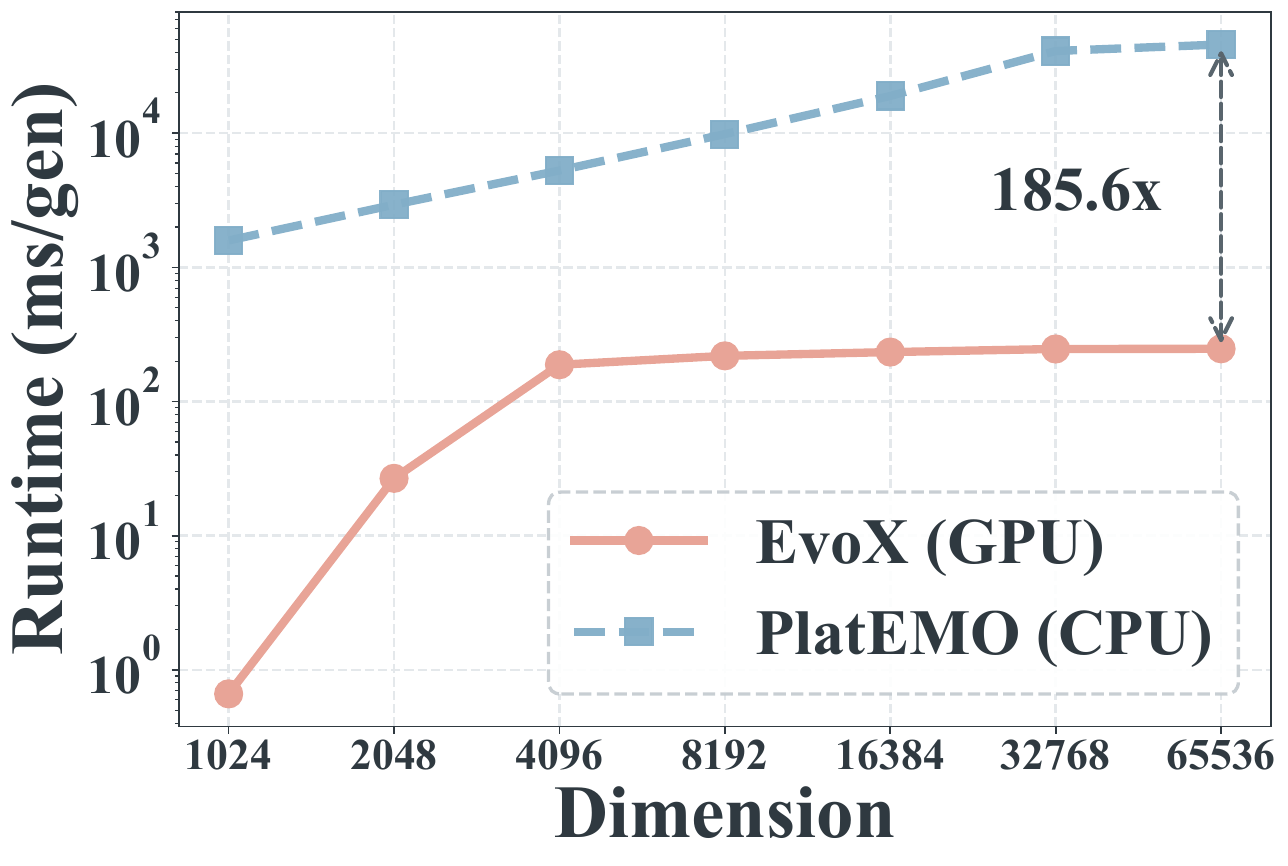}}
\\[-1mm]
\subfloat[SIBEA: varying $N$]{\includegraphics[width=0.22\textwidth]{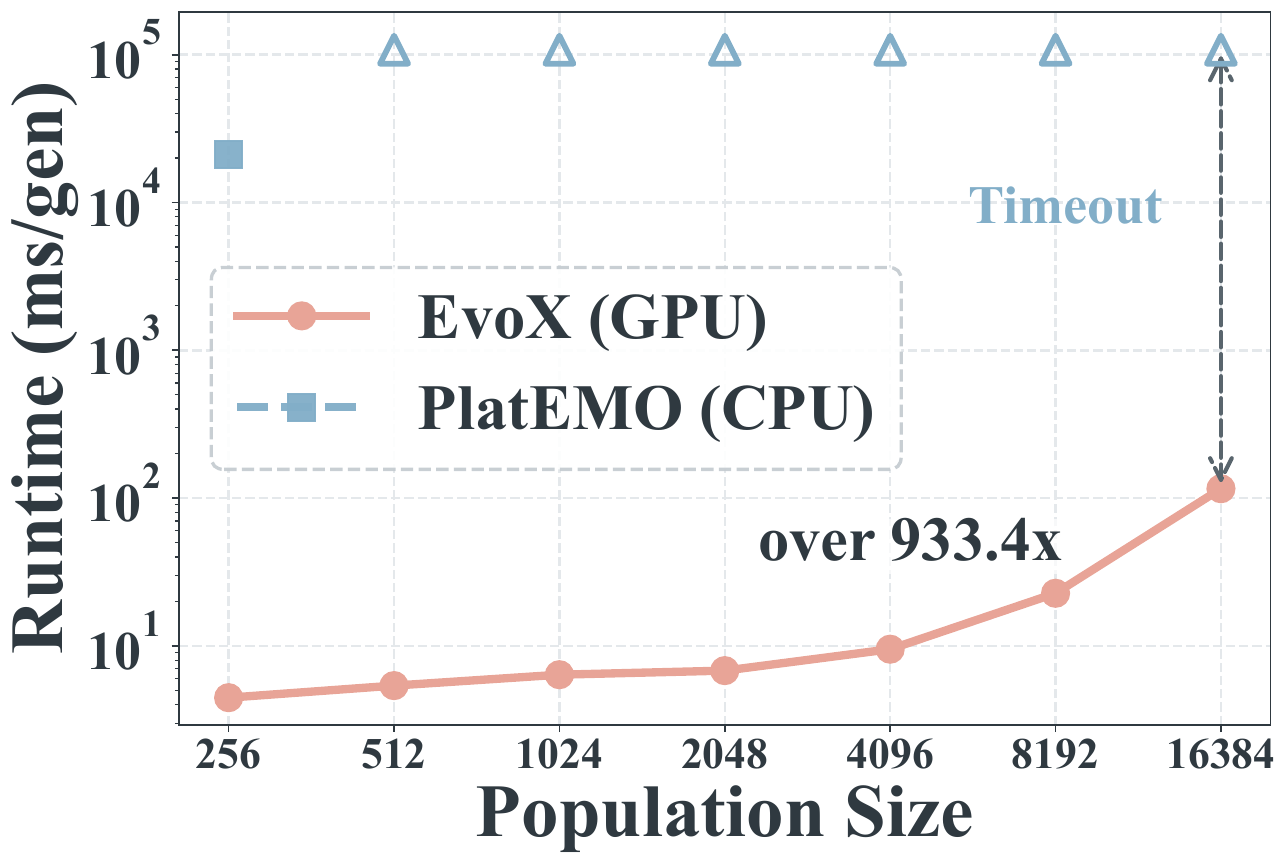}}
\hfill
\subfloat[SIBEA: varying $D$]{\includegraphics[width=0.22\textwidth]{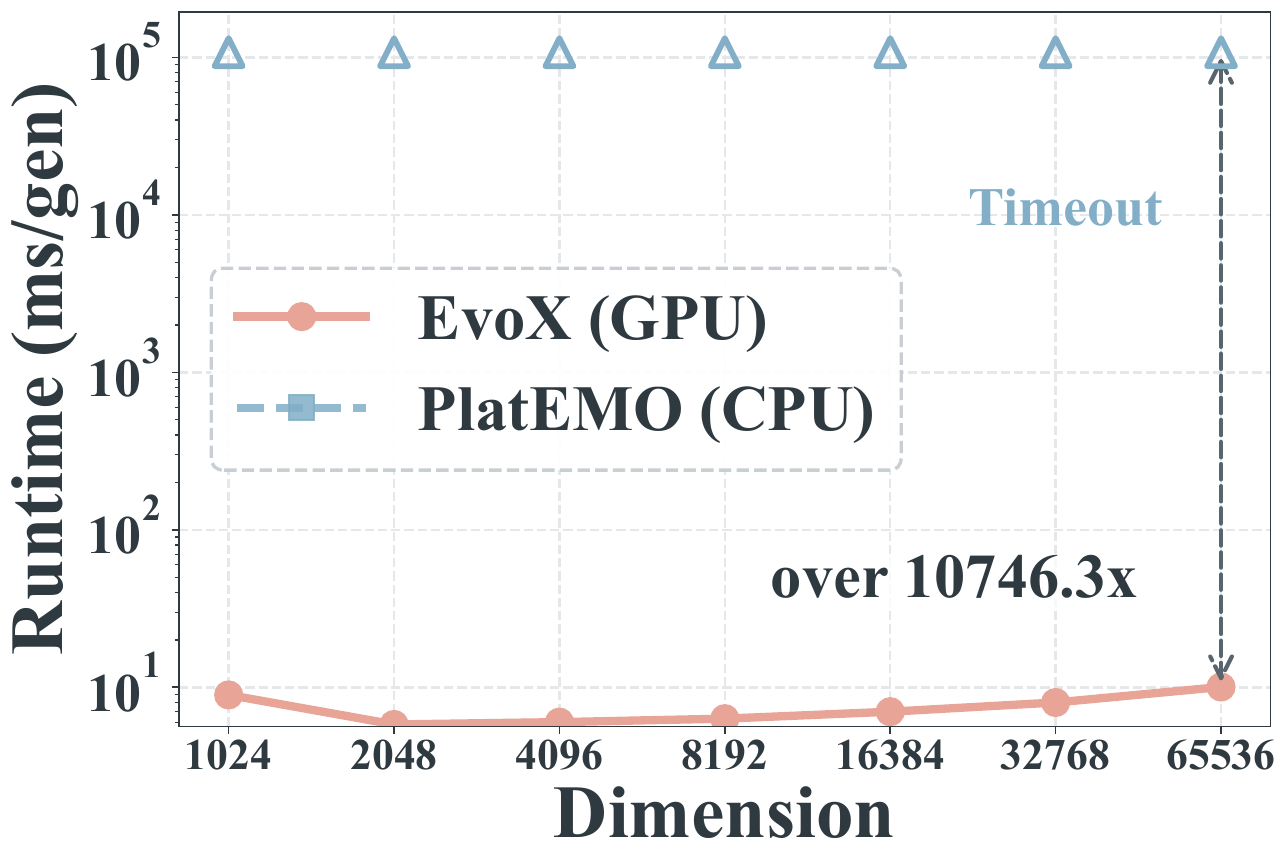}}
\hfill
\subfloat[SMPSO: varying $N$]{\includegraphics[width=0.22\textwidth]{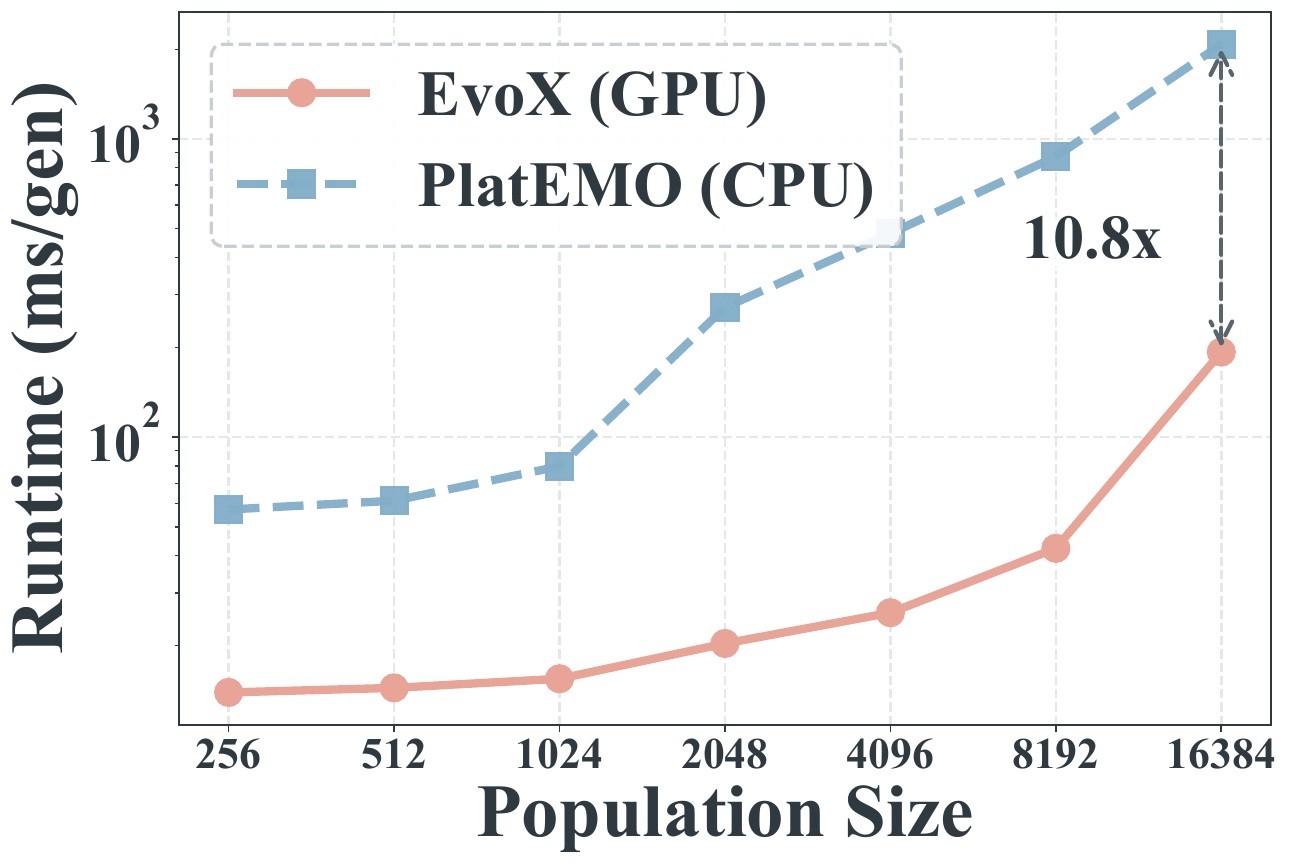}}
\hfill
\subfloat[SMPSO: varying $D$]{\includegraphics[width=0.22\textwidth]{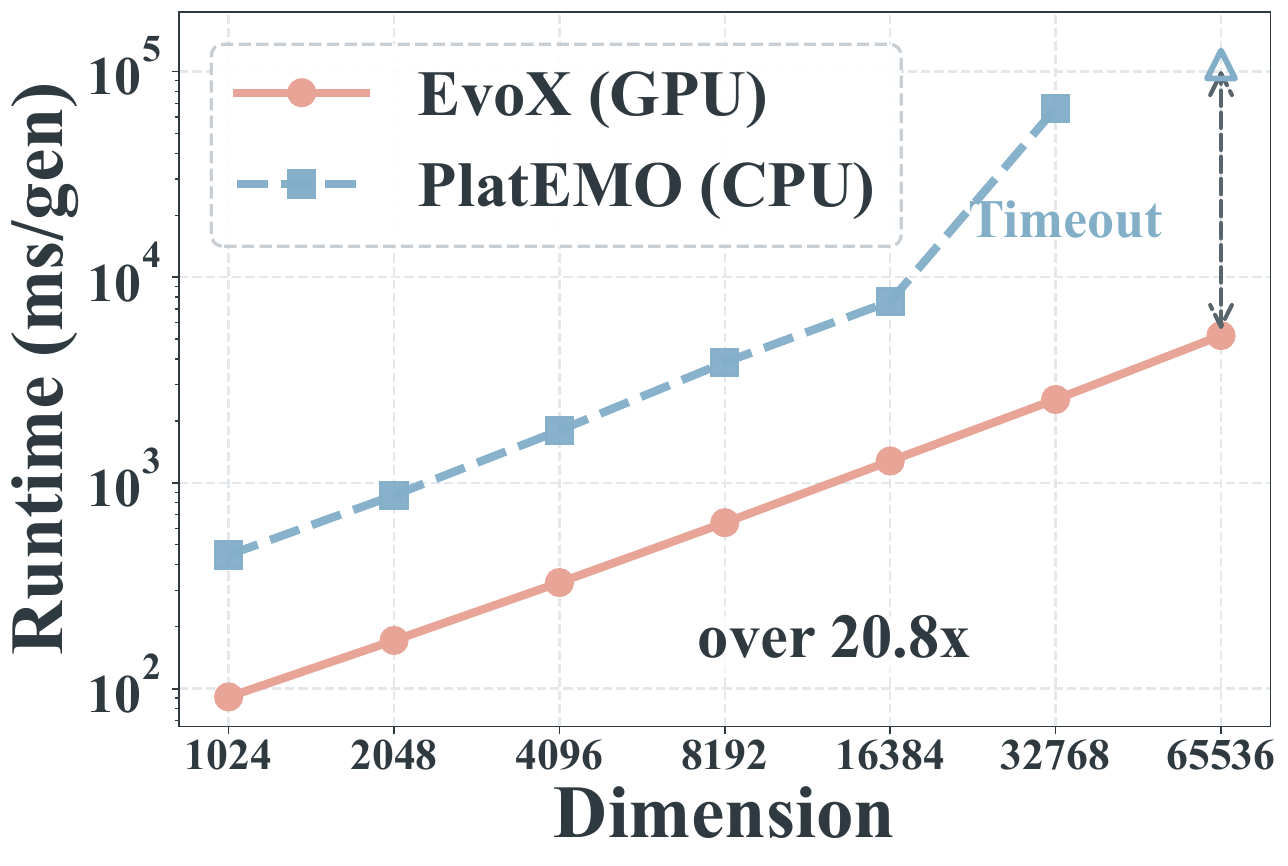}}
\\[-1mm]
\subfloat[S-NSGA-II: varying $N$]{\includegraphics[width=0.22\textwidth]{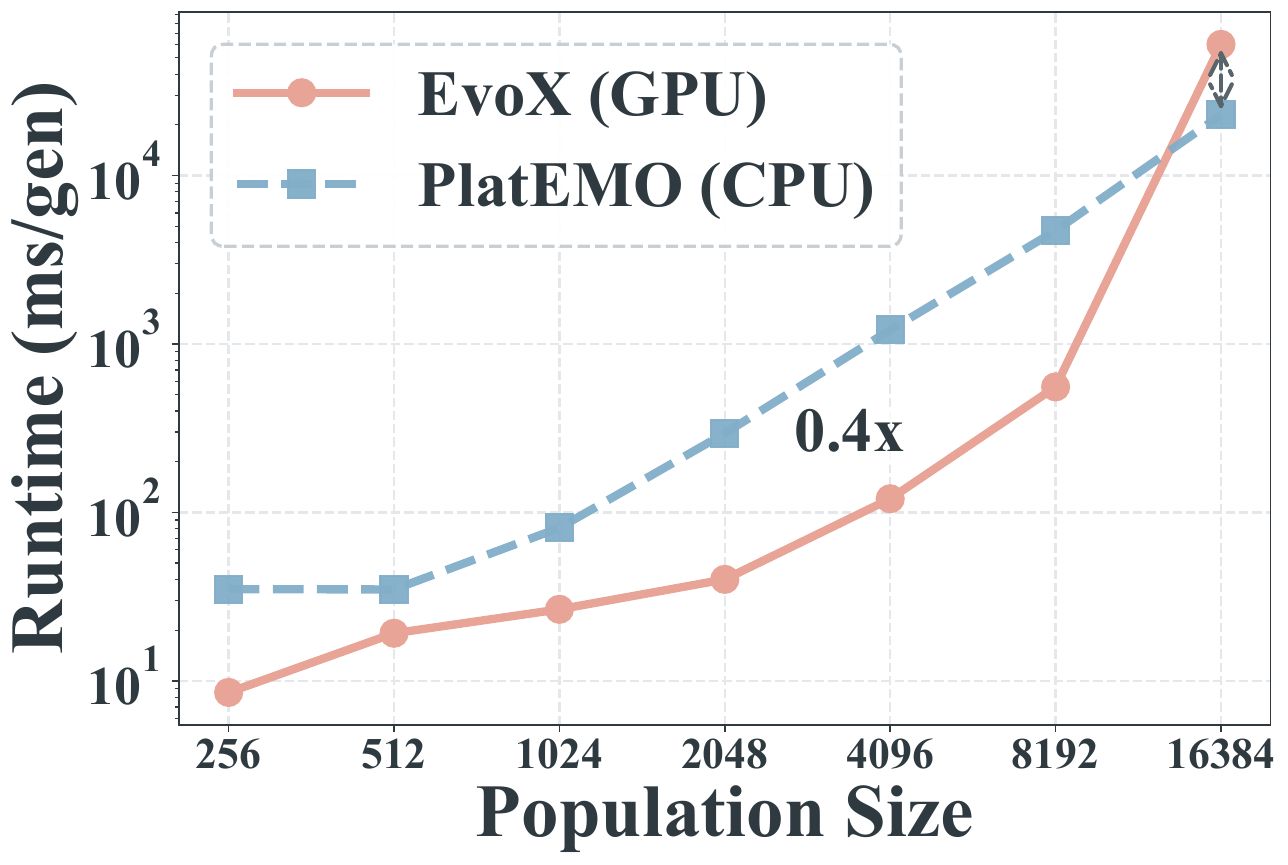}}
\hfill
\subfloat[S-NSGA-II: varying $D$]{\includegraphics[width=0.22\textwidth]{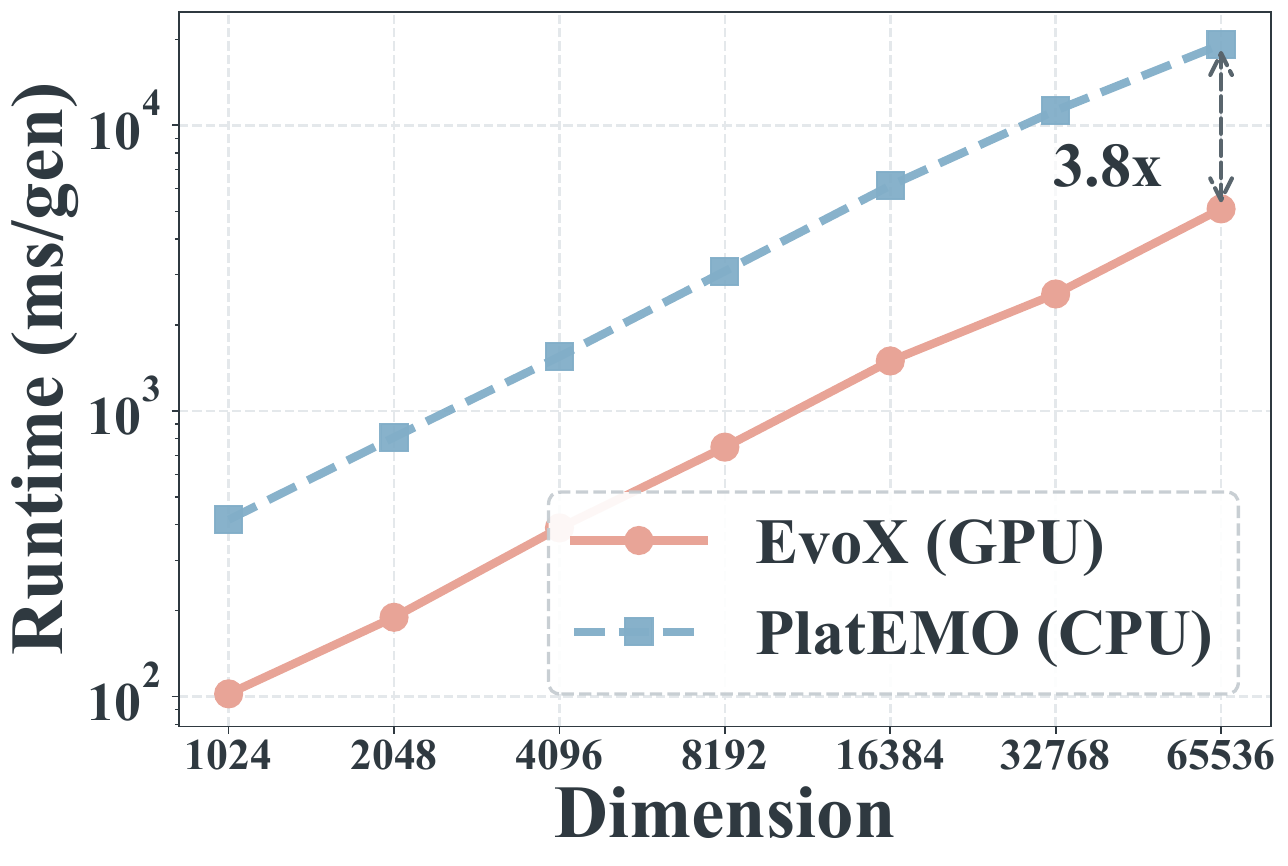}}
\hfill
\subfloat[SparseEA: varying $N$]{\includegraphics[width=0.22\textwidth]{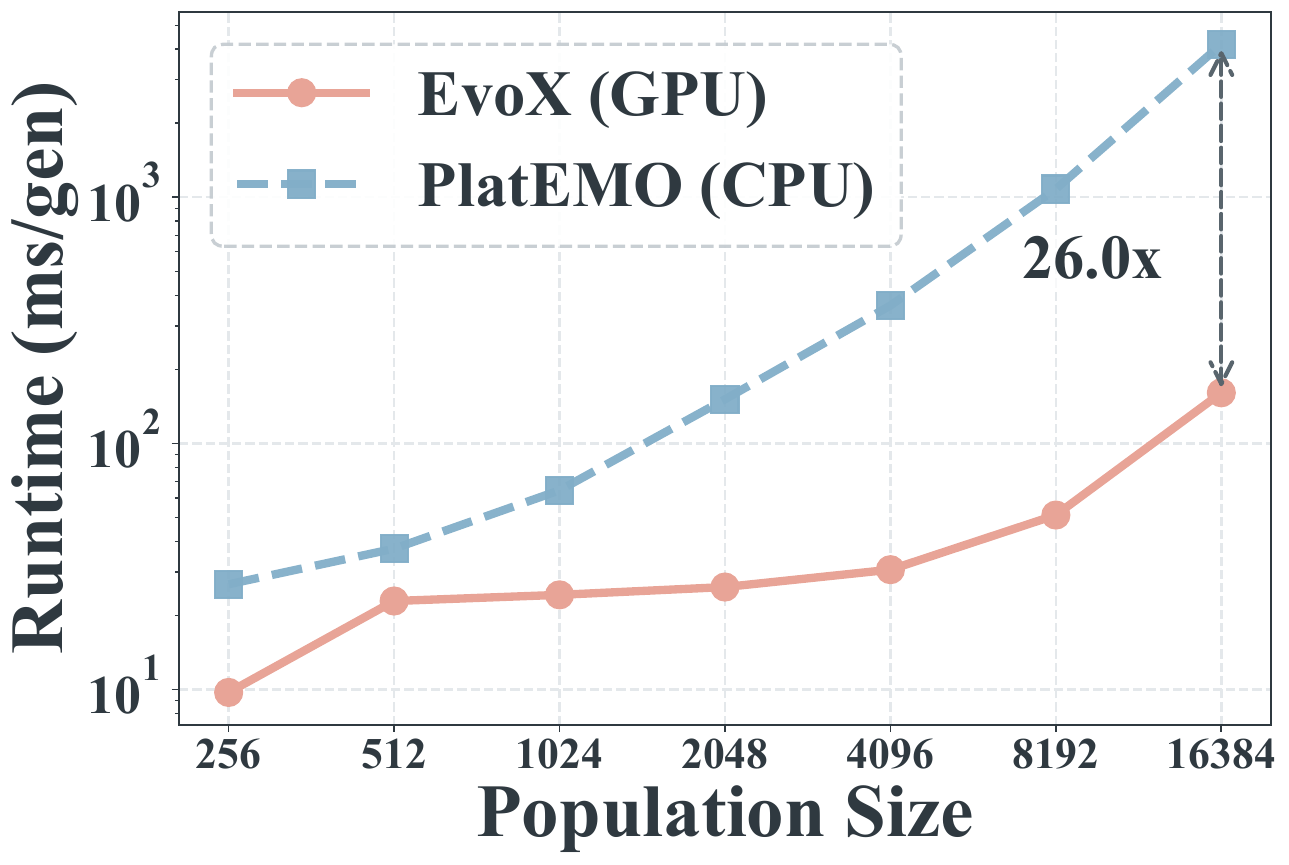}}
\hfill
\subfloat[SparseEA: varying $D$]{\includegraphics[width=0.22\textwidth]{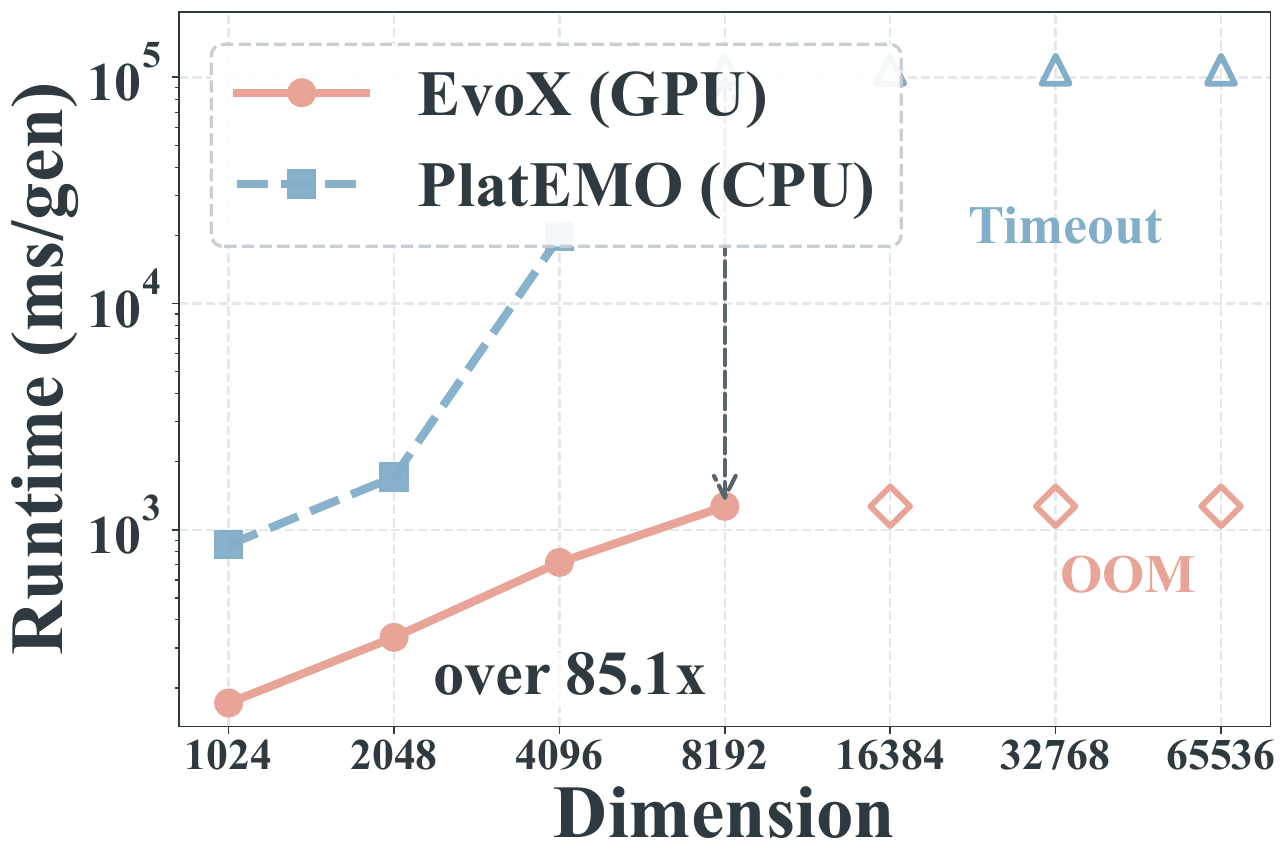}}
\\[-1mm]
\subfloat[SparseEA2: varying $N$]{\includegraphics[width=0.22\textwidth]{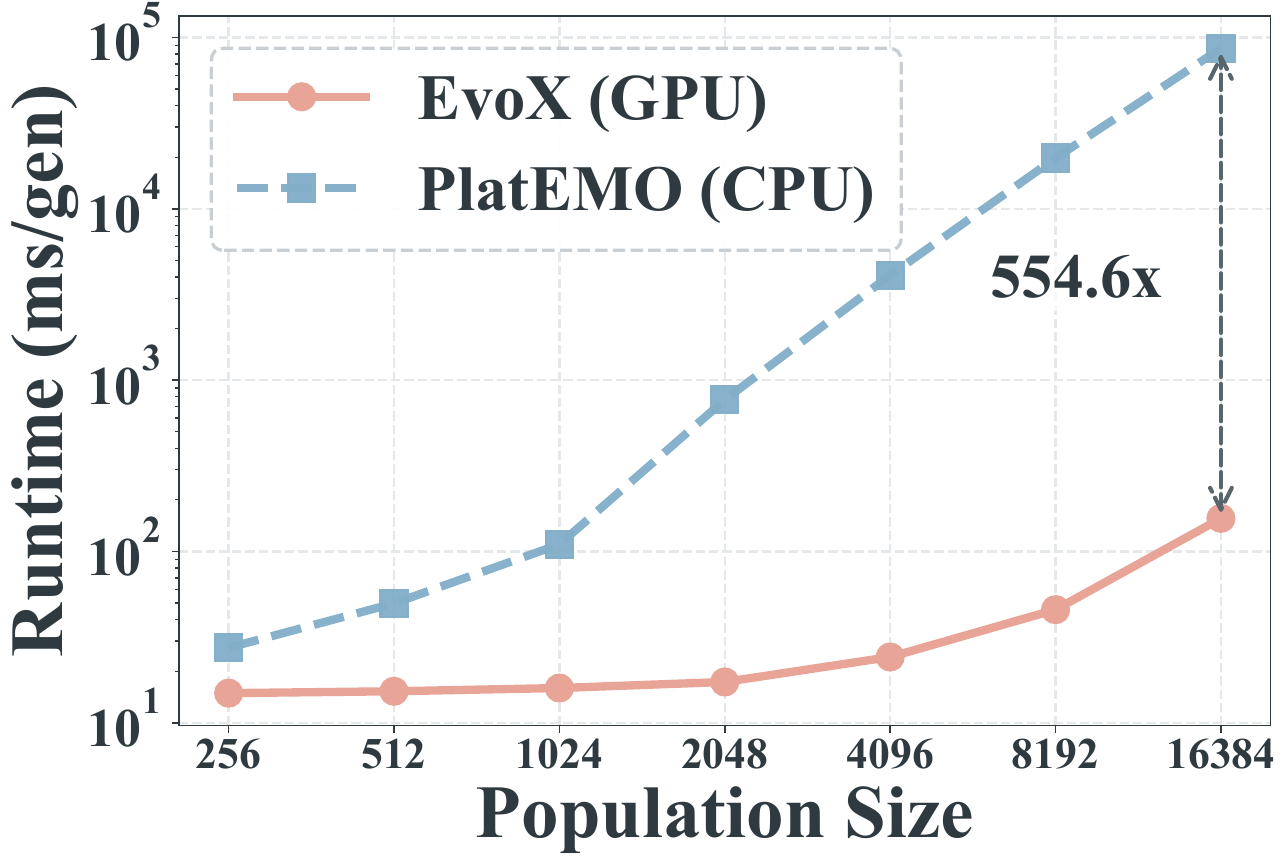}}
\hfill
\subfloat[SparseEA2: varying $D$]{\includegraphics[width=0.22\textwidth]{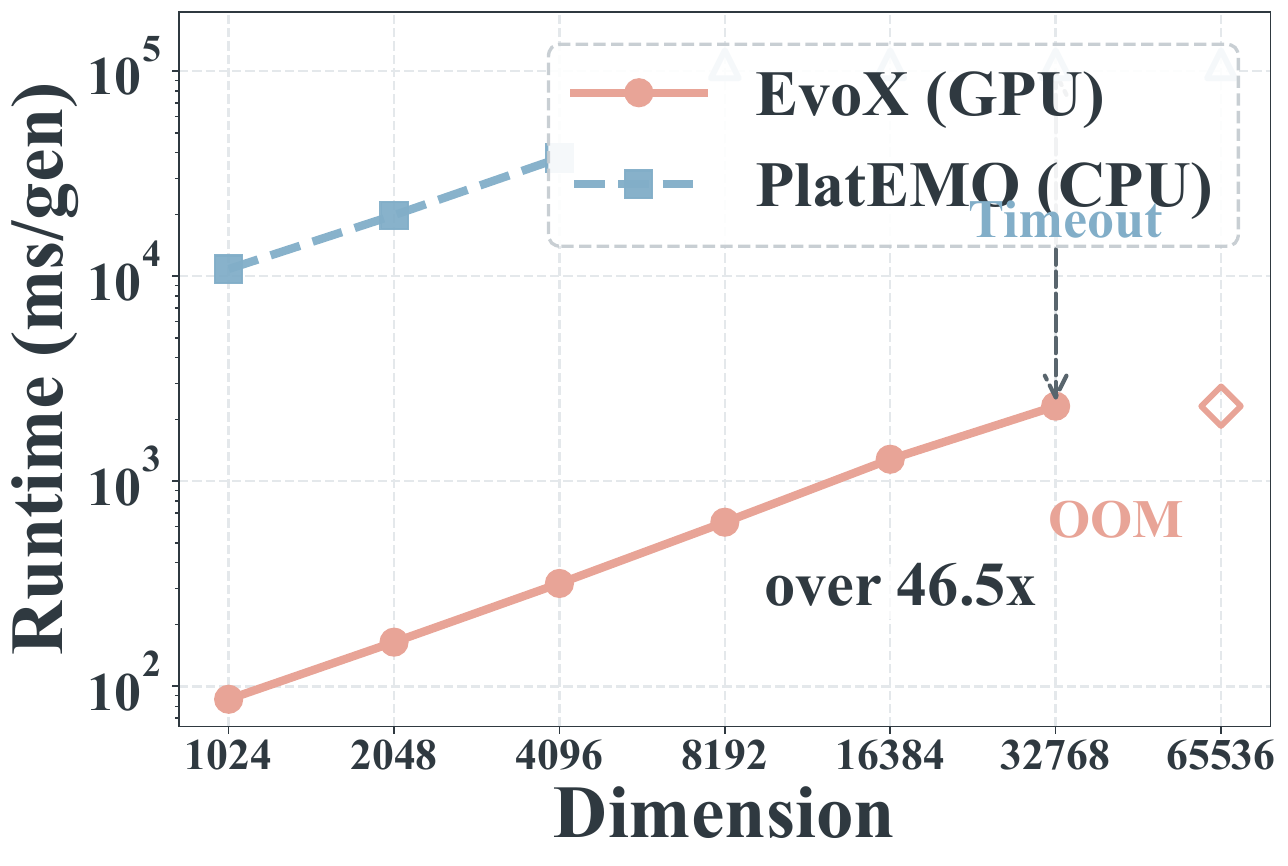}}
\hfill
\subfloat[SPEA-R: varying $N$]{\includegraphics[width=0.22\textwidth]{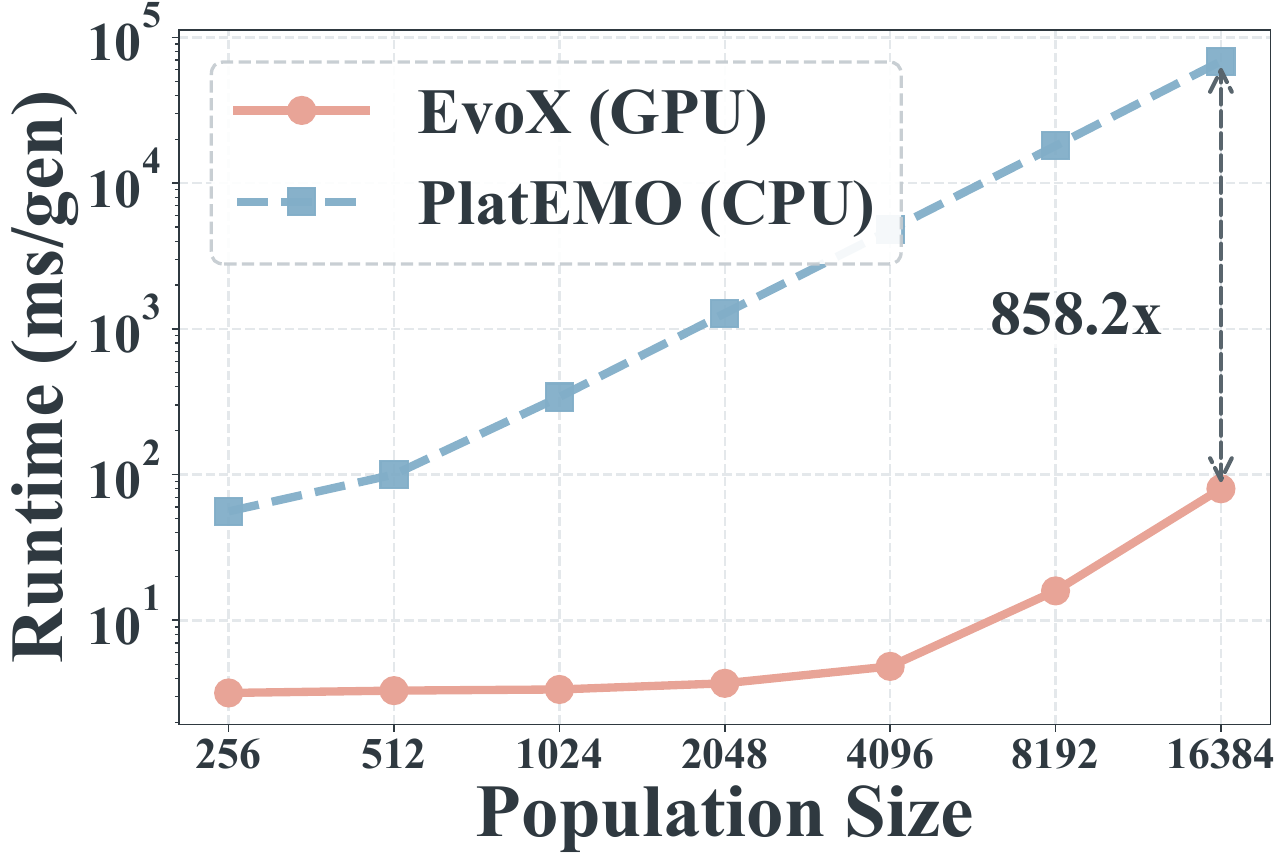}}
\hfill
\subfloat[SPEA-R: varying $D$]{\includegraphics[width=0.22\textwidth]{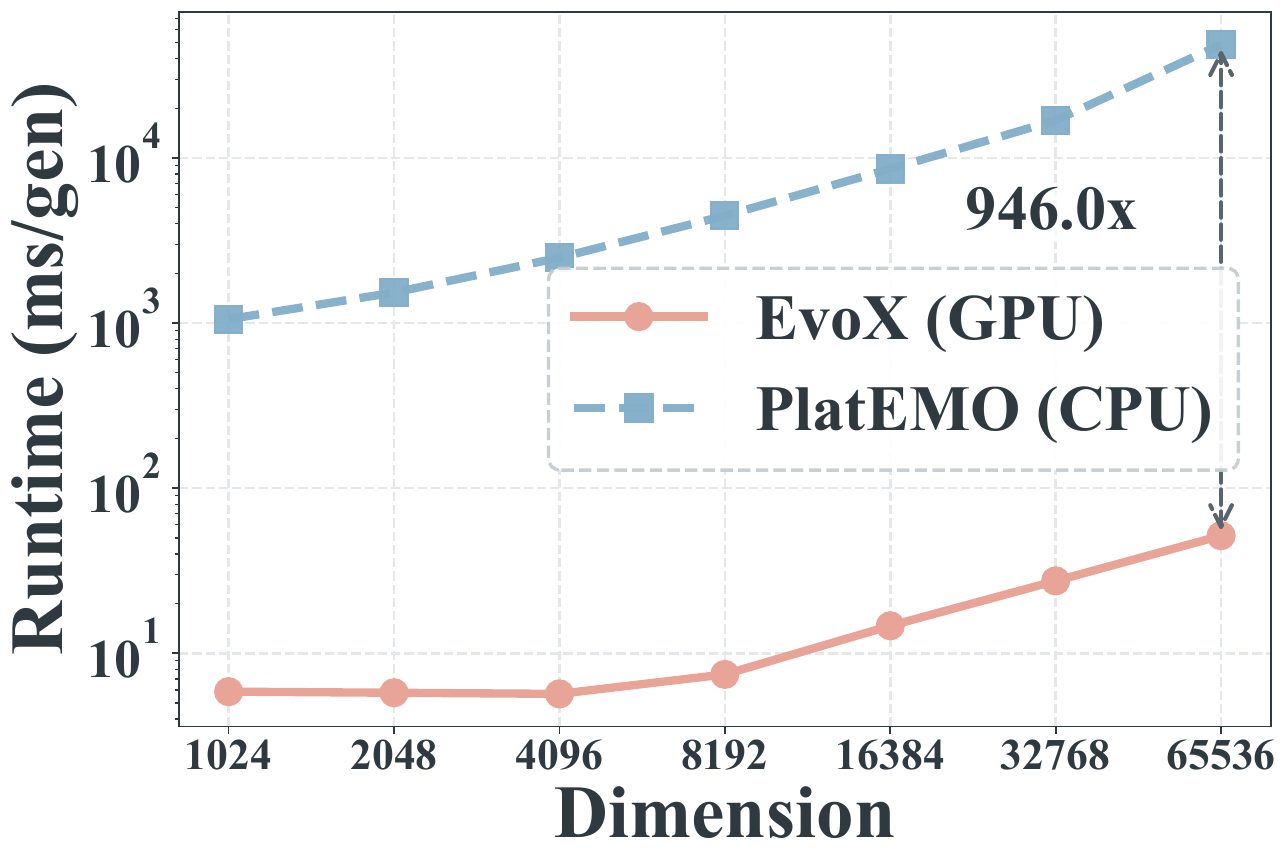}}
\\[-1mm]
\subfloat[SSCEA: varying $N$]{\includegraphics[width=0.22\textwidth]{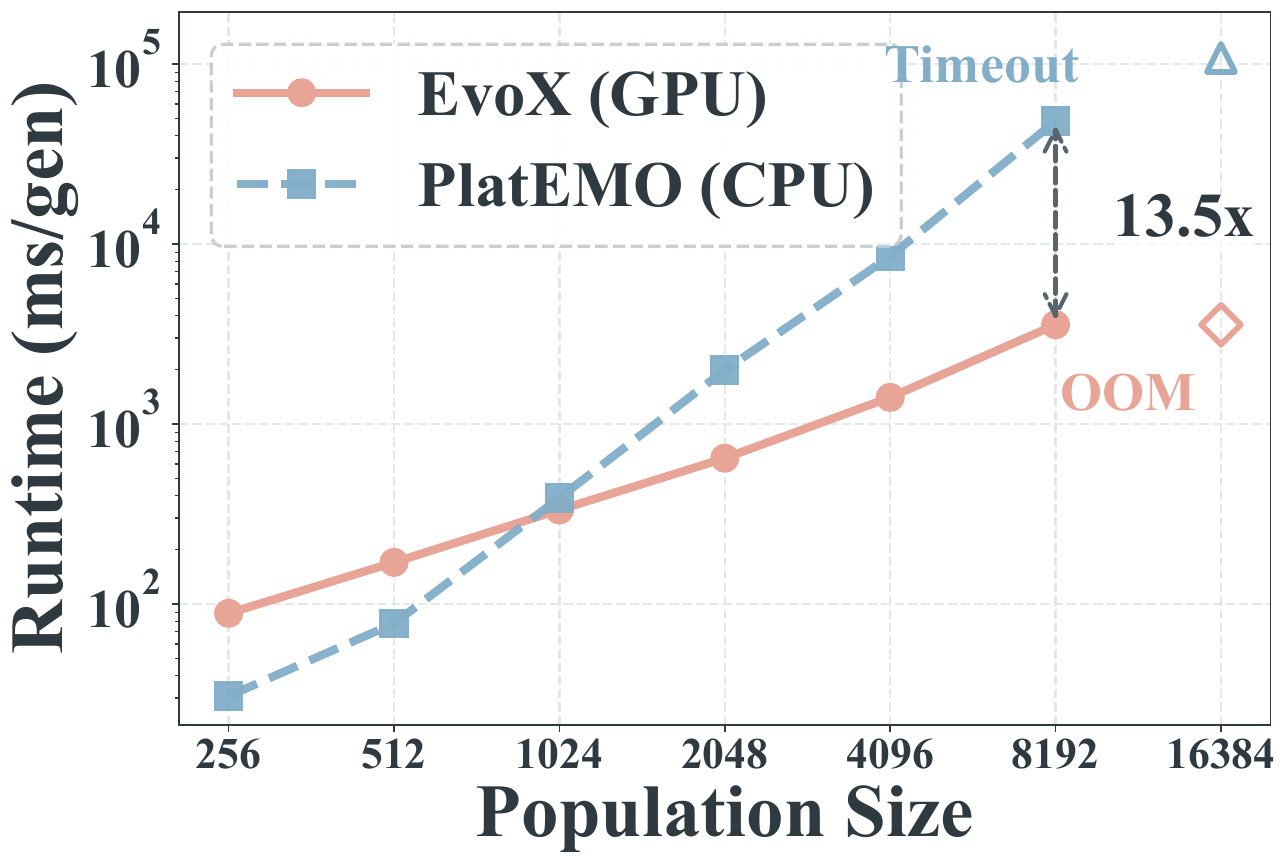}}
\hfill
\subfloat[SSCEA: varying $D$]{\includegraphics[width=0.22\textwidth]{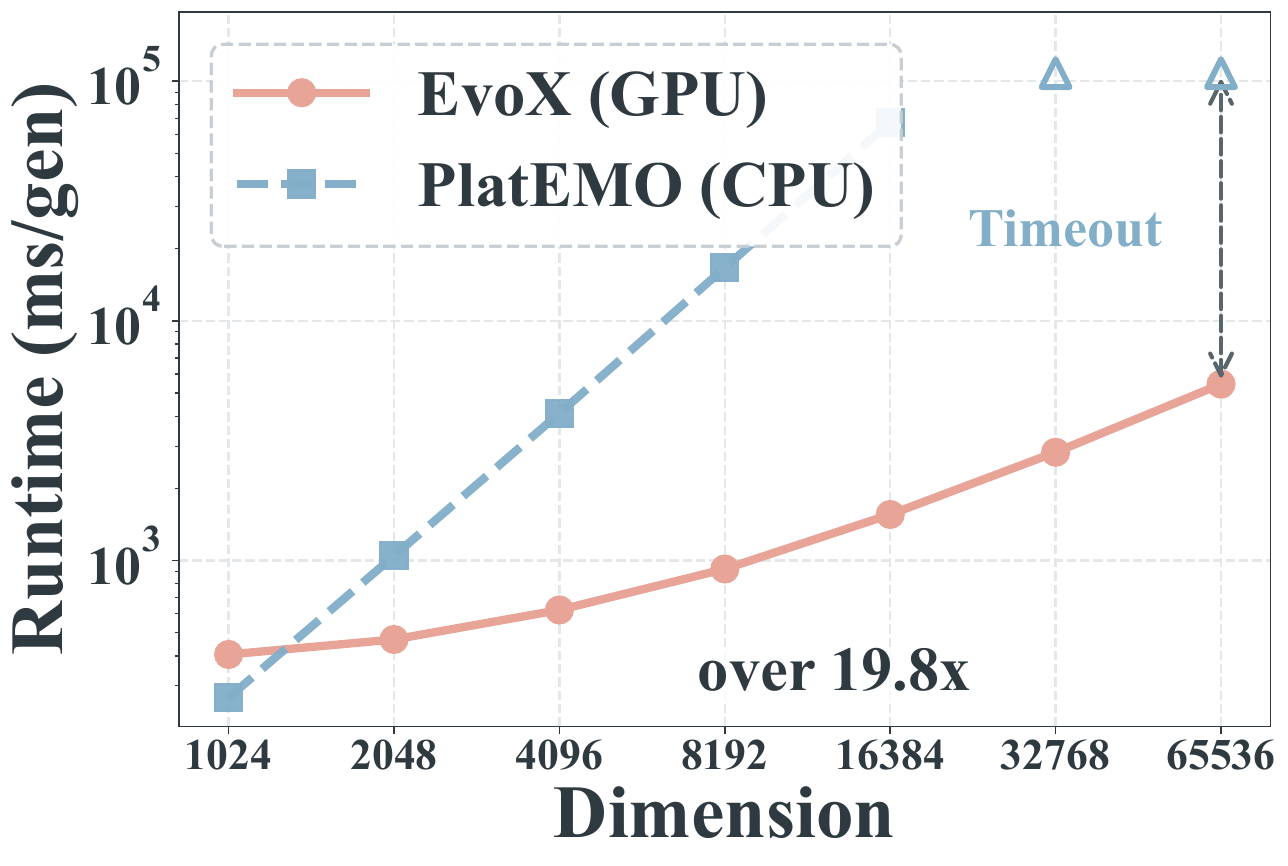}}
\hfill
\subfloat[t-DEA: varying $N$]{\includegraphics[width=0.22\textwidth]{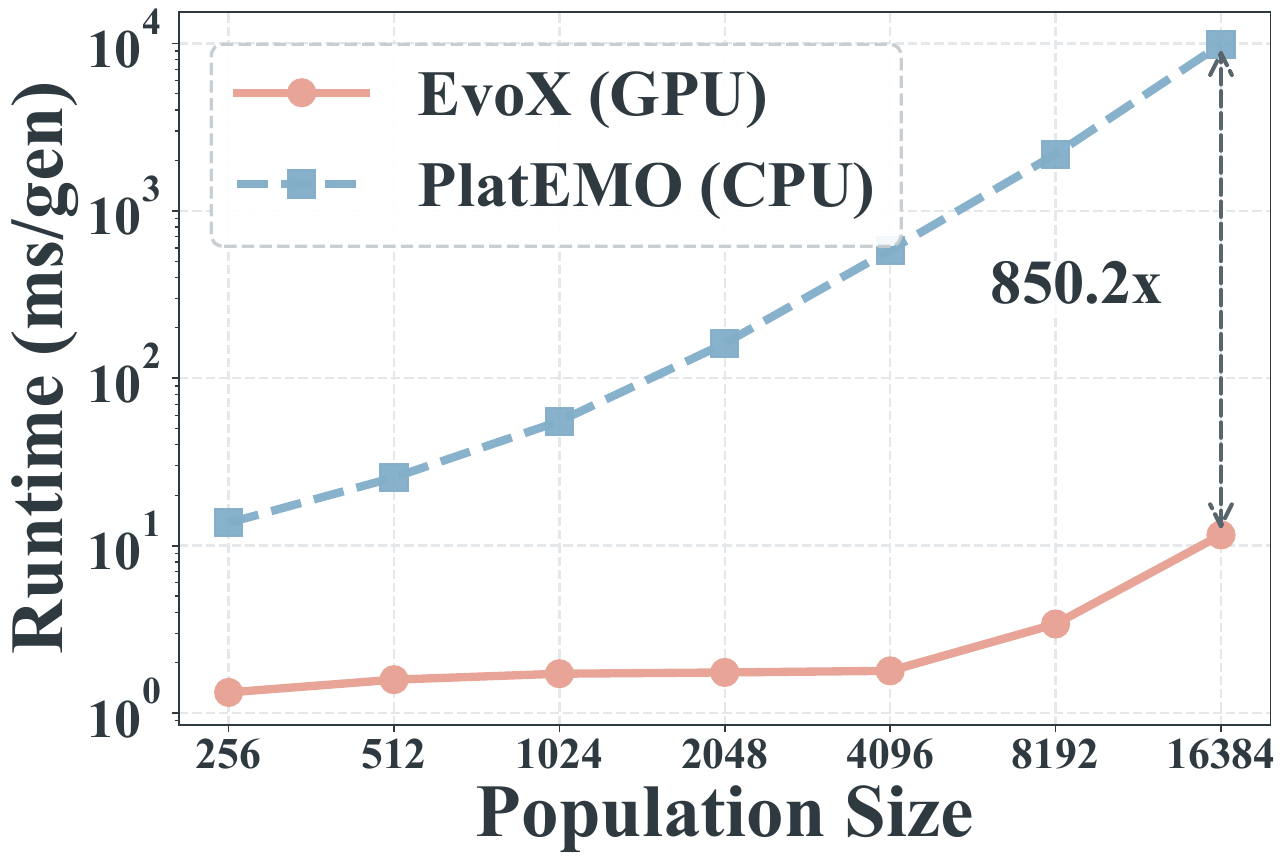}}
\hfill
\subfloat[t-DEA: varying $D$]{\includegraphics[width=0.22\textwidth]{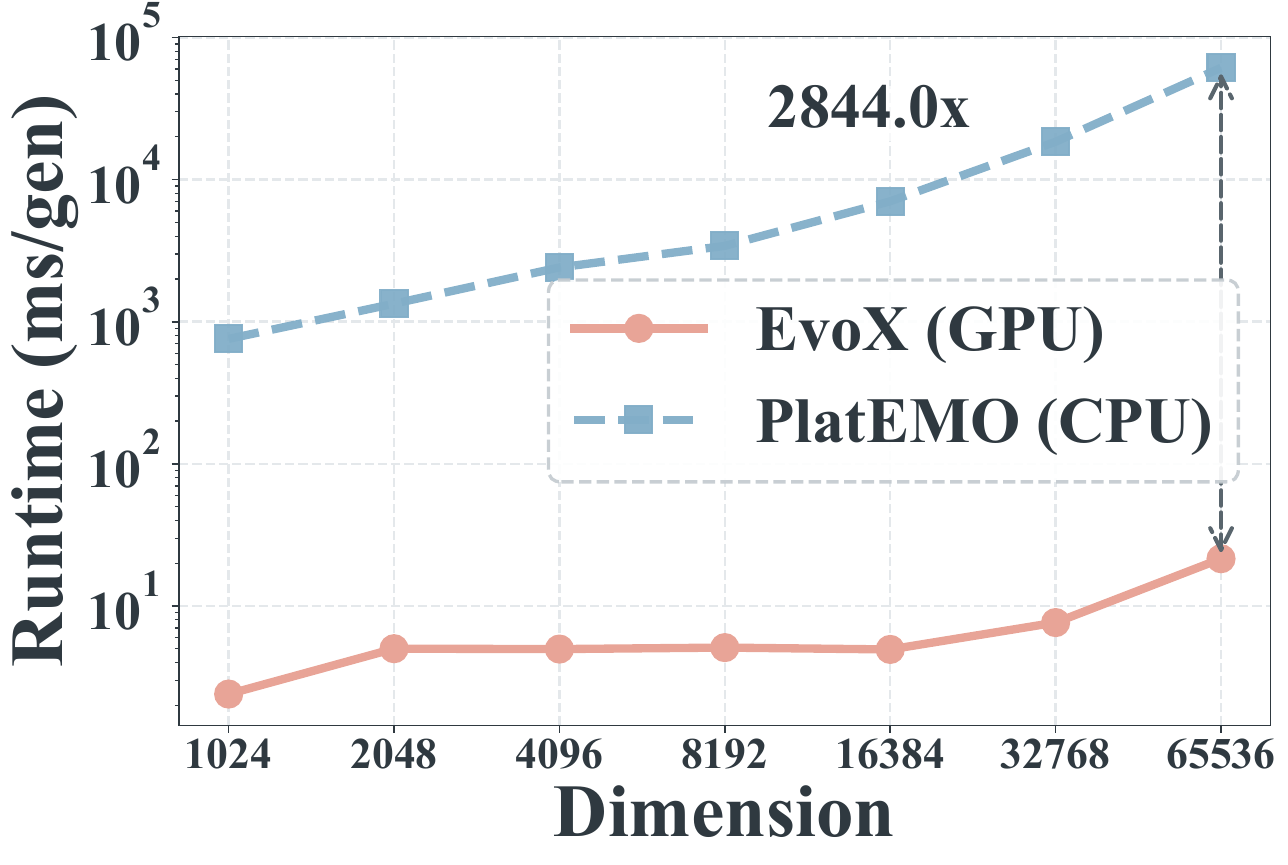}}
\caption{Complete population-size ($N$) and decision-dimension ($D$) scaling results for PICEA-g through t-DEA (Part IV of V). Adjacent panels report the two scaling axes for each algorithm.}
\label{fig:supp_scaling_curves_04}
\end{figure}

\begin{figure}[p]
\centering
\subfloat[tDEA-CPBI: varying $N$]{\includegraphics[width=0.22\textwidth]{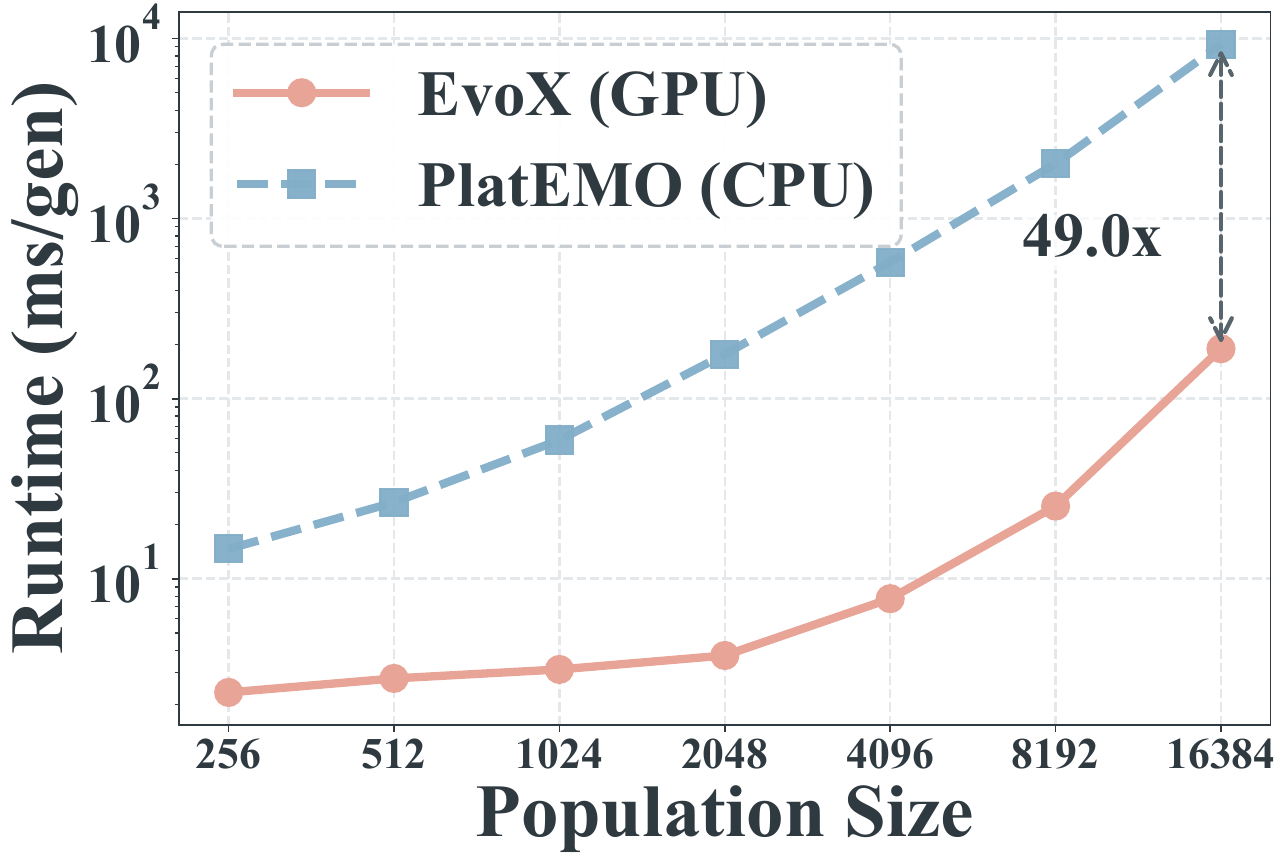}}
\hfill
\subfloat[tDEA-CPBI: varying $D$]{\includegraphics[width=0.22\textwidth]{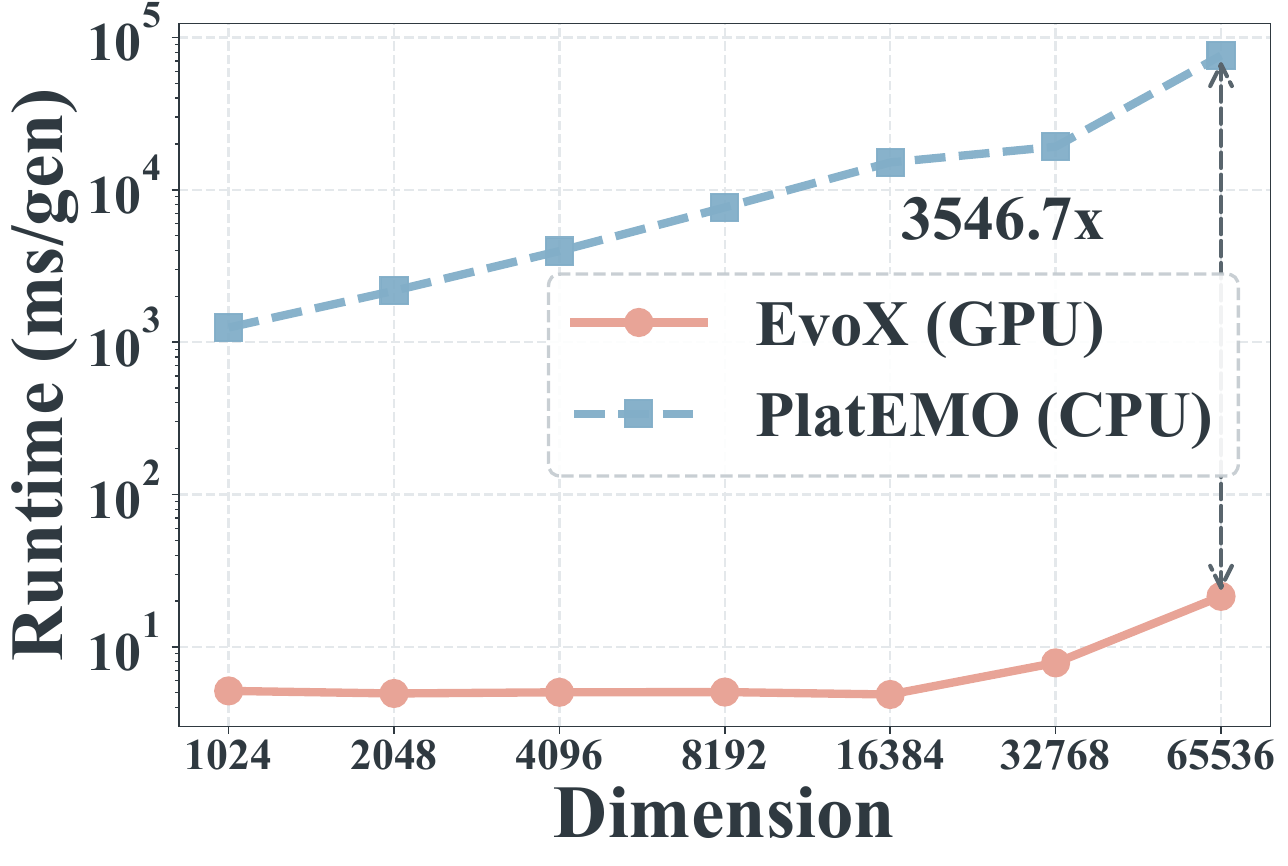}}
\hfill
\subfloat[TELSO: varying $N$]{\includegraphics[width=0.22\textwidth]{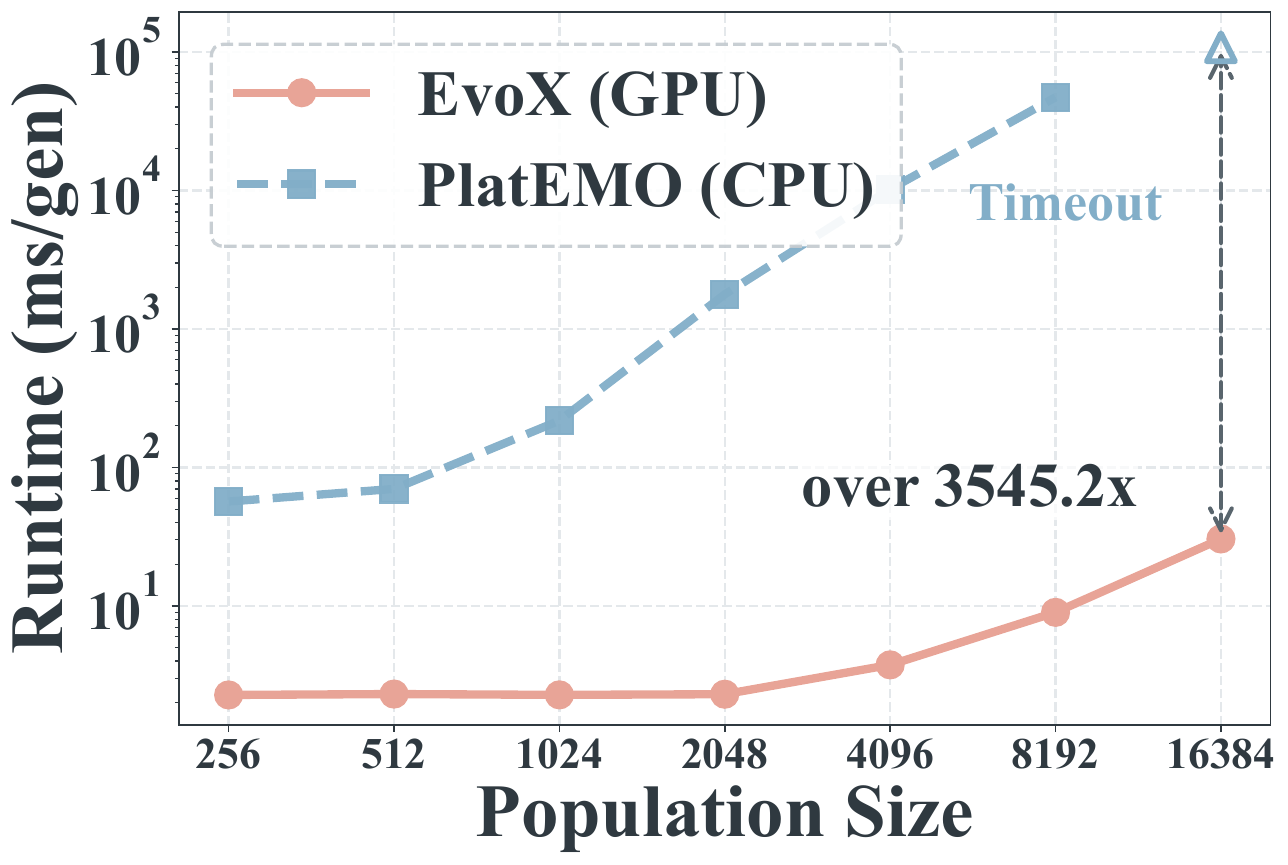}}
\hfill
\subfloat[TELSO: varying $D$]{\includegraphics[width=0.22\textwidth]{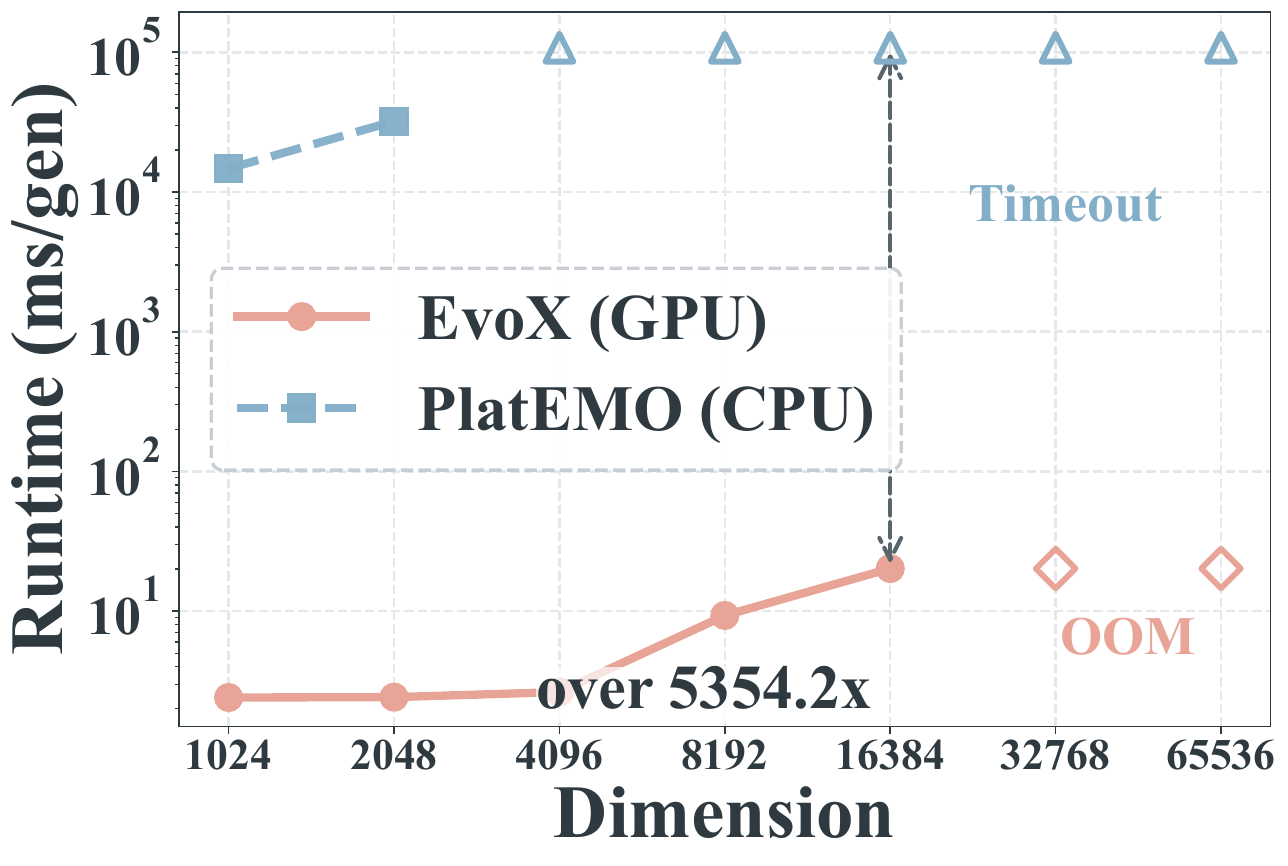}}
\\[-1mm]
\subfloat[TS-NSGA-II: varying $N$]{\includegraphics[width=0.22\textwidth]{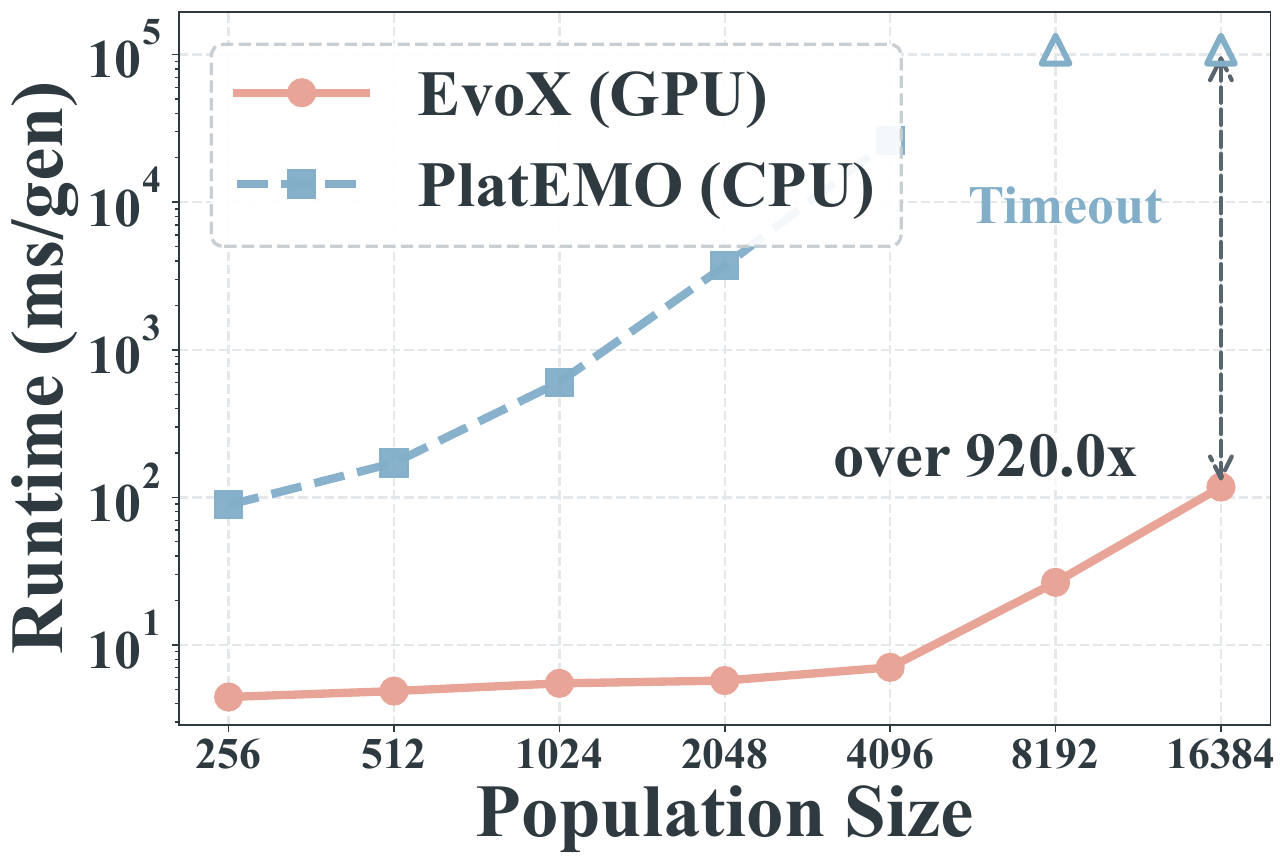}}
\hfill
\subfloat[TS-NSGA-II: varying $D$]{\includegraphics[width=0.22\textwidth]{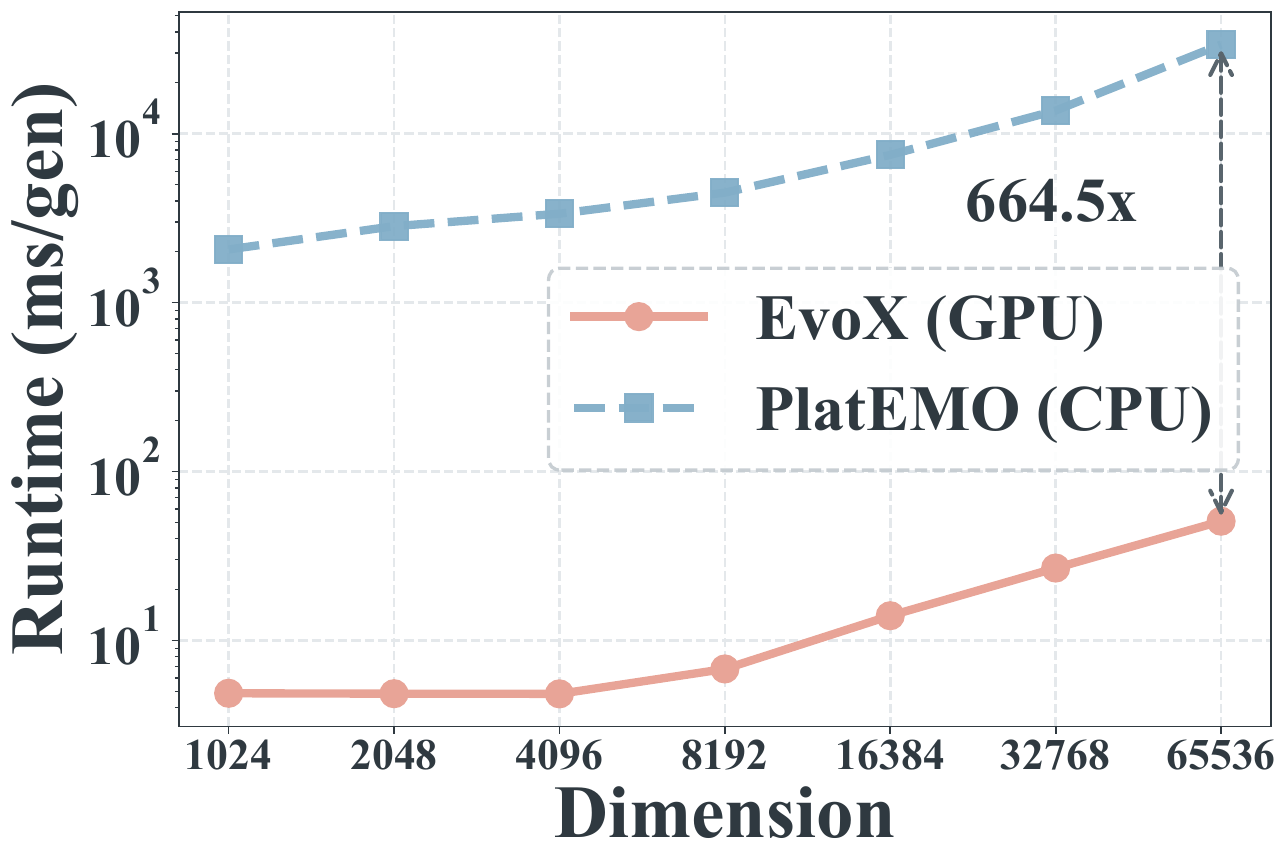}}
\hfill
\subfloat[TS-SparseEA: varying $N$]{\includegraphics[width=0.22\textwidth]{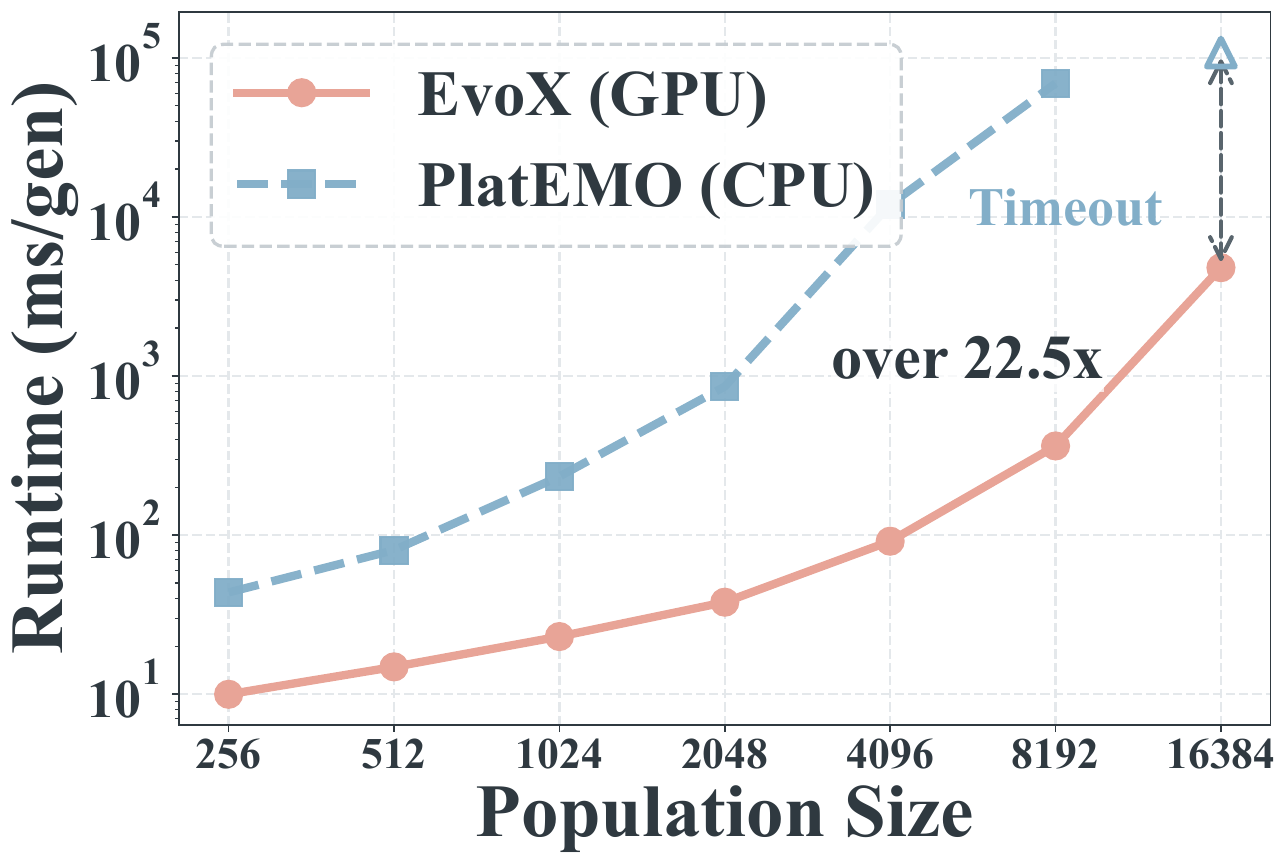}}
\hfill
\subfloat[TS-SparseEA: varying $D$]{\includegraphics[width=0.22\textwidth]{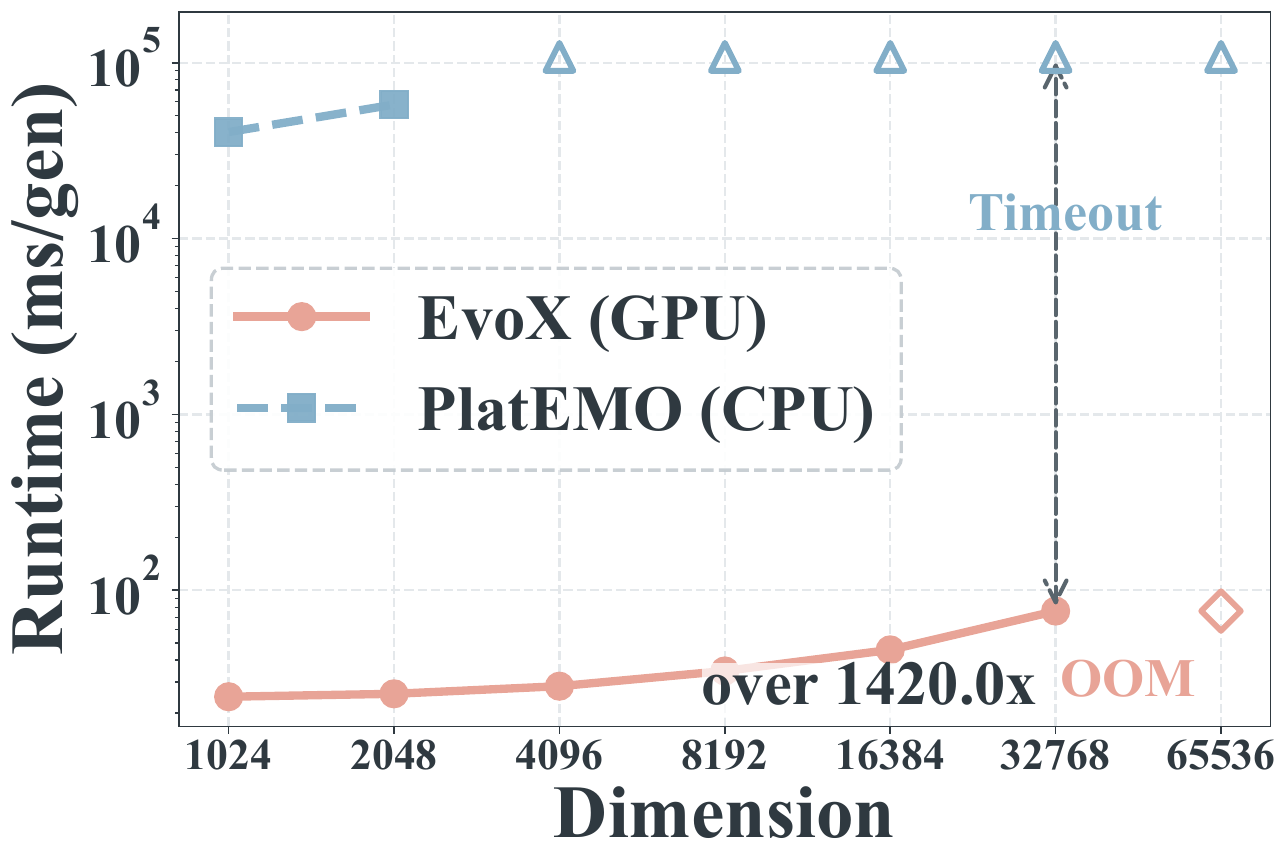}}
\\[-1mm]
\subfloat[Two\_Arch2: varying $N$]{\includegraphics[width=0.22\textwidth]{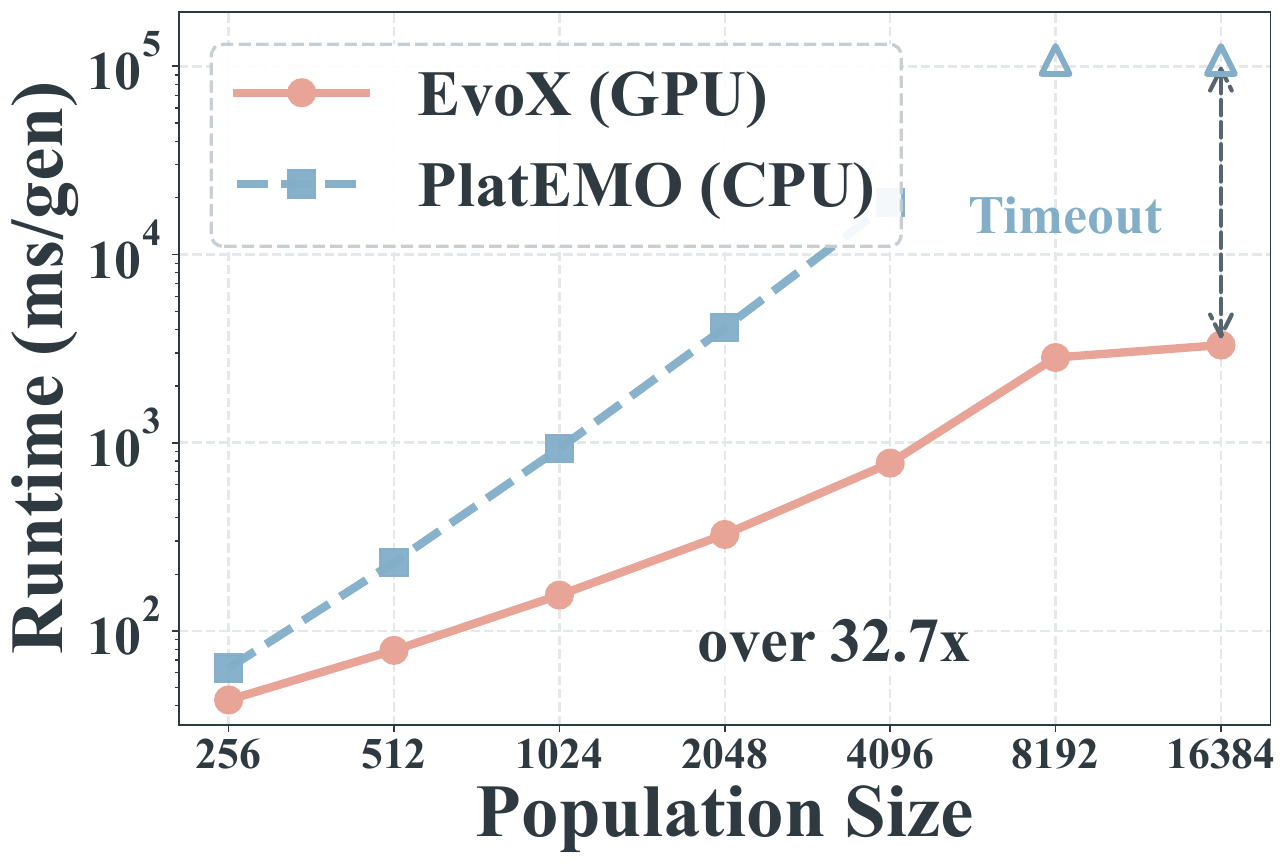}}
\hfill
\subfloat[Two\_Arch2: varying $D$]{\includegraphics[width=0.22\textwidth]{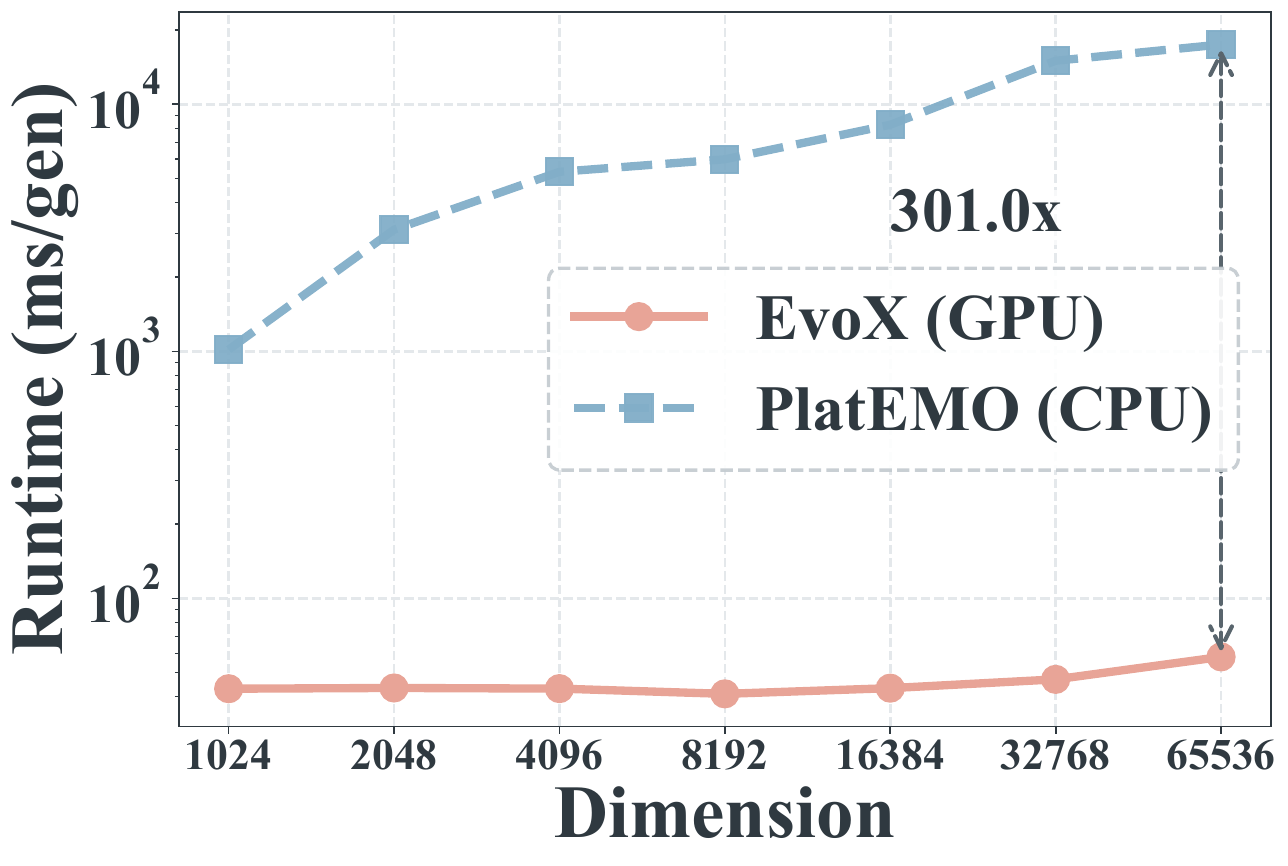}}
\hfill
\subfloat[VaEA: varying $N$]{\includegraphics[width=0.22\textwidth]{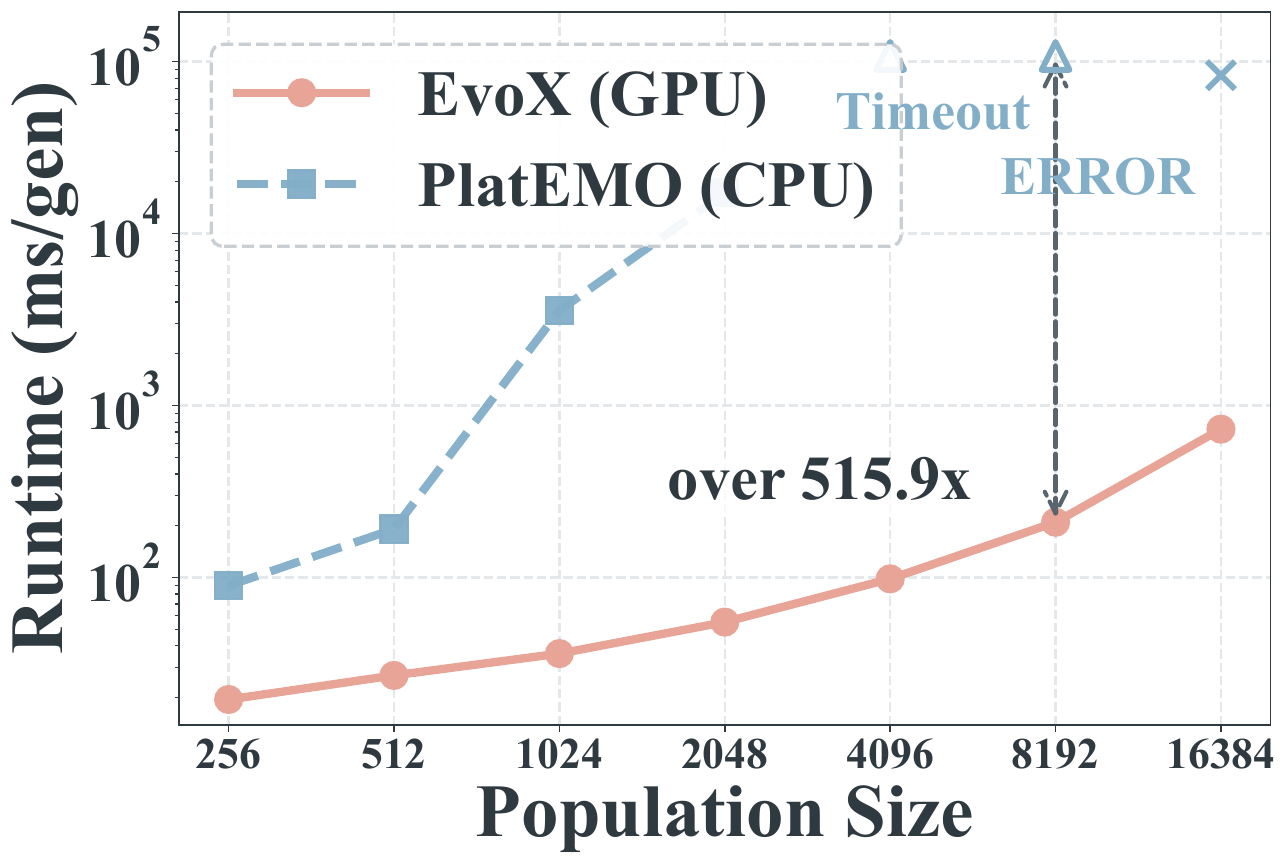}}
\hfill
\subfloat[VaEA: varying $D$]{\includegraphics[width=0.22\textwidth]{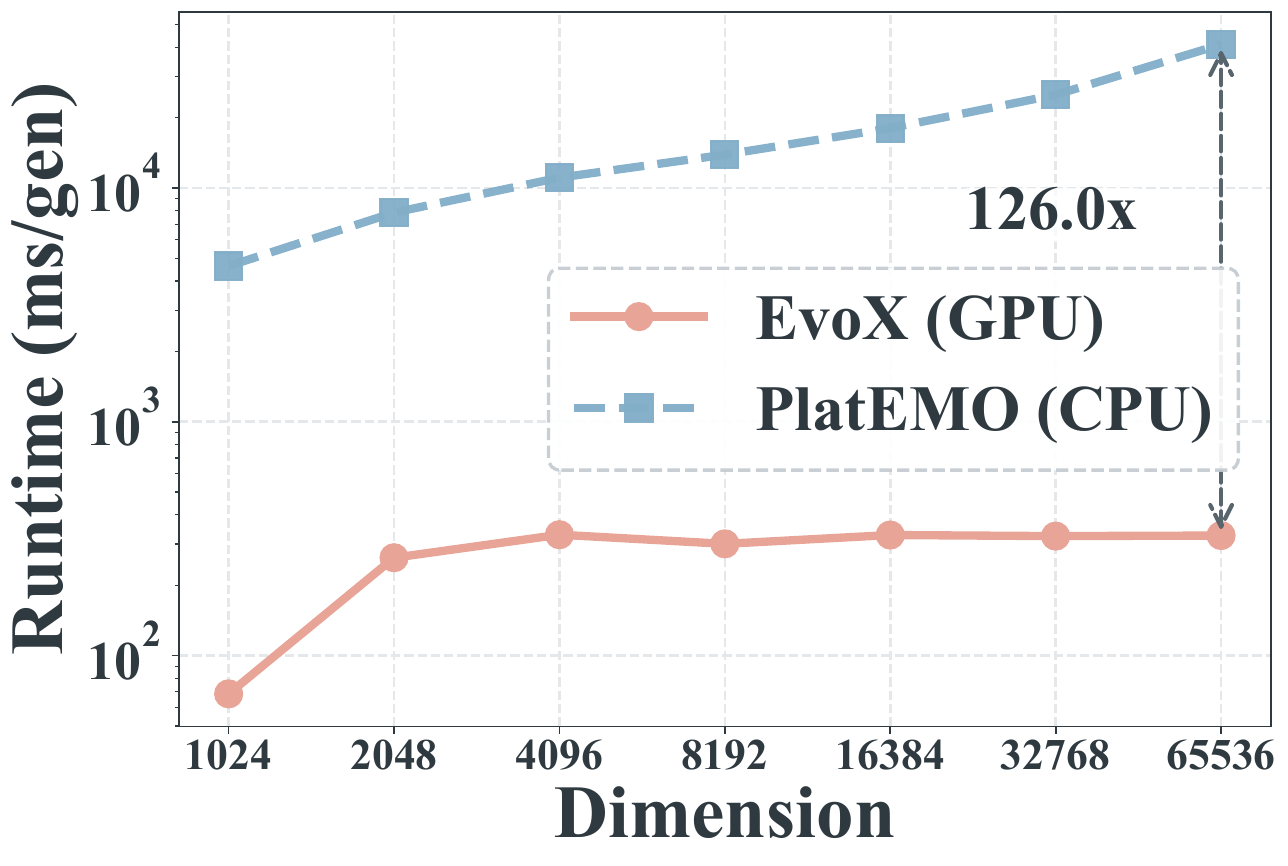}}
\\[-1mm]
\subfloat[WASF-GA: varying $N$]{\includegraphics[width=0.22\textwidth]{figures/exp3_scaling/curves/scaling_pop_comparison_WASF-GA_pop.pdf}}
\hfill
\subfloat[WASF-GA: varying $D$]{\includegraphics[width=0.22\textwidth]{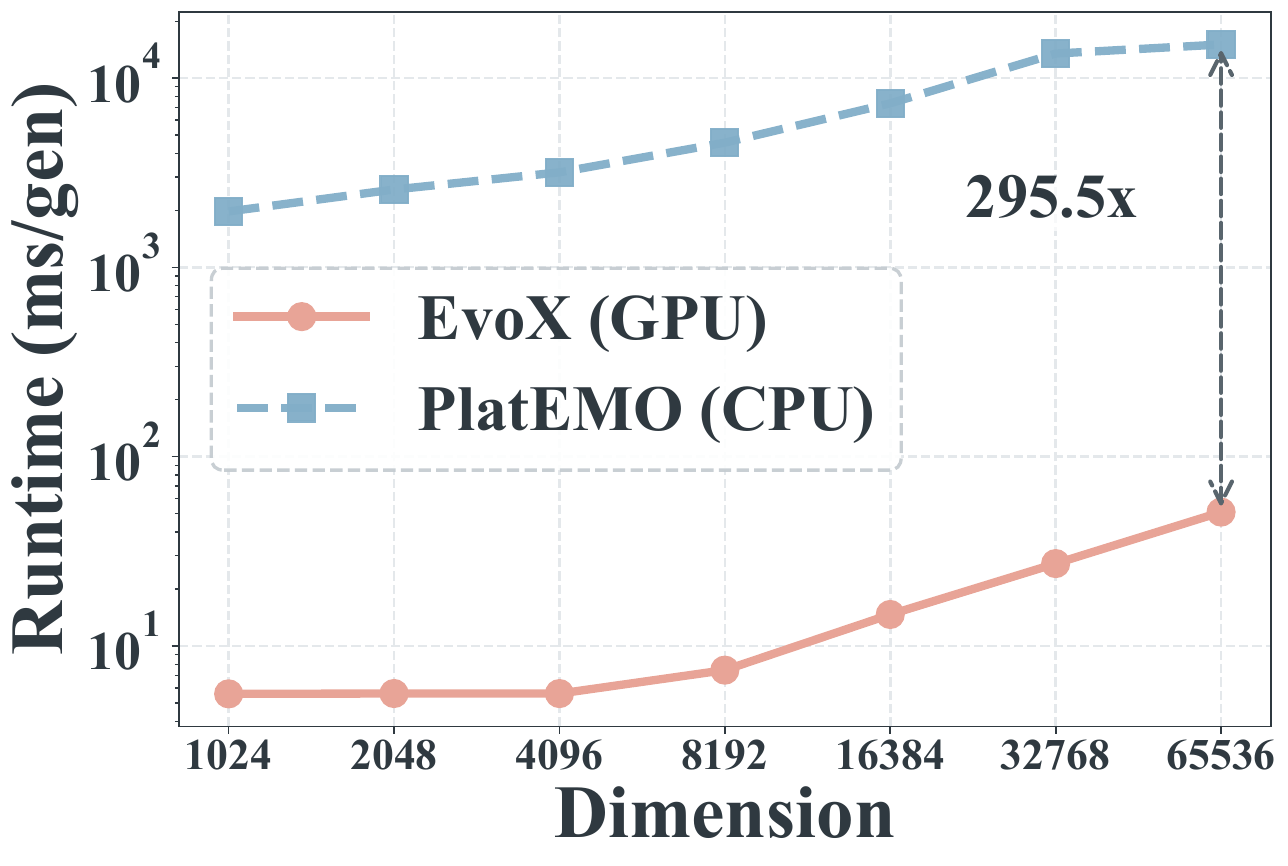}}
\hfill
\subfloat[WOF: varying $N$]{\includegraphics[width=0.22\textwidth]{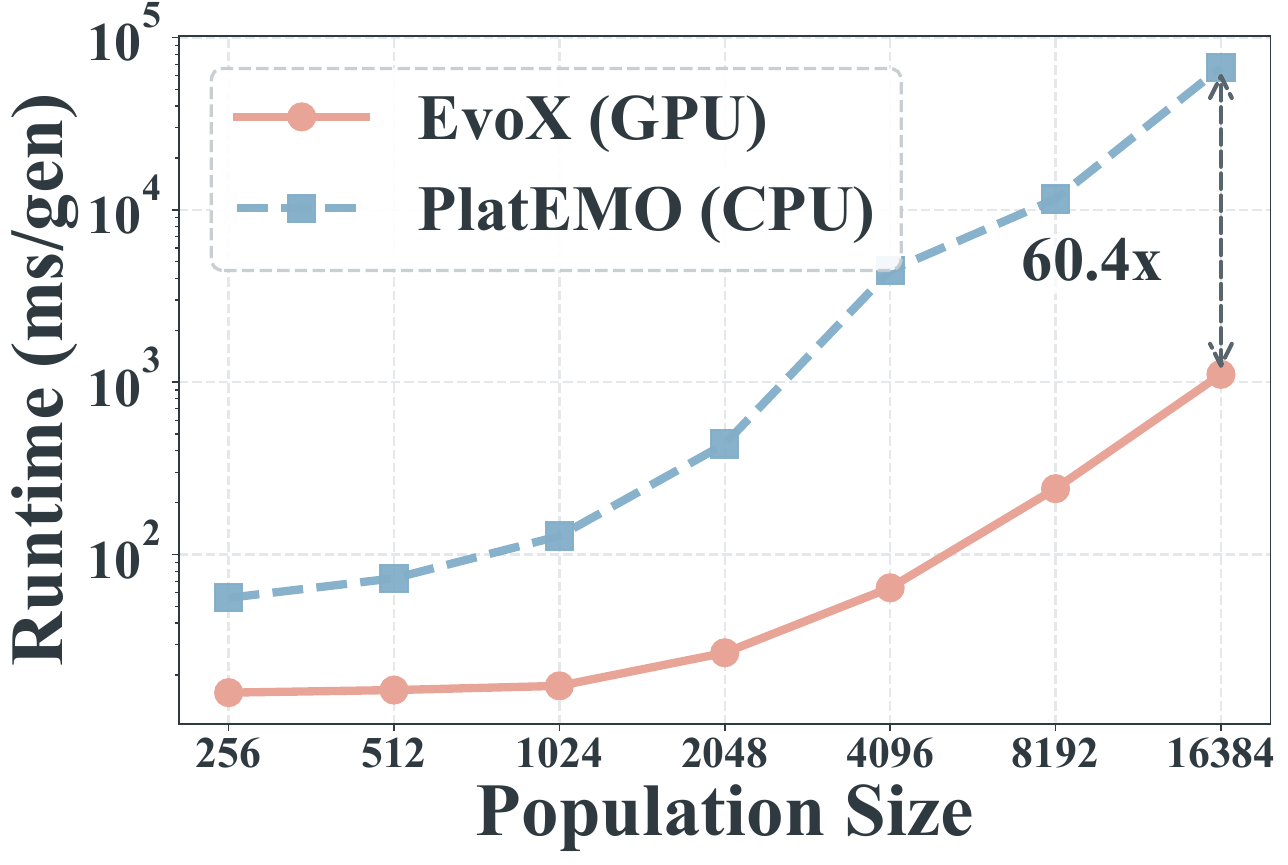}}
\hfill
\subfloat[WOF: varying $D$]{\includegraphics[width=0.22\textwidth]{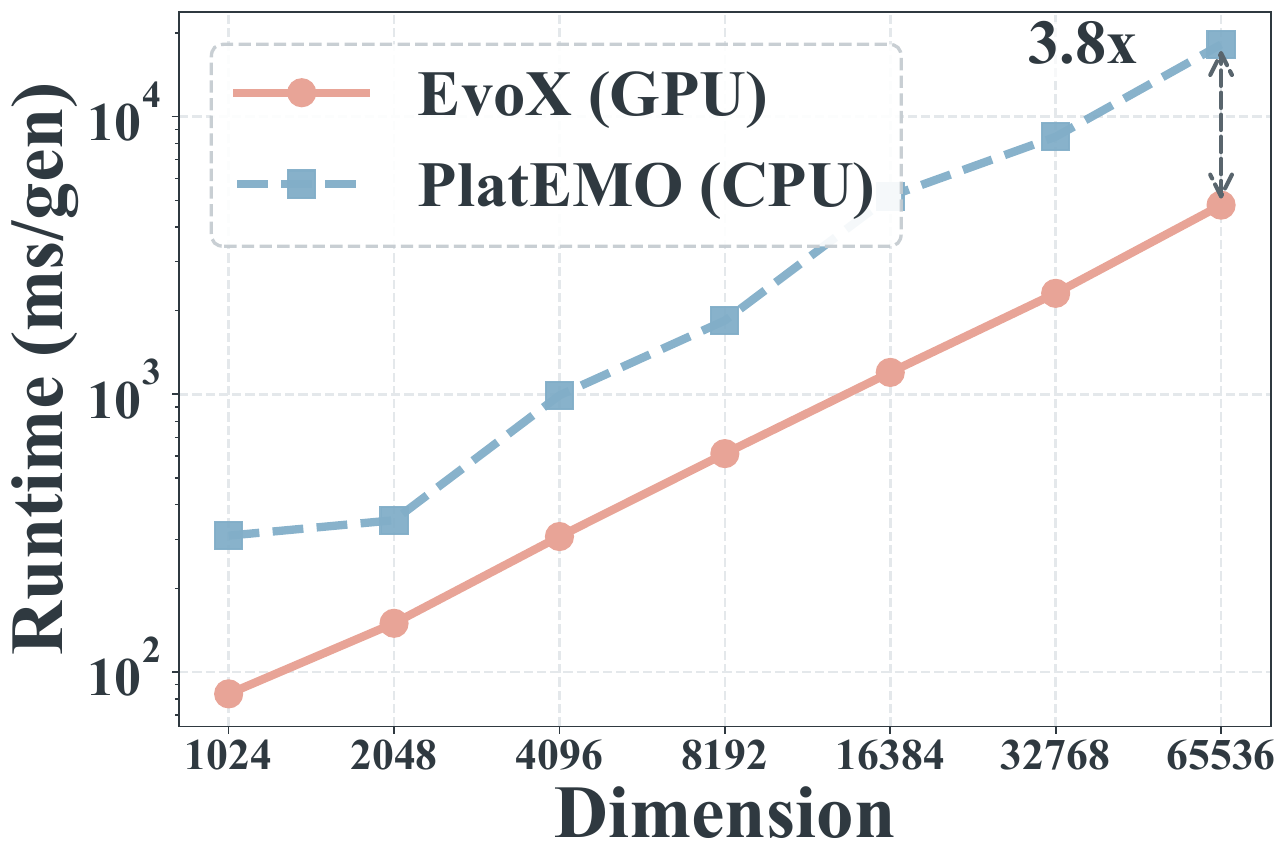}}
\caption{Complete population-size ($N$) and decision-dimension ($D$) scaling results for tDEA-CPBI through WOF (Part V of V). Adjacent panels report the two scaling axes for each algorithm.}
\label{fig:supp_scaling_curves_05}
\end{figure}

Figs.~\ref{fig:supp_scaling_curves_01}--\ref{fig:supp_scaling_curves_05} report the complete population- and dimension-scaling curves for all 48 algorithms. PlatEMO timeouts are displayed at 108000 ms/gen, and a lower-bound speedup is computed only when the corresponding EvoX runtime is valid. Out-of-memory (OOM) events are marked by a diamond, and execution errors are marked by a cross. For an EvoX OOM event, the marker is placed at the preceding valid runtime to identify the failing scale without treating it as a valid measurement. Table~\ref{tab:supp_scaling_exceptions} subsequently lists every exception record, including its scale, implementation side, and exception type.

\clearpage

\clearpage

\begingroup
\scriptsize
\setlength{\tabcolsep}{3pt}
\begin{longtable}{p{0.10\textwidth} p{0.17\textwidth} p{0.16\textwidth} p{0.18\textwidth} p{0.17\textwidth} p{0.10\textwidth} c}
\multicolumn{7}{c}{\label{tab:supp_scaling_exceptions}\normalfont\footnotesize TABLE~\thetable}\\[-0.2ex]
\multicolumn{7}{c}{\normalfont\footnotesize\scshape Runtime-Scaling Exceptions Across Population Size and Decision Dimension.}\\[0.5ex]
\toprule
Scaling & Algorithm & Scale axis & Scale value & Implementation & Exception & Count \\
\midrule
\endfirsthead
\multicolumn{7}{c}{\normalfont\footnotesize TABLE~\thetable\ (Continued)}\\[-0.2ex]
\multicolumn{7}{c}{\normalfont\footnotesize\scshape Runtime-Scaling Exceptions Across Population Size and Decision Dimension.}\\[0.5ex]
\toprule
Scaling & Algorithm & Scale axis & Scale value & Implementation & Exception & Count \\
\midrule
\endhead
\midrule
\multicolumn{7}{r}{Continued on next page}\\
\endfoot
\bottomrule
\endlastfoot
Population & BCE-IBEA & Population Size & 16384 & EvoX (GPU) & OOM & 1 \\
Population & BCE-MOEA-D & Population Size & 16384 & PlatEMO (CPU) & ERROR & 1 \\
Population & BiGE & Population Size & 16384 & PlatEMO (CPU) & TIMEOUT & 1 \\
Population & CMOEA-MS & Population Size & 16384 & EvoX (GPU) & OOM & 1 \\
Population & CMOEA-MS & Population Size & 8192, 16384 & PlatEMO (CPU) & TIMEOUT & 2 \\
Population & CMOPSO & Population Size & 16384 & PlatEMO (CPU) & TIMEOUT & 1 \\
Population & CoMMEA & Population Size & 2048, 4096, 8192, 16384 & PlatEMO (CPU) & ERROR & 4 \\
Population & DM-MOEA & Population Size & 16384 & PlatEMO (CPU) & TIMEOUT & 1 \\
Population & EFR-RR & Population Size & 16384 & EvoX (GPU) & OOM & 1 \\
Population & EFR-RR & Population Size & 16384 & PlatEMO (CPU) & TIMEOUT & 1 \\
Population & GWASF-GA & Population Size & 16384 & EvoX (GPU) & OOM & 1 \\
Population & GWASF-GA & Population Size & 16384 & PlatEMO (CPU) & TIMEOUT & 1 \\
Population & MaOEA-CSS & Population Size & 1024, 2048, 4096, 8192, 16384 & PlatEMO (CPU) & TIMEOUT & 5 \\
Population & MOEA-D-AWA & Population Size & 16384 & PlatEMO (CPU) & TIMEOUT & 1 \\
Population & MOEA-D-DCWV & Population Size & 16384 & PlatEMO (CPU) & TIMEOUT & 1 \\
Population & MOEA-D-DYTS & Population Size & 16384 & PlatEMO (CPU) & ERROR & 1 \\
Population & MOEA-D-PaS & Population Size & 16384 & PlatEMO (CPU) & TIMEOUT & 1 \\
Population & MOEA-D-URAW & Population Size & 8192, 16384 & PlatEMO (CPU) & ERROR & 2 \\
Population & NSBiDiCo & Population Size & 16384 & EvoX (GPU) & OOM & 1 \\
Population & PICEA-g & Population Size & 1024, 2048, 4096, 8192, 16384 & PlatEMO (CPU) & ERROR & 5 \\
Population & PREA & Population Size & 16384 & PlatEMO (CPU) & ERROR & 1 \\
Population & SIBEA & Population Size & 512, 1024, 2048, 4096, 8192, 16384 & PlatEMO (CPU) & TIMEOUT & 6 \\
Population & SSCEA & Population Size & 16384 & EvoX (GPU) & OOM & 1 \\
Population & SSCEA & Population Size & 16384 & PlatEMO (CPU) & TIMEOUT & 1 \\
Population & TELSO & Population Size & 16384 & PlatEMO (CPU) & TIMEOUT & 1 \\
Population & TS-NSGA-II & Population Size & 8192, 16384 & PlatEMO (CPU) & TIMEOUT & 2 \\
Population & TS-SparseEA & Population Size & 16384 & PlatEMO (CPU) & TIMEOUT & 1 \\
Population & Two\_Arch2 & Population Size & 8192, 16384 & PlatEMO (CPU) & TIMEOUT & 2 \\
Population & VaEA & Population Size & 16384 & PlatEMO (CPU) & ERROR & 1 \\
Population & VaEA & Population Size & 4096, 8192 & PlatEMO (CPU) & TIMEOUT & 2 \\
Dimension & CoMMEA & Dimension & 1024, 2048, 4096, 8192, 16384, 32768, 65536 & PlatEMO (CPU) & ERROR & 7 \\
Dimension & DM-MOEA & Dimension & 16384, 32768, 65536 & EvoX (GPU) & OOM & 3 \\
Dimension & DM-MOEA & Dimension & 16384, 32768, 65536 & PlatEMO (CPU) & ERROR & 3 \\
Dimension & e-MOEA & Dimension & 65536 & PlatEMO (CPU) & OOM & 1 \\
Dimension & MaOEA-CSS & Dimension & 1024, 2048, 4096, 8192, 16384, 32768, 65536 & PlatEMO (CPU) & TIMEOUT & 7 \\
Dimension & OSP-NSDE & Dimension & 65536 & EvoX (GPU) & OOM & 1 \\
Dimension & OSP-NSDE & Dimension & 1024, 2048, 4096, 8192, 16384, 32768, 65536 & PlatEMO (CPU) & TIMEOUT & 7 \\
Dimension & PICEA-g & Dimension & 1024, 2048, 4096, 8192, 16384, 32768, 65536 & PlatEMO (CPU) & ERROR & 7 \\
Dimension & SIBEA & Dimension & 1024, 2048, 4096, 8192, 16384, 32768, 65536 & PlatEMO (CPU) & TIMEOUT & 7 \\
Dimension & SMPSO & Dimension & 65536 & PlatEMO (CPU) & TIMEOUT & 1 \\
Dimension & SparseEA & Dimension & 16384, 32768, 65536 & EvoX (GPU) & OOM & 3 \\
Dimension & SparseEA & Dimension & 8192, 16384, 32768, 65536 & PlatEMO (CPU) & TIMEOUT & 4 \\
Dimension & SparseEA2 & Dimension & 65536 & EvoX (GPU) & OOM & 1 \\
Dimension & SparseEA2 & Dimension & 8192, 16384, 32768, 65536 & PlatEMO (CPU) & TIMEOUT & 4 \\
Dimension & SSCEA & Dimension & 32768, 65536 & PlatEMO (CPU) & TIMEOUT & 2 \\
Dimension & TELSO & Dimension & 32768, 65536 & EvoX (GPU) & OOM & 2 \\
Dimension & TELSO & Dimension & 4096, 8192, 16384, 32768, 65536 & PlatEMO (CPU) & TIMEOUT & 5 \\
Dimension & TS-SparseEA & Dimension & 65536 & EvoX (GPU) & OOM & 1 \\
Dimension & TS-SparseEA & Dimension & 4096, 8192, 16384, 32768, 65536 & PlatEMO (CPU) & TIMEOUT & 5 \\
\end{longtable}
\endgroup

\clearpage
\section{Transfer to External MOEA Implementations}

\subsection{Purpose, Sources, and Matched Inclusion}

This experiment tests whether EvoCoCo transfers beyond the current PlatEMO v4.14 setting to external or legacy MATLAB/Octave implementations. The sources differ from the retained PlatEMO benchmark in coding conventions, helper routines, and algorithm organization. Nine algorithms use standalone implementations, whereas LSMaODE was developed for an earlier PlatEMO interface and produces compatibility errors when executed directly under PlatEMO v4.14. The experiment therefore evaluates transfer across heterogeneous source styles, including both standalone and legacy framework-based implementations. Table~\ref{tab:supp_non_platemo_sources} lists the ten source MOEAs, the normalized identifiers used in the result files, and the corresponding GitHub repositories. Multiple DEMO variants originate from the same repository.

\begin{table}[H]
\centering
\caption{External and legacy MATLAB/Octave algorithms and source repositories used to evaluate transfer to external MOEA implementations.}
\label{tab:supp_non_platemo_sources}
\scriptsize
\setlength{\tabcolsep}{4pt}
\begin{tabular}{p{0.18\textwidth} p{0.37\textwidth} p{0.35\textwidth}}
\toprule
Result identifier & Source implementation & GitHub repository \\
\midrule
DEMO & DEMO MATLAB/Octave implementation~\cite{RobicFilipic2005DEMO} & \url{https://github.com/fcampelo/DEMO} \\
DEMO\_IBEA & DEMO variant using IBEA-style selection~\cite{RobicFilipic2005DEMO,ZitzlerKunzli2004IBEA} & \url{https://github.com/fcampelo/DEMO} \\
DEMO\_PBEA & DEMO variant using PBEA-style selection~\cite{RobicFilipic2005DEMO,ThieleEtAl2009PBEA} & \url{https://github.com/fcampelo/DEMO} \\
ISDE\_plus & ISDE+ MATLAB implementation~\cite{PamulapatiEtAl2019ISDEPlus} & \url{https://github.com/RammohanMallipeddi/Matlab-code-for-ISDE-} \\
LSMaODE & Implementation developed for an earlier PlatEMO interface~\cite{ZhangEtAl2023LSMaODE} & \url{https://github.com/MaOEA/LSMaODE} \\
PAR\_DEMO\_IND & PAR-DEMO variant using indicator-based selection~\cite{GoulartCampelo2016Preference} & \url{https://github.com/fcampelo/DEMO} \\
PAR\_DEMO\_NDS & PAR-DEMO variant using nondominated sorting~\cite{GoulartCampelo2016Preference} & \url{https://github.com/fcampelo/DEMO} \\
RTEA & RTEA MATLAB implementation~\cite{FieldsendEverson2015RTEA} & \url{https://github.com/fieldsend/ieee_tec_2014_rtea} \\
R\_DEMO & R-DEMO MATLAB/Octave implementation~\cite{RobicFilipic2005DEMO,DebEtAl2006ReferencePoint} & \url{https://github.com/fcampelo/DEMO} \\
Two\_Arch2 & Two\_Arch2 MATLAB implementation~\cite{TwoArch22015} & \url{https://github.com/HandingWang/Two_Arch2} \\
\bottomrule
\end{tabular}
\end{table}

EvoCoCo and the one-shot baseline each contribute five attempts for every source, which yields $10\times5=50$ matched generated implementations per method. The EvoCoCo directory also contains two additional R-DEMO attempts, \texttt{R\_DEMO\_run6.py} and \texttt{R\_DEMO\_run7.py}. They are excluded because the one-shot directory has no corresponding attempts. All aggregate and algorithm-level comparisons below therefore use identical denominators.

\subsection{Generation and Evaluation Protocol}

EvoCoCo uses a staged workflow comprising the Analysis Agent, Rule Retriever, Blueprint Agent, parallel Tensorization Agents, Repair Agent, and Selection Agent. Table~\ref{tab:supp_non_platemo_pipeline} summarizes these roles and their recorded artifacts. Every matched run uses the same six tensorization strategies: Broadcasting, Einsum Optimization, Masked Operations, In-Place Efficiency, Advanced Operations, and Tensorized Iterative Selection.

\begin{table}[H]
\centering
\caption{EvoCoCo workflow stages used in the experiment evaluating transfer to external MOEA implementations.}
\label{tab:supp_non_platemo_pipeline}
\scriptsize
\setlength{\tabcolsep}{5pt}
\begin{tabular}{l p{0.68\textwidth}}
\toprule
Stage & Function \\
\midrule
Analysis Agent & Analyze source algorithmic semantics and transformation requirements \\
Rule Retriever & Retrieve task-relevant implementation and tensorization rules \\
Blueprint Agent & Construct the target transformation blueprint \\
Tensorization Agents & Generate six candidates using distinct tensorization strategies \\
Repair Agent & Repair candidates using static and runtime feedback \\
Selection Agent & Select the final candidate from validated branches \\
\bottomrule
\end{tabular}
\end{table}

The 50 matched EvoCoCo runs produce 300 branch records, of which 248 are marked successful and 52 failed. The recorded runtime-repair counts are 189 branches with no repair, 39 with one repair, and 72 with two repairs. Each branch record contains two static-fix attempts. Table~\ref{tab:supp_non_platemo_generation_cost} reports generation-pipeline cost rather than optimizer runtime; corresponding per-run statistics are unavailable for the one-shot outputs.

\begin{table}[H]
\centering
\caption{Generation diagnostics for the 50 matched EvoCoCo conversion attempts, including branch usage, repair activity, and final candidate selection.}
\label{tab:supp_non_platemo_generation_cost}
\scriptsize
\setlength{\tabcolsep}{6pt}
\begin{tabular}{l r}
\toprule
Diagnostic & Value \\
\midrule
Average total tokens per run & 229,003 \\
Median total tokens per run & 191,598.5 \\
Minimum total tokens per run & 125,561 \\
Maximum total tokens per run & 592,376 \\
Average total LLM time per run & 375.92 s \\
Median total LLM time per run & 350.44 s \\
Minimum total LLM time per run & 157.12 s \\
Maximum total LLM time per run & 627.53 s \\
\bottomrule
\end{tabular}
\end{table}

The generated implementations target EvoX/PyTorch and use DTLZ2 with population size $N=100$, three objectives, and $D=12$. During migration, each generated implementation undergoes a lightweight 50-generation validation run. The corresponding demo block performs 49 update iterations after initialization. This run is used only for execution validation and repair and is not included in the final evaluation. For the final benchmark, complete algorithms are independently evaluated for 100 generations under the same problem setting. An executable algorithm is considered converged when its final IGD after 100 generations is below 0.25. Average IGD and runtime are not used for the main comparison because the one-shot method converges on only nine attempts; conditioning on this small and method-dependent success set would obscure the reliability difference.

\subsection{Aggregate Reliability and Failure Stages}

Across the 50 matched attempts per method, EvoCoCo achieves syntax, execution, and convergence pass rates of 100\%, 96\%, and 62\%, respectively, compared with 70\%, 34\%, and 18\% for one-shot generation. Table~\ref{tab:supp_non_platemo_failures} decomposes these outcomes into mutually exclusive failure stages and reports both percentages and counts. EvoCoCo remains near the top of the evaluation sequence through syntax and execution and loses most unsuccessful cases only at convergence, whereas the one-shot baseline loses many attempts before optimization outcomes are assessed.

The four mutually exclusive outcomes in Table~\ref{tab:supp_non_platemo_failures} distinguish incomplete or unparsable artifacts, runtime failures among syntactically valid files, convergence failures among executable files, and converged implementations. EvoCoCo eliminates syntax-level failures and reduces runtime failures from 18 to two. Its remaining failures are mainly executable implementations that do not meet the final-IGD threshold.

\begin{table}[H]
\centering
\caption{Failure-stage taxonomy for the matched conversions of external or legacy implementations. Cells report percentages, with counts in parentheses; categories are mutually exclusive and assigned by the earliest unsuccessful evaluation stage.}
\label{tab:supp_non_platemo_failures}
\scriptsize
\setlength{\tabcolsep}{3.5pt}
\begin{tabular}{l c c c c}
\toprule
Method & Syntax/incomplete & Runtime failure & Convergence failure & Converged \\
\midrule
EvoCoCo & 0\% (0/50) & 4\% (2/50) & 34\% (17/50) & 62\% (31/50) \\
One-shot & 30\% (15/50) & 36\% (18/50) & 16\% (8/50) & 18\% (9/50) \\
\bottomrule
\end{tabular}
\end{table}

\subsection{Algorithm-Level Outcomes}

Table~\ref{tab:supp_non_platemo_algorithms} gives the complete algorithm-level decomposition. EvoCoCo obtains at least one converged implementation for nine of the ten sources and achieves five of five converged attempts for DEMO and Two-Arch2. R-DEMO is the only source with no converged EvoCoCo attempt; four of its five matched outputs execute but remain above the convergence threshold. One-shot generation obtains convergence only for DEMO and ISDE+.

\begin{table}[H]
\centering
\caption{Algorithm-level execution and convergence counts in the experiment evaluating transfer to external MOEA implementations. Each algorithm contributes five matched attempts per method.}
\label{tab:supp_non_platemo_algorithms}
\scriptsize
\setlength{\tabcolsep}{3.5pt}
\begin{tabular}{l cccc cccc}
\toprule
& \multicolumn{4}{c}{EvoCoCo} & \multicolumn{4}{c}{One-shot} \\
\cmidrule(lr){2-5}\cmidrule(lr){6-9}
Algorithm & Syn. & Run. & Conv. fail & Conv. & Syn. & Run. & Conv. fail & Conv. \\
\midrule
DEMO & 0 & 0 & 0 & 5 & 1 & 0 & 0 & 4 \\
DEMO\_IBEA & 0 & 0 & 1 & 4 & 0 & 0 & 5 & 0 \\
DEMO\_PBEA & 0 & 0 & 1 & 4 & 0 & 5 & 0 & 0 \\
ISDE\_plus & 0 & 1 & 0 & 4 & 0 & 0 & 0 & 5 \\
LSMaODE & 0 & 0 & 1 & 4 & 3 & 1 & 1 & 0 \\
PAR\_DEMO\_IND & 0 & 0 & 4 & 1 & 4 & 1 & 0 & 0 \\
PAR\_DEMO\_NDS & 0 & 0 & 3 & 2 & 5 & 0 & 0 & 0 \\
RTEA & 0 & 0 & 3 & 2 & 0 & 3 & 2 & 0 \\
R\_DEMO & 0 & 1 & 4 & 0 & 0 & 5 & 0 & 0 \\
Two\_Arch2 & 0 & 0 & 0 & 5 & 2 & 3 & 0 & 0 \\
\bottomrule
\end{tabular}
\end{table}

The algorithm-level counts show that EvoCoCo's successes extend across nine sources, whereas one-shot success is concentrated in DEMO and ISDE+. This imbalance is another reason not to compare conditional IGD values without the accompanying reliability denominators.

\subsection{Truncation and Residual Failures}

Manual inspection identifies output truncation as the dominant syntax-failure pattern in one-shot generation. Many of the 15 syntax-failing files terminate inside a token, expression, import, function call, or formatted string rather than exhibiting an isolated local syntax error. Table~\ref{tab:supp_non_platemo_truncation} lists representative endings. These failures are concentrated in longer or structurally complex sources: all five PAR\_DEMO\_NDS attempts, four PAR\_DEMO\_IND attempts, three LSMaODE attempts, two Two\_Arch2 attempts, and one DEMO attempt are incomplete or unparsable.

\begin{table}[H]
\centering
\caption{Representative truncation patterns in incomplete one-shot conversion outputs.}
\label{tab:supp_non_platemo_truncation}
\scriptsize
\setlength{\tabcolsep}{4pt}
\begin{tabular}{l p{0.27\textwidth} p{0.35\textwidth}}
\toprule
File & Parser diagnostic & Truncated tail \\
\midrule
\texttt{Two\_Arch2\_run4.py} & Bracket never closed & \texttt{while curr\_pop.shape[0} \\
\texttt{PAR\_DEMO\_NDS\_run4.py} & Parenthesis never closed & \texttt{n\_objs=n\_} \\
\texttt{PAR\_DEMO\_IND\_run5.py} & Parenthesis never closed & \texttt{torch.where(mask, off\_pop, self.} \\
\texttt{LSMaODE\_run5.py} & Bracket never closed & \texttt{torch.clamp(off\_dec[rows, k\_} \\
\texttt{PAR\_DEMO\_NDS\_run2.py} & Invalid syntax & \texttt{from ev} \\
\texttt{DEMO\_run4.py} & Incomplete format specifier & \texttt{\{total\_time:.4} \\
\texttt{LSMaODE\_run3.py} & Parenthesis never closed & \texttt{p2[case\_b] - p} \\
\bottomrule
\end{tabular}
\end{table}

The two EvoCoCo runtime failures occur for ISDE+ and R-DEMO. Its other unsuccessful matched attempts are convergence failures: the generated implementations are complete and runnable but do not satisfy the convergence criterion. The experiment therefore supports a bounded transfer claim. Staged analysis, repair, and selection increase the probability of obtaining a usable transformation from external or legacy sources, but they do not guarantee convergence for every algorithm.

\FloatBarrier
\section{Detailed Ablation Results}

The ablation experiment identifies which EvoCoCo modules address different transformation failure mechanisms. Five variants are evaluated on a stratified 12-algorithm subset with five independent generation attempts per algorithm, which yields 60 runs per variant and 300 runs in total. All variants use Gemini 3 Flash and the same DTLZ2 validation protocol as the primary migration experiment. The subset is diagnostic rather than selected to maximize success: it contains stable-success controls, convergence-challenging algorithms, MOEA/D-family algorithms, runtime-sensitive cases, and structurally complex source implementations.

\subsection{Diagnostic Strata and Variant Design}

\begin{table}[H]
\centering
\caption{Diagnostic strata and algorithm composition of the stratified 12-algorithm ablation subset.}
\label{tab:ablation_diagnostic_strata}
\scriptsize
\setlength{\tabcolsep}{4pt}
\begin{tabular}{p{0.21\textwidth} p{0.31\textwidth} p{0.40\textwidth}}
\toprule
Stratum & Algorithms & Diagnostic purpose \\
\midrule
Stable-success control & BCE-IBEA, NSGA-II-SDR & Tests whether ablations damage algorithms transformed reliably by full EvoCoCo \\
Convergence-challenging & AGE-MOEA, GrEA, SMPSO, SparseEA & Tests whether executable transformations retain acceptable optimization outcomes \\
MOEA/D-family coverage & MOEA-D-AWA, MOEA-D-PaS & Covers decomposition-based mechanisms with implementation-sensitive details \\
Runtime-crash coverage & DM-MOEA, SIBEA & Covers runtime-sensitive tensor-shape and API failures \\
Structural/interface complexity & Two\_Arch2, WOF & Covers interface mismatch and structurally complex source implementations \\
\bottomrule
\end{tabular}
\end{table}

The full EvoCoCo pipeline serves as the reference. The other variants are denoted as w/o Blueprint Agent, w/o Repair Agent, w/o Rule Retrieval, and w/o Multi-Branch. The first three remove the Blueprint Agent, Repair Agent, and rule retrieval, respectively. The w/o Multi-Branch variant uses a single tensorization branch instead of generating and selecting among multiple branches. Syntax pass denotes parsable Python code; static pass additionally checks undefined symbols, forbidden imports, and incompatible patterns; execution pass denotes a candidate that runs under the benchmark harness; and convergence requires an executable run to meet the predefined final-IGD criterion. Average IGD is calculated only over converged runs.

\subsection{Aggregate Reliability and Failure Subtypes}

\begin{figure}[H]
\centering
\includegraphics[width=0.84\textwidth]{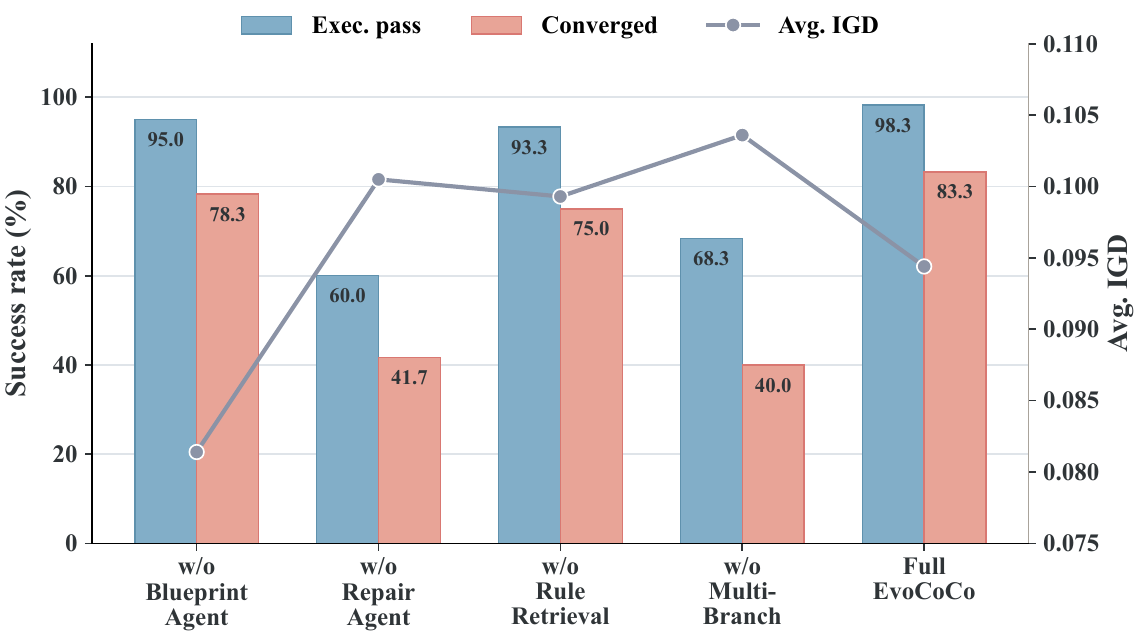}
\caption{Ablation performance on the stratified 12-algorithm subset. Bars show execution and convergence rates over 60 attempts per variant; the line shows mean IGD over converged attempts. Full EvoCoCo denotes the reference configuration.}
\label{fig:ablation_main_success_supp}
\end{figure}

Figure~\ref{fig:ablation_main_success_supp} summarizes the aggregate execution, convergence, and conditional-IGD results reported in the main paper. Full EvoCoCo achieves the highest execution and convergence rates. Removing the Repair Agent or multi-branch generation and selection produces the largest end-to-end degradation, whereas removing the Blueprint Agent or rule retrieval retains higher aggregate success but changes the underlying static and interface failure profile.

The failure analysis groups observed defects into four diagnostic subtypes. Static issues include unavailable symbols, forbidden imports, and incompatible code patterns. EvoX API misuse or missing API covers incorrect calls, missing lifecycle methods, and invalid state fields. Tensor shape mismatch covers incompatible reshape, indexing, broadcasting, or population/objective dimensions. Poor convergence denotes an executable implementation that does not meet the final-IGD criterion. These subtype counts record defects observed during generation and repair and are not mutually exclusive terminal outcomes. A candidate may therefore contain an intermediate static or runtime defect that is subsequently repaired and still contribute to the final execution or convergence pass rate.

\begin{table}[H]
\centering
\caption{Failure-subtype counts for the EvoCoCo ablation variants on the stratified 12-algorithm subset. Counts are aggregated over 60 conversion attempts per variant.}
\label{tab:ablation_failure_subtypes}
\scriptsize
\setlength{\tabcolsep}{4pt}
\begin{tabular}{l c c c c}
\toprule
Variant & Static & Poor conv. & API misuse/missing & Shape mismatch \\
\midrule
w/o Blueprint Agent & 14 & 10 & 2 & 0 \\
w/o Repair Agent & 2 & 12 & 11 & 7 \\
w/o Rule Retrieval & 18 & 11 & 1 & 2 \\
w/o Multi-Branch & 15 & 17 & 10 & 4 \\
Full EvoCoCo & 6 & 9 & 0 & 0 \\
\bottomrule
\end{tabular}
\end{table}

The subtype results explain why execution success alone is insufficient for interpreting the ablation. The Repair Agent ablation retains high static validity but records 11 API misuse or missing-API cases and seven tensor shape mismatches. These defects therefore emerge primarily during execution. The rule-retrieval ablation instead records 18 static issues but only three API or shape cases. This pattern indicates that retrieved implementation rules mainly prevent invalid dependencies, undefined symbols, and incompatible conventions before execution. Multi-branch generation and selection has the broadest effect: its removal produces the largest total subtype count and increases both runtime-level defects and poor-convergence outcomes.

\subsection{Complete Algorithm-Level Results}

Table~\ref{tab:ablation_subset_appendix} reports the complete algorithm-level results behind the compact main-paper table. The failure-profile column gives counts within each five-run algorithm block; runtime failures are non-executing runs, and convergence failures are executable runs that do not meet the predefined criterion.

\begingroup
\scriptsize
\setlength{\tabcolsep}{2.8pt}
\begin{longtable}{p{0.16\textwidth} p{0.20\textwidth} p{0.13\textwidth} c c c c c c p{0.18\textwidth}}
\multicolumn{10}{c}{\label{tab:ablation_subset_appendix}\normalfont\footnotesize TABLE~\thetable}\\[-0.2ex]
\multicolumn{10}{c}{\normalfont\footnotesize\scshape Algorithm-Level Ablation Results by Diagnostic Stratum.}\\[0.5ex]
\hline
Variant & Stratum & Algorithm & Syntax & Static & Exec. & Conv. & Avg. IGD & Time & Failure profile \\
\hline
\endfirsthead
\multicolumn{10}{c}{\normalfont\footnotesize TABLE~\thetable\ (Continued)}\\[-0.2ex]
\multicolumn{10}{c}{\normalfont\footnotesize\scshape Algorithm-Level Ablation Results by Diagnostic Stratum.}\\[0.5ex]
\hline
Variant & Stratum & Algorithm & Syntax & Static & Exec. & Conv. & Avg. IGD & Time & Failure profile \\
\hline
\endhead
Full EvoCoCo & Stable-success control & BCE-IBEA & 5/5 & 5/5 & 5/5 & 5/5 & 0.0782 & 14.28 & none \\
Full EvoCoCo & Stable-success control & NSGA-II-SDR & 5/5 & 5/5 & 5/5 & 5/5 & 0.0626 & 1.82 & none \\
Full EvoCoCo & Convergence-challenging & AGE-MOEA & 5/5 & 4/5 & 4/5 & 4/5 & 0.1166 & 2.43 & 1 static; 1 runtime \\
Full EvoCoCo & Convergence-challenging & GrEA & 5/5 & 5/5 & 5/5 & 3/5 & 0.0919 & 2.04 & 2 convergence \\
Full EvoCoCo & Convergence-challenging & SMPSO & 5/5 & 5/5 & 5/5 & 4/5 & 0.0999 & 0.44 & 1 convergence \\
Full EvoCoCo & Convergence-challenging & SparseEA & 5/5 & 4/5 & 5/5 & 2/5 & 0.1525 & 1.16 & 1 static; 3 convergence \\
Full EvoCoCo & MOEA/D-family coverage & MOEA-D-AWA & 5/5 & 4/5 & 5/5 & 5/5 & 0.0934 & 1.06 & 1 static \\
Full EvoCoCo & MOEA/D-family coverage & MOEA-D-PaS & 5/5 & 4/5 & 5/5 & 4/5 & 0.0789 & 1.19 & 1 static; 1 convergence \\
Full EvoCoCo & Runtime-crash coverage & DM-MOEA & 5/5 & 4/5 & 5/5 & 4/5 & 0.1529 & 1.16 & 1 static; 1 convergence \\
Full EvoCoCo & Runtime-crash coverage & SIBEA & 5/5 & 5/5 & 5/5 & 5/5 & 0.0816 & 5.86 & none \\
Full EvoCoCo & Structural/interface complexity & Two\_Arch2 & 5/5 & 5/5 & 5/5 & 5/5 & 0.0958 & 6.86 & none \\
Full EvoCoCo & Structural/interface complexity & WOF & 5/5 & 4/5 & 5/5 & 4/5 & 0.0714 & 2.70 & 1 static; 1 convergence \\
w/o Blueprint Agent & Stable-success control & BCE-IBEA & 5/5 & 2/5 & 4/5 & 3/5 & 0.0950 & 7.70 & 3 static; 1 runtime; 1 convergence \\
w/o Blueprint Agent & Stable-success control & NSGA-II-SDR & 5/5 & 4/5 & 5/5 & 5/5 & 0.0599 & 1.30 & 1 static \\
w/o Blueprint Agent & Convergence-challenging & AGE-MOEA & 5/5 & 4/5 & 5/5 & 5/5 & 0.0697 & 1.77 & 1 static \\
w/o Blueprint Agent & Convergence-challenging & GrEA & 5/5 & 5/5 & 5/5 & 0/5 & -- & 2.16 & 5 convergence \\
w/o Blueprint Agent & Convergence-challenging & SMPSO & 5/5 & 4/5 & 5/5 & 4/5 & 0.0886 & 0.88 & 1 static; 1 convergence \\
w/o Blueprint Agent & Convergence-challenging & SparseEA & 5/5 & 3/5 & 5/5 & 4/5 & 0.0777 & 3.77 & 2 static; 1 convergence \\
w/o Blueprint Agent & MOEA/D-family coverage & MOEA-D-AWA & 5/5 & 4/5 & 5/5 & 5/5 & 0.0634 & 4.44 & 1 static \\
w/o Blueprint Agent & MOEA/D-family coverage & MOEA-D-PaS & 5/5 & 4/5 & 5/5 & 5/5 & 0.0930 & 1.41 & 1 static \\
w/o Blueprint Agent & Runtime-crash coverage & DM-MOEA & 5/5 & 4/5 & 5/5 & 3/5 & 0.1202 & 0.77 & 1 static; 2 convergence \\
w/o Blueprint Agent & Runtime-crash coverage & SIBEA & 5/5 & 4/5 & 4/5 & 4/5 & 0.0703 & 1.08 & 1 static; 1 runtime \\
w/o Blueprint Agent & Structural/interface complexity & Two\_Arch2 & 5/5 & 3/5 & 4/5 & 4/5 & 0.1149 & 7.84 & 2 static; 1 runtime \\
w/o Blueprint Agent & Structural/interface complexity & WOF & 5/5 & 5/5 & 5/5 & 5/5 & 0.0690 & 1.08 & none \\
w/o Repair Agent & Stable-success control & BCE-IBEA & 5/5 & 5/5 & 4/5 & 3/5 & 0.1028 & 3.64 & 1 runtime; 1 convergence \\
w/o Repair Agent & Stable-success control & NSGA-II-SDR & 5/5 & 5/5 & 5/5 & 5/5 & 0.0628 & 0.95 & none \\
w/o Repair Agent & Convergence-challenging & AGE-MOEA & 5/5 & 5/5 & 3/5 & 2/5 & 0.1430 & 2.69 & 2 runtime; 1 convergence \\
w/o Repair Agent & Convergence-challenging & GrEA & 5/5 & 5/5 & 3/5 & 1/5 & 0.0677 & 6.82 & 2 runtime; 2 convergence \\
w/o Repair Agent & Convergence-challenging & SMPSO & 5/5 & 5/5 & 3/5 & 1/5 & 0.1126 & 0.83 & 2 runtime; 2 convergence \\
w/o Repair Agent & Convergence-challenging & SparseEA & 5/5 & 4/5 & 4/5 & 1/5 & 0.0864 & 4.03 & 1 static; 1 runtime; 3 convergence \\
w/o Repair Agent & MOEA/D-family coverage & MOEA-D-AWA & 5/5 & 5/5 & 0/5 & 0/5 & -- & -- & 5 runtime \\
w/o Repair Agent & MOEA/D-family coverage & MOEA-D-PaS & 5/5 & 4/5 & 2/5 & 2/5 & 0.1353 & 0.88 & 1 static; 3 runtime \\
w/o Repair Agent & Runtime-crash coverage & DM-MOEA & 5/5 & 5/5 & 5/5 & 4/5 & 0.1256 & 1.24 & 1 convergence \\
w/o Repair Agent & Runtime-crash coverage & SIBEA & 5/5 & 5/5 & 0/5 & 0/5 & -- & -- & 5 runtime \\
w/o Repair Agent & Structural/interface complexity & Two\_Arch2 & 5/5 & 5/5 & 3/5 & 3/5 & 0.1203 & 12.59 & 2 runtime \\
w/o Repair Agent & Structural/interface complexity & WOF & 5/5 & 5/5 & 4/5 & 3/5 & 0.0679 & 0.83 & 1 runtime; 1 convergence \\
w/o Rule Retrieval & Stable-success control & BCE-IBEA & 5/5 & 5/5 & 5/5 & 4/5 & 0.0876 & 11.67 & 1 convergence \\
w/o Rule Retrieval & Stable-success control & NSGA-II-SDR & 5/5 & 5/5 & 5/5 & 5/5 & 0.1595 & 5.47 & none \\
w/o Rule Retrieval & Convergence-challenging & AGE-MOEA & 5/5 & 4/5 & 5/5 & 4/5 & 0.1284 & 8.66 & 1 static; 1 convergence \\
w/o Rule Retrieval & Convergence-challenging & GrEA & 5/5 & 4/5 & 5/5 & 0/5 & -- & 9.23 & 1 static; 5 convergence \\
w/o Rule Retrieval & Convergence-challenging & SMPSO & 5/5 & 4/5 & 4/5 & 4/5 & 0.0958 & 0.80 & 1 static; 1 runtime \\
w/o Rule Retrieval & Convergence-challenging & SparseEA & 5/5 & 3/5 & 4/5 & 3/5 & 0.0985 & 1.14 & 2 static; 1 runtime; 1 convergence \\
w/o Rule Retrieval & MOEA/D-family coverage & MOEA-D-AWA & 5/5 & 3/5 & 5/5 & 5/5 & 0.0932 & 13.81 & 2 static \\
w/o Rule Retrieval & MOEA/D-family coverage & MOEA-D-PaS & 5/5 & 1/5 & 5/5 & 5/5 & 0.0796 & 5.22 & 4 static \\
w/o Rule Retrieval & Runtime-crash coverage & DM-MOEA & 5/5 & 1/5 & 5/5 & 3/5 & 0.0729 & 2.22 & 4 static; 2 convergence \\
w/o Rule Retrieval & Runtime-crash coverage & SIBEA & 5/5 & 5/5 & 3/5 & 2/5 & 0.1175 & 9.04 & 2 runtime; 1 convergence \\
w/o Rule Retrieval & Structural/interface complexity & Two\_Arch2 & 5/5 & 2/5 & 5/5 & 5/5 & 0.0908 & 26.49 & 3 static \\
w/o Rule Retrieval & Structural/interface complexity & WOF & 5/5 & 5/5 & 5/5 & 5/5 & 0.0712 & 1.04 & none \\
w/o Multi-Branch & Stable-success control & BCE-IBEA & 5/5 & 3/5 & 4/5 & 2/5 & 0.0885 & 17.43 & 2 static; 1 runtime; 2 convergence \\
w/o Multi-Branch & Stable-success control & NSGA-II-SDR & 5/5 & 4/5 & 4/5 & 2/5 & 0.0596 & 7.13 & 1 static; 1 runtime; 2 convergence \\
w/o Multi-Branch & Convergence-challenging & AGE-MOEA & 5/5 & 4/5 & 3/5 & 0/5 & -- & 3.36 & 1 static; 2 runtime; 3 convergence \\
w/o Multi-Branch & Convergence-challenging & GrEA & 5/5 & 5/5 & 3/5 & 0/5 & -- & 10.59 & 2 runtime; 3 convergence \\
w/o Multi-Branch & Convergence-challenging & SMPSO & 5/5 & 5/5 & 5/5 & 4/5 & 0.0960 & 1.83 & 1 convergence \\
w/o Multi-Branch & Convergence-challenging & SparseEA & 5/5 & 3/5 & 4/5 & 3/5 & 0.0711 & 6.67 & 2 static; 1 runtime; 1 convergence \\
w/o Multi-Branch & MOEA/D-family coverage & MOEA-D-AWA & 5/5 & 4/5 & 4/5 & 4/5 & 0.1347 & 13.42 & 1 static; 1 runtime \\
w/o Multi-Branch & MOEA/D-family coverage & MOEA-D-PaS & 5/5 & 2/5 & 4/5 & 3/5 & 0.0983 & 3.92 & 3 static; 1 runtime; 1 convergence \\
w/o Multi-Branch & Runtime-crash coverage & DM-MOEA & 5/5 & 5/5 & 2/5 & 1/5 & 0.1606 & 5.57 & 3 runtime; 1 convergence \\
w/o Multi-Branch & Runtime-crash coverage & SIBEA & 5/5 & 5/5 & 1/5 & 1/5 & 0.1329 & 1.46 & 4 runtime \\
w/o Multi-Branch & Structural/interface complexity & Two\_Arch2 & 5/5 & 2/5 & 3/5 & 3/5 & 0.0879 & 13.60 & 3 static; 2 runtime \\
w/o Multi-Branch & Structural/interface complexity & WOF & 5/5 & 3/5 & 4/5 & 1/5 & 0.2025 & 6.49 & 2 static; 1 runtime; 3 convergence \\
\hline
\multicolumn{10}{p{0.96\textwidth}}{\footnotesize Note: failure profile reports run counts within each five-run algorithm block. Runtime failures are non-executing runs; convergence failures are executable runs that do not satisfy the predefined convergence criterion. Avg. IGD and mean time are computed over converged and executable runs, respectively.}\\
\end{longtable}
\endgroup

The detailed rows localize these aggregate effects. Without the Repair Agent, all five MOEA-D-AWA and all five SIBEA attempts fail during execution; the complete pipeline executes all ten corresponding attempts. Removing multi-branch generation and selection is particularly harmful for DM-MOEA, SIBEA, and WOF, for which only one of five attempts converges. Rule retrieval and the Blueprint Agent mainly affect static validity and interface alignment, while the full pipeline removes the observed API-misuse and tensor-shape subtype failures in this diagnostic subset. These algorithm-level patterns are consistent with the aggregate trends in the main paper.

\end{document}